%% file: main.tex
\documentclass{article}

\usepackage{microtype}
\usepackage{graphicx}
\usepackage[labelformat=simple]{subcaption}
\usepackage{booktabs} 
\usepackage[utf8]{inputenc}
\usepackage[english]{babel}
 
\usepackage{amsthm}
\newtheorem{theorem}{Theorem}
\newtheorem{corollary}{Corollary}
\newtheorem{lemma}{Lemma}
\newtheorem{assumption}{Assumption}

\usepackage{hyperref}

\usepackage{amsmath,amsthm,amssymb}
\usepackage{bm}
\usepackage{sidecap}

\usepackage{makecell}
\usepackage{xcolor}

\usepackage{xfrac}

\usepackage[accepted]{icml2020}

\icmltitlerunning{C-ADAM: An Adaptive Compositional Solver}

\begin{document}

\twocolumn[
\icmltitle{Compositional ADAM: An Adaptive Compositional Solver}



\icmlsetsymbol{equal}{*}

\begin{icmlauthorlist}
\icmlauthor{Rasul Tutunov}{equal,to}
\icmlauthor{Minne Li}{equal,to,goo}
\icmlauthor{Alexander I. Cowen-Rivers}{equal,to}
\icmlauthor{Jun Wang}{goo}
\icmlauthor{Haitham Bou-Ammar}{to,goo}
\end{icmlauthorlist}

\icmlaffiliation{to}{Huawei Technologies Research and Development, London, UK.}
\icmlaffiliation{goo}{University College London, London, UK.}
\icmlcorrespondingauthor{Rasul Tutunov}{rasul.tutunov@huawei.com}
 
\icmlkeywords{Machine Learning, ICML}

\vskip 0.3in
]        

\printAffiliationsAndNotice{\icmlEqualContribution} 




\begin{abstract}
In this paper we present C-ADAM, the first adaptive solver for compositional problems involving a non-linear functional nesting of expected values. We prove that C-ADAM converges to a stationary point in $\mathcal{O}(\delta^{-2.25})$ first oracle calls with $\delta$ being a precision parameter. Moreover, we demonstrate the importance of our results by bridging, for the first time, model-agnostic meta-learning (MAML) and compositional optimisation showing the fastest known rates for deep network adaptation to-date. Finally, we validate our findings in a set of experiments from portfolio optimisation and meta-learning. Our results manifest significant sample complexity reductions compared to both standard and compositional solvers.  
\end{abstract}

\section{Introduction}
\label{introduction}
The availability of large data-sets coupled with improvements in optimisation algorithms and the growth in computing power has led to an unprecedented interest in machine learning and its applications in, for instance, medical imaging~\cite{faes2019automated}, autonomous self-driving~\cite{bojarski2016end}, fraud detection~\cite{wang2020deep}, computer games~\cite{mnih2015human, silver2017mastering,tian2018learning}, and many other fields. 

Fueling these successes are novel developments in non-convex optimisation, which is, by now, a wide-reaching field with algorithms involving zero \cite{hu2016bandit, shamir2017optimal, gabillon2019derivative}, first \cite{FirstOrder_review_1,beck2017first,ClassFirstOrder}, and second-order methods \cite{BoydBook,Nesterov2006,TutunovBJ16}. Despite such diversity, first-order algorithms are still highly regarded as the go-to technique when it comes to large-scale machine learning applications due to their ease of implementation, and memory and computational efficacy. First-order optimisers can further be split into gradient-based, momentum, and adaptive solvers, each varying in the process by which a decent direction is computed. On the one hand, momentum techniques~\cite{Nesterov, APG} handle the direction itself, while adaptive designs modify learning rates~\cite{ADAgrad,zeiler2012adadelta,RMSprop}. In spite of numerous theoretical developments~\cite{liu2018stochastically, fang2018spider,allenzhu2016variance}, ADAM -- an adaptive optimiser originally proposed in~\cite{kingma2014adam}, and then theoretically grounded in~\cite{NonAdaptiveGR,OnConvAdamand_Beyond} -- (arguably) retains state-of-the-art status in machine learning practice. 

Although this first wave of machine learning applications led to major breakthroughs and paradigm shifts, experts and practitioners are rapidly realising limitations of current models in that they require a large amount of training data and are slow to adapt to novel tasks even when these tasks involve minor changes in the data distribution. The quest for having fast and adaptable models re-surged interest in general and model-agnostic machine learning forms, such as meta~\cite{FinnAL17,fallah2019convergence}, lifelong~\cite{thrun1998learning,Ruvolo2013ELLA, Ammar1, Ammar3, Ammar2}, and few-shot learning~\cite{Few}. 

When investigated closer, one comes to realise that such key problems typically involve non-linear functional nesting of expected values. For instance, in both meta and lifelong learning, a learner attempts to acquire a model across multiple tasks, each exhibiting randomness in its data generating process. In other words, the goal of the agent is to find a set of deep network parameters that minimise an expectation overall tasks, where each task's loss function involves another expectation over training data\footnote{Please note that in model-agnostic meta-learning such an expectation can involve more than two layers. In this paper, however, we focus on the original presentation~\cite{FinnAL17} that performs one gradient step, and as such, only includes two nestings. Generalising to an $m$ nesting is an interesting avenue for future research.}. 

It is clear from the above that problems of this nature are compositional abiding by the form introduced in~\cite{wang2014stochastic}. Such a bridge, however, has not been formally established before, partially due to the discrepancy between communities and the lack of stochastic and adaptive compositional solvers that allow for scalable and effective algorithms (see Section~\ref{Sec:RelatedWork})\footnote{Though stochastic forms are proposed in literature, we would like to mention that these typically tackle relaxed versions that assume independent coupling random variables. }. With meta-learning as our motivation, we propose compositional ADAM (C-ADAM), the first, to the best of our knowledge, \emph{adaptive compositional optimiser} that provably converges to a first-order stationary point. Along the way, we also extend standard ADAM's proof from ~\cite{NonAdaptiveGR} to a more realistic setting of hyper-parameters closer to these used by practitioners (i.e., none of the $\beta$'s are set to zero). Apart from theoretical rigour, C-ADAM is also stochastic allowing for mini-batching and only executes one loop easing implementation.

\textit{Unlocking major associations}, we further derive a novel connection between model-agnostic meta-learning (MAML) and compositional optimisation allowing, for the first time, the application of compositional methods in large-scale machine learning problems beyond portfolio mean-variance trade-offs, and discrete state (or linear function approximation) reinforcement learning~\cite{wang2014stochastic}. Here, we show that MAML can be written as a nested expectation problem and derive corollaries from our results shedding light on MAML's convergence guarantees. Such an analysis reveals that our method achieves best-known convergence bounds to-date\footnote{We note that results we study on MAML handle the more realistic non-convex scenario compared to these in~\cite{OnlineMAML,golmantconvergence}.} for problems involving the fast adaptation of deep networks. 
Validating our theoretical discoveries, we, finally, conduct an in-depth empirical study demonstrating the C-ADAM outperforms others by significant margins. In fact, our results demonstrate that correctly handling correlations across data and tasks using C-ADAM, one can outperform ADAM (and others) in meta-learning.

\section{Related Work}\label{Sec:RelatedWork}
Since~\cite{wang2014stochastic} proposed a stochastic compositional gradient optimiser with $\mathcal{O}\left(\delta^{-4}\right)$ query complexity, much attention has been devoted towards more efficient and scalable solvers. Efforts in~\cite{Mendgi_2017}, for instance, led to further improvements achieving $\mathcal{O}(\delta^{-3.5})$ complexity by adopting an accelerated version of the approach in~\cite{wang2014stochastic}. Building on these results, further developments employed Nesterov acceleration techniques to arrive to an $\mathcal{O}(\delta^{-2.25})$ for gradient query complexities \cite{Lui2016}. Concurrently, other authors considered a \emph{relaxation} of the problem by studying a finite-sum alternative, i.e., a Monte-Carlo approximation to the stochastic objectives\footnote{We note that such forms are a special case of the problem considered in this work. Here, $m$ and $n$ denote the number of samples for the inner and outer nesting, respectively.}. Here, \citet{liu2017variance} examined variance reduction techniques to achieve a total complexity of $\mathcal{O}((m+n)^{0.8}\delta^{-1})$. Succeedingly, \citet{ZHuo2017} adapted variance reduction to proximal algorithms acquiring $\mathcal{O}((m+n)^{\frac{2}{3}}\delta^{-1})$ bound . A similar complexity bound was obtained in \cite{liu2018stochastically} by applying two kinds of variance reduction techniques. Such bound has been later improved to $\mathcal{O}((m+n)^{\frac{1}{2}}\delta^{-1})$ with  recursive gradient methods in~\cite{Hu_WenqingNIPS2019}.\footnote{In our related work review, we only focus on non-convex objectives. Notice that authors devoted lots of research to convex/strongly convex settings, e.g., \cite{lin2018improved,bedi2019nonparametric,liu2018stochastically,lian2016finitesum,liu2017variance}. }

\section{Compositional ADAM}

\subsection{Problem Definition, Notation \& Assumptions}\label{Sec:ProbDef}
We focus on the following two-level nested optimisation problem:
\begin{equation}
\label{Eq:Prob}
    \min_{\bm{x}\in\mathbb{R}^{p}}\mathcal{J}(\boldsymbol{x}) = \mathbb{E}_{\nu}\left[f_{\nu}\left(\mathbb{E}_{\omega}\left[g_{\omega}(\boldsymbol{x})\right] \right)\right], 
\end{equation}
where $\nu$ and $\omega$ are random variables sampled from $\nu \sim \mathcal{P}_{\nu}(\cdot)$ and $\omega \sim \mathcal{P}_{\omega}(\cdot)$ with $\mathcal{P}_{\nu}(\cdot)$ and $\mathcal{P}_{\omega}(\cdot)$ being unknown. Furthermore, for any $\nu$ and $\omega$, $f_\nu(\cdot):\mathbb{R}^{q}\to \mathbb{R}$ is a function mapping to real-values, while $g_\omega(\cdot):\mathbb{R}^{p}\to \mathbb{R}^{q}$ represents a map transforming the $p$-dimensional optimisation variable to a $q$-dimensional space. We make no restrictive assumptions on the probabilistic relationship between $\nu$ and $\omega$, which can be either dependent or independent.  

As our method utilises gradient information in performing updates, it is instructive to clearly state assumptions we exploit in designing compositional ADAM. Defining $\mathcal{J}(\bm{x}) = f\left(g(\bm{x})\right)$, with $f(\boldsymbol{y}) = \mathbb{E}_\nu[f_\nu(\boldsymbol{y})]$, and $g(\boldsymbol{x}) = \mathbb{E}_\omega[g_w(\boldsymbol{x})]$, gradients can be written as: $  \nabla_{\bm{x}} \mathcal{J}(\bm{x})= \nabla_{\bm{x}} g(\bm{x})^{\mathsf{T}} \nabla_{\bm{y}}f\left(g(\bm{x})\right)$. Analogous to most optimisation techniques, we prefer if gradient are Lipschitz continuous by that introducing regularity conditions that can aid in designing better behaved algorithms. Rather than restricting overall objectives, however, we realise that it suffices for gradients of $f_{\nu}(\cdot)$ and $g_{\omega}(\cdot)$ to be Lipschitz and bounded so as to achieve such results. As such, we assume: 
\begin{assumption}\label{assum_1}
We make the following assumptions:
  \begin{enumerate}
      \item We assume that $f_{\nu}(\cdot)$ is bounded above by $B_{f}$ and its gradients by $M_{f}$, i.e., $\forall \bm{y} \in \mathbb{R}^{q}$ and for any $\nu$, $\left|f_{\nu}(\bm{y})\right| \leq B_{f}$, and $\left|\left|\nabla f_{\nu}(\bm{y})\right|\right| \leq M_f$. 
      \item We assume that $f_{\nu}(\cdot)$ is $L_{f}$-smooth, i.e., for any $\bm{y}_{1},  \bm{y}_{2} \in \mathbb{R}^{q}$ and for any $\nu$, we have: $||\nabla f_\nu(\boldsymbol{y}_1) - \nabla f_\nu(\boldsymbol{y}_2)||_2 \le L_{f}||\boldsymbol{y}_1 - \boldsymbol{y}_2 ||_2$. 
    \item We assume that the mapping $g_{\omega}(\bm{x})$ is $M_{g}$- Lipschitz continuous, i.e., for any $(\bm{x}_{1}, \bm{x}_{2}) \in \mathbb{R}^{p}$ and for any $\omega$, we have:    $ \left|\left|g_{\omega}(\bm{x}_{1}) - g_{\omega}(\bm{x}_{2})\right|\right|_{2} \leq  M_{g}\left|\left|\bm{x}_{1} - \bm{x}_{2}\right|\right|_{2}$.
    \item We assume that the mapping $g_{\omega}(\bm{x})$ is $L_{g}$-smooth, i.e., for any $(\bm{x}_{1}, \bm{x}_{2}) \in \mathbb{R}^{p}$ and for any $\omega$, we have: $||\nabla g_{\omega}(\boldsymbol{x}_{1}) - \nabla g_{\omega}(\boldsymbol{x}_{2})||_2 \le L_{g}||\boldsymbol{x}_{1} - \boldsymbol{x}_{2} ||_2$. 
  \end{enumerate}
\end{assumption}

Given the above assumptions, we can now prove Lipschitz smoothness of\footnote{Due to space constraints, we relegate proofs to the appendix.} $\mathcal{J}(\bm{x})$: 
\begin{lemma}
If Assumption \ref{assum_1} holds, then $\mathcal{J}(\bm{x})$ is $L$-Lipschitz smooth, i.e., $    \left|\left|\nabla\mathcal{J}(\bm{x}_{1}) - \nabla\mathcal{J}(\bm{x}_{2}) \right|\right|_{2} \leq L \left|\left|\bm{x}_{1} - \bm{x}_{2}\right|\right|_{2}$, 
for all $(\bm{x}_{1}, \bm{x}_{2})\in \mathbb{R}^{p}$, with $L = M_{g}^{2}L_f + L_g M_f$.
\end{lemma}

We now focus on the process by which gradients are evaluated. As we assume that distributions $\mathcal{P}_{\nu}(\cdot)$ and $\mathcal{P}_{\omega}(\cdot)$ are unknown, we introduce two first-order oracles that can be queried to return gradients and function values. We presume $\text{Oracle}_{f}(\cdot, \cdot)$ and $\text{Oracle}_{g}(\cdot, \cdot)$, such that, at a time instance $t$, for any two fixed vectors $\bm{z}_{t} \in \mathbb{R}^{p}$, $\bm{y}_{t} \in \mathbb{R}^{q}$, and for any two integers, representing batch sizes, $K_{t}^{(1)}$ and $K_{t}^{(2)}$, these oracles return the following collection: 
\begin{align*}
    \text{Oracle}_{f}\left(\bm{y}_{t}, K_{t}^{(1)}\right) &= \left\{\langle \nu_{t_{i}}, \nabla f_{\nu_{t_{i}}}(\bm{y}_{t})\rangle\right\}_{i=1}^{K_{t}^{(1)}},  \\
    \text{Oracle}_{g}\left(\bm{z}_{t}, K_{t}^{(2)}\right) &= \left\{\langle \omega_{t_{i}}, g_{\omega_{t_{i}}}(\boldsymbol{z}_t), \nabla g_{\omega_{t_{i}}}(\bm{z}_{t})\rangle\right\}_{i=1}^{K_{t}^{(2)}}, 
\end{align*}
with $\{\nu_{t_{i}}\}_{i=1}^{K_{t}^{(1)}}$ and $\{\omega_{t_{i}}\}_{i=1}^{K_{t}^{(2)}}$ being identically independently distributed (i.i.d.).\footnote{Please notice that i.i.d. assumption here should be interpreted as follows: at each iteration $t$, samples $\nu_{t_i}$ and $\nu_{t_j}$ (respectively $w_{t_i}$ and $w_{t_j}$) for $i,j \in 1,\ldots, K^{(1)}_t$ (respectively $i,j \in 1,\ldots, K^{(2)}_t$) are i.i.d., but $\omega_{t_i}$ and $\nu_{t_i}$  might not be independent.}

We report our convergence complexity results in terms of the total number of calls to these first-order oracles. Similar to~\cite{wang2014stochastic}, we introduce one final set of assumptions needed to understand the randomness of the sampling process:
\begin{assumption}\label{assum_2}
For any time step $t$, $\text{Oracle}_{f}(\cdot, \cdot)$ and $\text{Oracle}_{g}(\cdot, \cdot)$ satisfy the following conditions for any $\bm{z} \in \mathbb{R}^{p}$ and $\bm{y} \in \mathbb{R}^{q}$:
\begin{enumerate}
    \item  Independent sample collections:  $$\left(\left\{\nu_{1_{i}}\right\}_{i=1}^{K_{1}^{(1)}},\left\{\omega_{1_{i}}\right\}_{i=1}^{K_{1}^{(2)}}
    \right), \dots, \left(\left\{\nu_{t_{i}}\right\}_{i=1}^{K_{t}^{(1)}},\left\{\omega_{t_{i}}\right\}_{i=1}^{K_{t}^{(1)}}
    \right);$$ 
    \item Oracles return unbiased gradient estimates for any $t$, $i$, and $j$: 
    \begin{align*}
        \mathbb{E}_{\omega_{t_{i}}}\left[g_{\omega_{t_{i}}}(\bm{z})\right] &= \mathbb{E}_{\omega}\left[g_{\omega}(\bm{z})\right], \\ 
        \mathbb{E}_{\omega_{t_{i}}, \nu_{t_{j}}}\left[\nabla g_{\omega_{t_{i}}}^{\mathsf{T}}(\bm{z})\nabla f_{\nu_{t_{j}}}(\bm{y})\right] &= \mathbb{E}_{\omega}\left[\nabla g^{\mathsf{T}}_{\omega}(\bm{z})\right] \\
        &\hspace{3em}\times\mathbb{E}_{\nu}\left[\nabla f_{\nu}(\bm{y})\right];
    \end{align*}
    
    \item Variance-bounded stochastic gradients for any $t$, $i$, and $j$: 
    \begin{align*}
        \mathbb{E}_{\nu_{t_{j}}}\left[\left|\left|\nabla f_{\nu_{t_{j}}}(\bm{y}) - \nabla \mathbb{E}_{\nu}\left[f_{\nu}(\bm{y})\right]\right|\right|_{2}^{2}\right] &\leq \sigma_{1}^{2}, \\        \mathbb{E}_{\omega_{t_{i}}}\left[\left|\left|\nabla g_{\omega_{t_{i}}}(\bm{z}) - \nabla \mathbb{E}_{\omega}\left[g_{\omega}(\bm{z})\right]\right|\right|_{2}^{2}\right] & \leq \sigma_{2}^{2}, \\
        \mathbb{E}_{\omega_{t_{i}}} \left[\left|\left|g_{\omega_{t_{i}}}(\bm{z}) - \mathbb{E}_{\omega}[g_{\omega}(\bm{z})]\right|\right|^{2}_{2}\right] &\leq \sigma_{3}^{2}.
    \end{align*}
\end{enumerate}
\end{assumption}
\subsection{Algorithmic Development \& Theoretical Results}
We now propose our algorithm providing its convergence properties. We summarise our solver, titled C-ADAM, in the pseudo-code of Algorithm~\ref{Algo:ADAM}.

On a high level, C-ADAM shares similarities with original ADAM~\citep{kingma2014adam} in that it exhibits both main, $\bm{x}_{t}$, and auxiliary variables $\bm{m}_{t}$ and $\bm{v}_{t}$. Similar to ADAM, at each iteration, we compute two parameters corresponding to a weighted combination of historical and current gradients, and a weighted combination of squared components that relate to variances. When performing parameter updates, however, we deviate from ADAM by introducing additional auxiliary variables essential for an extrapolation smoothing scheme that allows for fast estimation of the variances of $\mathbb{E}_\omega[g_{\omega}(\bm{x}_t)]$.

In addition to precision parameters and an initialisation, our algorithm acquires a schedule of learning rates and mini-batches as inputs. Given such inputs, we then execute a loop for $\mathcal{O}(\delta^{-\frac{4}{5}})$ steps to return a $\delta$-first-order stationary point to the problem in Equation~\ref{Eq:Prob}. The loop operates in two main phases. In the first, necessary variables needed for parameter updates are computed according to lines 5-6, while the second updates each of $\bm{x}_{t}$, $\bm{y}_{t}$, and $\bm{z}_{t}$. Similar to ADAM  $\bm{x}_{t}$ is updated with no additional calls to the oracle. Contrary to standard ADAM, on the other hand, our method makes another call to the oracle to sample $K_{t}^{(3)}$ function values before revising the value of $\bm{y}_{t}$, see lines 8-9. 
\begin{algorithm}[t!]
\caption{Compositional ADAM (C-ADAM)}
\label{Algo:ADAM}
\begin{algorithmic}[1]
\STATE \textbf{Inputs:} Initial variable $\bm{x}_{1} \in \mathbb{R}^{p}$, precision parameter $\delta \in (0, 1)$, positive constant $\epsilon$, running time $T= \mathcal{O}(\delta^{-\frac{5}{4}})$, learning rates $\left\{\left\langle\alpha_{t}, \beta_{t},\gamma^{(1)}_{t}, \gamma_{t}^{(2)}\right\rangle\right\}_{t=1}^{T}$, batch sizes $\left\{\left\langle K_{t}^{(1)}, K_{t}^{(2)}, K_{t}^{(3)}\right\rangle\right\}_{t=1}^{T}$ 
\STATE \textbf{Initialisation:} Initialise $\bm{z}_{1} = \bm{x}_{1}$, $\bm{y}_{1} = \boldsymbol{0} \in \mathbb{R}^{q}$, and $\bm{m}_{0} = \bm{v}_{0} = \boldsymbol{0} \in \mathbb{R}^{p}$
\STATE \textbf{for} $t=1$ to $T$ \textbf{do:}
\STATE \underline{\textbf{Computing Necessary Variables:}}
\STATE \hspace{2em} Call oracles to compute: 
\begin{align*}
    \overline{\nabla f_{t}(\bm{y}_{t})} &= \frac{1}{K_{t}^{(1)}} \sum_{i=1}^{K_{t}^{(1)}} \nabla f_{\nu_{t_{i}}}(\bm{y}_{t}) \\    \overline{\nabla g_{t}(\bm{x}_{t})} &= \frac{1}{K_{t}^{(2)}} \sum_{j=1}^{K_{t}^{(2)}} \nabla g_{\omega_{t_{j}}}(\bm{x}_{t})
\end{align*}
\STATE \hspace{2em} With $ \overline{\nabla \mathcal{J}(\bm{x}_{t})} = \overline{\nabla g_{t}(\bm{x}_{t})}^{\mathsf{T}} \overline{\nabla f_{t}(\bm{y}_{t})}$ compute:
\begin{align*}
    \bm{m}_{t} & = \gamma_{t}^{(1)}\bm{m}_{t-1} + \left(1-\gamma_{t}^{(1)}\right)\overline{\nabla \mathcal{J}(\bm{x}_{t})}\\ 
    \bm{v}_{t} & = \gamma_{t}^{(2)}\bm{v}_{t-1} + \left(1 - \gamma_{t}^{(2)}\right)\left[\overline{\nabla \mathcal{J}(\bm{x}_{t})}\right]^{2}
\end{align*}
\STATE \underline{\textbf{Variable Updates:}}
\STATE \hspace{2em} Update main and first auxiliary variable using: 
\begin{align*}
    \bm{x}_{t+1} & = \bm{x}_{t} - \alpha_{t} \frac{\bm{m}_{t}}{\sqrt{\bm{v}_{t}} + \epsilon} \\
    \bm{z}_{t+1} & = \left(1 -\frac{1}{\beta}_{t}\right)\bm{x}_{t} + \frac{1}{\beta_{t}}\bm{x}_{t+1}
    \end{align*}
\STATE \hspace{2em} To update the second auxiliary variable perform: 
\begin{equation*}
    \bm{y}_{t+1} = (1-\beta_{t})\bm{y}_{t} + \beta_{t}\overline{\bm{g}_{t}(\bm{z}_{t+1})}, 
\end{equation*}
with $\overline{\bm{g}_{t}(\bm{z}_{t+1})}=\frac{1}{K_{t}^{(3)}}\sum_{i=1}^{K_{t}^{(3)}}g_{\omega_{t_{i}}}(\bm{z}_{t+1})$ from $\text{Oracle}_{g}\left(\bm{z}_{t+1}, K_{t}^{(3)}\right)$ 
\STATE \textbf{Output:} Return a uniform sample from $\left\{\bm{x}_{t}\right\}_{t=1}^{T}$
\end{algorithmic}
\end{algorithm}

The remainder of this section is dedicated to providing theoretical guarantees for Algorithm~\ref{Algo:ADAM}. We, next, prove that our algorithm converges to a $\delta$-first-order stationary point after $T =\mathcal{O}(\delta^{-\frac{5}{4}})$ iterations: 
\begin{theorem}[Main Result]
Consider Algorithm~\ref{Algo:ADAM} with a parameter setup given by: $\alpha_{t} = \sfrac{C_{\alpha}}{t^{\frac{1}{5}}}, \beta_{t} = C_{\beta}, K_{t}^{(1)} = C_{1}t^{\frac{4}{5}}, K_{t}^{(2)} = C_{2}t^{\frac{4}{5}},  K_{t}^{(3)}  = C_{3} t^{\frac{4}{5}}, \gamma_{t}^{(1)}= C_{\gamma}\mu^{t}, \ \gamma_{2}^{(t)}  = 1 - \sfrac{C_{\alpha}}{t^{\frac{2}{5}}}(1 - C_{\gamma}\mu^{t})^{2}$, for some positive constants $C_{\alpha}, C_{\beta}, C_{1}, C_{2}, C_{3}, C_{\gamma}, \mu$ such that $C_{\beta} < 1$ and $\mu \in (0,1)$. For any $\delta \in (0,1)$, Algorithm~\ref{Algo:ADAM} outputs, in expectation, a $\delta$-approximate first-order stationary point $\tilde{\bm{x}}$ of $\mathcal{J}(\bm{x})$. That is: 
\begin{equation*}
    \mathbb{E}_{\textrm{total}}\left[\left|\left|\nabla\mathcal{J}(\tilde{\bm{x}})\right|\right|_{2}^{2}\right] \leq \delta,  
\end{equation*}
with ``total'' representing \emph{all} incurred randomness. 
Moreover, Algorithm~\ref{Algo:ADAM} acquires $\tilde{\bm{x}}$ with an overall oracle complexity for $\text{Oracle}_{f}(\cdot, \cdot)$ and $\text{Oracle}_{g}(\cdot, \cdot)$ of the order $\mathcal{O}\left(\delta^{-\frac{9}{4}}\right)$. 
\end{theorem}
\textbf{Proof Road-Map:} In achieving Theorem 1, we study the effect of the updates in lines 4-9 of Algorithm~\ref{Algo:ADAM} on the change in function value between two iterations. When attempting such an analysis, we realise the random effect introduced by gradient sub-sampling (e.g., line 5 in Algorithm~\ref{Algo:ADAM}). Consequently, we bound \emph{expected} loss value reduction with respect to all randomness induced at some iteration $t$. With this achieved, we then focus on bounding expected loss reduction with respect to all iterations $t=1, \dots, T$ exploiting the fact of independence across two successive updates. Throughout our proof, we also realise the need to study two additional components that we derive by generalising lemmas originally presented in~\cite{wang2014stochastic}. We note that our proof is not just a mere application of the results in~\cite{NonAdaptiveGR} to a compositional setting. In fact, our derivations are novel in that they follow alternative directions combining and generalising results from~\cite{wang2014stochastic, Mendgi_2017} with these from~\cite{NonAdaptiveGR}. Due to space constraints, all proofs can be found in the Appendix~A. 

\subsubsection{Preliminary Lemmas:}
First, we detail two essential lemmas that are adjusted from the work in~\cite{wang2014stochastic} to deal with our ADAM solver. 

\begin{lemma}[Auxiliary Variable Properties]\label{aux_lemma_1}
Consider auxiliary variable updates in lines 8 and 9 in Algorithm~\ref{Algo:ADAM}. 
Let $\mathbb{E}_{\text{total}}[\cdot]$ denote the expectation with respect to \emph{all} incurred randomness. For any $t$, the following holds: 
\begin{align*}
    &\mathbb{E}_{\text{total}}\left[\left|\left|g(\bm{x}_{t+1}) -\bm{y}_{t+1}\right|\right|^2_{2}\right] \leq \frac{L_{g}^{2}}{2}\mathbb{E}_{\text{total}}\left[\mathcal{D}_{t+1}^{2}\right]\\
    & \hspace{15em}+ 2 \mathbb{E}_{\text{total}}\left[\left|\left|\boldsymbol{\mathcal{E}}_{t+1}\right|\right|_{2}^{2}\right],
\end{align*}
with $g(\bm{x}_{t+1}) = \mathbb{E}_{\omega}[g_{\omega}(\bm{x}_{t+1})]$, and $\mathcal{D}_{t+1}$, $||\boldsymbol{\mathcal{E}}_{t+1}||^2_2$ satisfy the following recurrent inequalities:
\begin{align*}
    &\mathcal{D}_{t+1}  \leq  (1 - \beta_{t}) \mathcal{D}_{t} +\frac{2M_{g}^{2}M_{f}^{2}}{\epsilon^{2}}\frac{\alpha_{t}^{2}}{\beta_{t}} + \beta_{t}\mathcal{F}_{t}^{2},\\\nonumber
    &\mathbb{E}_{\text{total}}\left[\left|\left|\boldsymbol{\mathcal{E}}_{t+1}\right|\right|_{2}^{2}\right]  \leq \left(1 - \beta_{t}\right)^{2}\mathbb{E}_{\text{total}}\left[\left|\left|\boldsymbol{\mathcal{E}}_{t}\right|\right|_{2}^{2}\right] + \frac{\beta_{t}^{2}}{K_{t}^{(3)}} \sigma_{3}^{2},\\\nonumber
    &\mathcal{F}_{t}^{2} \leq (1 - \beta_{t-1})\mathcal{F}_{t-1}^{2} + \frac{4 M_{g}^{2}M_{f}^{2}}{\epsilon^{2}}\frac{\alpha_{t-1}^{2}}{\beta_{t-1}},
 \end{align*}
 and $\mathcal{D}_1 = 0,\ \  \mathbb{E}_{\text{total}}\left[||\boldsymbol{\mathcal{E}}_1||^2_2\right] = ||g(\boldsymbol{x}_1)||^2_2, \ \ \mathcal{F}_1 = 0$.
\end{lemma}

 Having established the first preliminary lemma depicting relationships between auxiliary variable updates, we now present a second one essential in our proof. 
\begin{lemma}[Recursion Property]
Let $\eta_t = \frac{C_{\eta}}{t^a}$, $\zeta_t = \frac{C_{\zeta}}{t^b}$, where $C_{\eta} > 1 + b - a$, $C_{\zeta} > 0$, $(b - a)\notin[-1,0]$ and $a\le 1$. Consider the following recurrent inequality: $
    A_{t+1} \le (1 - \eta_t + C_1\eta^2_t)A_t + C_2\zeta_t$, where $C_1, C_2 \ge 0$. Then, there is a constant $C_{A} > 0 $ such that  $A_t \le  \frac{C_{A}}{t^{b-a}}$.
\end{lemma}
We can easily adapt the above lemma to the specifics of our algorithm bounding expected norm differences between $g(\bm{x}_{t})$ and $\bm{y}_{t}$: 
\begin{corollary}
(Recursion \& Auxiliary Variables in Algorithm~\ref{Algo:ADAM}) 
Consider Algorithm \ref{Algo:ADAM} with step sizes given by: $\alpha_t = \frac{C_{\alpha}}{t^a}, \beta_t = \frac{C_{\beta}}{t^b}, \ \ \text{and} \ \  K^{(3)}_t = C_{3}t^{e}$,
for some constants $C_{\alpha},C_{\beta}, C_{3}, a,b,e > 0$ such that $(2a-2b)\notin [-1,0]$, and $b \le 1$. For $C_{\mathcal{D}},C_{\mathcal{E}}> 0$, we have: 
\begin{align*}
    \mathbb{E}_{\text{total}}\left[||g(\boldsymbol{x}_t) - \boldsymbol{y}_t||^2_2\right] \le  \frac{L^2_gC^2_{\mathcal{D}}}{2}\frac{1}{t^{4a-4b}} + 2C^2_{\mathcal{E}}\frac{1}{t^{b+e}}.
\end{align*}

\end{corollary}
\subsubsection{Steps in Main Proof:} 
With the above lemmas established, we now detail the essential steps needed to arrive at the statement of Theorem 1. As mentioned previously, our main analysis relies on the study of the change in the value of the compositional loss between two successive iterations. Previously in Lemma 1, we have shown that under our assumptions, the loss function is Lipschitz with a constant $L$. Hence, we write:
\begin{align*}
    \mathcal{J}(\bm{x}_{t+1}) \leq \mathcal{J}(\bm{x}_{t}) & + \nabla^{\mathsf{T}}\mathcal{J}(\bm{x}_{t})\Delta\bm{x}_{t+1}+ \frac{L}{2}\left|\left|\Delta\bm{x}_{t+1}\right|\right|_{2}^{2},
\end{align*}
with $\Delta\bm{x}_{t+1} = \bm{x}_{t+1} -\bm{x}_{t}$. 

Due to randomness induced by first-order oracles, such a change, has to be analysed in expectation with respect to all randomness incurred at iteration $t$ given a fixed  $\boldsymbol{x}_t$. We define such an expectation as $\mathbb{E}_{t}[\cdot] = \mathbb{E}_{K^{(1)}_t, K^{(2)}_t, K^{(3)}_t}\left[ \cdot \Big| \boldsymbol{x}_t\right]$ to consider all sampling performed at the $t^{th}$ iteration. With this, we note that gradients, and all primary and auxiliary variables are $t$-measurable, leading us to:
\begin{align*}
    \mathbb{E}_{t}\left[\mathcal{J}(\bm{x}_{t+1})\right] &\leq \mathcal{J}(\bm{x}_{t}) + \alpha_{t} \nabla^{\mathsf{T}}\mathcal{J}(\bm{x}_{t})\mathbb{E}_{t}\left[\frac{\bm{m}_{t}}{\sqrt{\bm{v}_{t}}+\epsilon}\right] \\ \nonumber
    &\hspace{5em} + \frac{L\alpha_{t}^{2}}{2}  \mathbb{E}_{t}\left[\left|\left|\frac{\bm{m}_{t}}{\sqrt{\bm{v}_{t}}+\epsilon}\right|\right|^2_2\right].  
\end{align*}
To achieve the convergence result, we bound each of the terms on the right-hand-side of the above equation. Using Assumptions 1 and 2 with the proposed setup of free-parameters, and applying the law of total expectation $\mathbb{E}_{\text{total}}\left[\mathbb{E}_t[\cdot]\right] = \mathbb{E}_{\text{total}}\left[\cdot\right]$, we get:
\begin{align*}
    &\mathbb{E}_{\text{total}}\left[\mathcal{J}(\boldsymbol{x}_{t+1}) - \mathcal{J}(\boldsymbol{x}_t)\right] \le \mathcal{O}(t^{-(a+c)}) + \mathcal{O}\left(\mu^tt^{-a}\right)- \\\nonumber
    &\mathcal{O}(t^{-a})\left(\mathbb{E}_{\text{total}}\left[||\nabla\mathcal{J}(\boldsymbol{x}_{t})||^2_2\right]  
    - \mathbb{E}_{\text{total}}\left[\left|\left|g(\boldsymbol{x}_t) - \boldsymbol{y}_t \right|\right|^2_2\right]\right).
\end{align*}
Hence, using Corollary 1 and re-grouping yields:
\begin{align*}
    &\mathbb{E}_{\text{total}}\left[\left|\left|\nabla\mathcal{J}(\boldsymbol{x}_t)\right|\right|^2_2\right] \le t^{-(5a-5b)} + t^{-(a+b+c)} + t^{-(a+c)} \\\nonumber
    &\hspace{5em} + \mu^{t}t^{-a} +\mathcal{O}\left(t^a\mathbb{E}_{\text{total}}[\mathcal{J}(\boldsymbol{x}_{t}) - \mathcal{J}(\boldsymbol{x}_{t+1})]\right).
\end{align*}
Considering the average over all iterations $T$, and using first-order concavity conditions for $f(t) = t^a$ (when $a\in(0,1)$) with the following setup for the constants: $a = 0.2; b = 0; c = e = 0.8$, eventually yields:
\begin{align*}
    \frac{1}{T}\sum_{t=1}^T\mathbb{E}_{\text{total}}\left[\left|\left|\nabla\mathcal{J}(\boldsymbol{x}_t)\right|\right|^2_2\right]\le \delta,
\end{align*}
with a total gradient sample complexity given by  $\mathcal{O}\left(\delta^{-\sfrac{9}{4}}\right)$ (i.e., result in Theorem 1).

\section{Use Case: Model-Agnostic Meta-Learning}\label{Sec:MAML}
In this section, we present a novel connection mapping model-agnostic meta-learning (MAML) to compositional optimisation. MAML, introduced in \cite{FinnAL17}, is an algorithm aiming at efficiently adapting deep networks to unseen problems by considering meta-updates performed on a set of tasks.

In its original form, MAML solves a meta-optimisation problem of the form:   
\begin{equation}
\label{Eq:MAML}
    \min_{\bm{x}} \mathcal{L}(\bm{x}) = \min_{\bm{x}} \sum_{ \mathcal{T}_{i} \sim \mathcal{P}_{\text{Tasks}}(\cdot)}\mathcal{L}_{\mathcal{T}_{i}}\left(\textrm{NN}\left({\bm{\Phi}(\bm{x};\mathcal{D}_{\mathcal{T}_{i}})}\right);\mathcal{D}_{\mathcal{T}_{i}}\right),
\end{equation}
where $\mathcal{T}_{i}$ represents the $T_{i}^{\text{th}}$ task, $\mathcal{P}_{\text{Tasks}}(\cdot)$ an unknown task distribution, and $\mathcal{D}_{\mathcal{T}_{i}}$ the data set for task $\mathcal{T}_{i}$. Furthermore, $\boldsymbol{\Phi}$ is a map operating on neural network parameters denoted by $\bm{x}$ to define a function typically encoded by a neural network or $\textrm{NN}(\cdot)$ in the above equation. Various algorithms from meta-learning have considered a variety of maps $\bm{\Phi}$ see \cite{finn2018probabilistic, kim2018bayesian, grant2018recasting, vuorio2018toward} and references therein. In this paper, we follow the original presentation in \cite{FinnAL17} that defines $\bm{\Phi}$ as a one-step gradient update, i.e.,
$\bm{\Phi}(\bm{x};\mathcal{D}_{\mathcal{T}_{i}}) = \bm{x} - \alpha \nabla \mathcal{L}_{\mathcal{T}_{i}}\left(\textrm{NN}(\bm{x});\mathcal{D}_{\mathcal{T}_i}\right)$. The remaining ingredient needed for fully defining the problem in Equation \ref{Eq:MAML} is the loss function for a task $\mathcal{T}_{i}$, i.e., $\mathcal{L}_{\mathcal{T}_{i}}(\cdot; \mathcal{D}_{\mathcal{T}_i})$. Such a loss, however, is task dependent, e.g., logistic loss in classification, and mean-squared error in regression. Keeping with the generality of the exposition, we will map to a compositional form, while assuming generic losses\footnote{Specifics on various machine learning tasks can be found in the appendix.} $\mathcal{L}_{\mathcal{T}_{i}}(\cdot; \mathcal{D}_{\mathcal{T}_i})$. 
\subsection{Compositional Meta-Learning: C-MAML}

\subsubsection{Stochastic Problem formulation:} To map MAML to compositional optimisation, we, first, rewrite the problem in Equation \ref{Eq:MAML} in an equivalent stochastic form. To do so, we start by considering $\text{Tasks}= \{\mathcal{T}_{1}, \mathcal{T}_{2}, \dots \}$ to be a collection of tasks (e.g., a set of regression or classification tasks) that can be observed by the learner. Furthermore, we assume a distribution, $\mathcal{P}_{\text{Tasks}} = \Delta\left(\{\mathcal{T}_{1}, \mathcal{T}_{2}, \dots\}\right)$, from which tasks are sampled over rounds. To evaluate loss functions and perform updates of Equation \ref{Eq:MAML}, we also assume a deterministically available data-set, $\mathcal{D}_{\mathcal{T}_{i}}$, for each task $\mathcal{T}_{i} \sim \Delta\left(\left\{\mathcal{T}_{1}, \mathcal{T}_{2}, \dots \right\}\right)$. As typical in machine learning, we suppose that an agent can not access all data for a task $\mathcal{T}_{i}$ but rather needs to employ a mini-batching scheme in performing updates. Signifying such a process, we introduce an additional random variable $\xi$ used for performing data sub-sampling according to an unknown distribution\footnote{We note that we do not assume that agents can access such a data distribution but can sample input-output pairs from $\mathcal{P}^{(\text{data})}_{{t}}(\cdot)$. } $\mathcal{P}^{(\text{data})}_{{\mathcal{T}_{i}}}(\cdot)$ over $\mathcal{D}_{\mathcal{T}_{i}}$ with $\mathcal{T}_{i}$ being the $i^{th}$ task. With these notations, we can write the gradient step update $\bm{\Phi}(\bm{x}; \mathcal{D}_{\mathcal{T}})$ for a given $\mathcal{T} \sim \Delta\left(\left\{\mathcal{T}_{1}, \mathcal{T}_{2}, \dots\right\}\right)$ as: 
\begin{align*}
    \bm{\Phi}(\bm{x}; \mathcal{D}_{\mathcal{T}})  = \mathbb{E}_{\xi \sim \mathcal{P}^{(\text{data})}_{\mathcal{T}}(\cdot)}\left[\bm{x} - \alpha \nabla \mathcal{L}_{t}\left(\textrm{NN}(\bm{x}); \xi\right) \Big| \textrm{t} = \mathcal{T}\right].
\end{align*}
Given a stochastic form of the update $\bm{\Phi}(\bm{x}; \mathcal{D}_{\mathcal{T}})$, we now focus on the problem in Equation \ref{Eq:MAML}. In \cite{FinnAL17}, the authors re-sample a new data-set from the same data distribution to compute the outer-loss. Hence, we introduce one more random variable $\xi^{\prime}$ to write: 
\begin{align}
\label{Eq:StoMAML}
    \min_{\bm{x}} \mathbb{E}_{\mathcal{T} }\Bigg[\mathbb{E}_{\xi^{\prime} \sim \mathcal{P}^{(\text{data})}_{\mathcal{T}}(\cdot)}\Bigg[\mathcal{L}_{t}\Bigg(\textrm{NN}(\bm{\Phi}(\bm{x}; \mathcal{D}_{\mathcal{T}})); \xi^{\prime}\Bigg)\Bigg| \textrm{t} = \mathcal{T}\Bigg]\Bigg].
\end{align}

In words, the stochastic formulation in Equation \ref{Eq:StoMAML} simply states that MAML attempts to minimise an expected (over tasks) loss, which itself involves a nested expectation over data mini-batches used to compute per-task errors and updates. 

Given such a stochastic formulation of MAML, we now consider two cases. The first involves one task, while the second handles a meta-learning scenario as we detail next. 

\subsubsection{Case I -- Single Task with Large Data-set:}
\label{subsubsec:single_maml}
Here, we assume the availability of only one task $\mathcal{T}_{1}$ whose data-set is large to motivate a data-streaming scenario. As such, the outer expectation in Equation~\ref{Eq:StoMAML} is non-existent, and the gradient map $\bm{\Phi}$ can be written as: 
\begin{equation*}
    \bm{\Phi}(\bm{x}; \mathcal{D}_{\mathcal{T}_{1}}) = \mathbb{E}_{\xi \sim \mathcal{P}^{(\text{data})}_{\mathcal{T}_1}(\cdot)}\left[\bm{x} - \alpha \nabla \mathcal{L}_{t}\left(\textrm{NN}(\bm{x}); \xi\right) \Big| \textrm{t} = \mathcal{T}_{1}\right].
\end{equation*}
Noting that the original form in Equation~\ref{Eq:Prob} can be rewritten as $\mathcal{J}(\bm{x}) = f(g(\bm{x}))$ (Section \ref{Sec:ProbDef}), we first define $g(\bm{x}) = \mathbb{E}_{\xi \sim \mathcal{P}^{(\text{data})}_{\mathcal{T}_{1}}(\cdot)}\left[\bm{x} - \alpha \nabla \mathcal{L}_{t}\left(\textrm{NN}(\bm{x}); \xi\right) \Big| \textrm{t} = \mathcal{T}_{1}\right]$. Subsequently, the objective in Equation \ref{Eq:StoMAML}, when considering one task $\mathcal{T}_{1}$, can be rewritten as: 
\begin{align}
\label{Eq:CompoMAML}
    \min_{\bm{x}} \mathbb{E}_{\xi^{\prime} \sim \mathcal{P}^{(\text{data})}_{\mathcal{T}_{1}}(\cdot)}\left[f_{\xi^{\prime}}\left(\mathbb{E}_{\xi \sim \mathcal{P}^{(\text{data})}_{\mathcal{T}_{1}}(\cdot)}\left[g_{\xi}(\bm{x})\right]\right)\right],
\end{align}
where $g_{\xi}(\bm{x}) = \bm{x} - \alpha \nabla \mathcal{L}_{t}\left(\textrm{NN}(\bm{x}); \xi\right)$, and $f_{\xi^{\prime}}(\bm{y})=\mathcal{L}_{\mathcal{T}_{1}}\left(\text{NN}(\bm{y}), \xi^{\prime}\right)$. Clearly, the problem in Equation~\ref{Eq:CompoMAML} is a special case of that in~\ref{Eq:Prob} and thus, one can employ Algorithm~\ref{Algo:ADAM} for finding a stationary point (see Section~\ref{Sec:ExperimentMAMLOne}). 

\subsubsection{Case II -- Multiple Tasks \& Meta-Updates:}
After showing single task MAML as a special case of compositional optimisation, we now generalise previous results to $K > 1$ tasks $\mathcal{T}_{1}, \dots, \mathcal{T}_{K}$. Sampling according to $\mathcal{P}_{\text{Tasks}}(\cdot)$, we can write a Monte-Carlo estimator of the loss in Equation~\ref{Eq:StoMAML} to arrive at the following optimisation problem: 
\begin{equation}
\label{Eq:GDUPDATES}
 \min_{\bm{x}} \hat{\mathcal{L}}(\bm{x}) =\min_{\bm{x}} \frac{1}{K}\sum_{j=1}^{K} \mathbb{E}_{\xi^{\prime}_{j}\sim \mathcal{P}^{(\text{data})}_{\mathcal{T}_{j}}(\cdot)} \Bigg[\gamma_{t}^{(j)}(\bm{x};\xi^{\prime}_{j})\Bigg| \textrm{t} = \mathcal{T}_{j}\Bigg],
\end{equation}
with $\gamma_{t}^{(j)}(\bm{x};\xi^{\prime}_{j})=\mathcal{L}_{t}\left(\textrm{NN}\left(\bm{\Phi}(\bm{x};\mathcal{D}_{\mathcal{T}_{j}})\right);\xi^{\prime}_{j}\right)$, and $\bm{\Phi}(\cdot) = \bm{\Phi}(\bm{x};\mathcal{D}_{\mathcal{T}_{j}})$ defined as: 
\begin{equation}
\label{Eq:Bloo}
\bm{\Phi}(\cdot) = \mathbb{E}_{\xi_{j} \sim \mathcal{P}^{(\text{data})}_{\mathcal{T}_j}(\cdot)}\left[\bm{x} - \alpha \nabla \mathcal{L}_{t}\left(\textrm{NN}(\bm{x}); \xi_j \right) \Big| \textrm{t} = \mathcal{T}_{j}\right].
\end{equation}
Now, consider $\bm{G}_{\xi_{1}, \dots, \xi_{K}}(\bm{x}) \in \mathbb{R}^{p \times K}$ to be a matrix with dimension of neural network parameters $p$ rows and total number of tasks $K$ columns, such that each column $j \in \{1, \dots, K\}$ is defined as: 
\begin{align*}
    \bm{G}_{\xi_{1}, \dots, \xi_{K}}(\bm{x})[:, j]= \bm{x} - \alpha \nabla \mathcal{L}_{\mathcal{T}_{j}}(\textrm{NN}(\bm{x}); \xi_{j}).
\end{align*}
Denoting $\bm{\xi} = (\xi_{1}, \dots, \xi_{K})$, where $\xi_{j} \sim \mathcal{P}_{\mathcal{T}_{j}}^{(\text{data})}(\cdot)$ for all $j \in \{1,\dots, K\}$, we define $\bm{G}(\bm{x}) = \mathbb{E}_{\bm{\xi}}\left[\bm{G}_{\bm{\xi}}(\bm{x})\right]$ as an expected map such that each of its $j$ columns is given by\footnote{Please note that for ease of notation we have denoted $\mathbb{E}_{\bm{\xi}}[\cdot] = \mathbb{E}_{\xi_{1}, \dots, \xi_{K}} = \mathbb{E}_{\xi_{1}}\left[\mathbb{E}_{\xi_{2}}\left[\dots \mathbb{E}_{\xi_{K}}[\cdot]\right]\right]$.}: 
\begin{align*}
    \bm{G}(\bm{x})[:, j] &= \mathbb{E}_{\xi_{j} \sim \mathcal{P}^{(\text{data})}_{\mathcal{T}_{j}}(\cdot)}\left[\bm{x} - \alpha \nabla \mathcal{L}_{t}(\textrm{NN}(\bm{x});\xi_{j})\Bigg| \textrm{t} = \mathcal{T}_{j}\right].
\end{align*}
Hence, one can write the gradient update map in Equation~\ref{Eq:Bloo} for any task $j$ simply as: $\bm{\Phi}(\bm{x};\mathcal{D}_{\mathcal{T}_{j}}) = \bm{G}(\bm{x})\bm{e}_{j}$ with $\bm{e}_{j}$ being the $j^{th}$ basis vector in a standard orthonormal basis in $\mathbb{R}^{K}$. Applying our newly introduced notions to the loss in Equation~\ref{Eq:GDUPDATES}, we get: 
\begin{align*}
    \hat{\mathcal{L}}(\bm{x})  &= \mathbb{E}_{\bm{\xi}^{\prime}}\left[\frac{1}{K}\sum_{j=1}^{K}\mathcal{L}_{t}\left(\textrm{NN}\left(\bm{G}(\bm{x})\bm{e}_{j}\right);\xi^{\prime}_{j}\right)|\textrm{t} = \mathcal{T}_{j}\right] \\ \nonumber
    & = \mathbb{E}_{\bm{\xi}^{\prime}}\left[f_{\bm{\xi}^{\prime}}\left(\mathbb{E}_{\bm{\xi}}\left[\bm{G}_{\bm{\xi}}(\bm{x})\right]\right)\right], \impliedby \text{\underline{Compositional Form}}
\end{align*}
with $\bm{\xi}^{\prime} = \left(\xi_{1}^{\prime}, \dots, \xi_{K}^{\prime}\right)$, and $f_{\bm{\xi}^{\prime}}(\bm{A}) = \frac{1}{K}\sum_{j=1}^{K}\mathcal{L}_{\mathcal{T}_{j}}(\textrm{NN}(\bm{A}[:,j]);\xi^{\prime}_{j}),$ with $\bm{A}[:,j]$ being the $j^{th}$ column of matrix $\bm{A} \in \mathbb{R}^{p \times K}$. It is again clear that multi-task MAML is another special case of compositional optimisation for which we can employ Algorithm~\ref{Algo:ADAM} to determine a stationary point.\\
\underline{\textbf{C-MAML Implications:}}
Connecting MAML and compositional optimisation sheds-light on interesting theoretical insights for meta-learning -- a topic gaining lots of attention in recent literature~\cite{fallah2019convergence}. Importantly, we realise that upon the application of C-ADAM to the bridge made in Section~\ref{Sec:MAML}, we achieve the fastest known complexity bound for meta-learning. Table~\ref{tab:sample_cmplx_maml} depicts these results demonstrating that the closest results to ours are recent bounds derived in~\cite{fallah2019convergence}. In Table~\ref{tab:sample_cmplx_maml}, we also note that methods proposed in~\cite{FinnAL17, NonAdaptiveGR} have no provided convergence guarantees (marked by N/A in the table) leaving such an analysis as an interesting avenue for future research\footnote{Furthermore, we note that other theoretical attempts aiming at understanding MAML~\cite{OnlineMAML,golmantconvergence} were not explicitly mentioned as these consider convex assumptions on loss functions abide using deep network models.}.   

  \begin{table}[t!]
      \centering   
      \caption{Theoretical sample complexity bounds comparing our solver C-MAML to state-of-the-art in current literature to achieve $||\nabla \hat{\mathcal{L}}(\bm{x})||^{2}_{2} \leq \delta$. It is clear that C-MAML achieves the best rate known so-far.}
\vskip 0.15in
            \begin{tabular}{cc} \toprule
            \thead{Algorithms} & \thead{Complexities} \\
    \midrule
    \textbf{C-MAML} (This paper) & $\mathcal{O}\left(\delta^{-2.25}\right)$ \\ \midrule
    \cite{fallah2019convergence} -- Alg. 1 & $\mathcal{O}(\delta^{-3})$ \\
    \cite{fallah2019convergence} -- Alg. 3 & $\mathcal{O}(\delta^{-3})$ \\ 
    \cite{kingma2014adam} in MAML & \textrm{N/A} \\ 
    \cite{FinnAL17} & \textrm{N/A} \\
    \bottomrule
\end{tabular}   
      \label{tab:sample_cmplx_maml}
\vskip -0.1in
  \end{table}

\section{Experiments}
We present an in-depth empirical study demonstrating that C-ADAM (Algorithm~\ref{Algo:ADAM}) outperforms state-of-the-art methods from both compositional and standard optimisation. We split this section into two. In the first, we experiment with more conventional portfolio optimisation tasks as presented in~\cite{Mendgi_2017}, while in the second, we focus on model-agnostic meta-learning building on our results from Section~\ref{Sec:MAML}. In the portfolio scenario, algorithms tasked to determine stationary points for sparse mean-variance trade-offs on real-world data-sets gathered by the center for research in security prices (CRSP) \footnote{\url{https://mba.tuck.dartmouth.edu/pages/faculty/ken.french/data\_library.html}} are studied. In MAML's case, on the other hand, we benchmark against regression tasks initially presented in~\cite{FinnAL17} demonstrating that C-ADAM achieves new state-of-the-art performance. All models were trained on a single NVIDIA GeForce GTX 1080 Ti GPU.
\begin{figure*}[t!]
\centering
\begin{subfigure}{0.245\textwidth}
\centering
\includegraphics[width=\textwidth]{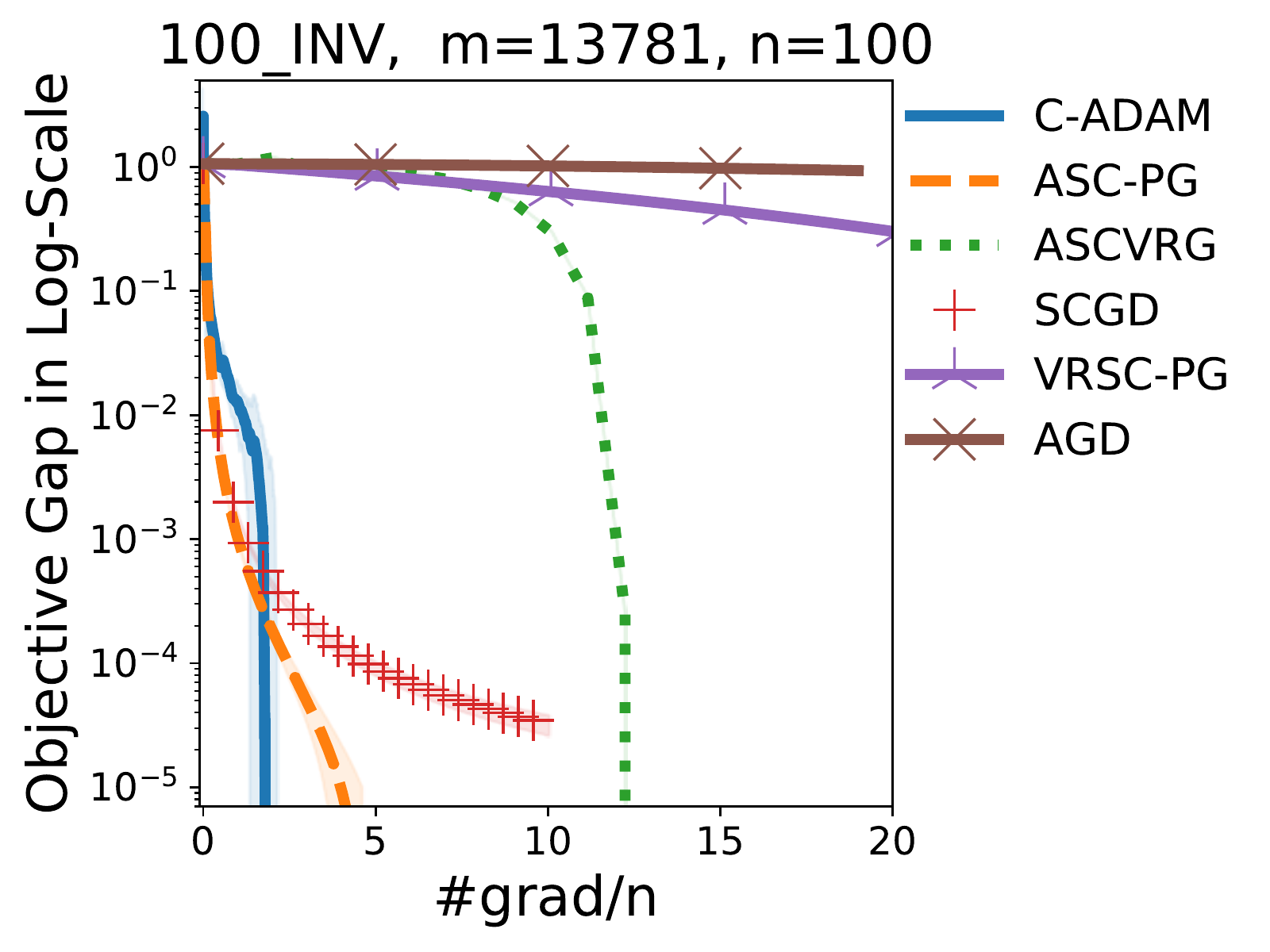}
\caption{}
\label{fig:100_inv}
\end{subfigure}
\begin{subfigure}{0.245\textwidth}
\centering
\includegraphics[width=\textwidth]{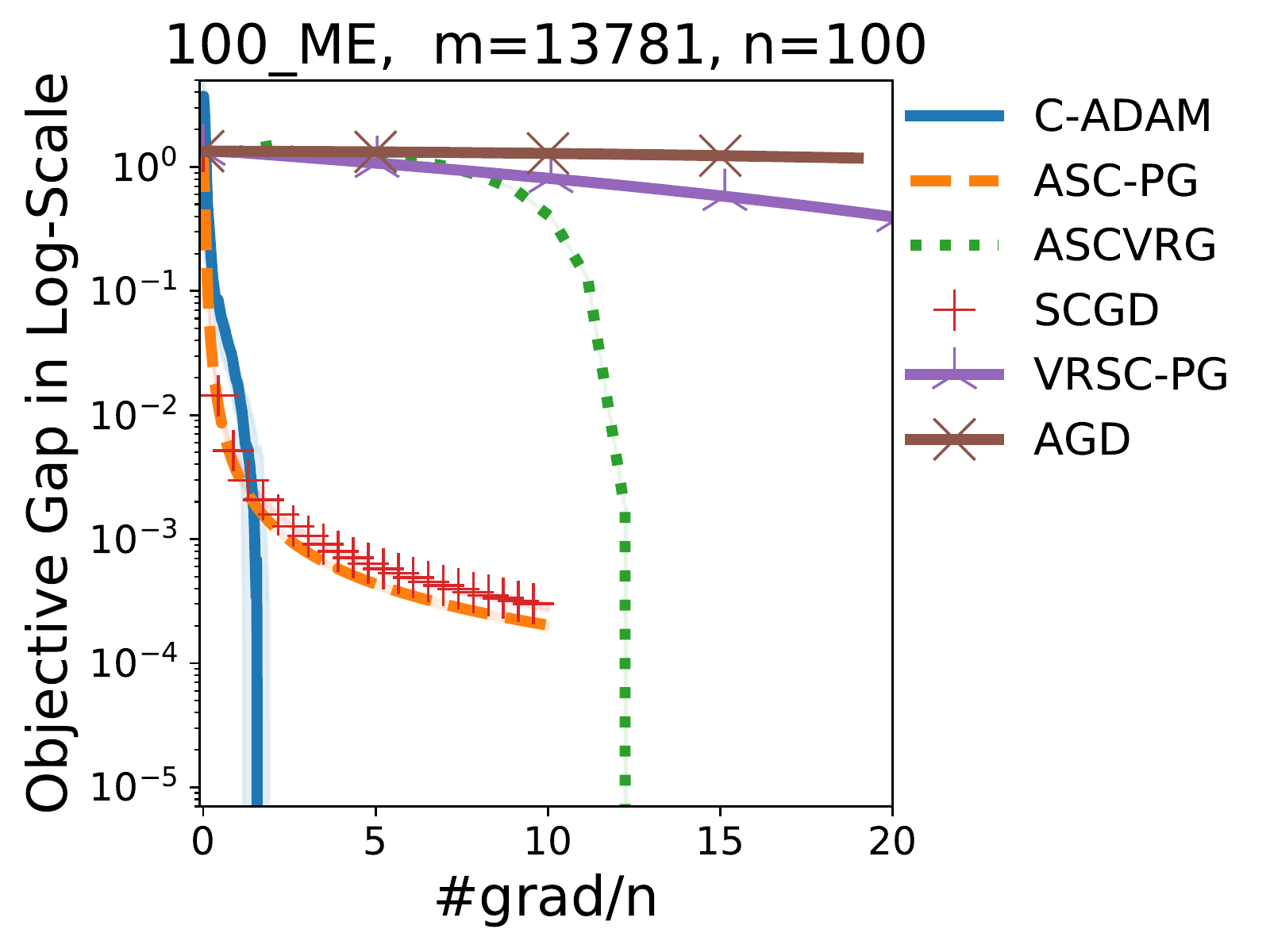}
\caption{}
\label{fig:100_me}
\end{subfigure}
\begin{subfigure}{0.245\textwidth}
\centering
\includegraphics[width=\textwidth]{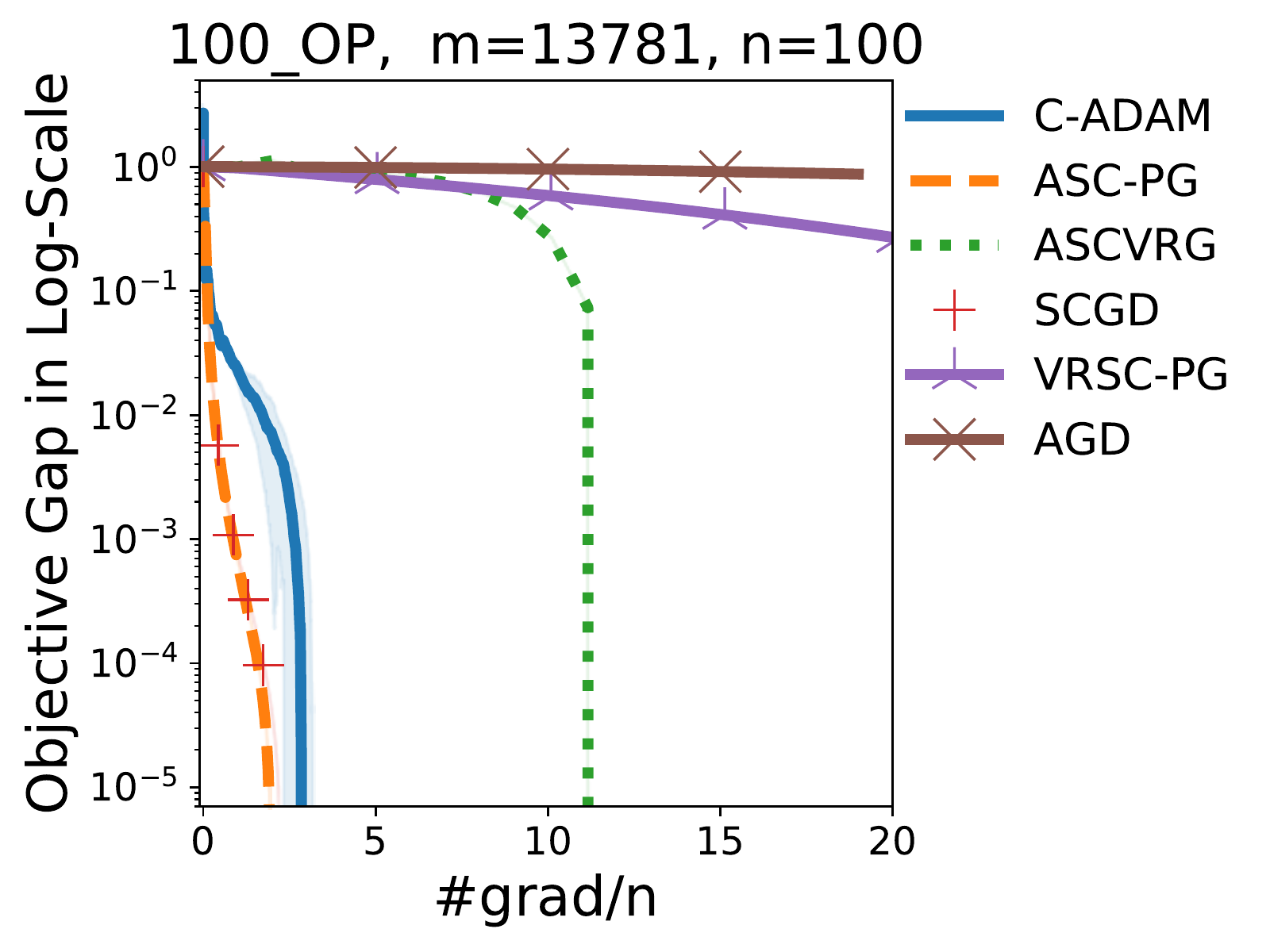}
\caption{}
\label{fig:100_op}
\end{subfigure}
\begin{subfigure}{0.245\textwidth}
\centering
\includegraphics[width=\textwidth]{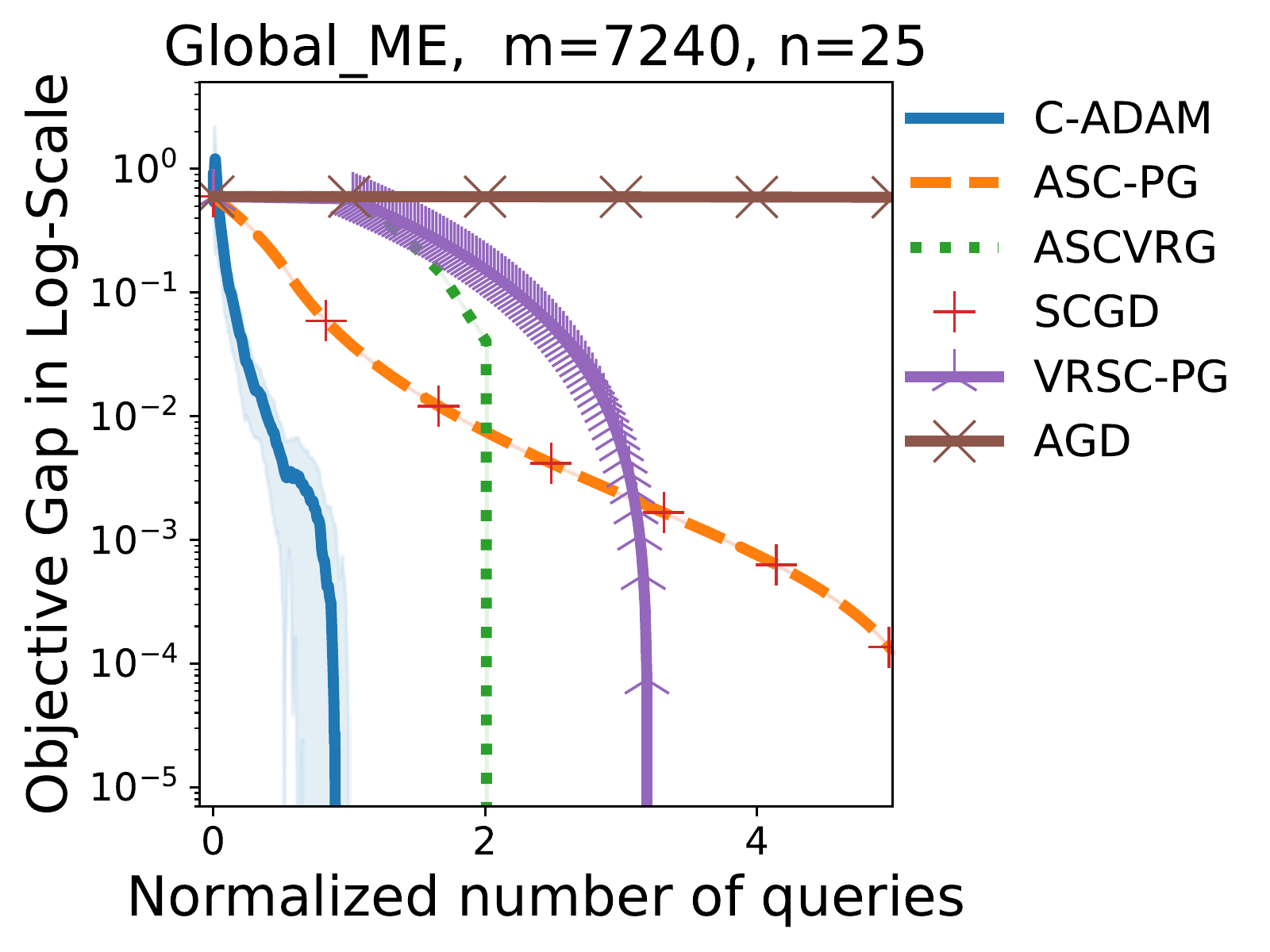}
\caption{}
\label{fig:ap_ex_jp_inv}
\end{subfigure}
\begin{subfigure}{0.245\textwidth}
\includegraphics[width=\textwidth]{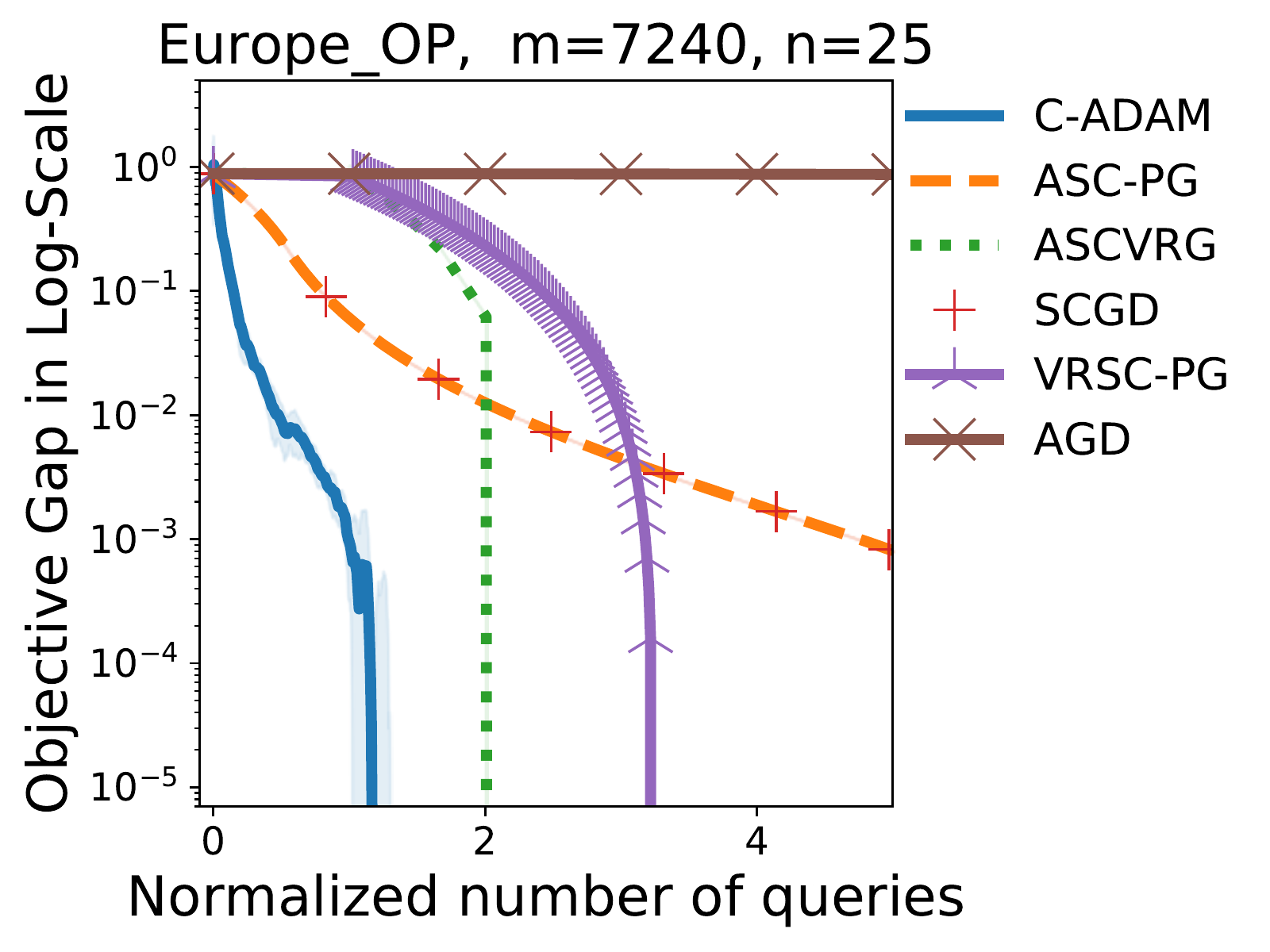}
\caption{}
\label{fig:n_am_me}
\end{subfigure}
\begin{subfigure}{0.245\textwidth}
\includegraphics[width=\textwidth]{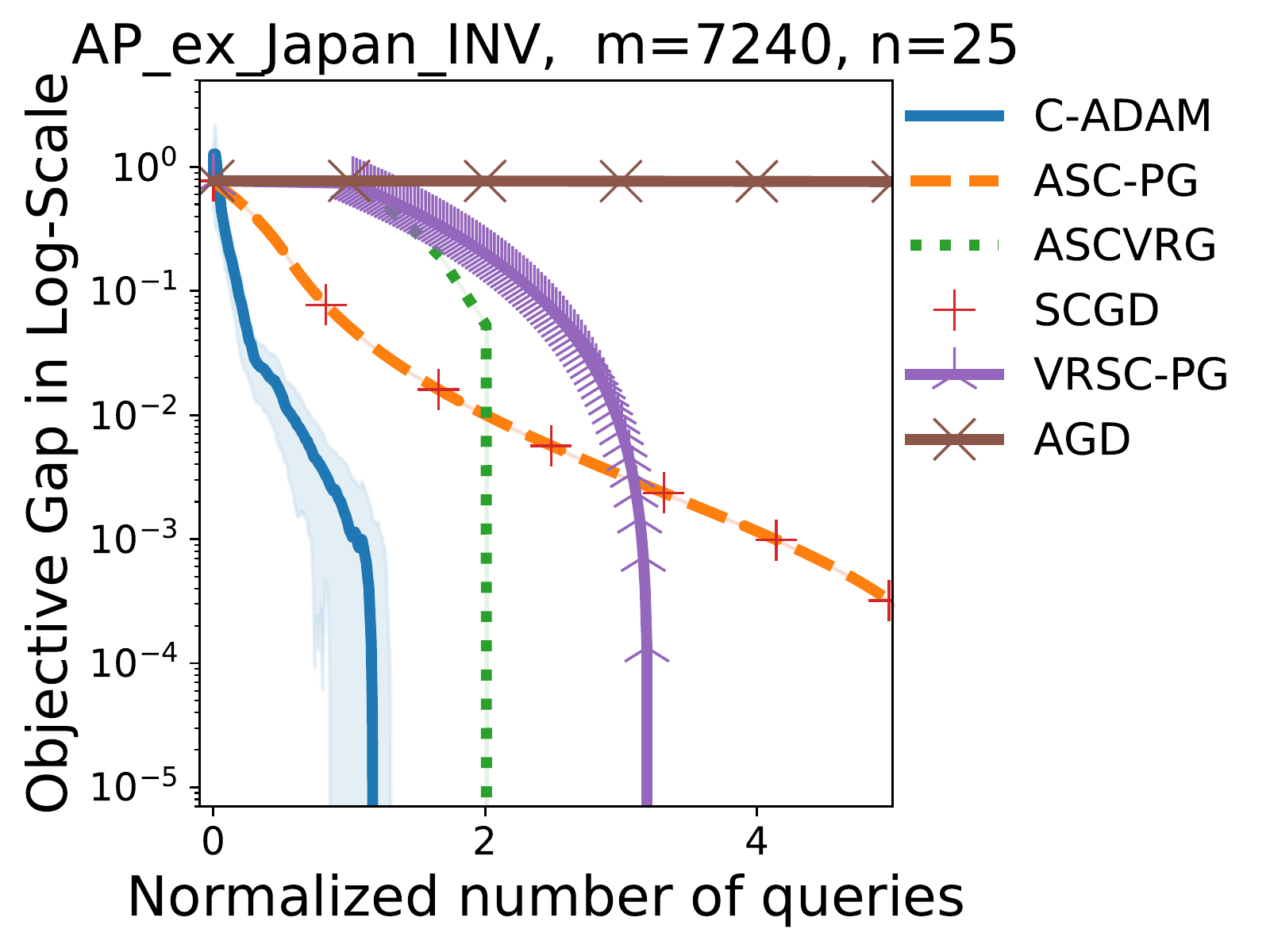}
\caption{}
\label{fig:jp_op}
\end{subfigure}
\begin{subfigure}{0.245\textwidth}
\includegraphics[width=\textwidth]{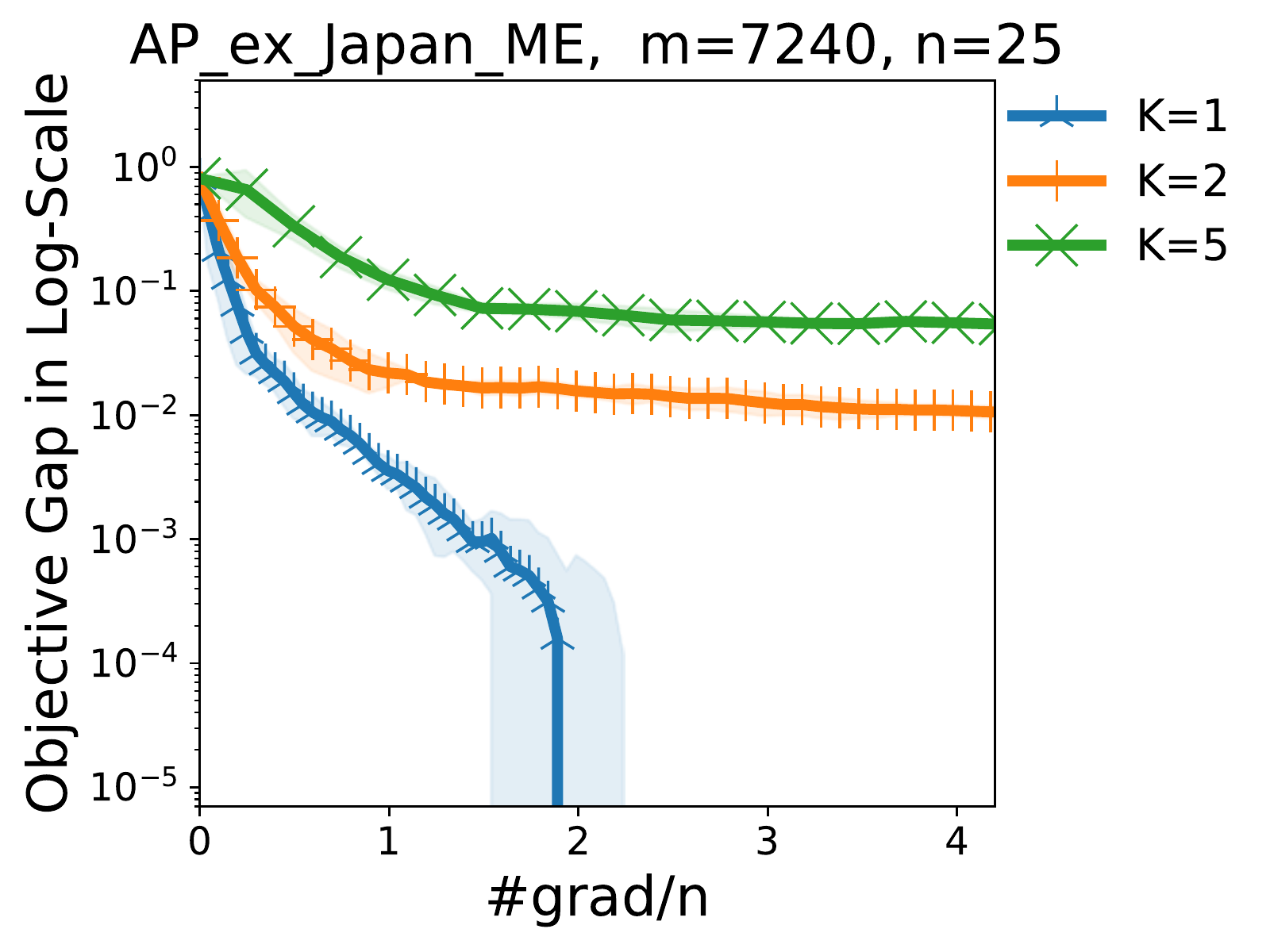}
\caption{}
\label{fig:abla_sample_size}
\end{subfigure}
\begin{subfigure}{0.245\textwidth}
\includegraphics[width=\textwidth]{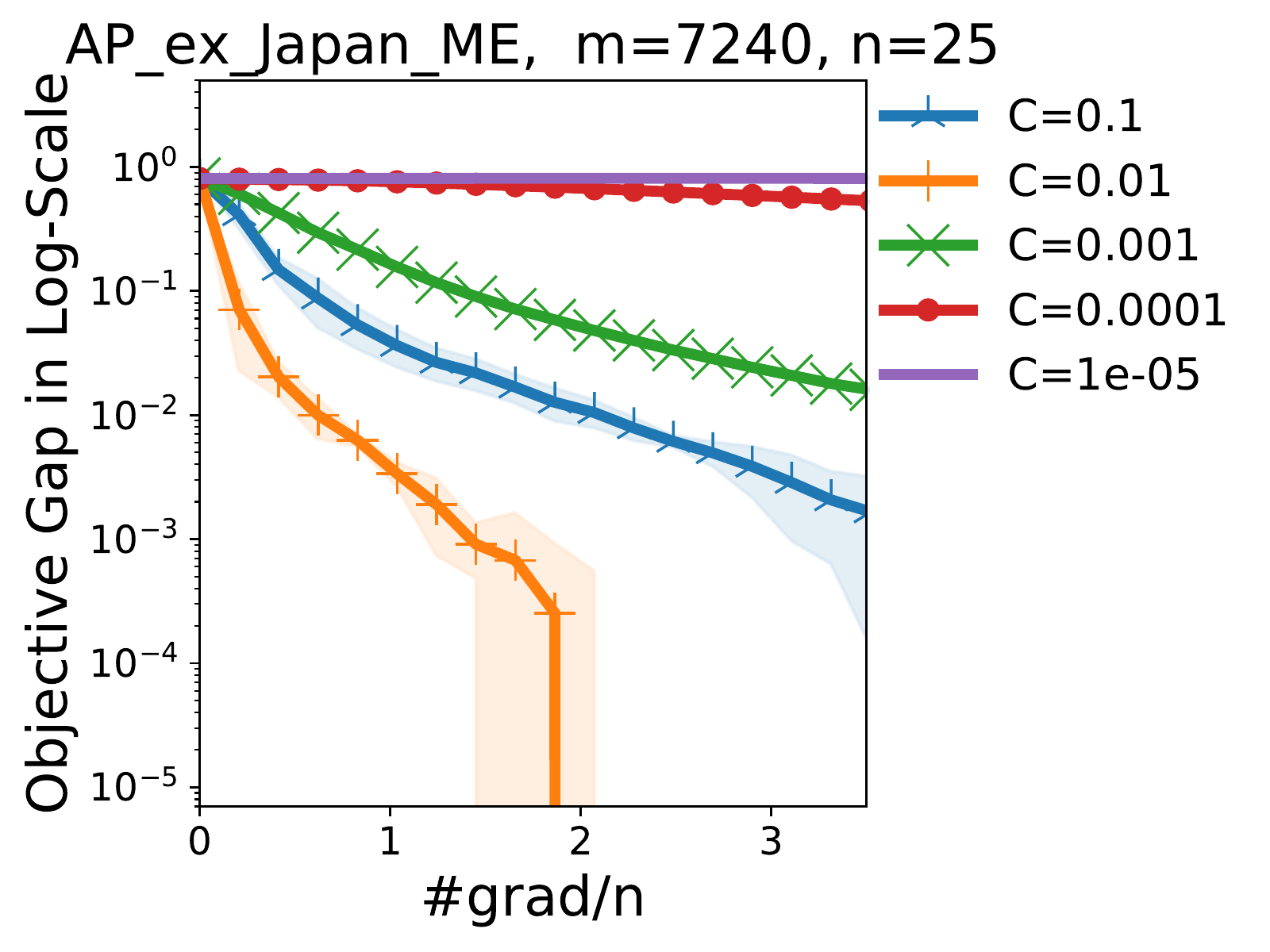}
\caption{}
\label{fig:abla_step_size}
\end{subfigure}
\begin{subfigure}{0.245\textwidth}
\includegraphics[width=\textwidth]{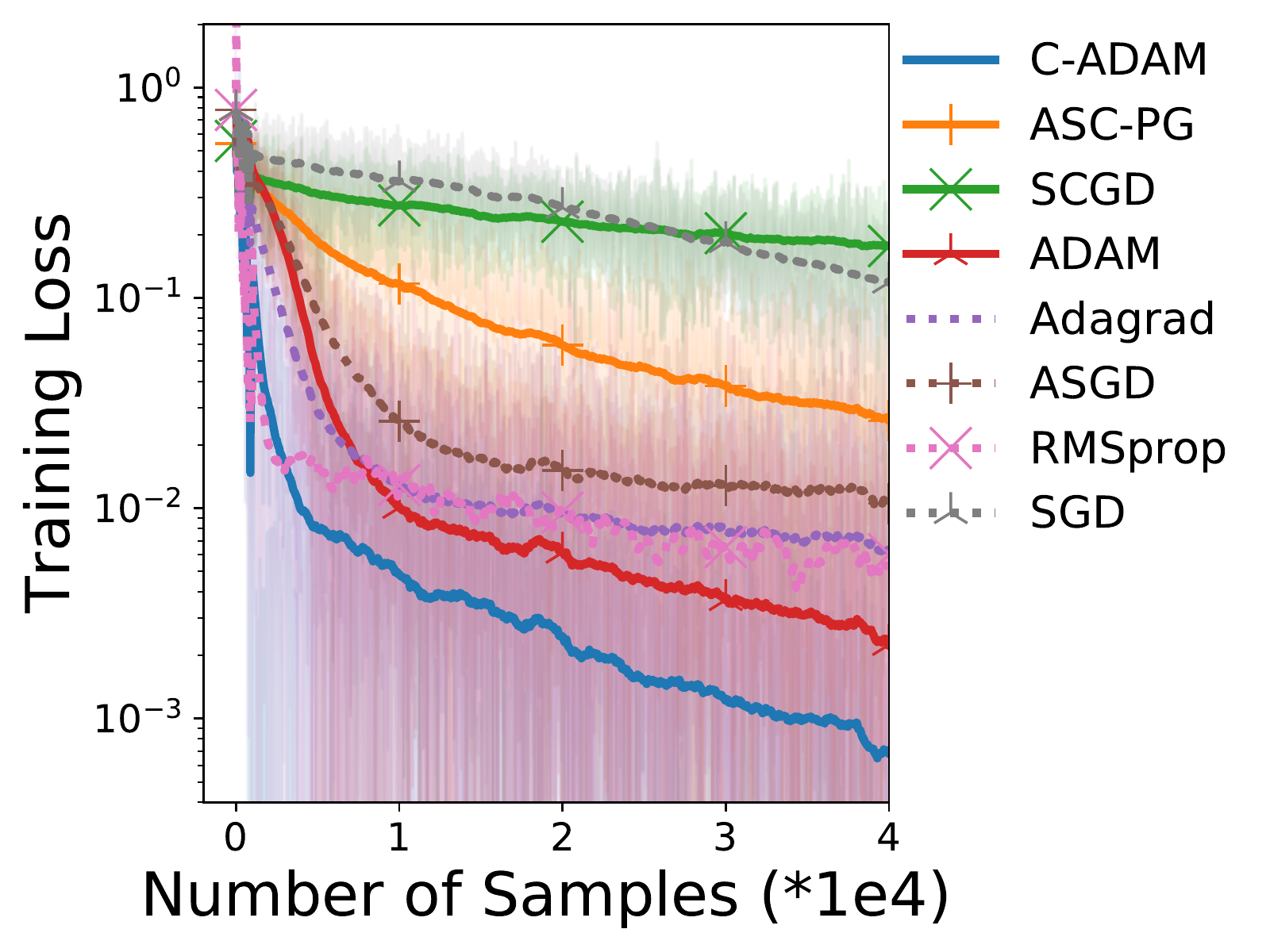}
\caption{}
\label{fig:reg1task_test}
\end{subfigure}
\begin{subfigure}{0.245\textwidth}
\includegraphics[width=\textwidth]{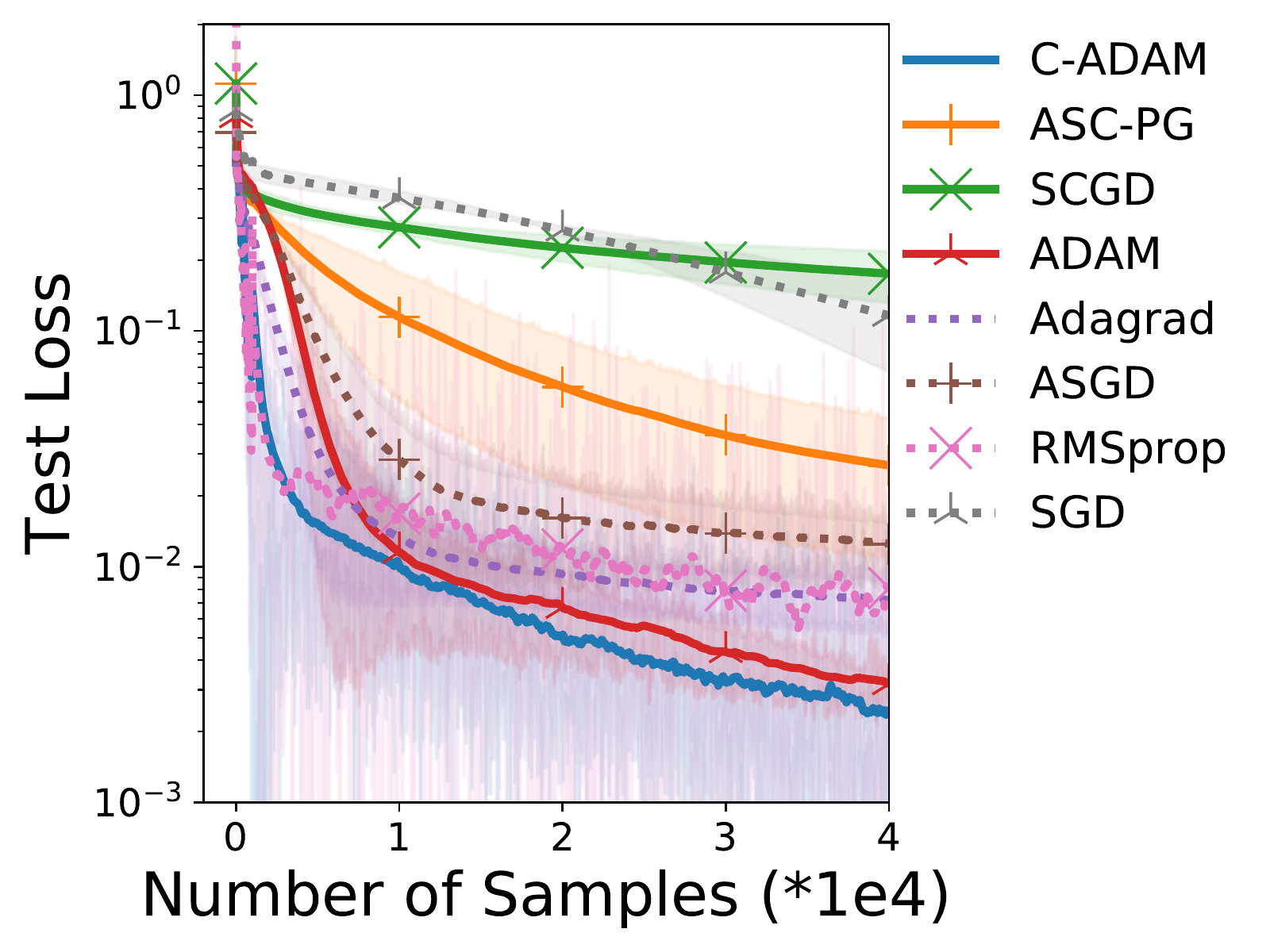}
\caption{}
\label{fig:reg1task_train}
\end{subfigure}
\begin{subfigure}{0.245\textwidth}
\includegraphics[width=\textwidth]{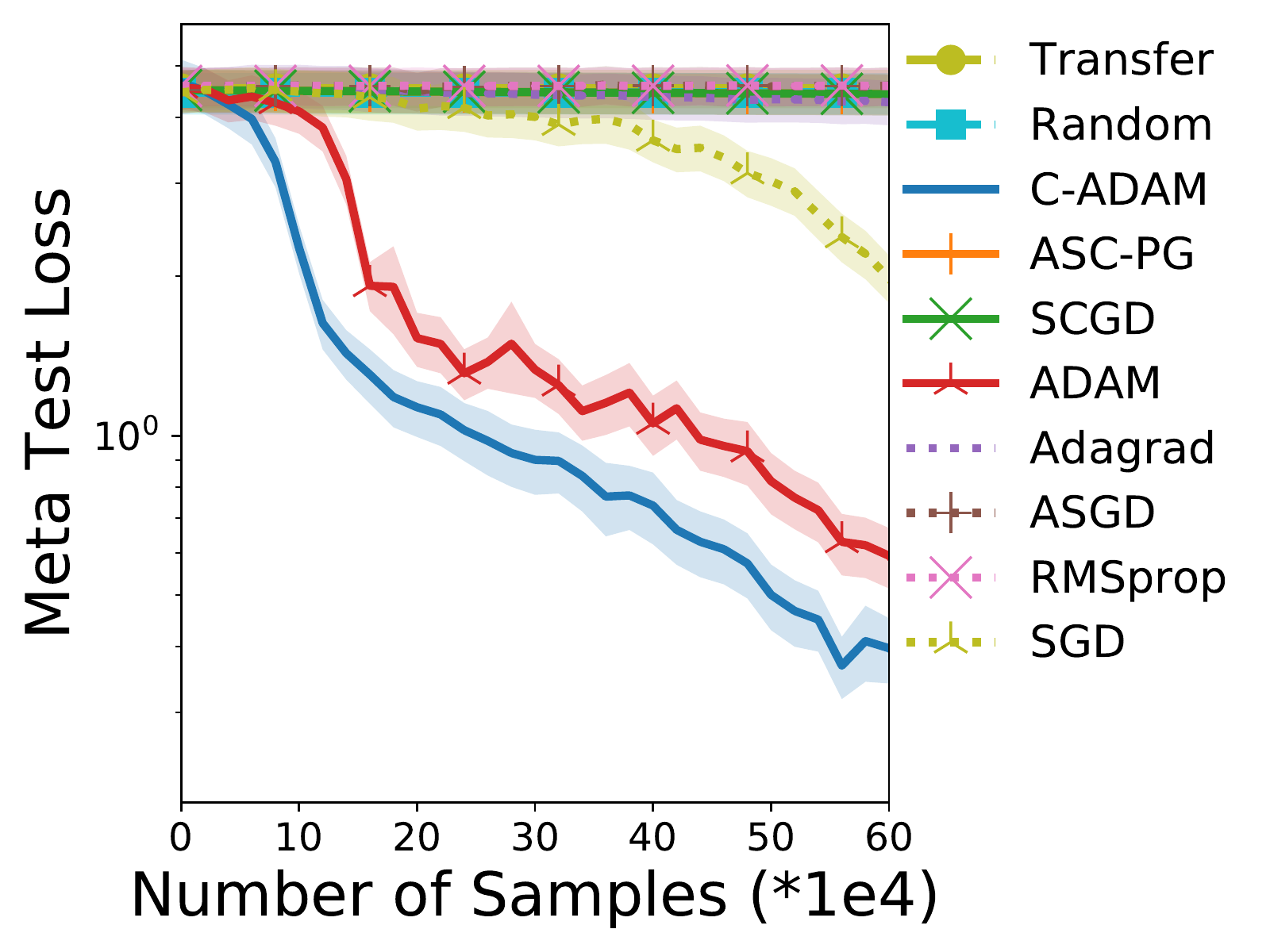}
\caption{}
\label{fig:reg_multi_task_fit_test}
\end{subfigure}
\begin{subfigure}{0.245\textwidth}
\includegraphics[width=\textwidth]{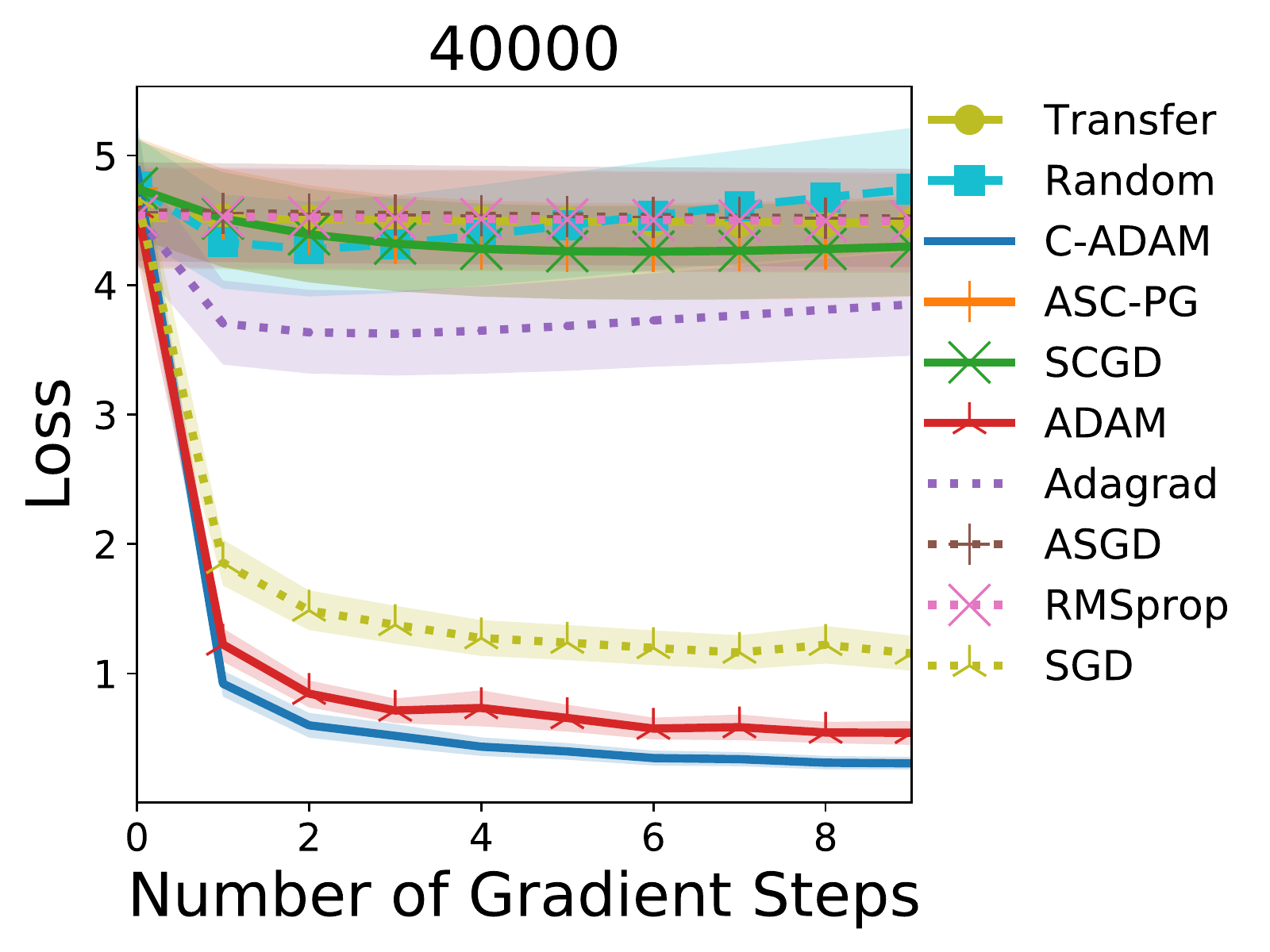}
\caption{}
\label{fig:reg_multi_task_finetune_40000}
\end{subfigure}
\caption{
\underline{(a, b, c):} C-ADAM's performance versus other compositional optimisation methods on the three large 100-portfolio data-sets demonstrating that our method significantly outperforms others in terms of convergence speeds.
\underline{(d, e, f):} [d]: C-ADAM's performance versus other composition methods on one data-set from BM, [e]: C-ADAM's performance on OP, and [f]: C-ADAM's performance on a data-set from Inv. In all cases, we see that C-ADAM outperforms other techniques significantly.
\underline{(g, h):} [g:] An ablation study demonstrating the effect of modifying the batch sizes -- $K_{t}^{(1)} = K_{t}^{(2)}= K_{t}^{(3)} = K$, and [h:] Effect of step-size -- $C_{\alpha} = C_{\beta} = C$.
\underline{(i, j):} Training loss curves of single-task compositional MAML compared to methods from compositional and standard optimisation. These results again demonstrate that C-ADAM outperforms others.
\underline{(k):} The meta test loss of multi-task compositional MAML compared to others demonstrating C-ADAM's performance.
\underline{(l):} Quantitative regression results showing the learning curve after training with $40000$ iterations on meta test-time. It is again clear that C-ADAM outperforms others. 
}
\label{fig:exp}
\end{figure*}

\begin{figure*}[h!]
\centering
\begin{subfigure}{0.245\textwidth}
\centering
\includegraphics[width=\textwidth]{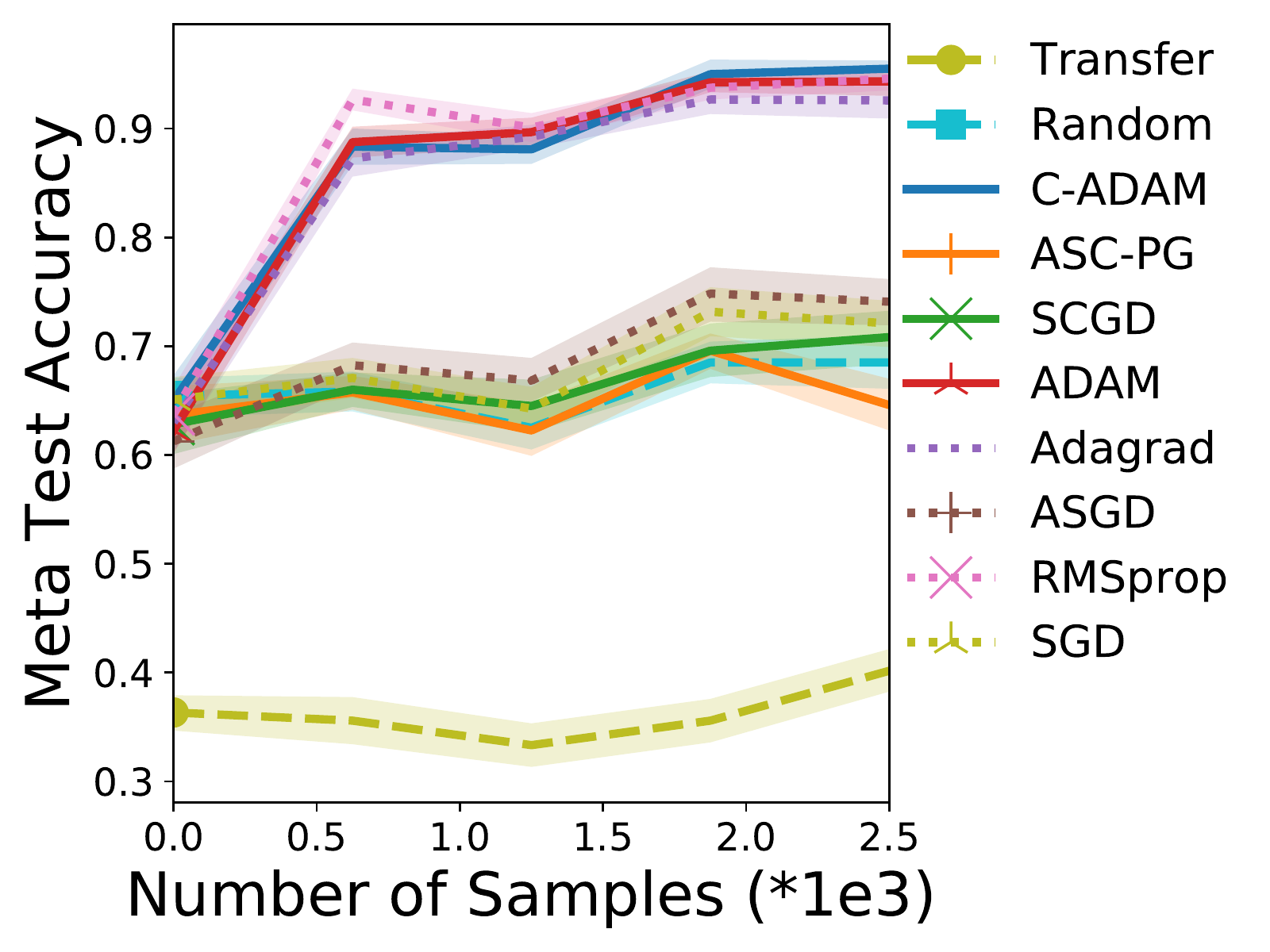}
\caption{}
\end{subfigure}
\begin{subfigure}{0.245\textwidth}
\centering
\includegraphics[width=\textwidth, trim={0 0 0 0.95cm },clip]{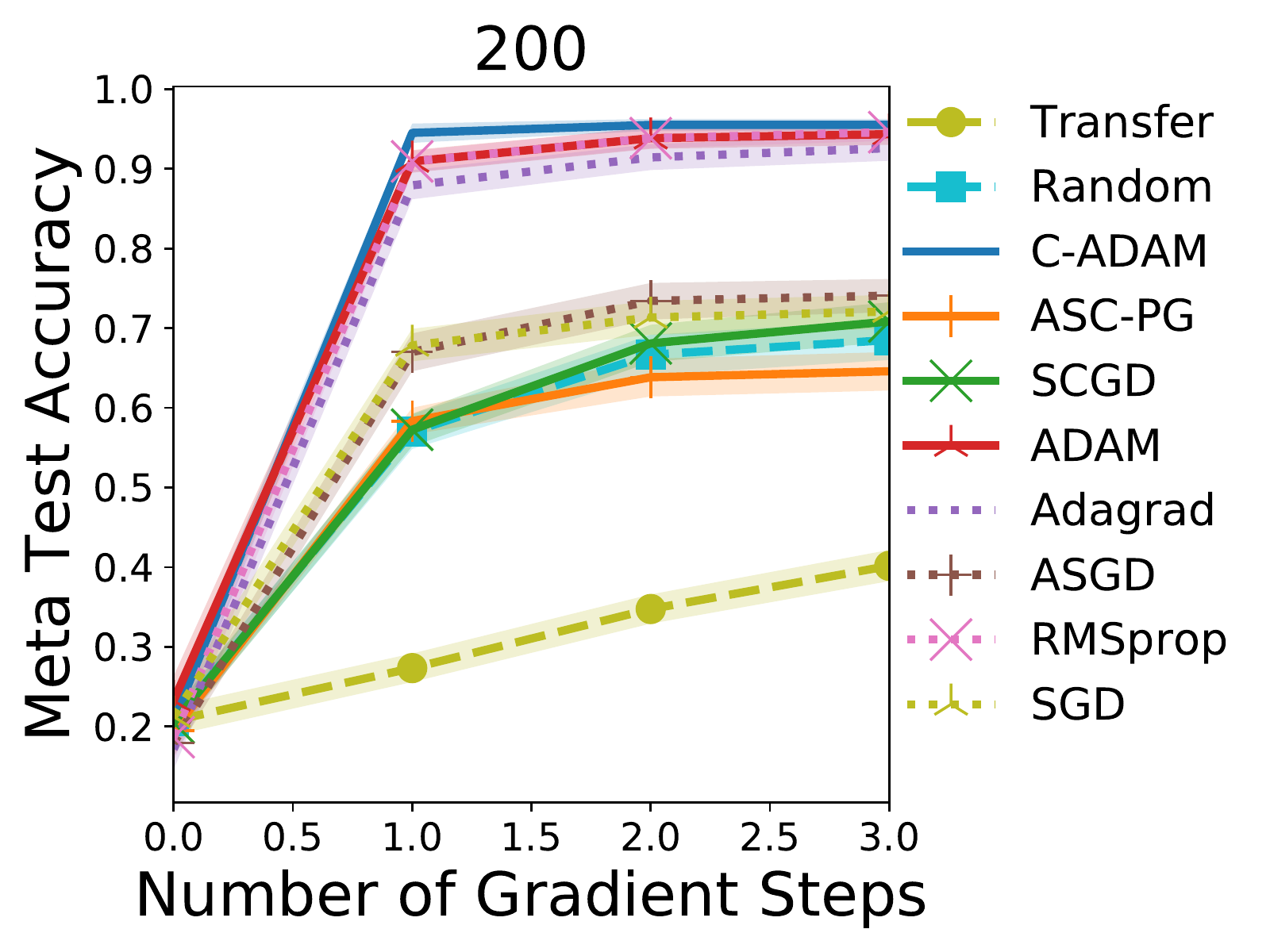}
\caption{}
\end{subfigure}
\begin{subfigure}{0.245\textwidth}
\centering
\includegraphics[width=\textwidth]{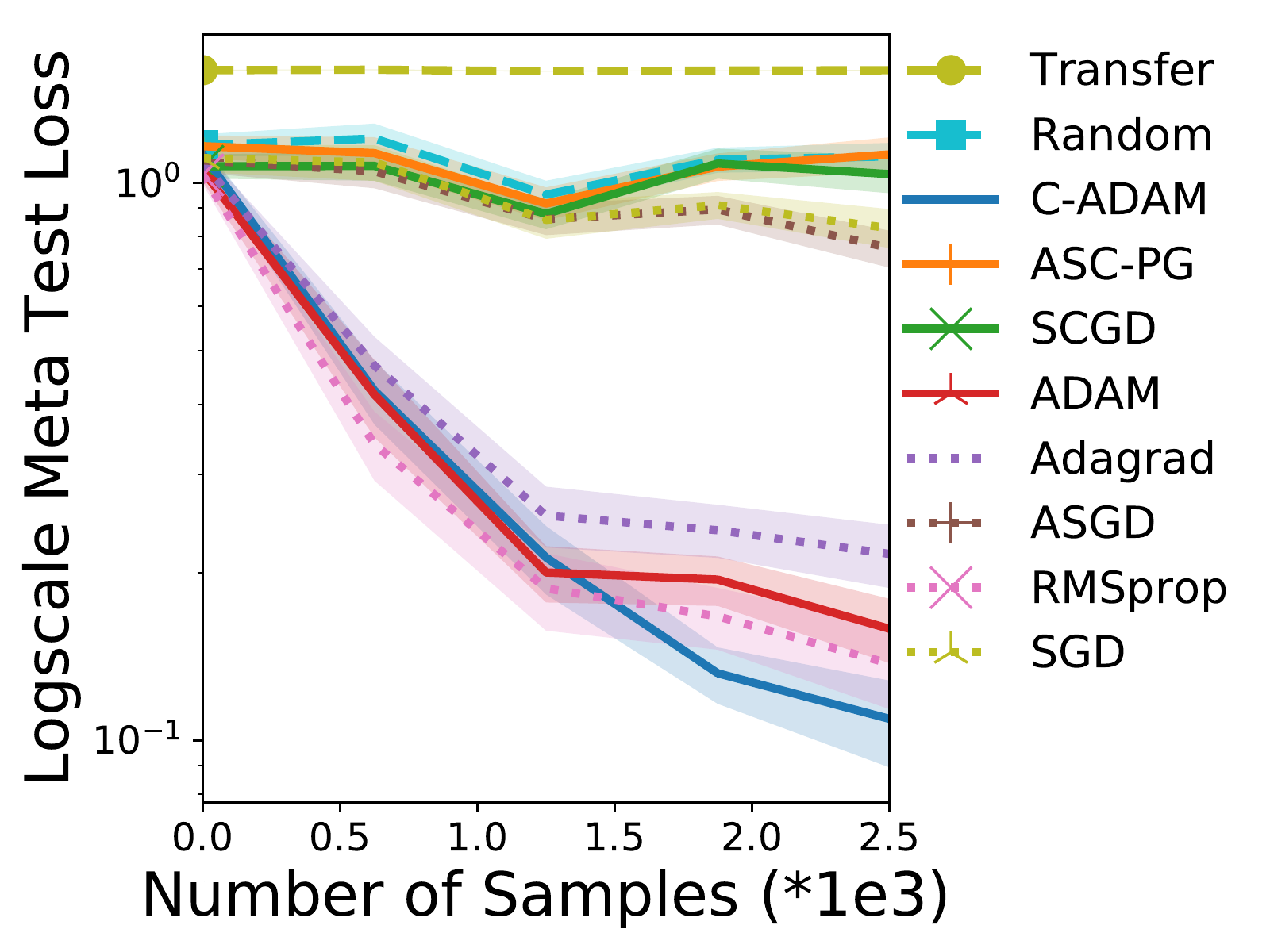}
\caption{}
\end{subfigure}
\begin{subfigure}{0.245\textwidth}
\centering
\includegraphics[width=\textwidth, trim={0 0 0 0.95cm },clip]{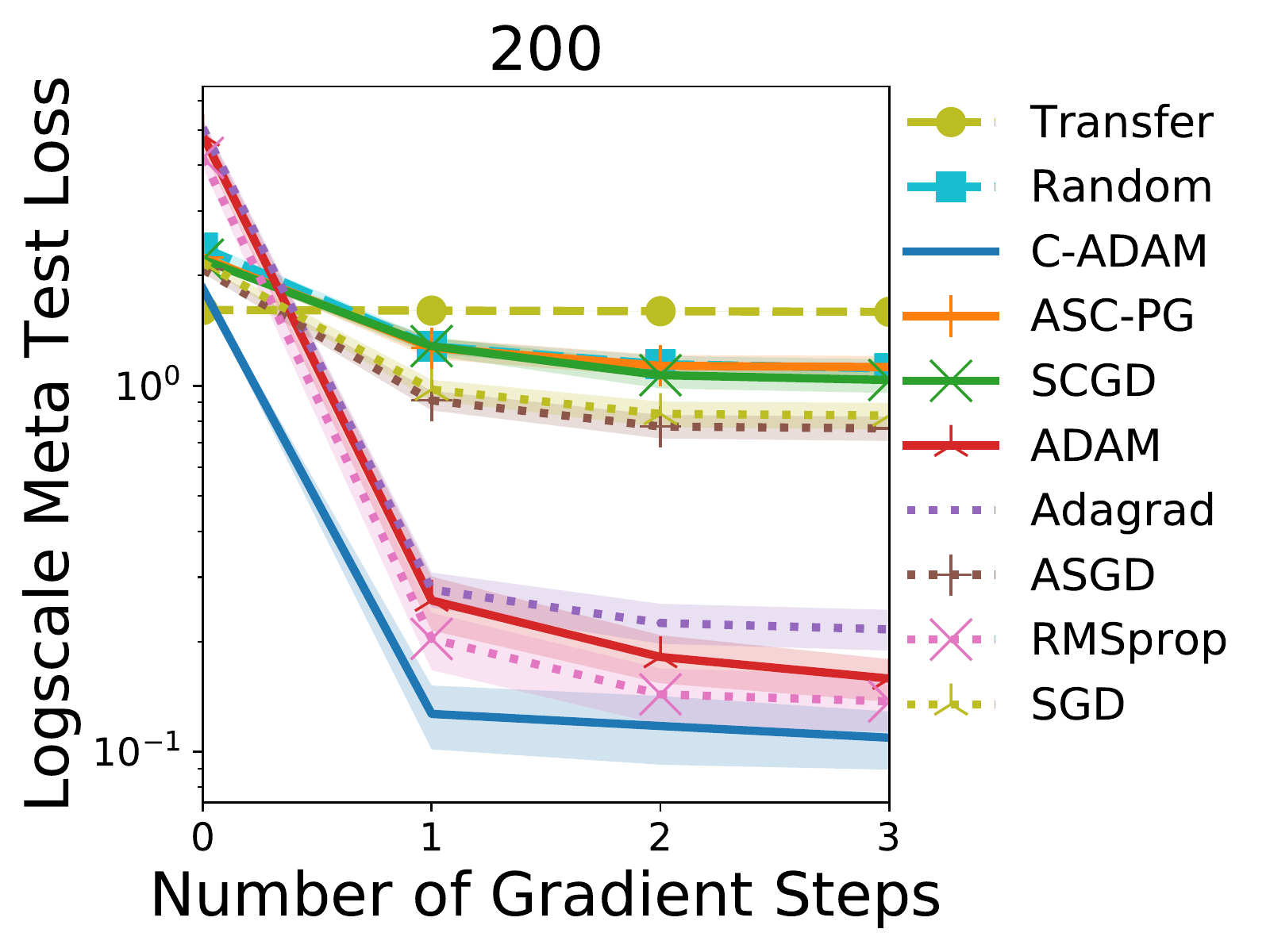}
\caption{}
\end{subfigure}
\begin{subfigure}{0.245\textwidth}
\centering
\includegraphics[width=\textwidth]{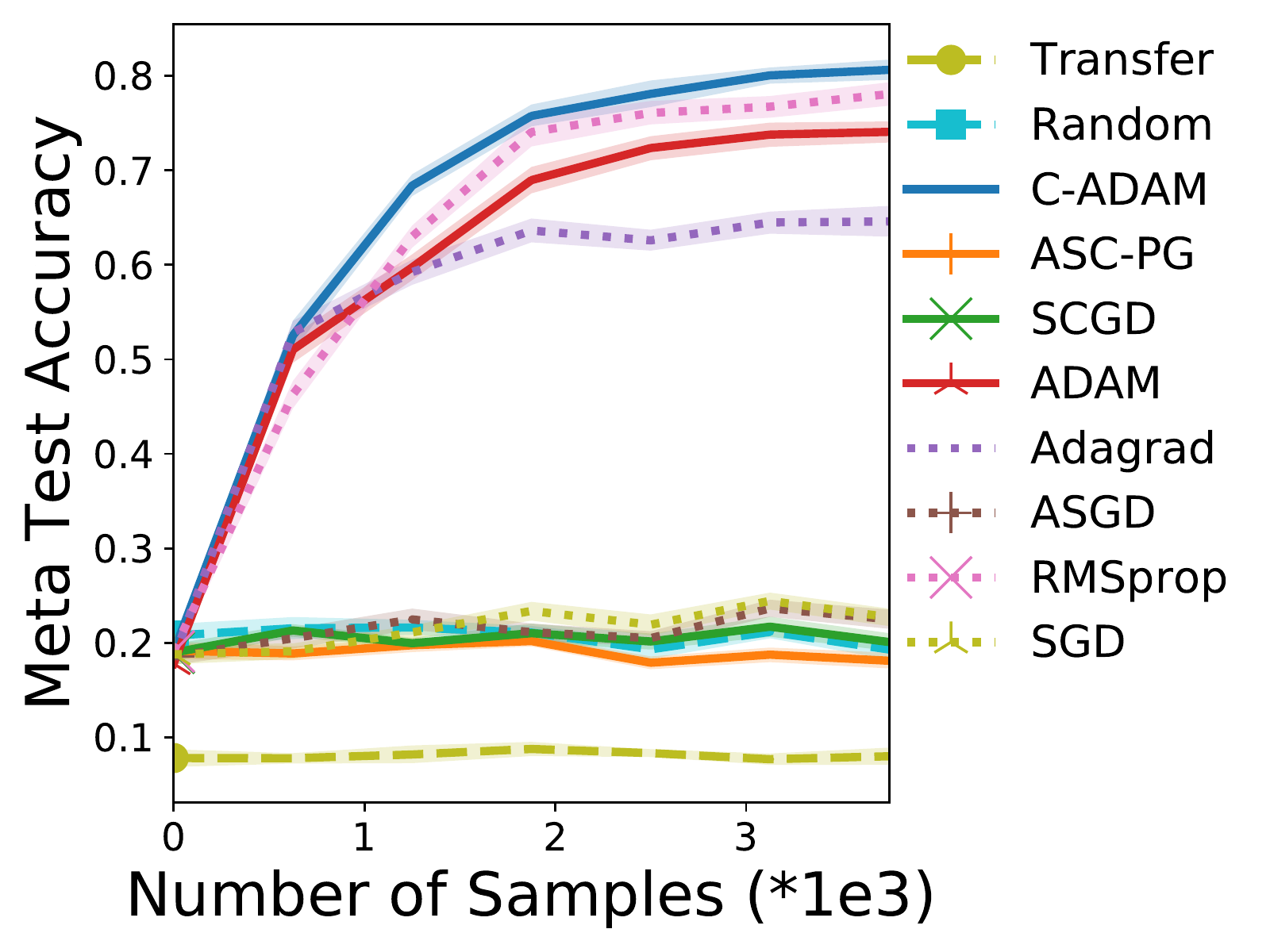}
\caption{}
\end{subfigure}
\begin{subfigure}{0.245\textwidth}
\centering
\includegraphics[width=\textwidth, trim={0 0 0 0.95cm },clip]{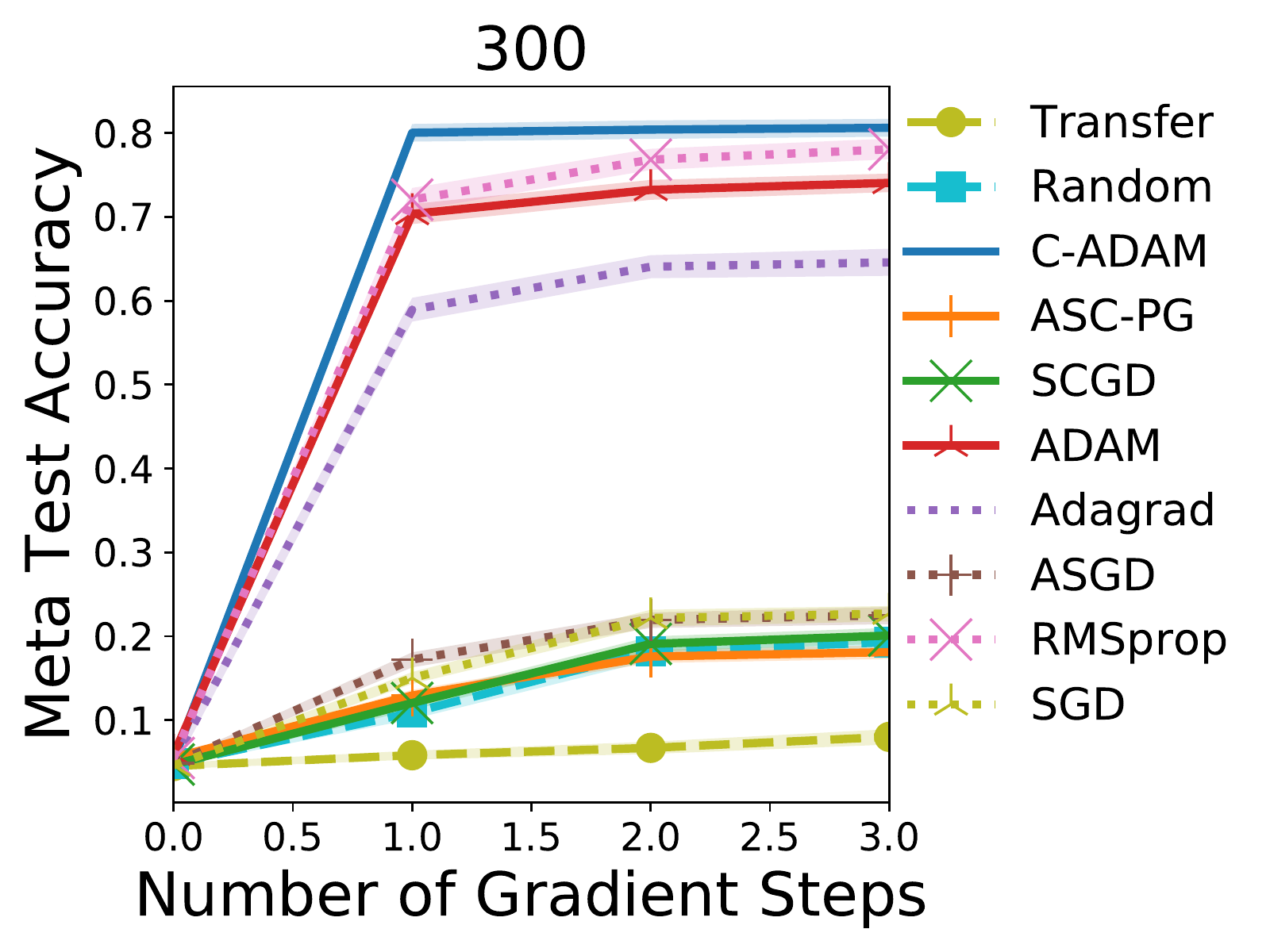}
\caption{}
\end{subfigure}
\begin{subfigure}{0.245\textwidth}
\centering
\includegraphics[width=\textwidth]{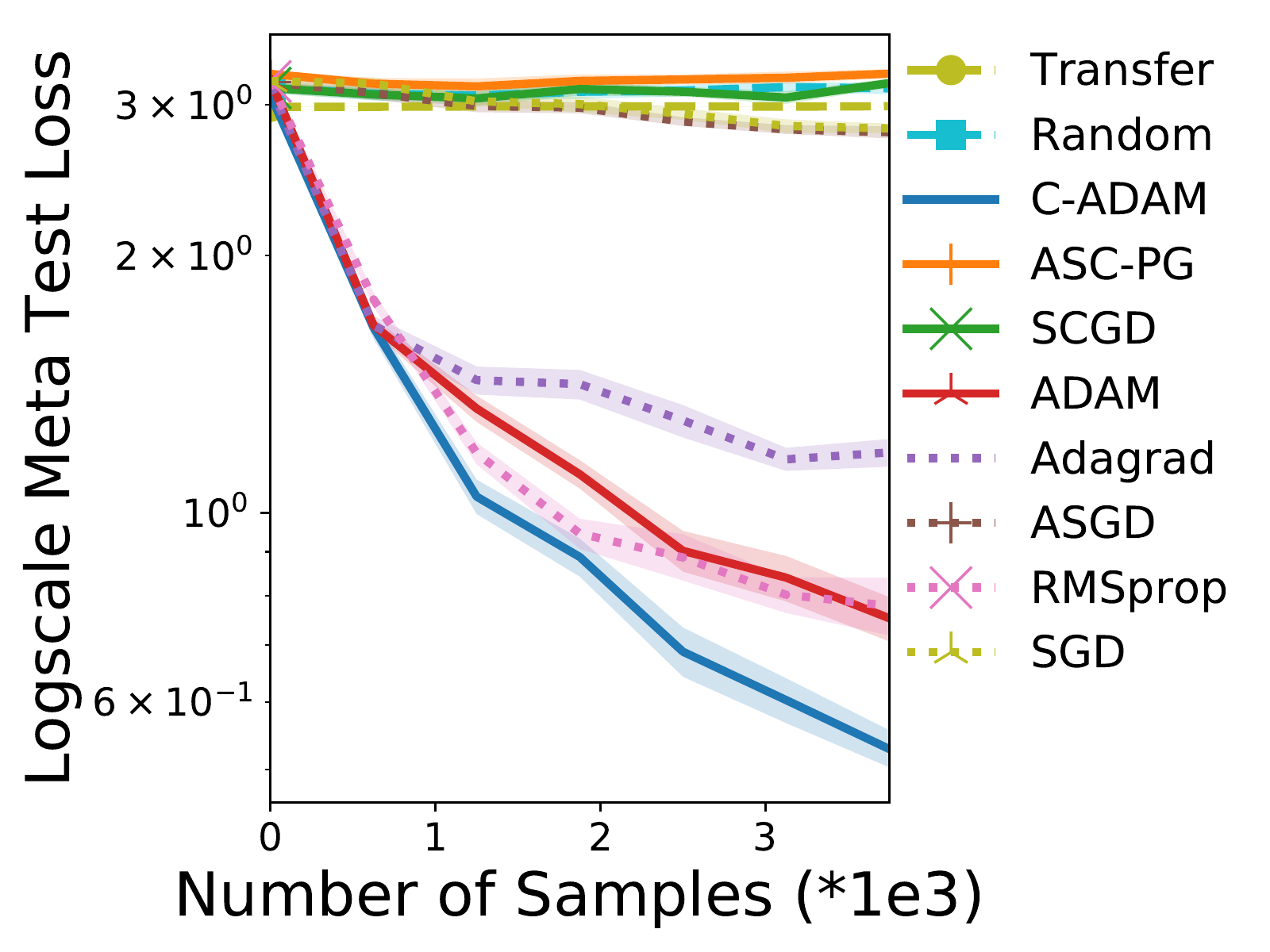}
\caption{}
\end{subfigure}
\begin{subfigure}{0.245\textwidth}
\centering
\includegraphics[width=\textwidth, trim={0 0 0 0.95cm },clip]{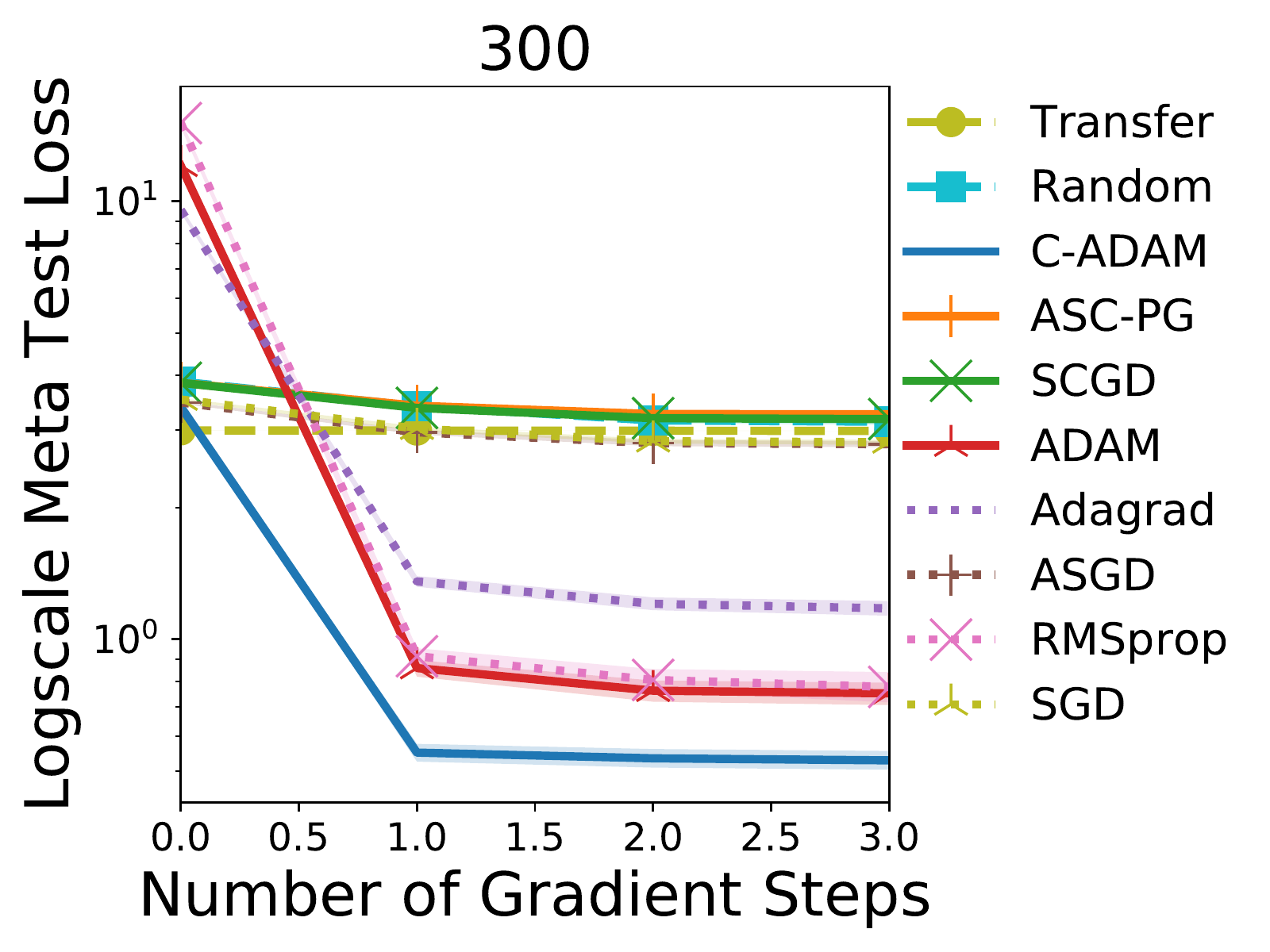}
\caption{}
\end{subfigure}
\begin{subfigure}{0.245\textwidth}
\centering
\includegraphics[width=\textwidth]{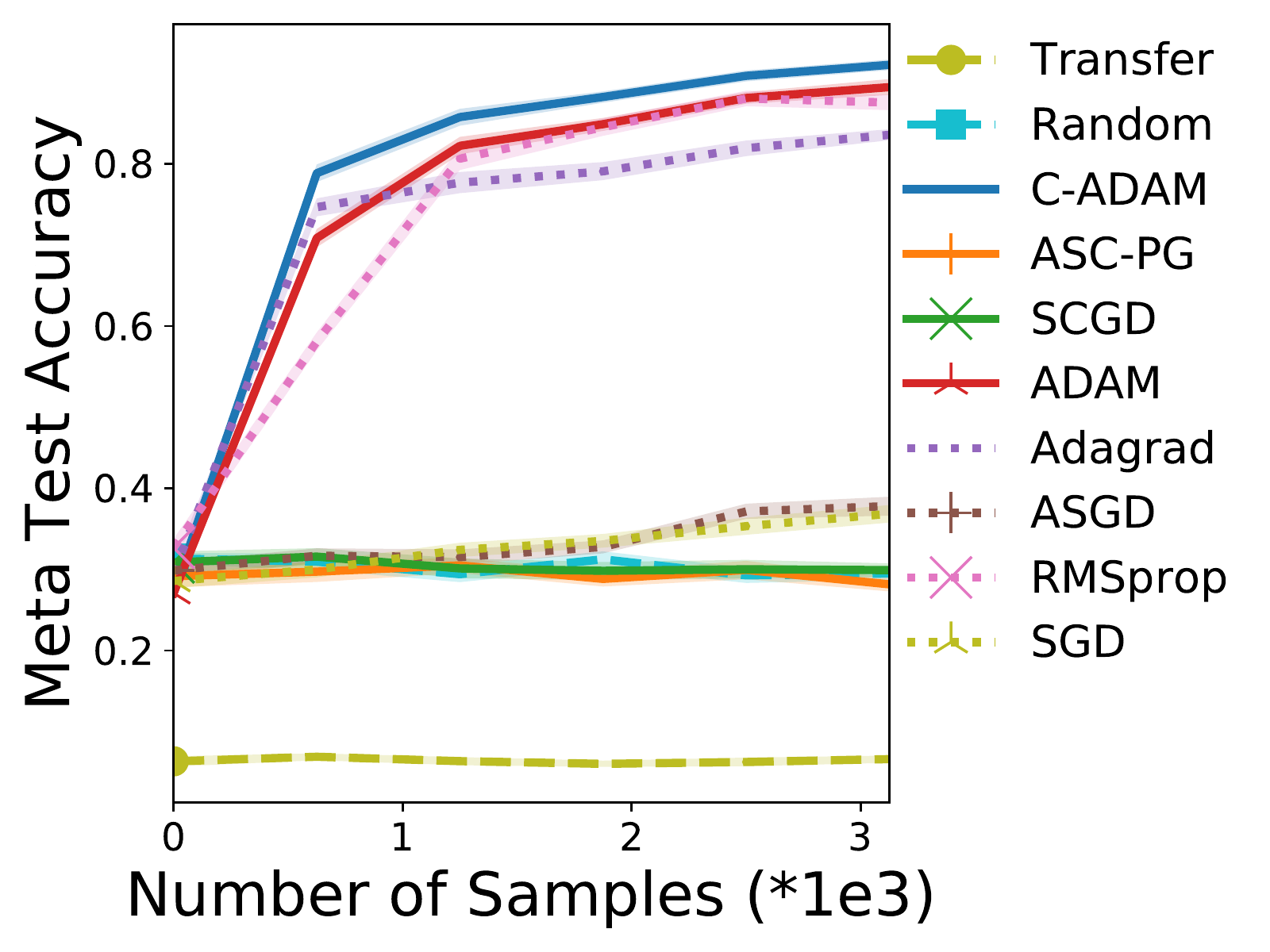}
\caption{}
\end{subfigure}
\begin{subfigure}{0.245\textwidth}
\centering
\includegraphics[width=\textwidth, trim={0 0 0 0.95cm },clip]{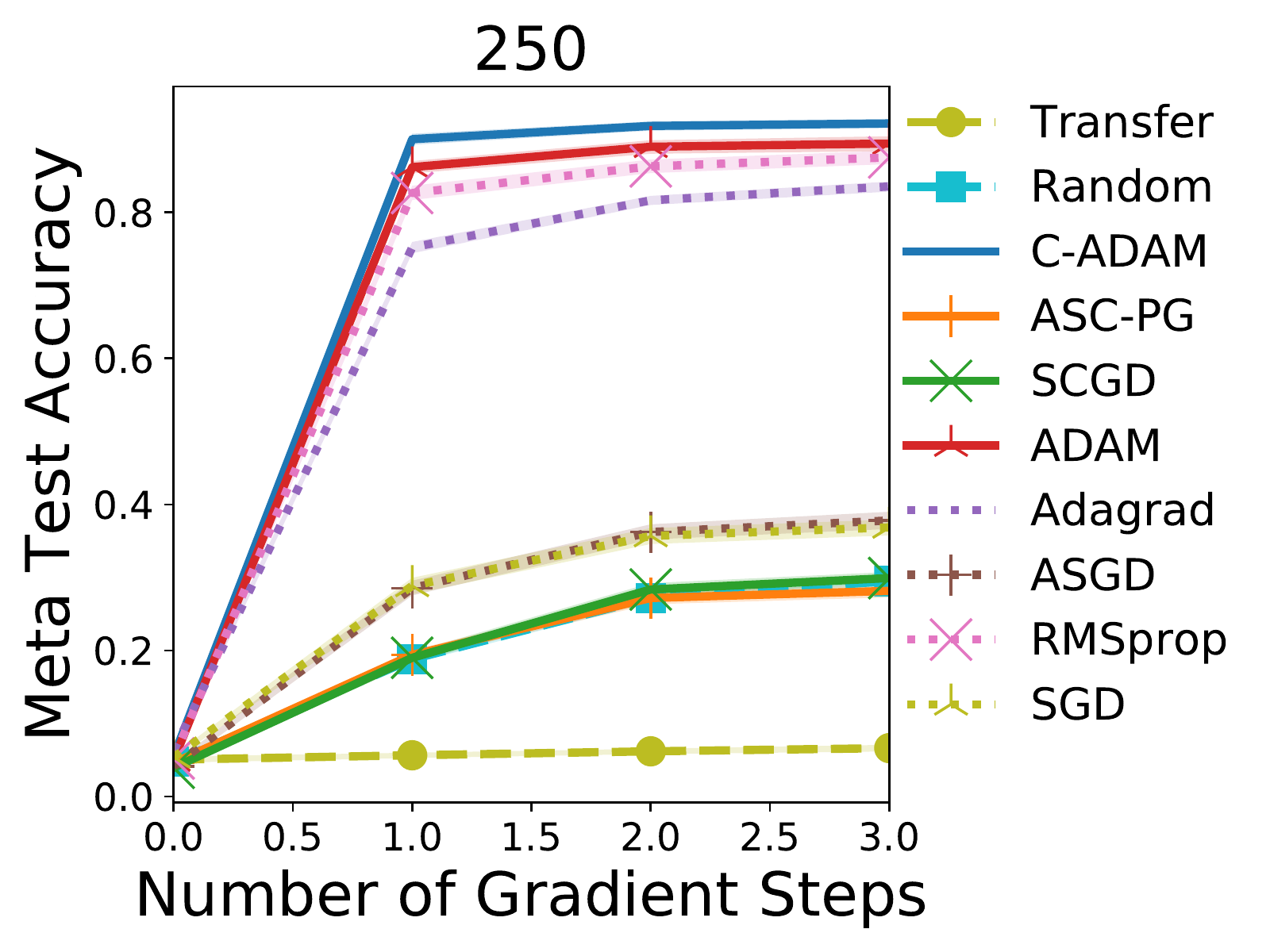}
\caption{}
\end{subfigure}
\begin{subfigure}{0.245\textwidth}
\centering
\includegraphics[width=\textwidth]{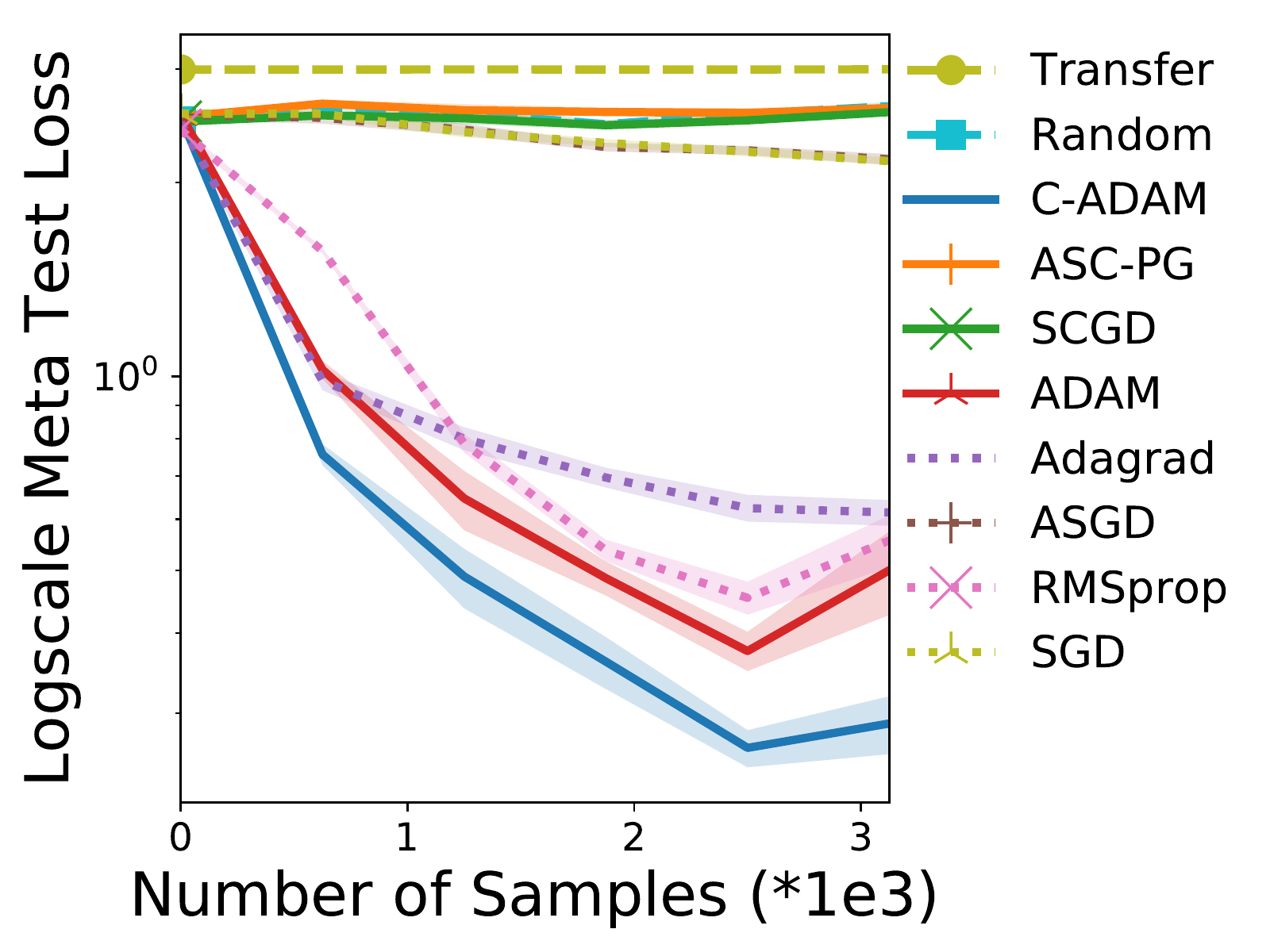}
\caption{}
\end{subfigure}
\begin{subfigure}{0.245\textwidth}
\centering
\includegraphics[width=\textwidth, trim={0 0 0 0.95cm },clip]{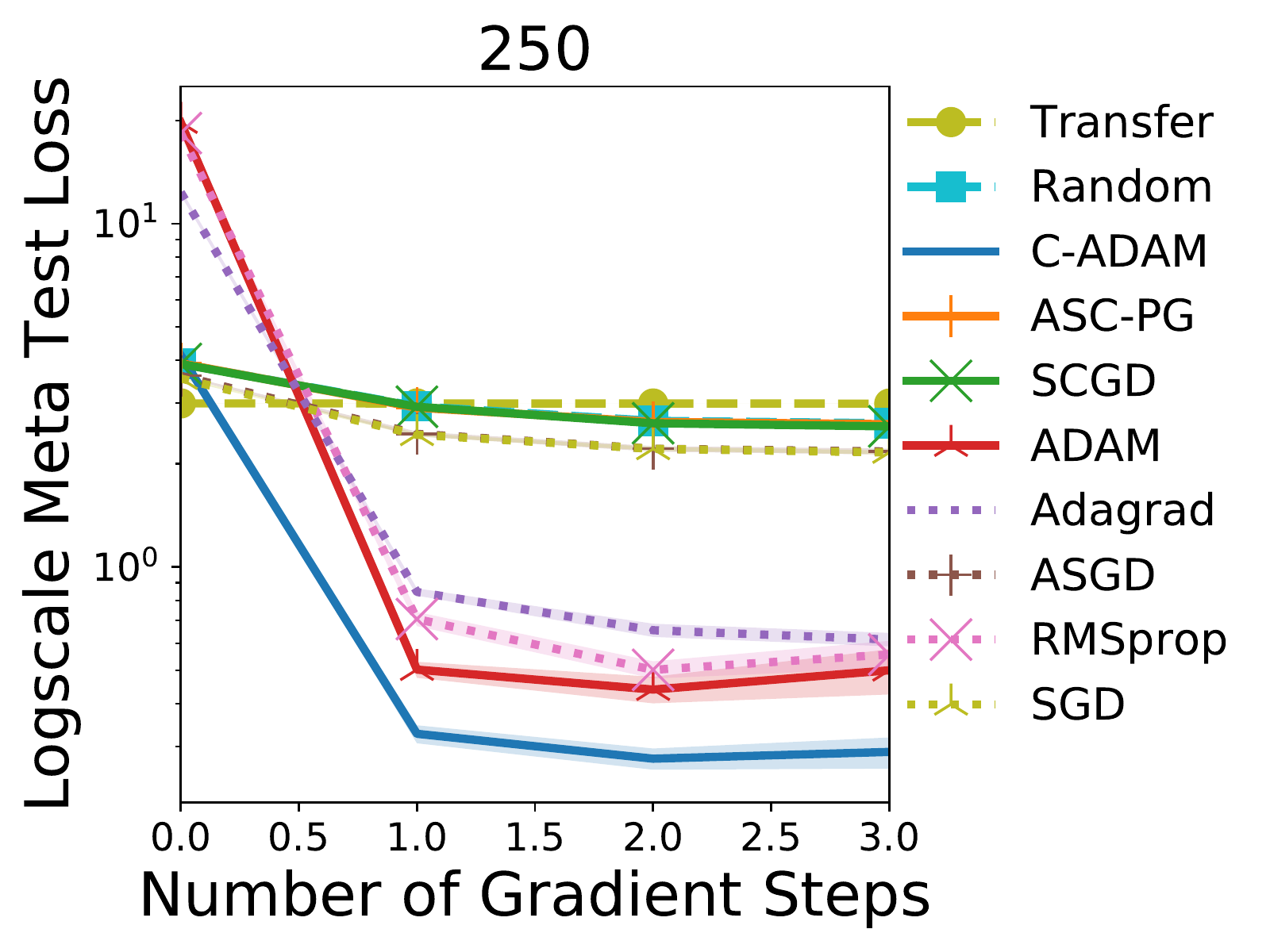}
\caption{}
\end{subfigure}
\caption{We show three scenarios on Omniglot data highlighting test statistics of C-MAML compared against other optimisation methods for varying N-shot K-way values using the convolutional model. (a)-(d) refer to the Omniglot 5-shot 5-way task, (e)-(h) Omniglot refers to 1-shot 20-way task, (i)-(l) refers to Omniglot 5-shot 20-way task.}
\label{fig:nshot_kway_omniglot_conv}
\end{figure*}

\subsection{Portfolio Mean-Variance}
We consider three large 100-portfolio data-sets ($m=13781, n=100$) and 18 region-based medium-sized sets with $m=7240$, and 25 assets as collected by CRSP to demonstrate book-to-market (BM), operating profitability (OP), and investment (Inv.). Given $n$ assets and reward vectors at $m$ time instances, the goal of sparse mean-variance optimisation~\citep{ravikumar2007sparse} is to maximise returns and control risk; a problem that can be mapped to a compositional form (see Appendix~B.1). 
We compared C-ADAM with a line of existing algorithms for compositional optimisation:
ASCVRG~\citep{lin2018improved}, VRSC-PG~\citep{ZHuo2017}, ASC-PG~\citep{Lui2016}, and SCGD~\citep{wang2014stochastic}.
Implementation of these baselines were provided by respective authors and optimised for mean-variance tasks. As a comparative metric between algorithms, we measured the optimality gap $\mathcal{J}(x) - \mathcal{J}^{*}$, versus number of samples used-per asset. Free parameters in Algorithm~\ref{Algo:ADAM} were set to $
C_{\alpha}=0.01, C_{\beta}=0.01, K_{t}^{(1)} = K_{t}^{(2)} = K_{t}^{(3)} = 1
$, and $C_{\gamma} = 1$ unless otherwise specified.
Figure~\ref{fig:exp}(a, b, c) shows that C-ADAM outperforms other algorithms on large data-sets, while Figure~\ref{fig:exp}(d, e, f) demonstrates that C-ADAM outperforms others on sample results from BM, OP, and Inv data-sets\footnote{Please note that due to space constraints full results (all demonstrating that C-ADAM outperforms others) on all 18 data-sets can be found in Appendix~B.1.}. \\
\textbf{Ablation Study:} Algorithm~\ref{Algo:ADAM} introduces free-parameters that need to be tuned for successful execution. To assess their importance, we conducted an ablation study on batch-sizes $K_{t}^{(1)}, K_{t}^{(2)}, K_{t}^{(3)}$, and step-sizes $C_{\alpha}, C_{\beta}$. We demonstrate the effects of varying these parameters on an asset data-set in Figure~\ref{fig:exp}(g, h). It is clear that C-ADAM performs best with a batch-size of 1 (Figure~\ref{fig:abla_sample_size}), and a small step size of 0.01 (Figure~\ref{fig:abla_step_size}). This, in turn, motivated our choice of these parameters in the portfolio experiments.  

\subsection{Compositional MAML}\label{Sec:ExperimentMAMLOne}
To validate theoretical guarantees achieved for MAML, we conduct two sets of experiments on the regression and classification problems originally introduced in~\cite{FinnAL17}.

\subsubsection{Compositional Regression Results}\label{Sec:ExperimentMAMLTwo}
After observing a set of tasks, the goal is to perform few-shot supervised regression on novel unobserved tasks when only a few-data points are available. Tasks vary in their data distribution, e.g., changing parameters for data generation distribution\footnote{Due to space constraints all hyper-parameters we used in our experiments can be found in Appendix~B.2.}.  

We use a neural network regressor with two hidden layers each, of size 40 with ReLU non-linearities. For all experiments, we use one meta step (one inner gradient update) with ten examples, i.e., 10-shot regression, and adopt an outer and inner learning rate (meta-learning rate) of $\alpha = 0.01$ for all algorithms unless stated otherwise. We compare C-ADAM with both compositional (ASC-PG and SCGD) and non-compositional optimisers (ADAM, Adagrad~\cite{ADAgrad}, ASGD~\cite{polyak1992acceleration}, RMSprop~\cite{RMSprop}, SGD) and demonstrate the performance in Figure~\ref{fig:exp}. In C-ADAM and ADAM, we set the outer learning rate to $0.001$. When dealing with only one task, i.e., streaming data points, C-ADAM achieves the best performance in training and test loss with respect to the number of samples, as shown in Figures 1(i) and (j).

Evaluating few-shot regression (multi-task scenario), we fine-tune the meta-learnt model using the same optimiser (SGD) on $M=10$ examples for each method. We also compare performance on two additional baseline models: (1) pre-training on all of the tasks, which entails training a single neural net to deal with multiple tasks at the same time (Transfer in Figure 1(k, l)), and (2) random initialization (Random in Figure 1(k, l)).

During fine-tuning, each gradient step is computed using the same $M=10$ data-points.
For each evaluation point, we report the result with $100$ test tasks and show the meta test loss (i.e., test loss on novel unseen tasks during training) after $M=10$ gradient steps in Figure~\ref{fig:reg_multi_task_fit_test}.
The meta-learnt model using C-ADAM optimiser is able to adapt during meta test quickly.
While models learnt with baseline optimisers can adapt to the meta test set after training for $40000$ iterations as shown in Figure~\ref{fig:reg_multi_task_finetune_40000},
C-ADAM's model still achieves the best convergence performance among all baselines, which
demonstrates both the advantage of our compositional formulation of MAML and the adaptive nature of C-ADAM.

\subsubsection{Compositional Classification Results}\label{Sec:ExperimentMAMLTwo}
Following a similar setup to MAML, we applied our algorithm to N-shot K-way image recognition on the Omniglot~\cite{lake2011one} and MiniImagenet~\cite{ravi2016optimization} data-set. In both data-sets, we use the same convolutional (Figure~\ref{fig:nshot_kway_omniglot_conv}) and  non-convolutional (Figure~\ref{fig:nshot_kway_omniglot}) model as in~\cite{FinnAL17}, trained with one meta step, and a meta batch size of 32 tasks. In Omniglot, we set all outer optimizer learning rates to 0.01 and a meta-learning rate of 0.4. For MiniImagenet, we set all outer optimizer learning rates to 0.001 and a meta-learning rate of 0.1. When comparing optimisers, we use the same learning rates and architecture. We evaluated models using three meta steps with the same meta step size as was used for training. Figure~\ref{fig:nshot_kway_omniglot_conv} highlights that C-MAML outperforms others across this suite of challenging image tasks. 

\section{Conclusions and Future Work}
We proposed C-ADAM, the first adaptive compositional solver. We validated our method both theoretically and empirically and provided the first connection between model-agnostic deep learning and compositional optimisation that attained best-known convergence bounds to-date. 

In future, we plan to investigate further applications of compositional optimisation such as robust single and multi-player type-objectives~\cite{graumoya2018balancing,mguni2018viscosity,abdullah2019wasserstein, wen2019modelling} as well as propose adaptive algorithms for problems involving nesting of more than two expected values.  

\bibliography{example_paper}
\bibliographystyle{icml2020}

\input{appendix.tex}
\end{document}

%% file: appendix.tex

\onecolumn
\appendix

\section{Theoretical Results}

In this part of the Appendix, we present proofs for all statements made in the main paper:

\subsection{Proof of Lemma 1}
\begin{lemma}\label{lemma_one}
If Assumptions \ref{assum_1}.1, \ref{assum_1}.3, and \ref{assum_1}.4 hold, then $\mathcal{J}(\boldsymbol{x})$ is $L$-Lipschitz smooth, i.e., 
\begin{equation*}
    \left|\left|\nabla_{\boldsymbol{x}}\mathcal{J}(\boldsymbol{x}_{1}) - \nabla_{\boldsymbol{x}}\mathcal{J}(\boldsymbol{x}_{2}) \right|\right|_{2} \leq L \left|\left|\boldsymbol{x}_{1} - \boldsymbol{x}_{2}\right|\right|_{2}, 
\end{equation*}  
for all $(\boldsymbol{x}_{1}, \boldsymbol{x}_{2})\in \mathbb{R}^{p}$, with $L = M_{g}^{2}L_f + L_g M_f$.
\end{lemma}
\begin{proof}
Assumption \ref{assum_1} implies that $||\nabla g_w(\boldsymbol{x})||_2 \le M_g$ for any $\boldsymbol{x}\in\mathbb{R}^p$. Hence, using Jensen inequality as well as property of the norm we have:
\begin{align*}
    &||\nabla\mathcal{J}(\boldsymbol{x}_1) - \nabla\mathcal{J}(\boldsymbol{x}_2)||_2 = \\\nonumber
    &||\mathbb{E}_{w}\left[\nabla g^{\mathsf{T}}_{w}(\boldsymbol{x_1})\right]\mathbb{E}_{v}\left[\nabla f_{v}\left(\mathbb{E}_{w}\left[g_{w}(\boldsymbol{x}_1)\right]\right)\right]  - \mathbb{E}_{w}\left[\nabla g^{\mathsf{T}}_{w}(\boldsymbol{x_2})\right]\mathbb{E}_{v}\left[\nabla f_{v}\left(\mathbb{E}_{w}\left[g_{w}(\boldsymbol{x}_2)\right]\right)\right]||_2\le\\\nonumber
    &||\mathbb{E}_{w}\left[\nabla g^{\mathsf{T}}_{w}(\boldsymbol{x_1})\right]\mathbb{E}_{v}\left[\nabla f_{v}\left(\mathbb{E}_{w}\left[g_{w}(\boldsymbol{x}_1)\right]\right)\right] - \mathbb{E}_{w}\left[\nabla g^{\mathsf{T}}_{w}(\boldsymbol{x_1})\right]\mathbb{E}_{v}\left[\nabla f_{v}\left(\mathbb{E}_{w}\left[g_{w}(\boldsymbol{x}_2)\right]\right) \right]||_2 + \\\nonumber
    &||\mathbb{E}_{w}\left[\nabla g^{\mathsf{T}}_{w}(\boldsymbol{x_1})\right]\mathbb{E}_{v}\left[\nabla f_{v}\left(\mathbb{E}_{w}\left[g_{w}(\boldsymbol{x}_2)\right]\right)\right] - \mathbb{E}_{w}\left[\nabla g^{\mathsf{T}}_{w}(\boldsymbol{x_2})\right]\mathbb{E}_{v}\left[\nabla f_{v}\left(\mathbb{E}_{w}\left[g_{w}(\boldsymbol{x}_2)\right]\right)\right]||_2\le \\\nonumber
    &\mathbb{E}_{w}\left[||\nabla g^{\mathsf{T}}_{w}(\boldsymbol{x}_1)||_2\right]||\mathbb{E}_{v}\left[\nabla f_{v}\left(\mathbb{E}_{w}\left[g_{w}(\boldsymbol{x}_1)\right]\right)\right] - \mathbb{E}_{v}\left[\nabla f_{v}\left(\mathbb{E}_{w}\left[g_{w}(\boldsymbol{x}_2)\right]\right)\right]||_2 + \\\nonumber
    &||\mathbb{E}_{w}\left[\nabla g^{\mathsf{T}}_{w}(\boldsymbol{x_1})\right] - \mathbb{E}_{w}\left[\nabla g^{\mathsf{T}}_{w}(\boldsymbol{x_2})\right]||_2\mathbb{E}_{v}\left[||\nabla f_{v}\left(\mathbb{E}_{w}\left[g_{w}(\boldsymbol{x}_2)\right]\right)||_2\right] \le\\\nonumber
    &M_g\mathbb{E}_{v}\left[||\nabla f_{v}\left(\mathbb{E}_{w}\left[g_{w}(\boldsymbol{x}_1)\right]\right) - \nabla f_{v}\left(\mathbb{E}_{w}\left[g_{w}(\boldsymbol{x}_2)\right]\right)||_2\right] + \\\nonumber
    &M_f\mathbb{E}_{w}\left[||\nabla g^{\mathsf{T}}_{w}(\boldsymbol{x}_1) - \nabla g^{\mathsf{T}}_{w}(\boldsymbol{x}_2)||_2\right] \le M_gL_f||\mathbb{E}_{w}\left[g_{w}(\boldsymbol{x}_1)\right] - \mathbb{E}_{w}\left[g_{w}(\boldsymbol{x}_2)\right]||_2 + \\\nonumber
    &M_f\mathbb{E}_{w}\left[||\nabla g^{\mathsf{T}}_{w}(\boldsymbol{x}_1) - \nabla g^{\mathsf{T}}_{w}(\boldsymbol{x}_2)||_2\right] \le M_gL_f\mathbb{E}_{w}\left[||g_{w}(\boldsymbol{x}_1) - g_{w}(\boldsymbol{x}_2)||_2\right] + \\\nonumber
    &M_f\mathbb{E}_{w}\left[||\nabla g^{\mathsf{T}}_{w}(\boldsymbol{x}_1) - \nabla g^{\mathsf{T}}_{w}(\boldsymbol{x}_2)||_2\right] \le M_gL_f\mathbb{E}_{w}\left[||g_{w}(\boldsymbol{x}_1) - g_{w}(\boldsymbol{x}_2)||_2\right] + M_fL_g||\boldsymbol{x}_1 - \boldsymbol{x}_2||_2 \le \\\nonumber
    &M^2_gL_f||\boldsymbol{x}_1 - \boldsymbol{x}_2||_2 + M_fL_g||\boldsymbol{x}_1 - \boldsymbol{x}_2||_2 = \left(M^2_gL_f + M_fL_g\right)||\boldsymbol{x}_1 - \boldsymbol{x}_2||_2 = L||\boldsymbol{x}_1 - \boldsymbol{x}_2||_2.
\end{align*}
which finishes the proof of the claim.
\end{proof}


\subsection{Proof of Lemma 2}
\begin{lemma}\label{lemma_2}
Consider auxiliary variable updates in lines 8 and 9 in Algorithm~\ref{Algo:ADAM}.
Let $\mathbb{E}_{\text{total}}[\cdot]$ denote the expectation with respect to \emph{all} incurred randomness. For any $t$, the following holds: 
\begin{align*}
    &\mathbb{E}_{\text{total}}\left[\left|\left|g(\boldsymbol{x}_{t+1}) -\boldsymbol{y}_{t+1}\right|\right|^2_{2}\right] \leq \frac{L_{g}^{2}}{2}\mathbb{E}_{\text{total}}\left[\mathcal{D}_{t+1}^{2}\right]\\
    & \hspace{15em}+ 2 \mathbb{E}_{\text{total}}\left[\left|\left|\boldsymbol{\mathcal{E}}_{t+1}\right|\right|_{2}^{2}\right],
\end{align*}
with $g(\boldsymbol{x}_{t+1}) = \mathbb{E}_{\omega}[g_{\omega}(\boldsymbol{x}_{t+1})]$, and $\mathcal{D}_{t+1}$, $||\boldsymbol{\mathcal{E}}_{t+1}||^2_2$ satisfy the following recurrent inequalities:
\begin{align*}
    &\mathcal{D}_{t+1}  \leq  (1 - \beta_{t}) \mathcal{D}_{t} +\frac{2M_{g}^{2}M_{f}^{2}}{\epsilon^{2}}\frac{\alpha_{t}^{2}}{\beta_{t}} + \beta_{t}\mathcal{F}_{t}^{2},\\\nonumber
    &\mathbb{E}_{\text{total}}\left[\left|\left|\boldsymbol{\mathcal{E}}_{t+1}\right|\right|_{2}^{2}\right]  \leq \left(1 - \beta_{t}\right)^{2}\mathbb{E}_{\text{total}}\left[\left|\left|\boldsymbol{\mathcal{E}}_{t}\right|\right|_{2}^{2}\right] + \frac{\beta_{t}^{2}}{K_{t}^{(3)}} \sigma_{3}^{2},\\\nonumber
    &\mathcal{F}_{t}^{2} \leq (1 - \beta_{t-1})\mathcal{F}_{t-1}^{2} + \frac{4 M_{g}^{2}M_{f}^{2}}{\epsilon^{2}}\frac{\alpha_{t-1}^{2}}{\beta_{t-1}},
 \end{align*}
 and $\mathcal{D}_1 = 0,\ \  \mathbb{E}_{\text{total}}\left[||\boldsymbol{\mathcal{E}}_1||^2_2\right] = ||g(\boldsymbol{x}_1)||^2_2, \ \ \mathcal{F}_1 = 0$.
\end{lemma}

\begin{proof}
 
 Let us introduce the sequence of coefficients $\{\theta\}^{t}_{j=0}$ such that
\begin{equation*}
    \theta^{(t)}_{j} = \begin{cases}
        \beta_{j}\prod_{i=j+1}^{t}(1 - \beta_{i}) &  \text{if  } j < t.\\
        \beta_{t} & \text{if  } j = t.
  \end{cases}
\end{equation*}
and we assume $\beta_0 = 1$ for simplicity. Denote $S_{t} = \sum_{j=0}^t\theta^{(t)}_{j}$, then:
\begin{align*}
    &S_{t} = \sum_{j=0}^t\theta^{(t)}_{j} = \\\nonumber
    &\beta_{t} + (1-\beta_{t})\beta_{t-1} + (1-\beta_{t})(1- \beta_{t-1})\beta_{t-2} + \cdots + (1-\beta_{t})(1- \beta_{t-1})\ldots (1 - \beta_1)\beta_{0} = \\\nonumber
    &\beta_{t} + (1 - \beta_{t})\left[\beta_{t-1} + (1- \beta_{t-1})\beta_{t-2} + \cdots + (1- \beta_{t-1})\ldots (1 - \beta_1)\beta_{0}\right] = \beta_{t} + (1 - \beta_{t})S_{t-1}
\end{align*}
and $S_1 = \beta_1 + (1 - \beta_1)\beta_0$. Since $\beta_0 = 1$ it implies $S_1 = S_2 = \ldots = S_t =  1$. By assuming $\overline{g_{0}(\boldsymbol{z}_1)} = \boldsymbol{0}_{q}$, one can represent $\boldsymbol{x}_{t+1}$ and $\boldsymbol{y}_{t+1}$ as a convex combinations of $\{\boldsymbol{z}_j\}^{t+1}_{j=1}$ and $\{\overline{g_{j}(\boldsymbol{z}_{j+1})}\}^{t}_{j=0}$ respectively:
\begin{align}
    &\boldsymbol{x}_{t+1} = \sum_{j=0}^{t}\theta^{(t)}_{j}\boldsymbol{z}_{j+1},\ \ \ \text{ and } \ \ \ \boldsymbol{y}_{t+1} = \sum_{j=0}^{t}\theta^{(t)}_{j}\overline{g_{j}(\boldsymbol{z}_{j+1})}
\end{align}
Hence, using Taylor expansion for function $g(\boldsymbol{z}_{j+1})$ around $\boldsymbol{x}_{t+1}$ we have:
\begin{align*}
    &\boldsymbol{y}_{t+1} = \sum_{j=0}^t\theta^{(t)}_{j}\overline{g_{j}(\boldsymbol{z}_{j+1})} = \sum_{j=0}^t\theta^{(t)}_{j}\left[g(\boldsymbol{z}_{j+1}) - g(\boldsymbol{z}_{j+1}) + \overline{g_{j}(\boldsymbol{z}_{j+1})}\right] = \sum_{j=0}^t\theta^{(t)}_{j}g(\boldsymbol{z}_{j+1}) + \\\nonumber
    &\sum_{j=0}^t\theta^{(t)}_{j}\left[\overline{g_{j}(\boldsymbol{z}_{j+1})} - g(\boldsymbol{z}_{j+1})\right] = \sum_{j=0}^t\theta^{(t)}_{j}\left(g(\boldsymbol{x}_{t+1}) + \nabla g(\boldsymbol{x}_{t+1})[\boldsymbol{z}_{j+1} - \boldsymbol{x}_{t+1}] + \mathcal{O}\left(||\boldsymbol{z}_{j+1} - \boldsymbol{x}_{t+1}||^2_2\right)\right)+ \\\nonumber
    &\sum_{j=0}^t\theta^{(t)}_{j}\left[\overline{g_{j}(\boldsymbol{z}_{j+1})} - g(\boldsymbol{z}_{j+1})\right] = g(\boldsymbol{x}_{t+1}) + \nabla g(\boldsymbol{x}_{t+1})\left[\sum_{j=0}^t\theta^{(t)}_{j}\boldsymbol{z}_{j+1} - \sum_{j=0}^t\theta^{(t)}_{j}\boldsymbol{x}_{k+1}\right] + \\\nonumber
    &\sum_{j=0}^t\theta^{(t)}_{j}\mathcal{O}\left(||\boldsymbol{z}_{j+1} - \boldsymbol{x}_{t+1}||^2_2\right) + \sum_{j=0}^t\theta^{(t)}_{j}\left[\overline{g_{j}(\boldsymbol{z}_{j+1})} - g(\boldsymbol{z}_{j+1})\right] = g(\boldsymbol{x}_{t+1}) + \sum_{j=0}^t\theta^{(t)}_{j}\left[\overline{g_{j}(\boldsymbol{z}_{j+1})} - g(\boldsymbol{z}_{j+1})\right] +\\\nonumber
    &\sum_{j=0}^t\theta^{(t)}_{j}\mathcal{O}\left(||\boldsymbol{z}_{j+1} - \boldsymbol{x}_{t+1}||^2_2\right)
\end{align*}
Therefore, using that $g(\cdot)$ is $L_g-$smooth function:
\begin{align*}
    ||\boldsymbol{y}_{t+1} - g(\boldsymbol{x}_{t+1})||_2 \le \frac{L_g}{2}\sum_{j=0}^{t}\theta^{(t)}_j||\boldsymbol{z}_{j+1} - \boldsymbol{x}_{t+1}||^2_2 + \left|\left|\sum_{j=0}^t\theta^{(t)}_{j}\left[\overline{g_{j}(\boldsymbol{z}_{j+1})} - g(\boldsymbol{z}_{j+1})\right]\right|\right|_2 
\end{align*}
and, applying $(a+b)^2\le 2a^2 + 2b^2$:
\begin{align*}
    ||\boldsymbol{y}_{t+1} - g(\boldsymbol{x}_{t+1})||^2_2 \le \frac{L^2_g}{2}\left(\underbrace{\sum_{j=0}^{t}\theta^{(t)}_j||\boldsymbol{z}_{j+1} - \boldsymbol{x}_{t+1}||^2_2}_{\mathcal{D}_{t+1}}\right)^2 + 2\left|\left|\underbrace{\sum_{j=0}^t\theta^{(t)}_{j}\left[\overline{g_{j}(\boldsymbol{z}_{j+1})} - g(\boldsymbol{z}_{j+1})\right]}_{\boldsymbol{\mathcal{E}}_{t+1}}\right|\right|^2_2 \ \ \
\end{align*}
Taking expectation $\mathbb{E}_{\text{total}}$ from both sides gives:
\begin{align}\label{expression_bound_1}
    \mathbb{E}_{\text{total}}\left[||\boldsymbol{y}_{t+1} - g(\boldsymbol{x}_{t+1})||^2_2\right] \le \frac{L^2_g}{2}\mathbb{E}_{\text{total}}\left[\mathcal{D}^2_{t+1}\right] + 2\mathbb{E}_{\text{total}}\left[\left|\left|\boldsymbol{\mathcal{E}}_{t+1}\right|\right|^2_2\right]
\end{align}
where 
\begin{align*}
    \mathcal{D}_{t+1} = \sum_{j=0}^{t} \theta_{j}^{(t)}\left|\left|\boldsymbol{z}_{j+1} - \boldsymbol{x}_{t+1}\right|\right|_{2}^{2}, \ \ \  \boldsymbol{\mathcal{E}}_{t+1} = \sum_{j=0}^{t} \theta_{j}^{(t)}\left[\overline{g_{j}(\boldsymbol{z}_{j+1})} - g(\boldsymbol{z}_{j+1})\right],  
\end{align*}
It is easy to see that $\mathcal{D}_1 = 0$ and $\boldsymbol{\mathcal{E}}_1 = g(\boldsymbol{z}_1)$ (using notational assumptions $\overline{g_{0}(\boldsymbol{z}_1)} = \boldsymbol{0}_{q}$ and $\beta_0 = 1$). Let us bound both terms in expression (\ref{expression_bound_1}). Due to $\theta^{t}_{j} = (1-\beta_t)\theta^{(t-1)}_{j}$ for $j<t$ for the expression $\mathcal{D}_{t+1}$ we have:
\begin{align*}
    &\mathcal{D}_{t+1} = \sum_{j=0}^{t}\theta^{(t)}_j||\boldsymbol{z}_{j+1} - \boldsymbol{x}_{t+1}||^2_2 = \\\nonumber
    &\sum_{j=0}^{t-1}\theta^{(t)}_j||\boldsymbol{z}_{j+1} - \boldsymbol{x}_{t+1}||^2_2 + \beta_t||\boldsymbol{z}_{t+1} - \boldsymbol{x}_{t+1}||^2_2 = (1 - \beta_t)\sum_{j=0}^{t-1}\theta^{(t-1)}_j||\boldsymbol{z}_{j+1} - \boldsymbol{x}_{t+1}||^2_2 + \beta_t||\boldsymbol{z}_{t+1} - \boldsymbol{x}_{t+1}||^2_2 = \\\nonumber
    &(1 - \beta_t)\sum_{j=0}^{t-1}\theta^{(t-1)}_j||\boldsymbol{z}_{j+1} - \boldsymbol{x}_{t+1}||^2_2 + \frac{(1 - \beta_t)^2}{\beta_t}||\boldsymbol{x}_{t+1} - \boldsymbol{x}_{t}||^2_2 = (1 - \beta_t)\sum_{j=0}^{t-1}\theta^{(t-1)}_j||\boldsymbol{z}_{j+1} - \boldsymbol{x}_{t}||^2_2 + \\\nonumber
    &\frac{(1 - \beta_t)^2}{\beta_t}||\boldsymbol{x}_{t+1} - \boldsymbol{x}_{t}||^2_2 + (1 - \beta_t)\sum_{j=0}^{t-1}\theta^{(t-1)}_j\left[||\boldsymbol{z}_{j+1} - \boldsymbol{x}_{t+1}||^2_2 - ||\boldsymbol{z}_{j+1} - \boldsymbol{x}_{t}||^2_2 \right] = (1 - \beta_t)\mathcal{D}_{t} + \\\nonumber
    &\frac{(1 - \beta_t)^2}{\beta_t}||\boldsymbol{x}_{t+1} - \boldsymbol{x}_{t}||^2_2 + (1 - \beta_t)\sum_{j=0}^{t-1}\theta^{(t-1)}_j\left[||\boldsymbol{z}_{j+1} - \boldsymbol{x}_{t+1}||_2 - ||\boldsymbol{z}_{j+1} - \boldsymbol{x}_{t}||_2\right]\times\\\nonumber
    &\left[||\boldsymbol{z}_{j+1} - \boldsymbol{x}_{t+1}||_2 + ||\boldsymbol{z}_{j+1} - \boldsymbol{x}_{t}||_2 \right] \le (1 - \beta_t)\mathcal{D}_{t} + \frac{(1 - \beta_t)^2}{\beta_t}||\boldsymbol{x}_{t+1} - \boldsymbol{x}_{t}||^2_2 + \\\nonumber
    &(1 - \beta_t)\sum_{j=0}^{t-1}\theta^{(t-1)}_j||\boldsymbol{x}_{t+1} - \boldsymbol{x}_t||_2\left[||\boldsymbol{x}_{t+1} - \boldsymbol{x}_t||_2 + 2||\boldsymbol{x}_{t} - \boldsymbol{z}_{j+1}||_2\right] = (1 - \beta_t)\mathcal{D}_{t} + \frac{(1 - \beta_t)^2}{\beta_t}||\boldsymbol{x}_{t+1} - \boldsymbol{x}_{t}||^2_2 + \\\nonumber
    &(1 - \beta_t)||\boldsymbol{x}_{t+1} - \boldsymbol{x}_t||^2_2 + 2(1 - \beta_t)||\boldsymbol{x}_{t+1} - \boldsymbol{x}_t||_2\sum_{j=0}^{t-1}\theta^{(t-1)}_j||\boldsymbol{x}_{t} - \boldsymbol{z}_{j+1}||_2 = (1 - \beta_t)\mathcal{D}_t + \\\nonumber
    &\frac{1-\beta_t}{\beta_t}||\boldsymbol{x}_{t+1} - \boldsymbol{x}_{t}||^2_2 + 2(1 - \beta_t)||\boldsymbol{x}_{t+1} - \boldsymbol{x}_t||_2\sum_{j=0}^{t-1}\theta^{(t-1)}_j||\boldsymbol{x}_{t} - \boldsymbol{z}_{j+1}||_2 
    \end{align*}
Applying $2ab \le \frac{1}{\beta_t}a^2 + \beta_tb^2$:
    \begin{align*}
    &\mathcal{D}_{t+1} \le \\\nonumber
    &(1 - \beta_t)\mathcal{D}_t + \frac{1-\beta_t}{\beta_t}||\boldsymbol{x}_{t+1} - \boldsymbol{x}_{t}||^2_2 + (1 - \beta_t)\left[\frac{||\boldsymbol{x}_{t+1} - \boldsymbol{x}_t||^2_2}{\beta_t} + \beta_t\left(\sum_{j=0}^{t-1}\theta^{(t-1)}_j||\boldsymbol{x}_{t} - \boldsymbol{z}_{j+1}||_2\right)^2\right] = \\\nonumber
    &(1 - \beta_t)\mathcal{D}_t + 2\frac{1-\beta_t}{\beta_t}||\boldsymbol{x}_{t+1} - \boldsymbol{x}_{t}||^2_2 + (1 - \beta_t)\beta_t\left(\sum_{j=0}^{t-1}\theta^{(t-1)}_j||\boldsymbol{x}_{t} - \boldsymbol{z}_{j+1}||_2\right)^2 \le \\\nonumber
    &(1 - \beta_t)\mathcal{D}_t + \frac{2}{\beta_t}||\boldsymbol{x}_{t+1} - \boldsymbol{x}_{t}||^2_2 + \beta_t\left(\underbrace{\sum_{j=0}^{t-1}\theta^{(t-1)}_j||\boldsymbol{x}_{t} - \boldsymbol{z}_{j+1}||_2}_{\mathcal{F}_{t}}\right)^2
\end{align*}
Applying the primal variable update with $||\overline{\nabla_{\boldsymbol{x}}\mathcal{J}(\cdot)}||_2 \le M_gM_f$ gives:
\begin{align}\label{d_t_expression}
    &\mathcal{D}_{t+1} \le \\\nonumber
    &(1 - \beta_t)\mathcal{D}_t + \frac{2\alpha^2_t}{\beta_t}\left|\left|\frac{\boldsymbol{m}_t}{\sqrt{\boldsymbol{v}_t} + \epsilon}\right|\right|^2_2 + \beta_t\mathcal{F}^2_t \le (1 - \beta_t)\mathcal{D}_t + \frac{2M^2_gM^2_f}{\epsilon^2}\frac{\alpha^2_t}{\beta_t} +  \beta_t\mathcal{F}^2_t
\end{align}
Next, for expression $\mathcal{F}_t$ we have (using $\theta^{t-1}_{j} = (1-\beta_{t-1})\theta^{(t-2)}_{j}$ for $j<t-1$):
\begin{align*}
    &\mathcal{F}_t = 
    \sum_{j=0}^{t-1}\theta^{(t-1)}_j||\boldsymbol{x}_{t} - \boldsymbol{z}_{j+1}||_2 = \\\nonumber
    &\sum_{j=0}^{t-2}\theta^{(t-1)}_j||\boldsymbol{x}_{t} - \boldsymbol{z}_{j+1}||_2 + \theta^{(t-1)}_{t-1}||\boldsymbol{x}_{t} - \boldsymbol{z}_{t}||_2 = \sum_{j=0}^{t-2}\theta^{(t-1)}_j||\boldsymbol{x}_{t} - \boldsymbol{z}_{j+1}||_2 + \beta_{t-1}||\boldsymbol{x}_{t} - \boldsymbol{z}_{t}||_2 = \\\nonumber
    &(1 - \beta_{t-1})\sum_{j=0}^{t-2}\theta^{(t-2)}_j||\boldsymbol{x}_{t} - \boldsymbol{z}_{j+1}||_2 + \beta_{t-1}||\boldsymbol{x}_{t} - \boldsymbol{z}_{t}||_2 \le (1 - \beta_{t-1})||\boldsymbol{x}_t - \boldsymbol{x}_{t-1}||_2 + \\\nonumber 
    &(1 - \beta_{t-1})\sum_{j=0}^{t-2}\theta^{(t-2)}_j\left[||\boldsymbol{x}_{t-1} - \boldsymbol{z}_{j+1}||_2 + ||\boldsymbol{x}_{t} - \boldsymbol{x}_{t-1}||_2\right] = (1 - \beta_{t-1})\left(\mathcal{F}_{t-1} + 2||\boldsymbol{x}_t - \boldsymbol{x}_{t-1}||_2\right) 
\end{align*}
and $\mathcal{F}_1 = 0$. Hence, applying $(a+b)^2 \le (1 + \alpha)a^2 + (1 + \frac{1}{\alpha})b^2$ for $\alpha = \beta_{t-1} > 0$ and using primal variable update with $||\overline{\nabla_{\boldsymbol{x}}\mathcal{J}(\cdot)}||_2 \le M_gM_f$:
\begin{align}\label{f_t_expression}
    &\mathcal{F}^2_{t} \le \\\nonumber
    &(1 + \beta_{t-1})(1 -  \beta_{t-1})^2\mathcal{F}^2_{t-1} + 4\left(1 + \frac{1}{\beta_{t-1}}\right)(1 - \beta_{t-1})^2||\boldsymbol{x}_t - \boldsymbol{x}_{t-1}||^2_2 \le \\\nonumber
    &(1 - \beta_{t-1})\mathcal{F}^2_{t-1} + \frac{4}{\beta_{t-1}}||\boldsymbol{x}_t - \boldsymbol{x}_{t-1}||^2_2 = (1 - \beta_{t-1})\mathcal{F}^2_{t-1} + \frac{4\alpha^2_{t-1}}{\beta_{t-1}}\left|\left|\frac{\boldsymbol{m}_{t-1}}{\sqrt{\boldsymbol{v}_{t-1}}+\epsilon}\right|\right|^2_2 \le \\\nonumber
    &(1 - \beta_{t-1})\mathcal{F}^2_{t-1} + \frac{4M^2_gM^2_f}{\epsilon^2}\frac{\alpha^2_{t-1}}{\beta_{t-1}}
\end{align}
Finally, for  $\boldsymbol{\mathcal{E}}_{t+1}$ we have (using $\theta^{t}_{j} = (1-\beta_t)\theta^{(t-1)}_{j}$ for $j<t$):
\begin{align*}
    &\boldsymbol{\mathcal{E}}_{t+1} = \\\nonumber &\sum_{j=0}^t\theta^{(t)}_{j}\left[\overline{g_j(\boldsymbol{z}_{j+1})} - g(\boldsymbol{z}_{j+1})\right]  = \sum_{j=0}^{t-1}\theta^{(t)}_{j}\left[\overline{g_j(\boldsymbol{z}_{j+1})} - g(\boldsymbol{z}_{j+1})\right] + \beta_t[\overline{g_t(\boldsymbol{z}_{t+1})} - g(\boldsymbol{z}_{t+1})] = \\\nonumber
    &\sum_{j=0}^{t-1}(1 - \beta_t)\theta^{(t-1)}_{j}\left[\overline{g_j(\boldsymbol{z}_{j+1})} - g(\boldsymbol{z}_{j+1})\right] + \beta_t[\overline{g_t(\boldsymbol{z}_{t+1})} - g(\boldsymbol{z}_{t+1})] = \\\nonumber
    &(1 - \beta_t)\sum_{j=0}^{t-1}\theta^{(t-1)}_{j}\left[\overline{g_j(\boldsymbol{z}_{j+1})} - g(\boldsymbol{z}_{j+1})\right] + \beta_t\left[\overline{g_t(\boldsymbol{z}_{t+1})} - g(\boldsymbol{z}_{t+1})\right] = \\\nonumber 
    &(1 - \beta_t)\boldsymbol{\mathcal{E}}_{t} + \beta_t\left[\overline{g_t(\boldsymbol{z}_{t+1})} - g(\boldsymbol{z}_{t+1})\right]
\end{align*}
Due to the fact, that all samplings done at iteration $t_1$ are independent from samplings done at iteration $t_2\neq t_1$, then consider expectation with all randomness induced at iteration $t$ (with fixed iterative value $\boldsymbol{x}_t$):
\begin{equation}\label{expec_prop_new}
    \mathbb{E}_{t}\left[\cdot\right] = \mathbb{E}_{K^{(1)}_{t},K^{(2)}_{t},K^{(3)}_{t}}\left[\cdot |\boldsymbol{x}_{t}\right] 
\end{equation}
for any $t$. Using that $\boldsymbol{\mathcal{E}}_t$ is independent from the randomness induced at iteration $t$ we have: 
\begin{align*}
    &\mathbb{E}_t\left[\left|\left|\boldsymbol{\mathcal{E}}_{t+1}\right|\right|^2_2\right] = \\\nonumber
    &\mathbb{E}_t\left[\left((1 - \beta_t)\boldsymbol{\mathcal{E}}^{\mathsf{T}}_{t} + \beta_t\left[\overline{g_t(\boldsymbol{z}_{t+1})} - g(\boldsymbol{z}_{t+1})\right]^{\mathsf{T}}\right)\left((1 - \beta_t)\boldsymbol{\mathcal{E}}_{t} + \beta_t\left[\overline{g_t(\boldsymbol{z}_{t+1})} - g(\boldsymbol{z}_{t+1})\right]\right) \right] = \\\nonumber
    &(1- \beta_t)^2\left|\left|\boldsymbol{\mathcal{E}}_{t}\right|\right|^2_2 + 2\beta_t(1 - \beta_t)\boldsymbol{\mathcal{E}}^{\mathsf{T}}_{t}\mathbb{E}_t\left[\overline{g_t(\boldsymbol{z}_{t+1})} - g(\boldsymbol{z}_{t+1})\right] + \beta^2_t\mathbb{E}_t\left[\left|\left|\overline{g_t(\boldsymbol{z}_{t+1})} - g(\boldsymbol{z}_{t+1})\right|\right|^2_2\right] = \\\nonumber
    &(1- \beta_t)^2\left|\left|\boldsymbol{\mathcal{E}}_{t}\right|\right|^2_2 + \beta^2_t\mathbb{E}_t\left[\left|\left|\overline{g_t(\boldsymbol{z}_{t+1})} - g(\boldsymbol{z}_{t+1})\right|\right|^2_2\right]
\end{align*}
where $\mathbb{E}_t\left[\overline{g_t(\boldsymbol{z}_{t+1})} - g(\boldsymbol{z}_{t+1})\right] = \mathbb{E}_{K^{(1)}_t, K^{(2)}_t, K^{(3)}_t}\left[\overline{g_t(\boldsymbol{z}_{t+1})} - g(\boldsymbol{z}_{t+1})\right] = \mathbb{E}_{K^{(1)}_t, K^{(2)}_t}[\boldsymbol{0}_q] = \boldsymbol{0}_q$ due to Assumption \ref{assum_2}.2. Assumption \ref{assum_2}.3 implies $\mathbb{E}_{t}\left[\left|\left|\overline{g_t(\boldsymbol{z}_{t+1})} - g(\boldsymbol{z}_{t+1})\right|\right|^2_2\right] \le \frac{1}{K^{(3)}_t}\sigma^2_3$, therefore,
\begin{align*}
    &\mathbb{E}_t\left[\left|\left|\boldsymbol{\mathcal{E}}_{t+1}\right|\right|^2_2\right] \le (1- \beta_t)^2\left|\left|\boldsymbol{\mathcal{E}}_{t}\right|\right|^2_2 + \frac{\beta^2_t}{K^{(3)}_t}\sigma^2_3
\end{align*}
Taking expectation $\mathbb{E}_{total}$ from both sides of the above inequality and using (\ref{expec_prop_new}) and the law of total expectation, we have:
\begin{align}\label{bound_on_e_term}
    &\mathbb{E}_{total}\left[\left|\left|\boldsymbol{\mathcal{E}}_{t+1}\right|\right|^2_2\right] \le (1- \beta_t)^2\mathbb{E}_{total}\left[\left|\left|\boldsymbol{\mathcal{E}}_{t}\right|\right|^2_2\right] + \frac{\beta^2_t}{K^{(3)}_t}\sigma^2_3
\end{align}
Combining (\ref{expression_bound_1}), (\ref{d_t_expression}), (\ref{f_t_expression}), and (\ref{bound_on_e_term}) gives the statement of the Lemma.
 \end{proof}


\subsection{Proof of Lemma 3}

\begin{lemma}\label{lemma_3}
Let $\eta_t = \frac{C_{\eta}}{t^a}$, $\zeta_t = \frac{C_{\zeta}}{t^b}$, where $C_{\eta} > 1 + b - a$, $C_{\zeta} > 0$, $(b - a)\notin[-1,0]$ and $a\le 1$. Consider the following recurrent inequality:
\begin{equation*}
    A_{t+1} \le (1 - \eta_t + C_1\eta^2_t)A_t + C_2\zeta_t,
\end{equation*}
where $C_1, C_2 \ge 0$. Then, there is a constant $C_{A} > 0 $ such that  $A_t \le  \frac{C_{A}}{t^{b-a}}$.
\end{lemma}

\begin{proof}
We adopted this proof from \cite{Mendgi_2017} and added to appendix to make the narration of the paper self-contained.\\

Let us introduce constant $C_{A}$ such that
\begin{equation*}
    C_{A} = \max_{t\le (C_1C^2_{\eta})^{\frac{1}{a}}+1}A_{t}t^{b - a} + \frac{C_2C_{\zeta}}{C_{\eta} - 1 - b + a}
\end{equation*}
The claim will be proved by induction. Consider two cases here:
\begin{enumerate}
    \item \textbf{If $t \le (C_1C^2_{\eta})^{\frac{1}{a}}$: } Then, by from the definition of constant $C_{A}$ it follows immediately:
    \begin{align*}
        A_{t} \le C_{A}t^{a-b} = \frac{C_{A}}{t^{b-a}}
    \end{align*}
    \item \textbf{If $t > (C_1C^2_{\eta})^{\frac{1}{a}}$: } Assume that $A_t \le \frac{C_{A}}{t^{b-a}}$ for some  $t > (C_1C^2_{\eta})^{\frac{1}{a}}$. Hence:
    \begin{align}\label{A_t_expression_1}
        &A_{t+1} \le \\\nonumber
        &(1 - \eta_t + C_1\eta^2_t)A_t + C_2\zeta_t = \left(1 - \frac{C_{\eta}}{t^a} + C_1\frac{C^2_{\eta}}{t^{2a}}\right)A_t + C_2\frac{C_{\zeta}}{t^b} \le \\\nonumber
        &\left(1 - \frac{C_{\eta}}{t^a} + C_1\frac{C^2_{\eta}}{t^{2a}}\right)\frac{C_{A}}{t^{b-a}} + C_2\frac{C_{\zeta}}{t^b} = \frac{C_{A}}{t^{b-a}} - \frac{C_AC_{\eta}}{t^{b}} + \frac{C_1C_AC^2_{\eta}}{t^{a + b}} + \frac{C_2C_{\zeta}}{t^{b}} = \\\nonumber
        &\frac{C_A}{(t+1)^{b-a}} - C_A\left[\underbrace{\frac{1}{(t+1)^{b-a}} - \frac{1}{t^{b-a}} + \frac{C_{\eta}}{t^{b}} - \frac{C_1C^2_{\eta}}{t^{a+b}}}_{\Delta_{t+1}}\right] + \frac{C_2C_{\zeta}}{t^b} = \\\nonumber
        &\frac{C_A}{(t+1)^{b-a}} - C_A\Delta_{t+1} + \frac{C_2C_{\zeta}}{t^b} = \frac{C_A}{(t+1)^{b-a}} - \Delta_{t+1}\left(C_A - \frac{C_2C_{\zeta}}{\Delta_{t+1}t^b}\right)
    \end{align}
    Since function $f(t) = \frac{1}{t^{c}}$ is convex for $c\notin [-1,0]$ and $t>0$ one can apply the first order condition of convexity:
    \begin{align*}
        f(t+1) \ge f(t) + f^{'}(t) \ \ \Longrightarrow \ \ \ \frac{1}{(t+1)^{c}} \ge \frac{1}{t^c} - c\frac{1}{t^{c+1}}
    \end{align*}
    Hence, using $a\le 1$ and $t > (C_1C^2_{\eta})^{\frac{1}{a}}$ for $\Delta_{t+1}$ we have:
    \begin{align*}
        &\Delta_{t+1} = \frac{1}{(t+1)^{b-a}} - \frac{1}{t^{b-a}} + \frac{C_{\eta}}{t^{b}} - \frac{C_1C^2_{\eta}}{t^{a+b}} \ge -(b-a)\frac{1}{t^{b-a+1}} + \frac{C_{\eta}}{t^{b}} - \frac{C_1C^2_{\eta}}{t^{a+b}} \ge \\\nonumber
        &-\frac{b-a}{t^{b-a+1}} + \frac{C_{\eta}}{t^{b}} - \frac{1}{t^{b}} \ge -\frac{b-a}{t^{b}} + \frac{C_{\eta}}{t^{b}} - \frac{1}{t^{b}} = \left(C_{\eta} - 1 - b + a\right)\frac{1}{t^b} > 0
    \end{align*}
    Moreover,
    \begin{align*}
        \frac{C_2C_{\zeta}}{\Delta_{t+1}t^b} \le \frac{C_2C_{\zeta}}{C_{\eta} - 1 - b + a} \le C_A
    \end{align*}
    Combining these two result in (\ref{A_t_expression_1}) gives:
    \begin{equation*}
        A_{t+1} \le \frac{C_A}{(t+1)^{b-a}} - \Delta_{t+1}\left(C_A - \frac{C_2C_{\zeta}}{\Delta_{t+1}t^b}\right) \le \frac{C_A}{(t+1)^{b-a}} 
    \end{equation*}
    which proves the induction step.
\end{enumerate}
\end{proof}


\subsection{Proof of Corollary 1}
\begin{corollary}\label{corollary_1}
Consider Algorithm \ref{Algo:ADAM} with step sizes given by: $\alpha_t = \frac{C_{\alpha}}{t^a}, \beta_t = \frac{C_{\beta}}{t^b}, \ \ \text{and} \ \  K^{(3)}_t = C_{3}t^{e}$,
for some constants $C_{\alpha},C_{\beta}, C_{3}, a,b,e > 0$ such that $(2a-2b)\notin [-1,0]$, and $b \le 1$. For $C_{\mathcal{D}},C_{\mathcal{E}}> 0$, we have: 
\begin{align*}
    \mathbb{E}_{total}\left[||g(\boldsymbol{x}_t) - \boldsymbol{y}_t||^2_2\right] \le  \frac{L^2_gC^2_{\mathcal{D}}}{2}\frac{1}{t^{4a-4b}} + 2C^2_{\mathcal{E}}\frac{1}{t^{b+e}},
\end{align*}
for some constants $C_{\mathcal{D}},C_{\mathcal{E}}> 0$.
\end{corollary}

\begin{proof}
Using $\alpha_t = \frac{C_{\alpha}}{t^a}$, $\beta_t = \frac{C_{\beta}}{t^b}$ in the recurrent inequalities for  $\mathcal{F}^2_t, \mathcal{D}_{t}$ and $\mathbb{E}_{total}\left[\left|\left|\boldsymbol{\mathcal{E}}_{t+1}\right|\right|^2_2\right]$ gives:

\begin{enumerate}
    \item For  $\mathcal{F}^2_{t+1}$:
          \begin{align*}
          &\mathcal{F}^2_{t+1} \le \\\nonumber
          &(1- \beta_{t})\mathcal{F}^2_{t} + \frac{4M^2_gM^2_f}{\epsilon^2}\frac{\alpha^2_{t}}{\beta_{t}} = \left(1 - \frac{C_{\beta}}{t^b}\right)\mathcal{F}^2_{t} + \frac{4M^2_gM^2_fC^2_{\alpha}}{\epsilon^2C_{\beta}}\frac{1}{t^{2a-b}}
          \end{align*}
          and applying Lemma \ref{lemma_3} gives 
          \begin{equation}\label{F_t_expression_assymp}
          \mathcal{F}^2_{t} \le \frac{C_{\mathcal{F}}}{t^{2a-2b}} \ \ \ \ \ \text{ where } \ \ \ \ \ C_{\mathcal{F}} = \frac{4M^2_gM^2_fC^2_{\alpha}}{\epsilon^2C_{\beta}(C_{\beta} - 1 - 2a + 2b)}
          \end{equation}
    
    \item For $\mathcal{D}_{t+1}$:
        \begin{align*}
        &\mathcal{D}_{t+1} \le \\\nonumber
        &\left(1 - \frac{C_{\beta}}{t^{b}}\right)\mathcal{D}_{t} + \frac{2M^2_gM^2_fC^2_{\alpha}}{\epsilon^2C_{\beta}}\frac{1}{t^{2a-b}} +  C_{\beta}C_{\mathcal{F}}\frac{1}{t^{2a-b}} = \\\nonumber
        &\left(1 - \frac{C_{\beta}}{t^{b}}\right)\mathcal{D}_{t} + \left[\frac{2M^2_gM^2_fC^2_{\alpha}}{\epsilon^2C_{\beta}} + C_{\beta}C_{\mathcal{F}}\right] \frac{1}{t^{2a-b}}
        \end{align*}
        and applying Lemma \ref{lemma_3} gives 
        \begin{equation}\label{D_t_expression_assymp}
        \mathcal{D}_{t} \le \frac{C_{\mathcal{D}}}{t^{2a-2b}} \ \ \ \ \ \text{ where } \ \ \ \ \ C_{\mathcal{D}} = \frac{2M^2_gM^2_fC^2_{\alpha} + \epsilon^2C^2_{\beta}C_{\mathcal{F}}}{\epsilon^2C_{\beta}(C_{\beta} - 1 - 2a + 2b)}
        \end{equation}
    
    \item For $\mathbb{E}_{total}\left[\left|\left|\boldsymbol{\mathcal{E}}_{t+1}\right|\right|^2_2\right]$:
    \begin{align*}
        &\mathbb{E}_{total}\left[\left|\left|\boldsymbol{\mathcal{E}}_{t+1}\right|\right|^2_2\right] \le (1- \beta_t)^2\mathbb{E}_{total}\left[\left|\left|\boldsymbol{\mathcal{E}}_{t}\right|\right|^2_2\right] + \frac{\beta^2_t}{K^{(3)}_t}\sigma^2_3 = \\\nonumber
        &\left(1 - \frac{2C_{\beta}}{t^{b}} + \frac{C^2_{\beta}}{t^{2b}}\right)\mathbb{E}_{total}\left[\left|\left|\boldsymbol{\mathcal{E}}_{t}\right|\right|^2_2\right] + \frac{C^2_{\beta}\sigma^2_3}{C_3}\frac{1}{t^{2b + e}} = \\\nonumber
        &\left(1 - \frac{\tilde{C}_{\beta}}{t^{b}} + \frac{\tilde{C}^2_{\beta}}{4t^{2b}}\right)\mathbb{E}_{total}\left[\left|\left|\boldsymbol{\mathcal{E}}_{t}\right|\right|^2_2\right] + \frac{C^2_{\beta}\sigma^2_3}{C_3}\frac{1}{t^{2b + e}}
    \end{align*}
     and applying Lemma \ref{lemma_3} gives:
     \begin{equation}\label{mathbb_e_expres_assymp}
         \mathbb{E}_{total}\left[\left|\left|\boldsymbol{\mathcal{E}}_{t}\right|\right|^2_2\right] \le \frac{C_{\mathcal{E}}}{t^{b + e}} \ \ \text{ where } \ \ C_{\mathcal{E}} = \max_{t \le (C^2_{\beta})^{\frac{1}{b}} + 1}\mathbb{E}_{total}\left[\left|\left|\boldsymbol{\mathcal{E}}_{t}\right|\right|^2_2\right]t^{b+e} + \frac{C^2_{\beta}\sigma^2_3}{C_3(2C_{\beta} - 1 - b - e)}
     \end{equation}
\end{enumerate}
Next, combining results (\ref{D_t_expression_assymp}) and (\ref{mathbb_e_expres_assymp}) in the bound of Lemma \ref{lemma_2} gives:
\begin{align*}
    &\mathbb{E}_{total}\left[||g(\boldsymbol{x}_t) - \boldsymbol{y}_t||^2_2\right] \le \\\nonumber
    &\frac{L^2_g}{2}\mathbb{E}_{total}\left[\mathcal{D}^2_{t}\right] + 2\mathbb{E}_{total}\left[\left|\left|\boldsymbol{\mathcal{E}}_{t}\right|\right|^2_2\right] \le \frac{L^2_gC^2_{\mathcal{D}}}{2}\frac{1}{t^{4a-4b}} + 2C^2_{\mathcal{E}}\frac{1}{t^{b+e}}
\end{align*}
\end{proof}


\subsection{Proof of Main Theorem 1}

\begin{theorem}
Consider Algorithm~\ref{Algo:ADAM} with a parameter setup given by: $\alpha_{t} = \sfrac{C_{\alpha}}{t^{\frac{1}{5}}}, \beta_{t} = C_{\beta}, K_{t}^{(1)} = C_{1}t^{\frac{4}{5}}, K_{t}^{(2)} = C_{2}t^{\frac{4}{5}},  K_{t}^{(3)}  = C_{3} t^{\frac{4}{5}}, \gamma_{t}^{(1)}= C_{\gamma}\mu^{t}, \ \gamma_{2}^{(t)}  = 1 - \sfrac{C_{\alpha}}{t^{\frac{2}{5}}}(1 - C_{\gamma}\mu^{t})^{2}$, for some positive constants $C_{\alpha}, C_{\beta}, C_{1}, C_{2}, C_{3}, C_{\gamma}, \mu$ such that $C_{\beta} < 1$ and $\mu \in (0,1)$. For any $\delta \in (0,1)$, Algorithm~\ref{Algo:ADAM} outputs, in expectation, a $\delta$-approximate first-order stationary point $\tilde{\bm{x}}$ of $\mathcal{J}(\bm{x})$. That is: 
\begin{equation*}
    \mathbb{E}_{\text{total}}\left[\left|\left|\nabla_{\bm{x}}\mathcal{J}(\tilde{\bm{x}})\right|\right|_{2}^{2}\right] \leq \delta,  
\end{equation*}
with ``total'' representing \emph{all} incurred randomness. 
Moreover, Algorithm~\ref{Algo:ADAM} acquires $\tilde{\bm{x}}$ with an overall oracle complexity for $\text{Oracle}_{f}(\cdot, \cdot)$ and $\text{Oracle}_{g}(\cdot, \cdot)$ of the order $\mathcal{O}\left(\delta^{-\frac{9}{4}}\right)$. 
\end{theorem}

\begin{proof}

Let us study the change of the function between two consecutive iterations. Using, that function $\mathcal{J}(\cdot)$ is Lipschitz continuous (see Lemma  \ref{lemma_one}):
\begin{align}\label{First_diff_equation}
    &\mathcal{J}(\boldsymbol{x}_{t+1}) \le \mathcal{J}(\boldsymbol{x}_{t}) + \nabla \mathcal{J}(\boldsymbol{x}_t)^{\mathsf{T}}(\boldsymbol{x}_{t+1} - \boldsymbol{x}_t) + \frac{L}{2}||\boldsymbol{x}_{t+1} - \boldsymbol{x}_t||^{2}_2 = \\\nonumber
    &\mathcal{J}(\boldsymbol{x}_t) - \alpha_t\sum_{i=1}^p\left[\nabla\mathcal{J}(\boldsymbol{x}_t)\right]_i\frac{[\boldsymbol{m}_t]_i}{\sqrt{[\boldsymbol{v}_t]_i} + \epsilon} + \frac{L\alpha^2_t}{2}\sum_{i=1}^p\frac{[\boldsymbol{m}_t]^2_i}{\left(\sqrt{[\boldsymbol{v}_t]_i} + \epsilon\right)^2}
\end{align}
Now, let us introduce mathematical expectation with respect to all randomness at iteration $t$ given a fixed iterative value $\boldsymbol{x}_t$ as $\mathbb{E}_{t}\left[\cdot\right]= \mathbb{E}_{K^{(1)}_t, K^{(2)}_t, K^{(3)}_t}\left[ \cdot \Big| \boldsymbol{x}_t\right]$. This expectation taking into account all samplings which is done on iteration $t$ of Algorithm \ref{Algo:ADAM}. Then, it is easy to see that the following variables will be $t-\text{ measurable}$: $\title{\nabla}\mathcal{J}(\boldsymbol{x}_t), \boldsymbol{m}_t, \boldsymbol{v}_t, \boldsymbol{x}_{t+1}, \boldsymbol{z}_{t+1}, \boldsymbol{y}_{t+1}$. On the other hand, the following variables will be independent from randomness introduced at iteration $t$: $\nabla\mathcal{J}(\boldsymbol{x}_{t-1}), \boldsymbol{m}_{t-1}, \boldsymbol{v}_{t-1}, \boldsymbol{x}_{t}, \boldsymbol{z}_{t}, \boldsymbol{y}_{t}$. Hence, taking expectation $\mathbb{E}_{t}\left[\right]$ from the both sides of equation (\ref{First_diff_equation}) gives:
\begin{align}\label{Secon_expectation}
    &\mathbb{E}_t\left[\mathcal{J}(\boldsymbol{x}_{t+1})\right] \le \mathcal{J}(\boldsymbol{x}_t) - \alpha_t\sum_{i=1}^p\left[\nabla\mathcal{J}(\boldsymbol{x}_t)\right]_i\mathbb{E}_t\left[\frac{[\boldsymbol{m}_t]_i}{\sqrt{[\boldsymbol{v}_t]_i} + \epsilon}\right] + \frac{L\alpha^2_t}{2}\sum_{i=1}^p\mathbb{E}_t\left[\frac{[\boldsymbol{m}_t]^2_i}{\left(\sqrt{[\boldsymbol{v}_t]_i} + \epsilon\right)^2}\right]
\end{align}
Now, let us focus on the second term in the above expression:
\begin{align*}
    &\sum_{i=1}^p\left[\nabla\mathcal{J}(\boldsymbol{x}_t)\right]_i\mathbb{E}_t\left[\frac{[\boldsymbol{m}_t]_i}{\sqrt{[\boldsymbol{v}_t]_i} + \epsilon}\right] =\sum_{i=1}^p\left[\nabla\mathcal{J}(\boldsymbol{x}_t)\right]_i\mathbb{E}_t\left[\frac{[\boldsymbol{m}_t]_i}{\sqrt{[\boldsymbol{v}_t]_i} + \epsilon} - \frac{[\boldsymbol{m}_t]_i}{\sqrt{\gamma^{(2)}_t[\boldsymbol{v}_{t-1}]_i} + \epsilon} + \frac{[\boldsymbol{m}_t]_i}{\sqrt{\gamma^{(2)}_t[\boldsymbol{v}_{t-1}]_i} + \epsilon}\right]  \\\nonumber
    &=\underbrace{\sum_{i=1}^p\left[\nabla\mathcal{J}(\boldsymbol{x}_t)\right]_i\mathbb{E}_t\left[\frac{[\boldsymbol{m}_t]_i}{\sqrt{\gamma^{(2)}_t[\boldsymbol{v}_{t-1}]_i} + \epsilon}\right]}_{\mathcal{A}} + \underbrace{\sum_{i=1}^p\left[\nabla\mathcal{J}(\boldsymbol{x}_t)\right]_i\mathbb{E}_t\left[\frac{[\boldsymbol{m}_t]_i}{\sqrt{[\boldsymbol{v}_{t}]_i} + \epsilon} - \frac{[\boldsymbol{m}_t]_i}{\sqrt{\gamma^{(2)}_t[\boldsymbol{v}_{t-1}]_i} + \epsilon}\right]}_{\mathcal{B}}
\end{align*}
Please notice, from Assumptions \ref{assum_1}.3 and \ref{assum_1}.4 it follows immediately:
\begin{align}\label{bound_expressions}
    &||\nabla\mathcal{J}(\boldsymbol{x}_t)||_2 \le \mathbb{E}_w\left[\left|\left|\nabla g_w(\boldsymbol{x}_t)^{\mathsf{T}} \right|\right|_2\right]\mathbb{E}_v\left[\left|\left|\nabla f_v(\mathbb{E}_w[g_w(\boldsymbol{x}_t)]) \right|\right|_2\right]\le M_gM_f\\\nonumber
    &||\overline{\nabla \mathcal{J}(\boldsymbol{x}_{t})}||_2 \le \left|\left|\overline{\nabla g_{t}(\boldsymbol{x}_{t})}^{\mathsf{T}} \right|\right|_2\left|\left|\overline{\nabla f_{t}(\boldsymbol{y}_{t})} \right|\right|_2\le M_gM_f\\\nonumber
    &\text{ and applying induction: }\\\nonumber
    &||\boldsymbol{m}_t||_2 \le \gamma^{(1)}_tM_gM_f + \left(1 - \gamma^{(1)}_t\right)M_gM_f = M_gM_f,\ \ \ \ \ \ \forall t\\\nonumber
    &||\boldsymbol{v}_t||_2 \le \gamma^{(2)}_tM^2_gM^2_f + \left(1 - \gamma^{(2)}_t\right)M^2_gM^2_f = M^2_gM^2_f,\ \ \ \ \ \ \forall t
\end{align}
Now, let us apply (\ref{bound_expressions}) for the  expression $\mathcal{A}$:
\begin{align*}
    &\mathcal{A} = \sum_{i=1}^p\left[\nabla\mathcal{J}(\boldsymbol{x}_t)\right]_i\frac{\mathbb{E}_t[\boldsymbol{m}_t]_i}{\sqrt{\gamma^{(2)}_t[\boldsymbol{v}_{t-1}]_i} + \epsilon} = \sum_{i=1}^p\left[\nabla\mathcal{J}(\boldsymbol{x}_t)\right]_i\frac{\mathbb{E}_t\left[\gamma^{(1)}_t[\boldsymbol{m}_{t-1}]_i + \left(1 - \gamma^{(1)}_t\right)[\overline{\nabla \mathcal{J}(\boldsymbol{x}_{t})}]_i\right]}{\sqrt{\gamma^{(2)}_t[\boldsymbol{v}_{t-1}]_i} + \epsilon} =\\\nonumber
    &\sum_{i=1}^p\left[\nabla\mathcal{J}(\boldsymbol{x}_t)\right]_i\frac{\gamma^{(1)}_t[\boldsymbol{m}_{t-1}]_i + \left(1 - \gamma^{(1)}_t\right)\mathbb{E}_t\left[[\overline{\nabla \mathcal{J}(\boldsymbol{x}_{t})}]_i\right]}{\sqrt{\gamma^{(2)}_t[\boldsymbol{v}_{t-1}]_i} + \epsilon} = \gamma^{(1)}_t\nabla\mathcal{J}(\boldsymbol{x}_t)^{\mathsf{T}}\frac{\boldsymbol{m}_{t-1}}{\sqrt{\gamma^{(2)}_t\boldsymbol{v}_{t-1}}+ \epsilon} + \\\nonumber
    &\left(1 - \gamma^{(1)}_t\right)\nabla\mathcal{J}(\boldsymbol{x}_t)^{\mathsf{T}}\frac{\mathbb{E}_t\left[\overline{\nabla g_{t}(\boldsymbol{x}_{t})}^{\mathsf{T}}\overline{\nabla f_{t}(\boldsymbol{y}_{t})}\right]}{\sqrt{\gamma^{(2)}_t\boldsymbol{v}_{t-1}}+ \epsilon} = \gamma^{(1)}_t\nabla\mathcal{J}(\boldsymbol{x}_t)^{\mathsf{T}}\frac{\boldsymbol{m}_{t-1}}{\sqrt{\gamma^{(2)}_t\boldsymbol{v}_{t-1}}+ \epsilon} + \\\nonumber
    &\left(1 - \gamma^{(1)}_t\right)\nabla\mathcal{J}(\boldsymbol{x}_t)^{\mathsf{T}}\frac{\mathbb{E}_t\left[\nabla\mathcal{J}(\boldsymbol{x}_t) - \nabla\mathcal{J}(\boldsymbol{x}_t) + \overline{\nabla g_{t}(\boldsymbol{x}_{t})}^{\mathsf{T}}\overline{\nabla f_{t}(\boldsymbol{y}_{t})}\right]}{\sqrt{\gamma^{(2)}_t\boldsymbol{v}_{t-1}}+ \epsilon} = \gamma^{(1)}_t\nabla\mathcal{J}(\boldsymbol{x}_t)^{\mathsf{T}}\frac{\boldsymbol{m}_{t-1}}{\sqrt{\gamma^{(2)}_t\boldsymbol{v}_{t-1}}+ \epsilon} + \\\nonumber
    &\left(1 - \gamma^{(1)}_t\right)\sum_{i=1}^{p}\frac{[\nabla\mathcal{J}(\boldsymbol{x}_{t})]^2_i}{\sqrt{\gamma^{(2)}_t[\boldsymbol{v}_{t-1}]_i} + \epsilon} - \left(1 - \gamma^{(1)}_t\right)\nabla\mathcal{J}(\boldsymbol{x}_t)^{\mathsf{T}}\frac{\mathbb{E}_t\left[\nabla\mathcal{J}(\boldsymbol{x}_t) - \overline{\nabla g_{t}(\boldsymbol{x}_{t})}^{\mathsf{T}}\overline{\nabla f_{t}(\boldsymbol{y}_{t})} \right]}{\sqrt{\gamma^{(2)}_t\boldsymbol{v}_{t-1}}+ \epsilon} = \\\nonumber
    &\gamma^{(1)}_t\nabla\mathcal{J}(\boldsymbol{x}_t)^{\mathsf{T}}\frac{\boldsymbol{m}_{t-1}}{\sqrt{\gamma^{(2)}_t\boldsymbol{v}_{t-1}}+ \epsilon} + \left(1 - \gamma^{(1)}_t\right)\sum_{i=1}^{p}\frac{[\nabla\mathcal{J}(\boldsymbol{x}_{t})]^2_i}{\sqrt{\gamma^{(2)}_t[\boldsymbol{v}_{t-1}]_i} + \epsilon} - \\\nonumber
    &\left(1 - \gamma^{(1)}_t\right)\underbrace{\frac{\mathbb{E}_t\left[\nabla\mathcal{J}(\boldsymbol{x}_t)^{\mathsf{T}}\left(\nabla\mathcal{J}(\boldsymbol{x}_t) - \overline{\nabla g_{t}(\boldsymbol{x}_{t})}^{\mathsf{T}}\overline{\nabla f_{t}(\boldsymbol{y}_{t})} \right)\right]}{\sqrt{\gamma^{(2)}_t\boldsymbol{v}_{t-1}}+ \epsilon}}_{\mathcal{A}1}
\end{align*}
Using Assumption \ref{assum_2}.2 we have:
\begin{align*}
    &\mathcal{A}1 = \frac{\mathbb{E}_t\left[\nabla\mathcal{J}(\boldsymbol{x}_t)^{\mathsf{T}}\left(\nabla\mathcal{J}(\boldsymbol{x}_t) - \overline{\nabla g_{t}(\boldsymbol{x}_{t})}^{\mathsf{T}}\overline{\nabla f_t(g(\boldsymbol{x}_t))} \right)\right]}{\sqrt{\gamma^{(2)}_t\boldsymbol{v}_{t-1}}+ \epsilon} + \frac{\mathbb{E}_t\left[\nabla\mathcal{J}(\boldsymbol{x}_t)^{\mathsf{T}}\left(\overline{\nabla g_{t}(\boldsymbol{x}_{t})}^{\mathsf{T}}\overline{\nabla f_t(g(\boldsymbol{x}_t))} - \overline{\nabla g_{t}(\boldsymbol{x}_{t})}^{\mathsf{T}}\overline{\nabla f_{t}(\boldsymbol{y}_{t})} \right)\right]}{\sqrt{\gamma^{(2)}_t\boldsymbol{v}_{t-1}}+ \epsilon} = \\\nonumber
    &\frac{\mathbb{E}_t\left[\nabla\mathcal{J}(\boldsymbol{x}_t)^{\mathsf{T}}\left(\overline{\nabla g_{t}(\boldsymbol{x}_{t})}^{\mathsf{T}}\overline{\nabla f_t(g(\boldsymbol{x}_t))} - \overline{\nabla g_{t}(\boldsymbol{x}_{t})}^{\mathsf{T}}\overline{\nabla f_{t}(\boldsymbol{y}_{t})} \right)\right]}{\sqrt{\gamma^{(2)}_t\boldsymbol{v}_{t-1}}+ \epsilon} = \frac{\mathbb{E}_t\left[\nabla\mathcal{J}(\boldsymbol{x}_t)^{\mathsf{T}}\overline{\nabla g_{t}(\boldsymbol{x}_{t})}^{\mathsf{T}}\left(\overline{\nabla f_t(g(\boldsymbol{x}_t))} - \overline{\nabla f_{t}(\boldsymbol{y}_{t})} \right)\right]}{\sqrt{\gamma^{(2)}_t\boldsymbol{v}_{t-1}}+ \epsilon} = \\\nonumber
    &\mathbb{E}_t\left[\sum_{i=1}^p\frac{[\nabla\mathcal{J}(\boldsymbol{x}_t)]_i}{\sqrt{\sqrt{\gamma^{(2)}_t[\boldsymbol{v}_{t-1}]_i} + \epsilon}}\frac{\left[\overline{\nabla g_{t}(\boldsymbol{x}_{t})}^{\mathsf{T}}\overline{\nabla f_t(g(\boldsymbol{x}_t))} - \overline{\nabla g_{t}(\boldsymbol{x}_{t})}^{\mathsf{T}}\overline{\nabla f_{t}(\boldsymbol{y}_{t})}\right]_i}{\sqrt{\sqrt{\gamma^{(2)}_t[\boldsymbol{v}_{t-1}]_i} + \epsilon}}\right] \le \frac{1}{2}\mathbb{E}_t\left[\sum_{i=1}^p\frac{[\nabla\mathcal{J}(\boldsymbol{x}_t)]^2_i}{\sqrt{\gamma^{(2)}_t[\boldsymbol{v}_{t-1}]_i} + \epsilon}\right] + \\\nonumber
    &\frac{1}{2}\mathbb{E}_t\left[\sum_{i=1}^p\frac{\left[\overline{\nabla g_{t}(\boldsymbol{x}_{t})}^{\mathsf{T}}\overline{\nabla f_t(g(\boldsymbol{x}_t))} - \overline{\nabla g_{t}(\boldsymbol{x}_{t})}^{\mathsf{T}}\overline{\nabla f_{t}(\boldsymbol{y}_{t})}\right]^2_i}{\sqrt{\gamma^{(2)}_t[\boldsymbol{v}_{t-1}]_i} + \epsilon}\right] \le \frac{1}{2}\mathbb{E}_t\left[\sum_{i=1}^p\frac{[\nabla\mathcal{J}(\boldsymbol{x}_t)]^2_i}{\sqrt{\gamma^{(2)}_t[\boldsymbol{v}_{t-1}]_i} + \epsilon}\right] + \\\nonumber
    &\frac{1}{2\epsilon}\mathbb{E}_{t}\left[\left|\left|\overline{\nabla g_{t}(\boldsymbol{x}_{t})}^{\mathsf{T}}\overline{\nabla f_t(g(\boldsymbol{x}_t))} - \overline{\nabla g_{t}(\boldsymbol{x}_{t})}^{\mathsf{T}}\overline{\nabla f_{t}(\boldsymbol{y}_{t})} \right|\right|^2_2\right] \le \frac{1}{2}\mathbb{E}_t\left[\sum_{i=1}^p\frac{[\nabla\mathcal{J}(\boldsymbol{x}_t)]^2_i}{\sqrt{\gamma^{(2)}_t[\boldsymbol{v}_{t-1}]_i} + \epsilon}\right] + \\\nonumber &\frac{1}{2\epsilon}\mathbb{E}_t\left[\left|\left|\overline{\nabla g_{t}(\boldsymbol{x}_{t})}^{\mathsf{T}} \right|\right|^2_2\left|\left|\overline{\nabla f_t(g(\boldsymbol{x}_t))} - \overline{\nabla f_{t}(\boldsymbol{y}_{t})} \right|\right|^2_2\right] \le \frac{1}{2}\mathbb{E}_t\left[\sum_{i=1}^p\frac{[\nabla\mathcal{J}(\boldsymbol{x}_t)]^2_i}{\sqrt{\gamma^{(2)}_t[\boldsymbol{v}_{t-1}]_i} + \epsilon}\right] + \\\nonumber
    &\frac{1}{2\epsilon}M^2_g\mathbb{E}_t\left[\left|\left|\overline{\nabla f_t(g(\boldsymbol{x}_t))} - \overline{\nabla f_t(\boldsymbol{y}_t)} \right|\right|^2_2\right] \le
    \frac{1}{2}\mathbb{E}_t\left[\sum_{i=1}^p\frac{[\nabla\mathcal{J}(\boldsymbol{x}_t)]^2_i}{\sqrt{\gamma^{(2)}_t[\boldsymbol{v}_{t-1}]_i} + \epsilon}\right] + \\\nonumber &\frac{1}{2\epsilon}M^2_g\mathbb{E}_t\left[\left|\left|\overline{\nabla f_t(g(\boldsymbol{x}_t))} - \overline{\nabla f_{t}(\boldsymbol{y}_{t})} \right|\right|^2_2\right] \le  \frac{1}{2}\mathbb{E}_t\left[\sum_{i=1}^p\frac{[\nabla\mathcal{J}(\boldsymbol{x}_t)]^2_i}{\sqrt{\gamma^{(2)}_t[\boldsymbol{v}_{t-1}]_i} + \epsilon}\right] + \\\nonumber &\frac{1}{2K^{(1)}_t\epsilon}M^2_g\sum_{a=1}^{K^{(1)}_t}\mathbb{E}_t\left[\left|\left| \nabla f_{t_a}(g(\boldsymbol{x}_t)) - \nabla f_{t_a}(\boldsymbol{y}_t)\right|\right|^2_2\right] \le \frac{1}{2}\mathbb{E}_t\left[\sum_{i=1}^p\frac{[\nabla\mathcal{J}(\boldsymbol{x}_t)]^2_i}{\sqrt{\gamma^{(2)}_t[\boldsymbol{v}_{t-1}]_i} + \epsilon}\right] + \\\nonumber &\frac{1}{2\epsilon}M^2_gL^2_f\mathbb{E}_t\left[\left|\left|g(\boldsymbol{x}_t) - \boldsymbol{y}_t \right|\right|^2_2\right] = \frac{1}{2}\sum_{i=1}^p\frac{[\nabla\mathcal{J}(\boldsymbol{x}_t)]^2_i}{\sqrt{\gamma^{(2)}_t[\boldsymbol{v}_{t-1}]_i} + \epsilon}
    + \frac{1}{2\epsilon}M^2_gL^2_f\left|\left|g(\boldsymbol{x}_t) - \boldsymbol{y}_t \right|\right|^2_2
\end{align*}
Hence, we arrive at the following expression for $\mathcal{A}$:
\begin{align}\label{mathcalAexpression}
    &-\mathcal{A} = -\gamma^{(1)}_t\nabla\mathcal{J}(\boldsymbol{x}_t)^{\mathsf{T}}\frac{\boldsymbol{m}_{t-1}}{\sqrt{\gamma^{(2)}_t\boldsymbol{v}_{t-1}}+ \epsilon} - \left(1 - \gamma^{(1)}_t\right)\sum_{i=1}^{p}\frac{[\nabla\mathcal{J}(\boldsymbol{x}_{t})]^2_i}{\sqrt{\gamma^{(2)}_t[\boldsymbol{v}_{t-1}]_i} + \epsilon} + \left(1 - \gamma^{(1)}_t\right)\mathcal{A}1 \le \\\nonumber
    &-\gamma^{(1)}_t\nabla\mathcal{J}(\boldsymbol{x}_t)^{\mathsf{T}}\frac{\boldsymbol{m}_{t-1}}{\sqrt{\gamma^{(2)}_t\boldsymbol{v}_{t-1}}+ \epsilon} - \left(1 - \gamma^{(1)}_t\right)\sum_{i=1}^{p}\frac{[\nabla\mathcal{J}(\boldsymbol{x}_{t})]^2_i}{\sqrt{\gamma^{(2)}_t[\boldsymbol{v}_{t-1}]_i} + \epsilon} + \\\nonumber
    &\left(1 - \gamma^{(1)}_t\right)\left[\frac{1}{2}\sum_{i=1}^p\frac{[\nabla\mathcal{J}(\boldsymbol{x}_t)]^2_i}{\sqrt{\gamma^{(2)}_t[\boldsymbol{v}_{t-1}]_i} + \epsilon}
    + \frac{1}{2\epsilon}M^2_gL^2_f\left|\left|g(\boldsymbol{x}_t) - \boldsymbol{y}_t \right|\right|^2_2\right] =\\\nonumber
    &-\gamma^{(1)}_t\nabla\mathcal{J}(\boldsymbol{x}_t)^{\mathsf{T}}\frac{\boldsymbol{m}_{t-1}}{\sqrt{\gamma^{(2)}_t\boldsymbol{v}_{t-1}}+ \epsilon} - \frac{\left(1 - \gamma^{(1)}_t\right)}{2}\sum_{i=1}^{p}\frac{[\nabla\mathcal{J}(\boldsymbol{x}_{t})]^2_i}{\sqrt{\gamma^{(2)}_t[\boldsymbol{v}_{t-1}]_i} + \epsilon}  
    + \frac{\left(1 - \gamma^{(1)}_t\right)}{2\epsilon}M^2_gL^2_f\left|\left|g(\boldsymbol{x}_t) - \boldsymbol{y}_t \right|\right|^2_2
\end{align}
Now, let us consider more carefully the second term:
\begin{align*}
    &-\mathcal{B} = -\sum_{i=1}^p\left[\nabla\mathcal{J}(\boldsymbol{x}_t)\right]_i\mathbb{E}_t\left[\frac{[\boldsymbol{m}_t]_i}{\sqrt{[\boldsymbol{v}_{t}]_i} + \epsilon} - \frac{[\boldsymbol{m}_t]_i}{\sqrt{\gamma^{(2)}_t[\boldsymbol{v}_{t-1}]_i} + \epsilon}\right] \le \sum_{i=1}^p\left|\left[\nabla\mathcal{J}(\boldsymbol{x}_t)\right]_i\right|\left|\mathbb{E}_t\left[\underbrace{\frac{[\boldsymbol{m}_t]_i}{\sqrt{[\boldsymbol{v}_{t}]_i} + \epsilon} - \frac{[\boldsymbol{m}_t]_i}{\sqrt{\gamma^{(2)}_t[\boldsymbol{v}_{t-1}]_i} + \epsilon}}_{\mathcal{B}1}\right]\right|
\end{align*}
For the expression $\mathcal{B}1$ we have:
\begin{align*}
    &\mathcal{B}1 = \frac{[\boldsymbol{m}_t]_i}{\sqrt{[\boldsymbol{v}_{t}]_i} + \epsilon} - \frac{[\boldsymbol{m}_t]_i}{\sqrt{\gamma^{(2)}_t[\boldsymbol{v}_{t-1}]_i} + \epsilon} \le |[\boldsymbol{m}_t]_i|\times\left|\frac{1}{\sqrt{[\boldsymbol{v}_t]_i} + \epsilon} - \frac{1}{\sqrt{\gamma^{(2)}_t[\boldsymbol{v}_{t-1}]_i} + \epsilon}\right| = \\\nonumber
    &\frac{|[\boldsymbol{m}_t]_i|}{(\sqrt{[\boldsymbol{v}_t]_i}+ \epsilon)(\sqrt{\gamma^{(2)}_{t}[\boldsymbol{v}_{t-1}]_i} + \epsilon)}\times\left|\frac{(1-\gamma^{(2)}_t)[\overline{\nabla \mathcal{J}(\boldsymbol{x}_{t})}]^2_i}{\sqrt{[\boldsymbol{v}_{t}]_i} + \sqrt{\gamma^{(2)}_t[\boldsymbol{v}_{t-1}]_i}}\right| = \frac{(1-\gamma^{(2)}_t)|[\boldsymbol{m}_t]_i|}{(\sqrt{[\boldsymbol{v}_t]_i}+ \epsilon)(\sqrt{\gamma^{(2)}_{t}[\boldsymbol{v}_{t-1}]_i} + \epsilon)}\times\\\nonumber
    &\frac{[\overline{\nabla \mathcal{J}(\boldsymbol{x}_{t})}]^2_i}{\sqrt{\gamma^{(2)}_{t}[\boldsymbol{v}_{t-1}]_i + (1 - \gamma^{(2)}_{t})[\tilde{\nabla}\mathcal{J}(\boldsymbol{x}_t)]^2_i} + \sqrt{\gamma^{(2)}_t[\boldsymbol{v}_{t-1}]_i}} \le \frac{\sqrt{1 - \gamma^{(2)}_t}|[\boldsymbol{m}_t]_i||[\overline{\nabla \mathcal{J}(\boldsymbol{x}_{t})}]_i|}{\epsilon\left(\sqrt{\gamma^{(2)}_t[\boldsymbol{v}_{t-1}]_i} +\epsilon\right)} \le \\\nonumber
    &\sqrt{1 - \gamma^{(2)}_t}\frac{|\gamma^{(1)}_t[\boldsymbol{m}_{t-1}]_i| + (1 - \gamma^{(1)}_t)|[\overline{\nabla \mathcal{J}(\boldsymbol{x}_{t})}]_i|}{\epsilon\left(\sqrt{\gamma^{(2)}_t[\boldsymbol{v}_{t-1}]_i} +\epsilon\right)}|[\overline{\nabla \mathcal{J}(\boldsymbol{x}_{t})}]_i| = \frac{(1 - \gamma^{(1)}_t)\sqrt{1 - \gamma^{(2)}_t}}{\epsilon\left(\sqrt{\gamma^{(2)}_t[\boldsymbol{v}_{t-1}]_i} +\epsilon\right)}[\overline{\nabla \mathcal{J}(\boldsymbol{x}_{t})}]^2_i + \\\nonumber 
    &\frac{\gamma^{(1)}_t\sqrt{1 - \gamma^{(2)}_t}}{\epsilon\left(\sqrt{\gamma^{(2)}_t[\boldsymbol{v}_{t-1}]_i} +\epsilon\right)}|[\overline{\nabla \mathcal{J}(\boldsymbol{x}_{t})}]_i||[\boldsymbol{m}_{t-1}]_i| \le 2\frac{(1 - \gamma^{(1)}_t)\sqrt{1 - \gamma^{(2)}_t}}{\epsilon\left(\sqrt{\gamma^{(2)}_t[\boldsymbol{v}_{t-1}]_i} +\epsilon\right)}[\overline{\nabla \mathcal{J}(\boldsymbol{x}_{t})}]^2_i + \\\nonumber
    &\frac{\left(\gamma^{(1)}_t\right)^2\sqrt{1 - \gamma^{(2)}_t}}{\epsilon(1 - \gamma^{(1)}_t)\left(\sqrt{\gamma^{(2)}_t[\boldsymbol{v}_{t-1}]_i} +\epsilon\right)}[\boldsymbol{m}_{t-1}]^2_i
\end{align*}
where in the last step we use $|ab| \le \alpha a^2 + \frac{1}{\alpha}b^2$ for $\alpha = \frac{1 - \gamma^{(1)}_t}{\gamma^{(1)}_t}$. Therefore, for the term $\mathcal{B}$ we have:
\begin{align}
    &-\mathcal{B} \le \sum_{i=1}^p|[\nabla \mathcal{J}(\boldsymbol{x}_t)]_i|\mathbb{E}_t\left[2\frac{(1 - \gamma^{(1)}_t)\sqrt{1 - \gamma^{(2)}_t}}{\epsilon\left(\sqrt{\gamma^{(2)}_t[\boldsymbol{v}_{t-1}]_i} +\epsilon\right)}[\tilde{\nabla}\mathcal{J}(\boldsymbol{x}_t)]^2_i + \frac{\left(\gamma^{(1)}_t\right)^2\sqrt{1 - \gamma^{(2)}_t}}{\epsilon(1 - \gamma^{(1)}_t)\left(\sqrt{\gamma^{(2)}_t[\boldsymbol{v}_{t-1}]_i} +\epsilon\right)}[\boldsymbol{m}_{t-1}]^2_i\right] = \\\nonumber
    &\frac{\left(\gamma^{(1)}_t\right)^2\sqrt{1 - \gamma^{(2)}_t}}{\epsilon(1 - \gamma^{(1)}_t)}\sum_{i=1}^{p}\frac{|[\nabla\mathcal{J}(\boldsymbol{x}_t)]_i|[\boldsymbol{m}_{t-1}]^2_i}{\left(\sqrt{\gamma^{(2)}_t[\boldsymbol{v}_{t-1}]_i} + +\epsilon\right)} + \frac{(1 - \gamma^{(1)}_t)\sqrt{1 - \gamma^{(2)}_t}}{\epsilon}\sum_{i=1}^p\frac{|[\nabla\mathcal{J}(\boldsymbol{x}_t)]_i|}{\left(\sqrt{\gamma^{(2)}_t[\boldsymbol{v}_{t-1}]_i} +\epsilon\right)}\mathbb{E}_t\left[[\overline{\nabla \mathcal{J}(\boldsymbol{x}_{t})}]^2_i\right] 
\end{align}

Using that $[\boldsymbol{v}_{t}]_i \ge 0$ for any $t$ and applying (\ref{bound_expressions}) we get immediately:
\begin{align*}
    &-\mathcal{B}\le \frac{\left(\gamma^{(1)}_t\right)^2\sqrt{1 - \gamma^{(2)}_t}}{\epsilon(1 - \gamma^{(1)}_t)}\sum_{i=1}^{p}\frac{|[\nabla\mathcal{J}(\boldsymbol{x}_t)]_i|[\boldsymbol{m}_{t-1}]^2_i}{\left(\sqrt{\gamma^{(2)}_t[\boldsymbol{v}_{t-1}]_i} + \epsilon\right)} + \frac{(1 - \gamma^{(1)}_t)\sqrt{1 - \gamma^{(2)}_t}}{\epsilon}\sum_{i=1}^p\frac{|[\nabla\mathcal{J}(\boldsymbol{x}_t)]_i|}{\left(\sqrt{\gamma^{(2)}_t[\boldsymbol{v}_{t-1}]_i} +\epsilon\right)}\mathbb{E}_t\left[[\overline{\nabla \mathcal{J}(\boldsymbol{x}_{t})}]^2_i\right] \le \\\nonumber
    &\frac{M_gM_f\left(\gamma^{(1)}_t\right)^2\sqrt{1 - \gamma^{(2)}_t}}{\epsilon (1 - \gamma^{(1)}_t)}\sum_{i=1}^{p}\frac{|[\nabla\mathcal{J}(\boldsymbol{x}_t)]_i[\boldsymbol{m}_{t-1}]_i|}{\sqrt{\gamma^{(2)}_{t}[\boldsymbol{v}_{t-1}]_i} + \epsilon} + \frac{M_gM_f(1 - \gamma^{(1)}_t)\sqrt{1 - \gamma^{(2)}_t}}{\epsilon}\sum_{i=1}^p\mathbb{E}_t\left[\frac{[\overline{\nabla \mathcal{J}(\boldsymbol{x}_{t})}]^2_i}{\sqrt{\gamma^{(2)}_{t}[\boldsymbol{v}_{t-1}]_i} + \epsilon}\right]
\end{align*}
Finally, we have the bound for the second term in the expression (\ref{Secon_expectation}):
\begin{align}\label{second_term_expression}
    &-\sum_{i=1}^p\left[\nabla\mathcal{J}(\boldsymbol{x}_t)\right]_i\mathbb{E}_t\left[\frac{[\boldsymbol{m}_t]_i}{\sqrt{[\boldsymbol{v}_t]_i} + \epsilon}\right] =  -\mathcal{A} - \mathcal{B}\le\\\nonumber
    &-\gamma^{(1)}_t\nabla\mathcal{J}(\boldsymbol{x}_t)^{\mathsf{T}}\frac{\boldsymbol{m}_{t-1}}{\sqrt{\gamma^{(2)}_t\boldsymbol{v}_{t-1}}+ \epsilon} - \frac{\left(1 - \gamma^{(1)}_t\right)}{2}\sum_{i=1}^{p}\frac{[\nabla\mathcal{J}(\boldsymbol{x}_{t})]^2_i}{\sqrt{\gamma^{(2)}_t[\boldsymbol{v}_{t-1}]_i} + \epsilon}  
    + \frac{\left(1 - \gamma^{(1)}_t\right)}{2\epsilon}M^2_gL^2_f\left|\left|g(\boldsymbol{x}_t) - \boldsymbol{y}_t \right|\right|^2_2 + \\\nonumber
    &\frac{M_gM_f\left(\gamma^{(1)}_t\right)^2\sqrt{1 - \gamma^{(2)}_t}}{\epsilon (1 - \gamma^{(1)}_t)}\sum_{i=1}^{p}\frac{|[\nabla\mathcal{J}(\boldsymbol{x}_t)]_i[\boldsymbol{m}_{t-1}]_i|}{\sqrt{\gamma^{(2)}_{t}[\boldsymbol{v}_{t-1}]_i} + \epsilon} + \frac{M_gM_f(1 - \gamma^{(1)}_t)\sqrt{1 - \gamma^{(2)}_t}}{\epsilon}\sum_{i=1}^p\mathbb{E}_t\left[\frac{[\overline{\nabla \mathcal{J}(\boldsymbol{x}_{t})}]^2_i}{\sqrt{\gamma^{(2)}_{t}[\boldsymbol{v}_{t-1}]_i} + \epsilon}\right] 
\end{align}
Now, we can focus on the third term in the expression (\ref{Secon_expectation}). Applying that $[\boldsymbol{v}_t]_i \ge \gamma^{(2)}_{t}[\boldsymbol{v}_{t-1}]_i$ we have:
\begin{align*}
    &\sum_{i=1}^p\mathbb{E}_t\left[\frac{[\boldsymbol{m}_t]^2_i}{\left(\sqrt{[\boldsymbol{v}_t]_i} + \epsilon\right)^2}\right] \le \sum_{i=1}^p\mathbb{E}_t\left[\frac{[\boldsymbol{m}_t]^2_i}{\left(\sqrt{\gamma^{(2)}_t[\boldsymbol{v}_{t-1}]_i} + \epsilon\right)^2}\right] = \sum_{i=1}^p\frac{\mathbb{E}_t\left[[\boldsymbol{m}_t]^2_i\right]}{\left(\sqrt{\gamma^{(2)}_t[\boldsymbol{v}_{t-1}]_i} + \epsilon\right)^2} = \\\nonumber
    &\sum_{i=1}^p\frac{\mathbb{E}_t\left[[\gamma^{(1)}_t\boldsymbol{m}_{t-1} + \left(1 - \gamma^{(1)}_t\right)\overline{\nabla \mathcal{J}(\boldsymbol{x}_{t})}]^2_i\right]}{\left(\sqrt{\gamma^{(2)}_t[\boldsymbol{v}_{t-1}]_i} + \epsilon\right)^2} \le \sum_{i=1}^p\frac{2\left(\gamma^{(1)}_t\right)^2\mathbb{E}_t\left[[\boldsymbol{m}_{t-1}]^2_i\right] + 2\left(1 - \gamma^{(1)}_t\right)^2\mathbb{E}_t\left[[\overline{\nabla \mathcal{J}(\boldsymbol{x}_{t})}]^2_i\right]}{\left(\sqrt{\gamma^{(2)}_t[\boldsymbol{v}_{t-1}]_i} + \epsilon\right)^2} = \\\nonumber
    &2\left(\gamma^{(1)}_t\right)^2\sum_{i=1}^{p}\frac{[\boldsymbol{m}_{t-1}]^2_i}{\left(\sqrt{\gamma^{(2)}_t[\boldsymbol{v}_{t-1}]_i} + \epsilon\right)^2} + 2\left(1 - \gamma^{(1)}_t\right)^2\sum_{i=1}^p\frac{\mathbb{E}_t\left[[\overline{\nabla \mathcal{J}(\boldsymbol{x}_{t})}]^2_i\right]}{\left(\sqrt{\gamma^{(2)}_t[\boldsymbol{v}_{t-1}]_i} + \epsilon\right)^2} \le \\\nonumber
    &2\left(\gamma^{(1)}_t\right)^2\sum_{i=1}^{p}\frac{[\boldsymbol{m}_{t-1}]^2_i}{\left(\sqrt{\gamma^{(2)}_t[\boldsymbol{v}_{t-1}]_i} + \epsilon\right)^2} + 2\frac{\left(1 - \gamma^{(1)}_t\right)^2}{\epsilon}\sum_{i=1}^p\frac{\mathbb{E}_t\left[[\overline{\nabla \mathcal{J}(\boldsymbol{x}_{t})}]^2_i\right]}{\sqrt{\gamma^{(2)}_t[\boldsymbol{v}_{t-1}]_i} + \epsilon}\le\\\nonumber
    &\frac{2\left(\gamma^{(1)}_t\right)^2}{\epsilon}\sum_{i=1}^{p}\frac{[\boldsymbol{m}_{t-1}]^2_i}{\sqrt{\gamma^{(2)}_t[\boldsymbol{v}_{t-1}]_i} + \epsilon} + 2\frac{\left(1 - \gamma^{(1)}_t\right)^2}{\epsilon}\sum_{i=1}^p\frac{\mathbb{E}_t\left[[\overline{\nabla \mathcal{J}(\boldsymbol{x}_{t})}]^2_i\right]}{\sqrt{\gamma^{(2)}_t[\boldsymbol{v}_{t-1}]_i} + \epsilon}
\end{align*}
Hence, we arrive at the following expression for change in the function between two consecutive interaction:
\begin{align}\label{new_change_function}
    &\mathbb{E}_t\left[\mathcal{J}(\boldsymbol{x}_{t+1})\right] - \mathcal{J}(\boldsymbol{x}_t) \le \\\nonumber
    &-\alpha_t\gamma^{(1)}_t\nabla\mathcal{J}(\boldsymbol{x}_t)^{\mathsf{T}}\frac{\boldsymbol{m}_{t-1}}{\sqrt{\gamma^{(2)}_t\boldsymbol{v}_{t-1}}+ \epsilon} - \alpha_t\frac{\left(1 - \gamma^{(1)}_t\right)}{2}\sum_{i=1}^{p}\frac{[\nabla\mathcal{J}(\boldsymbol{x}_{t})]^2_i}{\sqrt{\gamma^{(2)}_t[\boldsymbol{v}_{t-1}]_i} + \epsilon}  
    + \alpha_t\frac{\left(1 - \gamma^{(1)}_t\right)}{2\epsilon}M^2_gL^2_f\left|\left|g(\boldsymbol{x}_t) - \boldsymbol{y}_t \right|\right|^2_2 +\\\nonumber
    &\frac{\alpha_tM_gM_f\left(\gamma^{(1)}_t\right)^2\sqrt{1 - \gamma^{(2)}_t}}{\epsilon (1 - \gamma^{(1)}_t)}\sum_{i=1}^{p}\frac{|[\nabla\mathcal{J}(\boldsymbol{x}_t)]_i[\boldsymbol{m}_{t-1}]_i|}{\sqrt{\gamma^{(2)}_{t}[\boldsymbol{v}_{t-1}]_i} + \epsilon} + \frac{\alpha_tM_gM_f(1 - \gamma^{(1)}_t)\sqrt{1 - \gamma^{(2)}_t}}{\epsilon}\sum_{i=1}^p\frac{\mathbb{E}_t\left[[\overline{\nabla \mathcal{J}(\boldsymbol{x}_{t})}]^2_i\right]}{\sqrt{\gamma^{(2)}_{t}[\boldsymbol{v}_{t-1}]_i} + \epsilon} + \\\nonumber
    &\frac{L\alpha_t^2\left(\gamma^{(1)}_t\right)^2}{\epsilon}\sum_{i=1}^{p}\frac{[\boldsymbol{m}_{t-1}]^2_i}{\sqrt{\gamma^{(2)}_t[\boldsymbol{v}_{t-1}]_i} + \epsilon} + \frac{L\alpha_t^2\left(1 - \gamma^{(1)}_t\right)^2}{\epsilon}\sum_{i=1}^p\frac{\mathbb{E}_t\left[[\overline{\nabla \mathcal{J}(\boldsymbol{x}_{t})}]^2_i\right]}{\sqrt{\gamma^{(2)}_t[\boldsymbol{v}_{t-1}]_i} + \epsilon} = 
\end{align}
\begin{align*}
    &-\frac{\alpha_t\left(1 - \gamma^{(1)}_t\right)}{2}\sum_{i=1}^{p}\frac{[\nabla\mathcal{J}(\boldsymbol{x}_{t})]^2_i}{\sqrt{\gamma^{(2)}_t[\boldsymbol{v}_{t-1}]_i} + \epsilon} + \frac{\alpha_t\left(1 - \gamma^{(1)}_t\right)}{2\epsilon}M^2_gL^2_f\left|\left|g(\boldsymbol{x}_t) - \boldsymbol{y}_t \right|\right|^2_2 + \sum_{i=1}^p\frac{\mathbb{E}_t\left[[\overline{\nabla \mathcal{J}(\boldsymbol{x}_{t})}]^2_i\right]}{\sqrt{\gamma^{(2)}_t[\boldsymbol{v}_{t-1}]_i} + \epsilon}\times\\\nonumber
    &\left[\frac{L\alpha^2_t\left(1 - \gamma^{(1)}_t\right)^2}{\epsilon} + \frac{\alpha_tM_gM_f(1 - \gamma^{(1)}_t)\sqrt{1 - \gamma^{(2)}_t}}{\epsilon}\right]  -\alpha_t\gamma^{(1)}_t\nabla\mathcal{J}(\boldsymbol{x}_t)^{\mathsf{T}}\frac{\boldsymbol{m}_{t-1}}{\sqrt{\gamma^{(2)}_t\boldsymbol{v}_{t-1}}+ \epsilon} + \\\nonumber &\frac{L\alpha^2_t\left(\gamma^{(1)}_t\right)^2}{\epsilon}\sum_{i=1}^{p}\frac{[\boldsymbol{m}_{t-1}]^2_i}{\sqrt{\gamma^{(2)}_t[\boldsymbol{v}_{t-1}]_i} + \epsilon} + \frac{\alpha_tM_gM_f\left(\gamma^{(1)}_t\right)^2\sqrt{1 - \gamma^{(2)}_t}}{\epsilon (1 - \gamma^{(1)}_t)}\sum_{i=1}^{p}\frac{|[\nabla\mathcal{J}(\boldsymbol{x}_t)]_i[\boldsymbol{m}_{t-1}]_i|}{\sqrt{\gamma^{(2)}_{t}[\boldsymbol{v}_{t-1}]_i} + \epsilon}
\end{align*}
Please notice, from Assumption \ref{assum_2}.3 we immediately have the following bounds on the variances of $\overline{\nabla g_{t}(\boldsymbol{x})}$ and $\overline{\nabla f_t(\boldsymbol{y})}$:
\begin{align}\label{var_bounds}
    &\mathbb{E}_{t}\left[\left|\left|\overline{\nabla g_t(\boldsymbol{x})} - \nabla g(\boldsymbol{x})\right|\right|^2_2\right] \le \frac{1}{K^{(1)}_t}\sigma^2_2, \ \ \ \   \mathbb{E}_{t}\left[\left|\left|\overline{\nabla f_t(\boldsymbol{y})} - \nabla f(\boldsymbol{y})\right|\right|^2_2\right] \le \frac{1}{K^{(2)}_t}\sigma^2_1 
\end{align}
Now, let us focus on the term $\mathbb{E}_t\left[[\overline{\nabla \mathcal{J}(\boldsymbol{x}_{t})}]^2_i\right]$. Using Assumption \ref{assum_2}.2 we have:
\begin{align}\label{variance_est_1}
    &\mathbb{E}_t\left[\left([\overline{\nabla \mathcal{J}(\boldsymbol{x}_{t})}]_i - \left[\nabla g(\boldsymbol{x}_t)^{\mathsf{T}}\nabla f(\boldsymbol{y}_t)\right]_i\right)^2\right] = \mathbb{E}_t\left[[\overline{\nabla \mathcal{J}(\boldsymbol{x}_{t})}]^2_i\right] - \left[\nabla g(\boldsymbol{x}_t)^{\mathsf{T}}\nabla f(\boldsymbol{y}_t)\right]^2_i
\end{align}
From the other hand, using the properties of matrix $||\cdot||_2$\footnote{Particularly, we use that for any matrix $\boldsymbol{A}\in\mathbb{R}^{p\times q}$: $||\boldsymbol{A}||^2_2 = \sup_{\boldsymbol{v}: ||\boldsymbol{v}|| = 1 }\boldsymbol{v}^{\mathsf{T}}\boldsymbol{A}\boldsymbol{A}^{\mathsf{T}}\boldsymbol{v} \ge \sum_{r=1}^q[\boldsymbol{A}]^2_{ir}$ for any $i = 1,\ldots,p$. To see this, just take $\boldsymbol{v} = \boldsymbol{e}_i$} and Assumptions \ref{assum_1}.1, \ref{assum_1}.3 and \ref{assum_2}.3:
\begin{align*}
    &\mathbb{E}_t\left[\left([\overline{\nabla \mathcal{J}(\boldsymbol{x}_{t})}]_i - \left[\nabla g(\boldsymbol{x}_t)^{\mathsf{T}}\nabla f(\boldsymbol{y}_t)\right]_i\right)^2\right] = \mathbb{E}_t\left[\left([\overline{\nabla g_{t}(\boldsymbol{x}_{t})}^{\mathsf{T}}\overline{\nabla f_{t}(\boldsymbol{y}_{t})}]_i - \left[\nabla g(\boldsymbol{x}_t)^{\mathsf{T}}\nabla f(\boldsymbol{y}_t)\right]_i\right)^2\right] = \\\nonumber
    &\mathbb{E}_t\left[\left[\overline{\nabla g_{t}(\boldsymbol{x}_{t})}^{\mathsf{T}}\overline{\nabla f_{t}(\boldsymbol{y}_{t})} - \nabla g(\boldsymbol{x}_t)^{\mathsf{T}}\overline{\nabla f_{t}(\boldsymbol{y}_{t})} + \nabla g(\boldsymbol{x}_t)^{\mathsf{T}}\overline{\nabla f_{t}(\boldsymbol{y}_{t})} - \nabla g(\boldsymbol{x}_t)^{\mathsf{T}}\nabla f(\boldsymbol{y}_t)\right]^2_i\right] \le \\\nonumber
    &2\mathbb{E}_t\left[\left[\left(\overline{\nabla g_{t}(\boldsymbol{x}_{t})}^{\mathsf{T}} - \nabla g(\boldsymbol{x}_t)^{\mathsf{T}}\right)\overline{\nabla f_{t}(\boldsymbol{y}_{t})}\right]^2_i\right] + 2\mathbb{E}_t\left[\left[\nabla g(\boldsymbol{x}_t)^{\mathsf{T}}\left(\overline{\nabla f_{t}(\boldsymbol{y}_{t})} - \nabla f(\boldsymbol{y}_t)\right)\right]^2_i\right] = \\\nonumber 
    &2\mathbb{E}_t\left[\left(\sum_{r=1}^q\left[\left(\overline{\nabla g_{t}(\boldsymbol{x}_{t})}^{\mathsf{T}} - \nabla g(\boldsymbol{x}_t)^{\mathsf{T}}\right)\right]_{ir}[\overline{\nabla f_{t}(\boldsymbol{y}_{t})}]_r\right)^2\right] + 2\mathbb{E}_t\left[\left(\sum_{r=1}^q\left[\nabla g(\boldsymbol{x}_t)^{\mathsf{T}}\right]_{ir}\left[\overline{\nabla f_{t}(\boldsymbol{y}_{t})} - \nabla f(\boldsymbol{y}_t)\right]_r\right)^2\right] \le\\\nonumber
    &2q\mathbb{E}_t\left[\sum_{r=1}^q\left[\left(\overline{\nabla g_{t}(\boldsymbol{x}_{t})}^{\mathsf{T}} - \nabla g(\boldsymbol{x}_t)^{\mathsf{T}}\right)\right]^2_{ir}[\overline{\nabla f_{t}(\boldsymbol{y}_{t})}]^2_r\right] + 2q\mathbb{E}_t\left[\sum_{r=1}^q\left[\nabla g(\boldsymbol{x}_t)^{\mathsf{T}}\right]^2_{ir}\left[\overline{\nabla f_{t}(\boldsymbol{y}_{t})} - \nabla f(\boldsymbol{y}_t)\right]^2_r\right] \le 
\end{align*}
\begin{align*}
    &2q\mathbb{E}_t\left[\left|\left|\overline{\nabla g_{t}(\boldsymbol{x}_{t})}^{\mathsf{T}} - \nabla g(\boldsymbol{x}_t)^{\mathsf{T}}\right|\right|^2_2\sum_{r=1}^q[\overline{\nabla f_{t}(\boldsymbol{y}_{t})}]^2_r\right] + 2q\mathbb{E}_t\left[\left|\left|\overline{\nabla f_{t}(\boldsymbol{y}_{t})} - \nabla f(\boldsymbol{y}_t)\right|\right|^2_2\sum_{r=1}^q\left[\nabla g(\boldsymbol{x}_t)^{\mathsf{T}}\right]^2_{ir}\right] \le \\\nonumber
    &2q\mathbb{E}_t\left[\left|\left|\overline{\nabla g_{t}(\boldsymbol{x}_{t})}^{\mathsf{T}} - \nabla g(\boldsymbol{x}_t)^{\mathsf{T}}\right|\right|^2_2\left|\left|\overline{\nabla f_{t}(\boldsymbol{y}_{t})}\right|\right|^2_2\right] + 2q\mathbb{E}_t\left[\left|\left|\overline{\nabla f_{t}(\boldsymbol{y}_{t})} - \nabla f(\boldsymbol{y}_t)\right|\right|^2_2\left|\left|g(\boldsymbol{x}_t)^{\mathsf{T}}\right|\right|^2_2\right] \le \\\nonumber
    &2qM^2_f\mathbb{E}_t\left[\left|\left|\overline{\nabla g_{t}(\boldsymbol{x}_{t})}^{\mathsf{T}} - \nabla g(\boldsymbol{x}_t)^{\mathsf{T}}\right|\right|^2_2\right] + 2qM^2_g\mathbb{E}_t\left[\left|\left|\overline{\nabla f_{t}(\boldsymbol{y}_{t})} - \nabla f(\boldsymbol{y}_t)\right|\right|^2_2\right] \le \\\nonumber
    &2qM^2_f\mathbb{E}_t\left[\left|\left|\overline{\nabla g_{t}(\boldsymbol{x}_{t})} - \nabla g(\boldsymbol{x}_t)\right|\right|^2_2\right] + 2qM^2_g\mathbb{E}_t\left[\left|\left|\overline{\nabla f_{t}(\boldsymbol{y}_{t})} - \nabla f(\boldsymbol{y}_t)\right|\right|^2_2\right] \le \\\nonumber
    &2qM^2_f\frac{1}{K^{(1)}_t}\sigma^2_2 + 2qM^2_g\frac{1}{K^{(2)}_t}\sigma^2_1
\end{align*}
where in the last step we use (\ref{var_bounds}). Combining this result with (\ref{variance_est_1}) gives:
\begin{equation*}
    \mathbb{E}_t\left[[\overline{\nabla \mathcal{J}(\boldsymbol{x}_{t})}]^2_i\right]  \le 2qM^2_f\frac{1}{K^{(1)}_t}\sigma^2_2 + 2qM^2_g\frac{1}{K^{(2)}_t}\sigma^2_1 + \left[\nabla g(\boldsymbol{x}_t)^{\mathsf{T}}\nabla f(\boldsymbol{y}_t)\right]^2_i
\end{equation*}
Moreover,
\begin{align*}
    &\left[\nabla g(\boldsymbol{x}_t)^{\mathsf{T}}\nabla f(\boldsymbol{y}_t)\right]^2_i = \left[\nabla g(\boldsymbol{x}_t)^{\mathsf{T}}\nabla f(\boldsymbol{y}_t) - \nabla g(\boldsymbol{x}_t)^{\mathsf{T}}\nabla f(g(\boldsymbol{x}_t)) + \nabla g(\boldsymbol{x}_t)^{\mathsf{T}}\nabla f(g(\boldsymbol{x}_t))\right]^2_i = \\\nonumber
    &\left[\nabla g(\boldsymbol{x}_t)^{\mathsf{T}}\left[\nabla f(\boldsymbol{y}_t) - \nabla f(g(\boldsymbol{x}_t))\right] + \nabla g(\boldsymbol{x}_t)^{\mathsf{T}}\nabla f(g(\boldsymbol{x}_t))\right]^2_i \le 2\left[\nabla g(\boldsymbol{x}_t)^{\mathsf{T}}\nabla f(g(\boldsymbol{x}_t))\right]^2_i + \\\nonumber
    &2\left[\nabla g(\boldsymbol{x}_t)^{\mathsf{T}}\left[\nabla f(\boldsymbol{y}_t) - \nabla f(g(\boldsymbol{x}_t))\right]\right]^2_i \le 2\left[\nabla \mathcal{J}(\boldsymbol{x}_t)\right]^2_i + 2\left|\left|\nabla g(\boldsymbol{x}_t)^{\mathsf{T}}\right|\right|^2_2\left|\left|\nabla f(\boldsymbol{y}_t) - \nabla f(g(\boldsymbol{x}_t))\right|\right|^2_2 \le\\\nonumber
    &2\left[\nabla \mathcal{J}(\boldsymbol{x}_t)\right]^2_i + 2M^2_gL^2_f\left|\left|\boldsymbol{y}_t - g(\boldsymbol{x}_t)\right|\right|^2_2
\end{align*}
Hence, we have
\begin{align}\label{addition_expression_1}
    &\sum_{i=1}^p\frac{\mathbb{E}_t\left[[\overline{\nabla \mathcal{J}(\boldsymbol{x}_{t})}]^2_i\right]}{\sqrt{\gamma^{(2)}_t[\boldsymbol{v}_{t-1}]_i} + \epsilon} \le \sum_{i=1}^p\frac{2qM^2_f\frac{1}{K^{(1)}_t}\sigma^2_2 + 2qM^2_g\frac{1}{K^{(2)}_t}\sigma^2_1 + \left[\nabla g(\boldsymbol{x}_t)^{\mathsf{T}}\nabla f(\boldsymbol{y}_t)\right]^2_i}{\sqrt{\gamma^{(2)}_t[\boldsymbol{v}_{t-1}]_i} + \epsilon} = \\\nonumber
    &\frac{2pqM^2_f\sigma^2_2}{\epsilon K^{(1)}_{t}} + \frac{2pqM^2_g\sigma^2_1}{\epsilon K^{(2)}_{t}} + \sum_{i=1}^p\frac{2\left[\nabla \mathcal{J}(\boldsymbol{x}_t)\right]^2_i + 2M^2_gL^2_f\left|\left|\boldsymbol{y}_t - g(\boldsymbol{x}_t)\right|\right|^2_2}{\sqrt{\gamma^{(2)}_t[\boldsymbol{v}_{t-1}]_i} + \epsilon} \le \\\nonumber
    &\frac{2pqM^2_f\sigma^2_2}{\epsilon K^{(1)}_{t}} + \frac{2pqM^2_g\sigma^2_1}{\epsilon K^{(2)}_{t}} + 
    \frac{2pM^2_gL^2_f}{\epsilon}\left|\left|\boldsymbol{y}_t - g(\boldsymbol{x}_t)\right|\right|^2_2 + 
    2\sum_{i=1}^p\frac{\left[\nabla \mathcal{J}(\boldsymbol{x}_t)\right]^2_i}{\sqrt{\gamma^{(2)}_t[\boldsymbol{v}_{t-1}]_i} + \epsilon}
\end{align}
Hence,  expression (\ref{new_change_function}) can be simplified as follows:
\begin{align}\label{newest_change_function}
    &\mathbb{E}_t\left[\mathcal{J}(\boldsymbol{x}_{t+1})\right] - \mathcal{J}(\boldsymbol{x}_t) \le - \alpha_t\frac{\left(1 - \gamma^{(1)}_t\right)}{2}\sum_{i=1}^{p}\frac{[\nabla\mathcal{J}(\boldsymbol{x}_{t})]^2_i}{\sqrt{\gamma^{(2)}_t[\boldsymbol{v}_{t-1}]_i} + \epsilon}  
    + \alpha_t\frac{\left(1 - \gamma^{(1)}_t\right)}{2\epsilon}M^2_gL^2_f\left|\left|g(\boldsymbol{x}_t) - \boldsymbol{y}_t \right|\right|^2_2 +\\\nonumber
    &\left[\frac{L\alpha_t^2\left(1 - \gamma^{(1)}_t\right)^2}{\epsilon} + \frac{\alpha_tM_gM_f(1 - \gamma^{(1)}_t)\sqrt{1 - \gamma^{(2)}_t}}{\epsilon}\right]\sum_{i=1}^p\frac{\mathbb{E}_t\left[[\overline{\nabla \mathcal{J}(\boldsymbol{x}_{t})}]^2_i\right]}{\sqrt{\gamma^{(2)}_{t}[\boldsymbol{v}_{t-1}]_i} + \epsilon} - \alpha_t\gamma^{(1)}_t\nabla\mathcal{J}(\boldsymbol{x}_t)^{\mathsf{T}}\frac{\boldsymbol{m}_{t-1}}{\sqrt{\gamma^{(2)}_t\boldsymbol{v}_{t-1}}+ \epsilon} + \\\nonumber
    &\frac{L\alpha_t^2\left(\gamma^{(1)}_t\right)^2}{\epsilon}\sum_{i=1}^{p}\frac{[\boldsymbol{m}_{t-1}]^2_i}{\sqrt{\gamma^{(2)}_t[\boldsymbol{v}_{t-1}]_i} + \epsilon} + \frac{\alpha_tM_gM_f\left(\gamma^{(1)}_t\right)^2\sqrt{1 - \gamma^{(2)}_t}}{\epsilon (1 - \gamma^{(1)}_t)}\sum_{i=1}^{p}\frac{|[\nabla\mathcal{J}(\boldsymbol{x}_t)]_i[\boldsymbol{m}_{t-1}]_i|}{\sqrt{\gamma^{(2)}_{t}[\boldsymbol{v}_{t-1}]_i} + \epsilon} = \\\nonumber
    &- \alpha_t\frac{\left(1 - \gamma^{(1)}_t\right)}{2}\sum_{i=1}^{p}\frac{[\nabla\mathcal{J}(\boldsymbol{x}_{t})]^2_i}{\sqrt{\gamma^{(2)}_t[\boldsymbol{v}_{t-1}]_i} + \epsilon}  
    + \alpha_t\frac{\left(1 - \gamma^{(1)}_t\right)}{2\epsilon}M^2_gL^2_f\left|\left|g(\boldsymbol{x}_t) - \boldsymbol{y}_t \right|\right|^2_2 +\\\nonumber
    & \frac{\alpha_t\gamma^{(1)}_tM^2_gM^2_f}{\epsilon} +  \frac{L\alpha^2_t\left(\gamma^{(1)}_t\right)^2M^2_gM^2_f}{\epsilon^2} +  \frac{\alpha_tM^3_gM^3_f\left(\gamma^{(1)}_t\right)^2\sqrt{1 - \gamma^{(2)}_t}}{\epsilon^2 (1 - \gamma^{(1)}_t)} + \\\nonumber
    &\left[\frac{L\alpha_t^2\left(1 - \gamma^{(1)}_t\right)^2}{\epsilon} + \frac{\alpha_tM_gM_f(1 - \gamma^{(1)}_t)\sqrt{1 - \gamma^{(2)}_t}}{\epsilon}\right]\sum_{i=1}^p\frac{\mathbb{E}_t\left[[\overline{\nabla \mathcal{J}(\boldsymbol{x}_{t})}]^2_i\right]}{\sqrt{\gamma^{(2)}_{t}[\boldsymbol{v}_{t-1}]_i} + \epsilon}
\end{align}
where we used (\ref{bound_expressions}):
\begin{align*}
    &-\alpha_t\gamma^{(1)}_t\nabla\mathcal{J}(\boldsymbol{x}_t)^{\mathsf{T}}\frac{\boldsymbol{m}_{t-1}}{\sqrt{\gamma^{(2)}_t\boldsymbol{v}_{t-1}}+ \epsilon} \le \frac{\alpha_t\gamma^{(1)}_t}{\epsilon}||\nabla\mathcal{J}(\boldsymbol{x}_t)||_2||\boldsymbol{m}_{t-1}||_2 \le\frac{\alpha_t\gamma^{(1)}_tM^2_gM^2_f}{\epsilon},\\\nonumber
    &\frac{L\alpha^2_t\left(\gamma^{(1)}_t\right)^2}{\epsilon}\sum_{i=1}^{p}\frac{[\boldsymbol{m}_{t-1}]^2_i}{\sqrt{\gamma^{(2)}_t[\boldsymbol{v}_{t-1}]_i} + \epsilon} \le \frac{L\alpha^2_t\left(\gamma^{(1)}_t\right)^2M^2_gM^2_f}{\epsilon^2},
\end{align*}
\begin{align*}
    &\frac{\alpha_tM_gM_f\left(\gamma^{(1)}_t\right)^2\sqrt{1 - \gamma^{(2)}_t}}{\epsilon (1 - \gamma^{(1)}_t)}\sum_{i=1}^{p}\frac{|[\nabla\mathcal{J}(\boldsymbol{x}_t)]_i[\boldsymbol{m}_{t-1}]_i|}{\sqrt{\gamma^{(2)}_{t}[\boldsymbol{v}_{t-1}]_i} + \epsilon} \le \frac{\alpha_tM_gM_f\left(\gamma^{(1)}_t\right)^2\sqrt{1 - \gamma^{(2)}_t}}{\epsilon^2 (1 - \gamma^{(1)}_t)} \sum_{i=1}^p|[\nabla\mathcal{J}(\boldsymbol{x}_t)]_i[\boldsymbol{m}_{t-1}]_i| \le \\\nonumber
    &\frac{\alpha_tM_gM_f\left(\gamma^{(1)}_t\right)^2\sqrt{1 - \gamma^{(2)}_t}}{\epsilon^2 (1 - \gamma^{(1)}_t)}(||\nabla\mathcal{J}(\boldsymbol{x}_t)||^2_2 + ||\boldsymbol{m}_{t-1}||^2_2) \le \frac{\alpha_tM^3_gM^3_f\left(\gamma^{(1)}_t\right)^2\sqrt{1 - \gamma^{(2)}_t}}{\epsilon^2 (1 - \gamma^{(1)}_t)}
\end{align*}
Hence, applying (\ref{addition_expression_1}) in (\ref{newest_change_function}) gives:
\begin{align}\label{change_function_1}
    &\mathbb{E}_t\left[\mathcal{J}(\boldsymbol{x}_{t-1})\right] - \mathcal{J}(\boldsymbol{x}_t) \le \\\nonumber
    &- \alpha_t\frac{\left(1 - \gamma^{(1)}_t\right)}{2}\sum_{i=1}^{p}\frac{[\nabla\mathcal{J}(\boldsymbol{x}_{t})]^2_i}{\sqrt{\gamma^{(2)}_t[\boldsymbol{v}_{t-1}]_i} + \epsilon}  
    + \alpha_t\frac{\left(1 - \gamma^{(1)}_t\right)}{2\epsilon}M^2_gL^2_f\left|\left|g(\boldsymbol{x}_t) - \boldsymbol{y}_t \right|\right|^2_2 +\\\nonumber
    & \frac{\alpha_t\gamma^{(1)}_tM^2_gM^2_f}{\epsilon}\left[1 + \frac{L\alpha_t\gamma^{(1)}_t}{\epsilon} + \frac{M_gM_f\gamma^{(1)}_t\sqrt{1 - \gamma^{(2)}_t}}{\epsilon\left(1 - \gamma^{(1)}_t\right)}\right] + \\\nonumber
    &\frac{\alpha_t\left(1 - \gamma^{(1)}_t\right)}{\epsilon}\left[L\alpha_t\left(1 - \gamma^{(1)}_t\right) + M_gM_f\sqrt{1 - \gamma^{(2)}_t}\right]\sum_{i=1}^p\frac{\mathbb{E}_t\left[[\overline{\nabla \mathcal{J}(\boldsymbol{x}_{t})}]^2_i\right]}{\sqrt{\gamma^{(2)}_{t}[\boldsymbol{v}_{t-1}]_i} + \epsilon} \le \\\nonumber
    &-\alpha_t\frac{\left(1 - \gamma^{(1)}_t\right)}{2}\sum_{i=1}^{p}\frac{[\nabla\mathcal{J}(\boldsymbol{x}_{t})]^2_i}{\sqrt{\gamma^{(2)}_t[\boldsymbol{v}_{t-1}]_i} + \epsilon}  
    + \alpha_t\frac{\left(1 - \gamma^{(1)}_t\right)}{2\epsilon}M^2_gL^2_f\left|\left|g(\boldsymbol{x}_t) - \boldsymbol{y}_t \right|\right|^2_2 +\\\nonumber
    & \frac{\alpha_t\gamma^{(1)}_tM^2_gM^2_f}{\epsilon}\left[1 + \frac{L\alpha_t\gamma^{(1)}_t}{\epsilon} + \frac{M_gM_f\gamma^{(1)}_t\sqrt{1 - \gamma^{(2)}_t}}{\epsilon\left(1 - \gamma^{(1)}_t\right)}\right] + \frac{\alpha_t\left(1 - \gamma^{(1)}_t\right)}{\epsilon}\left[L\alpha_t\left(1 - \gamma^{(1)}_t\right) + M_gM_f\sqrt{1 - \gamma^{(2)}_t}\right]\times\\\nonumber
    &\left[\frac{2pqM^2_f\sigma^2_2}{\epsilon K^{(1)}_{t}} + \frac{2pqM^2_g\sigma^2_1}{\epsilon K^{(2)}_{t}} + 
    \frac{2pM^2_gL^2_f}{\epsilon}\left|\left|\boldsymbol{y}_t - g(\boldsymbol{x}_t)\right|\right|^2_2 + 
    2\sum_{i=1}^p\frac{\left[\nabla \mathcal{J}(\boldsymbol{x}_t)\right]^2_i}{\sqrt{\gamma^{(2)}_t[\boldsymbol{v}_{t-1}]_i} + \epsilon}\right]
\end{align}
Grouping the terms in (\ref{change_function_1}) gives:
\begin{align}\label{change_function_1_1}
    &\mathbb{E}_t\left[\mathcal{J}(\boldsymbol{x}_{t+1})\right] - \mathcal{J}(\boldsymbol{x}_t) \le -\frac{\alpha_t\left(1 - \gamma^{(1)}_t\right)}{2}\left[1 - \frac{4\left[L\alpha_t\left(1 - \gamma^{(1)}_t\right) + M_gM_f\sqrt{1 - \gamma^{(2)}_t}\right]}{\epsilon} \right]\sum_{i=1}^{p}\frac{[\nabla\mathcal{J}(\boldsymbol{x}_{t})]^2_i}{\sqrt{\gamma^{(2)}_t[\boldsymbol{v}_{t-1}]_i} + \epsilon}  
    +  \\\nonumber
    &\frac{\alpha_t\left(1 - \gamma^{(1)}_t\right)}{2\epsilon}M^2_gL^2_f\left[1 + \frac{4p\left[L\alpha_t\left(1 - \gamma^{(1)}_t\right) + M_gM_f\sqrt{1 - \gamma^{(2)}_t}\right]}{\epsilon} \right]\left|\left|g(\boldsymbol{x}_t) - \boldsymbol{y}_t \right|\right|^2_2 +\\\nonumber
    &\frac{2pq\alpha_t\left(1 - \gamma^{(1)}_t\right)}{\epsilon^2}\left[L\alpha_t\left(1 - \gamma^{(1)}_t\right) + M_gM_f\sqrt{1 - \gamma^{(2)}_t}\right]\left(\frac{M^2_f\sigma^2_2}{ K^{(1)}_{t}} + \frac{M^2_g\sigma^2_1}{K^{(2)}_{t}} \right) + \\\nonumber
    &\frac{\alpha_t\gamma^{(1)}_tM^2_gM^2_f}{\epsilon}\left[1 + \frac{\gamma^{(1)}_t\left[L\alpha_t\left(1 - \gamma^{(1)}_t\right) + M_gM_f\sqrt{1 - \gamma^{(2)}_t}\right]}{\epsilon\left(1  - \gamma^{(1)}_{t}\right)}\right] = 
\end{align}
\begin{align*}
    &-\frac{\alpha_t\left(1 - \gamma^{(1)}_t\right)}{2}\left[1 - 4\mathcal{C}_t \right]\sum_{i=1}^{p}\frac{[\nabla\mathcal{J}(\boldsymbol{x}_{t})]^2_i}{\sqrt{\gamma^{(2)}_t[\boldsymbol{v}_{t-1}]_i} + \epsilon}  
    +  \frac{\alpha_t\left(1 - \gamma^{(1)}_t\right)M^2_gL^2_f}{2\epsilon}\left[1 + 4p\mathcal{C}_t\right]\left|\left|g(\boldsymbol{x}_t) - \boldsymbol{y}_t \right|\right|^2_2 +\\\nonumber
    &\frac{2pq\alpha_t\mathcal{C}_t\left(1 - \gamma^{(1)}_t\right)}{\epsilon}\left(\frac{M^2_f\sigma^2_2}{ K^{(1)}_{t}} + \frac{M^2_g\sigma^2_1}{K^{(2)}_{t}} \right) + \frac{\alpha_t\gamma^{(1)}_tM^2_gM^2_f}{\epsilon}\left[1 + \frac{\gamma^{(1)}_t\mathcal{C}_t}{\left(1  - \gamma^{(1)}_{t}\right)}\right] \le
\end{align*}
\begin{align*}
    &-\frac{\alpha_t\left(1 - \gamma^{(1)}_t\right)}{2(M_gM_f + \epsilon)}\left[1 - 4\mathcal{C}_t \right]||\nabla\mathcal{J}(\boldsymbol{x}_{t})||^2_2 
    +  \frac{\alpha_t\left(1 - \gamma^{(1)}_t\right)M^2_gL^2_f}{2\epsilon}\left[1 + 4p\mathcal{C}_t\right]\left|\left|g(\boldsymbol{x}_t) - \boldsymbol{y}_t \right|\right|^2_2 +\\\nonumber
    &\frac{2pq\alpha_t\mathcal{C}_t\left(1 - \gamma^{(1)}_t\right)}{\epsilon}\left(\frac{M^2_f\sigma^2_2}{ K^{(1)}_{t}} + \frac{M^2_g\sigma^2_1}{K^{(2)}_{t}} \right) + \frac{\alpha_t\gamma^{(1)}_tM^2_gM^2_f}{\epsilon}\left[1 + \frac{\gamma^{(1)}_t\mathcal{C}_t}{\left(1  - \gamma^{(1)}_{t}\right)}\right]
\end{align*}
where we use notation $\mathcal{C}_{t} = \frac{L\alpha_t\left(1 - \gamma^{(1)}_t\right) + M_gM_f\sqrt{1 - \gamma^{(2)}_t}}{\epsilon}$ and $[\boldsymbol{v}_{t-1}]_i \le M^2_gM^2_f$.\\
Next, let us denote $\mathbb{E}_{total}\left[\right]$ be the expectation with respect to all randomness induced in all $T$ iterations of Algorithm \ref{Algo:ADAM}. Using the low of total expectation:
\begin{equation*}
    \mathbb{E}_{total}\left[\mathbb{E}_t\left[\zeta_t\right]\right] =  \mathbb{E}_{total}\left[\mathbb{E}_{K^{(1)}_t,K^{(2)}_t,K^{(3)}_t }\left[\zeta_t\Big| \boldsymbol{x}_t\right]\right] = \mathbb{E}_{total}\left[\zeta_t\right]
\end{equation*}
for any $t-$measurable\footnote{Random variable is called  $t-$measurable if its affected by the randomness induced in  the first $t$ rounds.} random variable $\zeta_t$. Hence, taking expectation $\mathbb{E}$ from both sides of (\ref{change_function_1_1}) gives:
\begin{align}\label{change_function_1_2}
    &\mathbb{E}_{total}\left[\mathcal{J}(\boldsymbol{x}_{t+1}) - \mathcal{J}(\boldsymbol{x}_t)\right] \le \\\nonumber
    &-\frac{\alpha_t\left(1 - \gamma^{(1)}_t\right)}{2(M_gM_f + \epsilon)}\left[1 - 4\mathcal{C}_t \right]\mathbb{E}_{total}\left[||\nabla\mathcal{J}(\boldsymbol{x}_{t})||^2_2\right]  
    +  \frac{\alpha_t\left(1 - \gamma^{(1)}_t\right)M^2_gL^2_f}{2\epsilon}\left[1 + 4p\mathcal{C}_t\right]\mathbb{E}_{total}\left[\left|\left|g(\boldsymbol{x}_t) - \boldsymbol{y}_t \right|\right|^2_2\right] +\\\nonumber
    &\frac{2pq\alpha_t\mathcal{C}_t\left(1 - \gamma^{(1)}_t\right)}{\epsilon}\left(\frac{M^2_f\sigma^2_2}{ K^{(1)}_{t}} + \frac{M^2_g\sigma^2_1}{K^{(2)}_{t}} \right) + \frac{\alpha_t\gamma^{(1)}_tM^2_gM^2_f}{\epsilon}\left[1 + \frac{\gamma^{(1)}_t\mathcal{C}_t}{\left(1  - \gamma^{(1)}_{t}\right)}\right]
\end{align}
Now, let $\alpha_t = \frac{C_{\alpha}}{t^a}$, $\beta_t = \frac{C_{\beta}}{t^b}$, $K^{(1)}_t = C_{1}t^{c}$, $K^{(2)}_t = C_{2}t^{c}$ and  $K^{(3)}_t = C_{3}t^{e}$ for some constants $C_{\alpha},C_{\beta}, C_1, C_2, C_{3}, a,b,c,e > 0$ such that $(2a-2b)\notin [-1,0]$, $b \le 1$. Following Corollary \ref{corollary_1} we have:
\begin{equation*}
    \mathbb{E}_{total}\left[\left|\left|g(\boldsymbol{x}_t) - \boldsymbol{y}_t \right|\right|^2_2\right] \le \frac{L^2_gC^2_{\mathcal{D}}}{2}\frac{1}{t^{4a-4b}} + 2C^2_{\mathcal{E}}\frac{1}{t^{b+e}}
\end{equation*}
for some constants $C_{\mathcal{D}},C_{\mathcal{E}}> 0$, and (\ref{change_function_1_2}) can be written as:
\begin{align}\label{change_function_1_4}
    &\mathbb{E}_{total}\left[\mathcal{J}(\boldsymbol{x}_{t+1}) - \mathcal{J}(\boldsymbol{x}_t)\right] \le \\\nonumber
    &-\frac{C_{\alpha}\left(1 - \gamma^{(1)}_t\right)}{2t^a(M_gM_f + \epsilon)}\left[1 - 4\mathcal{C}_t \right]\mathbb{E}_{total}\left[||\nabla\mathcal{J}(\boldsymbol{x}_{t})||^2_2\right]  
    +  \frac{C_{\alpha}\left(1 - \gamma^{(1)}_t\right)M^2_gL^2_f}{2t^a\epsilon}\left[1 + 4p\mathcal{C}_t\right]\left[\frac{L^2_gC^2_{\mathcal{D}}}{2}\frac{1}{t^{4a-4b}} + 2C^2_{\mathcal{E}}\frac{1}{t^{b+e}}\right] +\\\nonumber
    &\frac{2pqC_{\alpha}\mathcal{C}_t\left(1 - \gamma^{(1)}_t\right)}{t^a\epsilon}\left(\frac{M^2_f\sigma^2_2}{C_1t^c} + \frac{M^2_g\sigma^2_1}{C_2t^c} \right) + \frac{C_{\alpha}\gamma^{(1)}_tM^2_gM^2_f}{t^a\epsilon}\left[1 + \frac{\gamma^{(1)}_t\mathcal{C}_t}{\left(1  - \gamma^{(1)}_{t}\right)}\right]
\end{align}
with $\mathcal{C}_t = \frac{L\frac{C_{\alpha}}{t^a}\left(1 - \gamma^{(1)}_t\right) + M_gM_f\sqrt{1 - \gamma^{(2)}_t}}{\epsilon}$. By choosing $\gamma^{(2)}_t = 1 - \frac{C^2_{\alpha}}{t^{2a}}(1 - \gamma^{(1)}_t)^2$ such that $\sqrt{1 - \gamma^{(2)}_t} = \frac{C_{\alpha}}{t^a}(1 - \gamma^{(1)}_t)$ we have $\mathcal{C}_t = C_{\alpha}\frac{(L + M_gM_f)}{\epsilon}\frac{1 - \gamma^{(1)}_{t}}{t^a} \le C_{\alpha}\frac{(L + M_gM_f)}{\epsilon}$. By choosing $C_{\alpha} \le \frac{\epsilon}{8p(L + M_gM_f)}$ we have $8p\mathcal{C}_t \le 1$, hence, (\ref{change_function_1_4}) can be simplified as:
\begin{align}\label{change_function_1_5}
    &\mathbb{E}_{total}\left[\mathcal{J}(\boldsymbol{x}_{t+1}) - \mathcal{J}(\boldsymbol{x}_t)\right] \le \\\nonumber
    &-\frac{C_{\alpha}\left(1 - \gamma^{(1)}_t\right)}{2t^a(M_gM_f + \epsilon)}\left[1  - \frac{1}{2p} \right]\mathbb{E}_{total}\left[||\nabla\mathcal{J}(\boldsymbol{x}_{t})||^2_2\right]  
    +  \frac{C_{\alpha}\left(1 - \gamma^{(1)}_t\right)M^2_gL^2_f}{t^a\epsilon}\left[\frac{L^2_gC^2_{\mathcal{D}}}{2}\frac{1}{t^{4a-4b}} + 2C^2_{\mathcal{E}}\frac{1}{t^{b+e}}\right] +\\\nonumber
    &\frac{qC_{\alpha}(C_1M^2_g\sigma^2_1 + C_2M^2_f\sigma^2_2)}{2C_1C_2\epsilon}\frac{\left(1 - \gamma^{(1)}_t\right)}{t^{a + c}} + \frac{C_{\alpha}M^2_gM^2_f}{\epsilon}\left[1 + \frac{\gamma^{(1)}_t}{4p\left(1  - \gamma^{(1)}_{t}\right)}\right]\frac{\gamma^{(1)}_t}{t^a} \le \\\nonumber
    &-\frac{C_{\alpha}\left(1 - \gamma^{(1)}_t\right)}{4t^a(M_gM_f + \epsilon)}\mathbb{E}_{total}\left[||\nabla\mathcal{J}(\boldsymbol{x}_{t})||^2_2\right]  
    +  \frac{C_{\alpha}\left(1 - \gamma^{(1)}_t\right)M^2_gL^2_f}{t^a\epsilon}\left[\frac{L^2_gC^2_{\mathcal{D}}}{2}\frac{1}{t^{4a-4b}} + 2C^2_{\mathcal{E}}\frac{1}{t^{b+e}}\right] +\\\nonumber
    &\frac{qC_{\alpha}(C_1M^2_g\sigma^2_1 + C_2M^2_f\sigma^2_2)}{2C_1C_2\epsilon}\frac{\left(1 - \gamma^{(1)}_t\right)}{t^{a + c}} + \frac{C_{\alpha}M^2_gM^2_f}{\epsilon}\left[1 + \frac{\gamma^{(1)}_t}{4p\left(1  - \gamma^{(1)}_{t}\right)}\right]\frac{\gamma^{(1)}_t}{t^a}
\end{align}
Assuming that $\gamma^{(1)}_t = C_{\gamma}\mu^{t}$ for some $C_{\gamma}\in\left(0,\frac{1}{2}\right)$ and $\mu\in[0,1)$, then expression (\ref{change_function_1_5}):
\begin{align*}
    &\mathbb{E}_{total}\left[\mathcal{J}(\boldsymbol{x}_{t+1}) - \mathcal{J}(\boldsymbol{x}_t)\right] \le \\\nonumber
    &-\frac{C_{\alpha}}{8t^a(M_gM_f + \epsilon)}\mathbb{E}_{total}\left[||\nabla\mathcal{J}(\boldsymbol{x}_{t})||^2_2\right]  
    +  \frac{C_{\alpha}M^2_gL^2_f}{t^a\epsilon}\left[\frac{L^2_gC^2_{\mathcal{D}}}{2}\frac{1}{t^{4a-4b}} + 2C^2_{\mathcal{E}}\frac{1}{t^{b+e}}\right] +\\\nonumber
    &\frac{qC_{\alpha}(C_1M^2_g\sigma^2_1 + C_2M^2_f\sigma^2_2)}{2C_1C_2\epsilon}\frac{1}{t^{a + c}} + \frac{2C_{\alpha}M^2_gM^2_f}{\epsilon}\frac{\mu^{t}}{t^a}
\end{align*}
Finally, choosing 
\begin{equation*}
C_{\alpha} = \min\Big\{\frac{\epsilon}{8p(L + M_gM_f)}, \frac{e}{2M^2_gM^2_f}, \frac{2C_1C_2\epsilon}{q(C_1M^2_g\sigma^2_1 + C_2M^2_f\sigma^2_2)}, \frac{2\epsilon}{M^2_gL^2_fL^2_gC^2_{\mathcal{D}}}, \frac{\epsilon}{2M^2_gL^2_fC^2_{\mathcal{E}}}\Big\}    
\end{equation*}
we have:
\begin{align*}
    &\mathbb{E}_{total}\left[\mathcal{J}(\boldsymbol{x}_{t+1}) - \mathcal{J}(\boldsymbol{x}_t)\right] \le \\\nonumber
    &-\frac{C_{0}}{t^a}\mathbb{E}_{total}\left[||\nabla\mathcal{J}(\boldsymbol{x}_{t})||^2_2\right]  
    +  \frac{1}{t^{5a-4b}} + \frac{1}{t^{a + b + e}} + \frac{1}{t^{a + c}} + \frac{\mu^{t}}{t^a}
\end{align*}
where $C_0 = \frac{C_{\alpha}}{8(M_gM_f + \epsilon)}$. Therefore,
\begin{align}\label{change_function_1_8}
    &\mathbb{E}_{total}\left[||\nabla\mathcal{J}(\boldsymbol{x}_{t})||^2_2\right] \le \\\nonumber &\frac{t^a}{C_0}\mathbb{E}_{total}\left[\mathcal{J}(\boldsymbol{x}_{t}) - \mathcal{J}(\boldsymbol{x}_{t+1})\right] + \frac{1}{C_0t^{4a-4b}} + \frac{1}{C_0t^{b + e}} + \frac{1}{C_0t^{c}} + \frac{\mu^{t}}{C_0}
\end{align}
Taking summation in (\ref{change_function_1_8}) over $t=1,\ldots,T$ and dividing the result by $T$ gives:
\begin{align}\label{change_function_1_9}
    &\frac{\sum_{t=1}^T\mathbb{E}_{total}\left[||\nabla\mathcal{J}(\boldsymbol{x}_{t})||^2_2\right]}{T} \le \\\nonumber &\frac{\sum_{t=1}^Tt^a\mathbb{E}_{total}\left[\mathcal{J}(\boldsymbol{x}_{t}) - \mathcal{J}(\boldsymbol{x}_{t+1})\right]}{C_0T} + \frac{1}{C_0T}\sum_{t=1}^T\left[\frac{1}{t^{4a-4b}} + \frac{1}{t^{b + e}} + \frac{1}{t^{c}}\right] + \frac{1}{C_0T(1 - \mu)}
\end{align}
Notice, using first order concavity condition for function $f(t) = t^a$ (if $a\in(0,1)$) and Assumption \ref{assum_1}.1: 
\begin{align*}
    &\sum_{t=1}^Tt^a\mathbb{E}_{total}\left[\mathcal{J}(\boldsymbol{x}_{t}) - \mathcal{J}(\boldsymbol{x}_{t+1})\right] = \\\nonumber
    &\mathcal{J}(\boldsymbol{x}_1) + \sum_{t=2}^{T}\left[(t+1)^a - t^a\right]\mathbb{E}_{total}\left[\mathcal{J}(\boldsymbol{x}_{t})\right] \le B_f + \sum_{t=2}^Tat^{a-1}\mathbb{E}_{total}\left[\mathcal{J}(\boldsymbol{x}_{t})\right] \le \mathcal{J}(\boldsymbol{x}_1) + B_f\sum_{t=2}^Tat^{a-1}\le\\\nonumber
    &B_f + B_faT^{a}
\end{align*} 
Hence, for (\ref{change_function_1_9}) we have (for $4a - 4b \le 1$):
\begin{align}\label{change_function_1_10}
    &\frac{\sum_{t=1}^T\mathbb{E}_{total}\left[||\nabla\mathcal{J}(\boldsymbol{x}_{t})||^2_2\right]}{T} \le \frac{B_f}{C_0}\frac{1}{T} + \frac{aB_f}{C_0}\frac{1}{T^{1-a}} + \frac{1}{T}\mathcal{O}\left(T^{4b-4a+1}\mathbb{I}\{4a-4b\neq 1\}\right) +\\\nonumber
    &\frac{1}{T}\mathcal{O}\left(\log T\mathbb{I}\{4a-4b = 1\}\right) + \frac{1}{T}\mathcal{O}\left( T^{1-b-e}\mathbb{I}\{b+e \ne 1\}+ \log T\mathbb{I}\{b+e = 1\}\right)\\\nonumber
    &\frac{1}{T}\mathcal{O}\left( T^{1-c}\mathbb{I}\{c \ne 1\}+ \log T\mathbb{I}\{c = 1\}\right) + \frac{1}{C_0(1-\mu)}\frac{1}{T} = \\\nonumber
    &\left(\frac{B_f}{C_0} + \frac{1}{C_0(1-\mu)}\right)\frac{1}{T} + \frac{aB_f}{C_0}\frac{1}{T^{1-a}} + \mathcal{O}\left(\frac{1}{T^{4a-4b}}\mathbb{I}\{4a-4b\neq 1\} + \frac{1}{T^{b+e}}\mathbb{I}\{b+e \ne 1\} + \frac{1}{T^c}\mathbb{I}\{c \ne 1\}\right)+\\\nonumber
    &\mathcal{O}\left(\frac{\log T}{T}\right)\left[\mathbb{I}\{4a-4b =  1\} + \mathbb{I}\{b+e = 1\} + \mathbb{I}\{c = 1\}\right]
\end{align}
where $\mathbb{I}\{\textbf{\text{condition}}\} = 1 $ if $\textbf{\text{condition}}$ is satisfied, and $0$ otherwise. Notice, that oracle complexity per iteration of Algorithm \ref{Algo:ADAM} is given by $\mathcal{O}(t^{\max\{c,e\}})$. Hence, after $T$ iterations, the total first order oracles complexity is given by $\mathcal{O}\left(T^{1 + \max\{c,e\}}\right)$. Let us denote $\phi(a,b,c,e) = \min\{1-a,  4a-4b,  b+e,  c  \}$. Then, ignoring logarithmic factors $\log T$ we have:
\begin{equation*}
    \frac{1}{T}\sum_{t=1}^T\mathbb{E}_{total}\left[||\nabla\mathcal{J}(\boldsymbol{x}_{t})||^2_2\right]\le \mathcal{O}\left(\frac{1}{T^{\phi(a,b,c,e)}}\right) \le \delta
\end{equation*}
implies $T = \mathcal{O}\left(\frac{1}{\delta^{\frac{1}{\phi(a,b,c,e)}}}\right)$, and for the first oracles complexity we have the following expression $\Psi(a,b,e,c) = \frac{1}{\delta^{\frac{1 + \max\{c,e\}}{\phi(a,b,c,e)}}}$. Hence, we have the following optimisation problem to find the optimal setup of parameters $a,b,c,e$:
\begin{align}
    &\min_{a,b,c,e}\frac{1 + \max\{c,e\}}{\min\{1-a,  4a-4b,  b+e,  c  \}}\\\nonumber
    &s.t.\ \ a > b \\\nonumber
    &0\le a\le 1\\\nonumber
    &0\le b\le 1\\\nonumber
    &4a - 4b \le 1\\\nonumber
    &c \ge 0\\\nonumber
    &e \ge 0
\end{align}
Considering all possible cases (8 in total) it is easy to see that the optimal setup is given by: $a^* = \frac{1}{5}$, $b^* = 0$, $c^* = e^* =  \frac{4}{5}$, and overall complexity is given by:
\begin{equation*}
    \Psi(a^*,b^*,c^*,e^*) = \delta^{-\frac{9}{4}}
\end{equation*}
This implies that Algorithm \ref{Algo:ADAM} in expectation outputs $\delta-$approximate first order stationary point of function $\mathcal{J}(\boldsymbol{x})$ and requires $\mathcal{O}\left(\delta^{-\frac{9}{4}}\right)$ calls to the oracles $\mathcal{FOO}_f[\cdot,\cdot]$ and $\mathcal{FOO}_g[\cdot,\cdot]$.
\end{proof}

\section{Additional Experiment Details}
\begin{figure*}[h!]
\centering
\begin{subfigure}{0.325\textwidth}
\centering
\includegraphics[width=\textwidth]{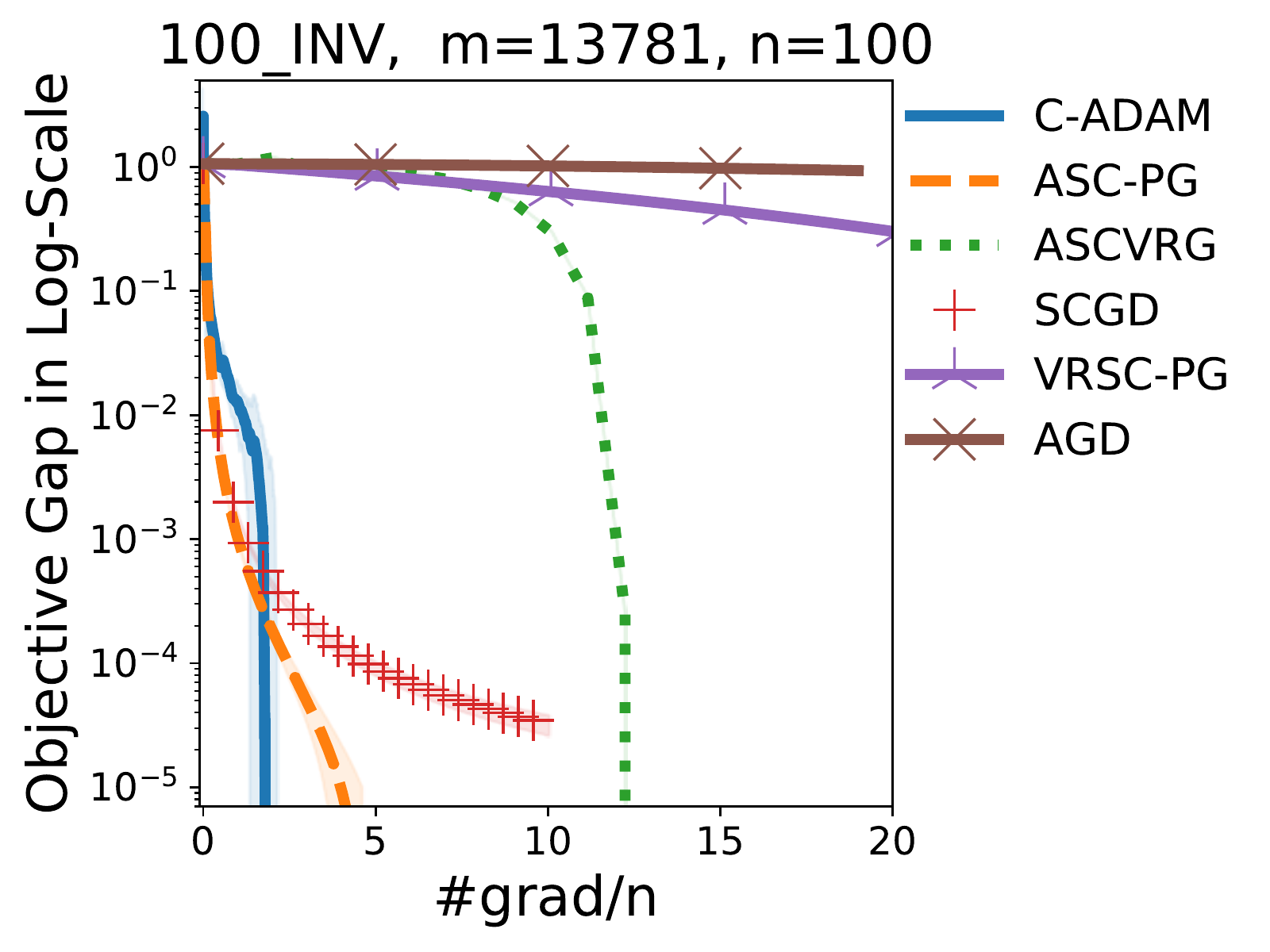}
\caption{}
\end{subfigure}
\begin{subfigure}{0.325\textwidth}
\centering
\includegraphics[width=\textwidth]{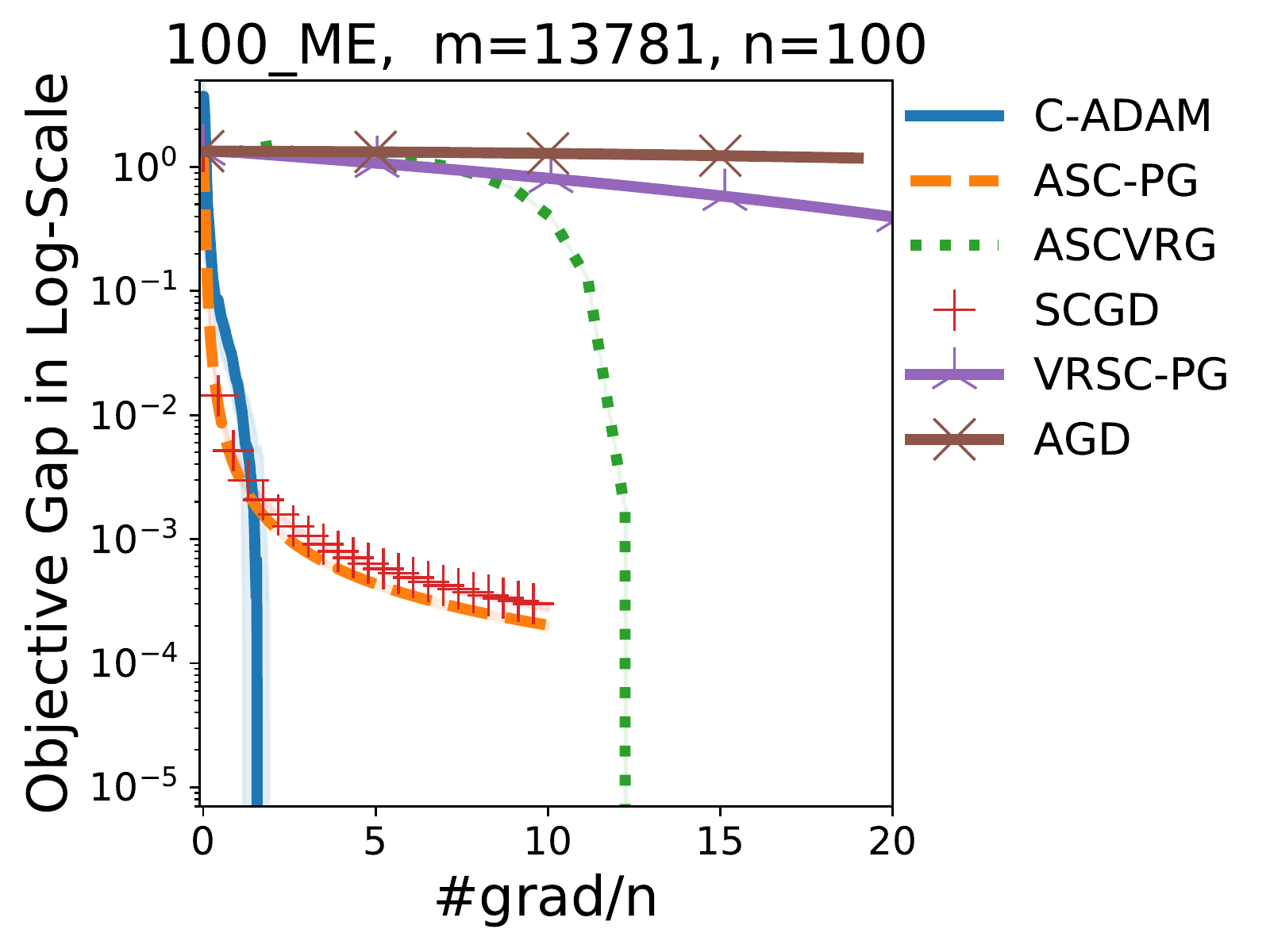}
\caption{}
\end{subfigure}
\begin{subfigure}{0.325\textwidth}
\centering
\includegraphics[width=\textwidth]{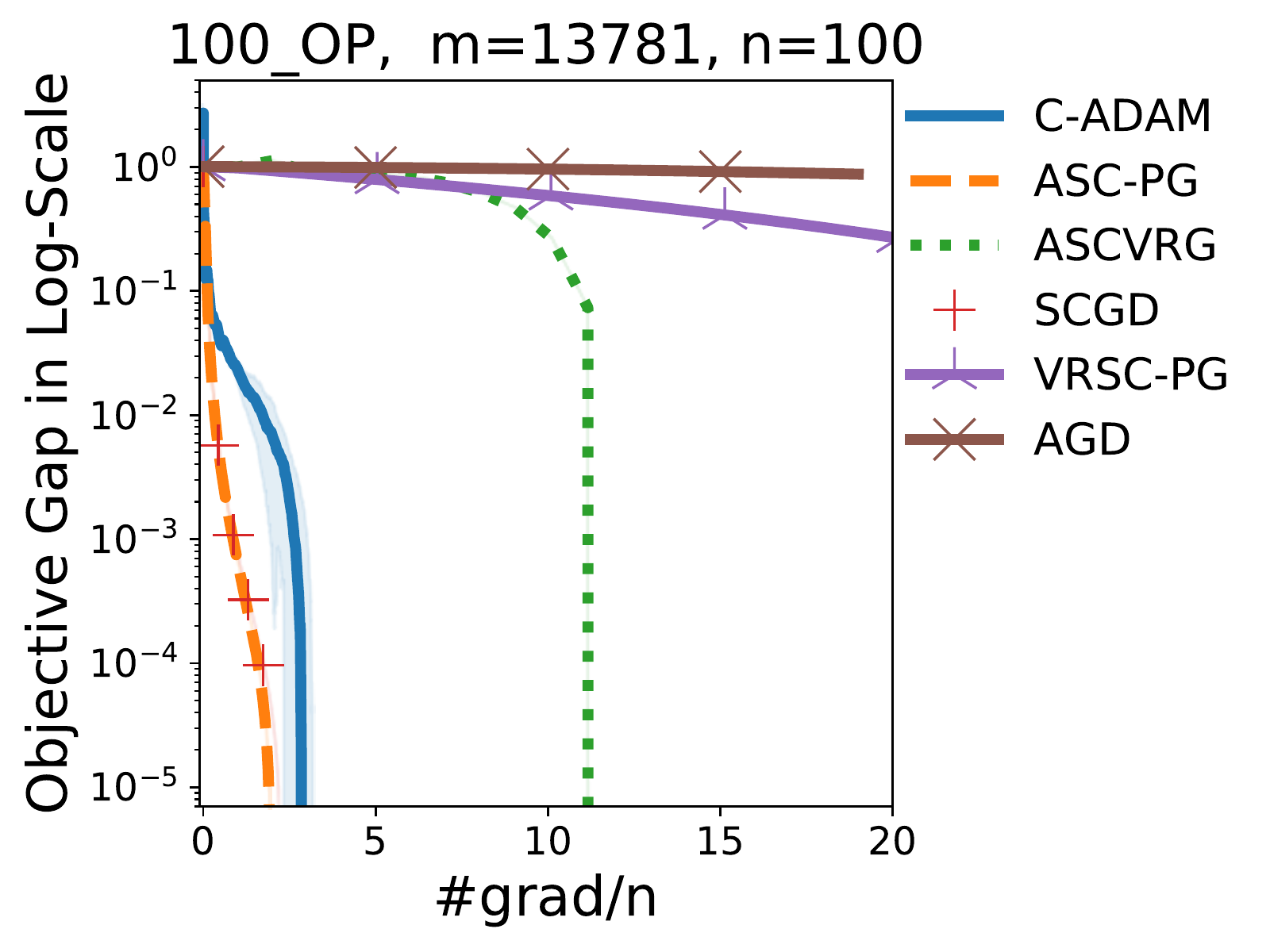}
\caption{}
\end{subfigure}
\caption{
Performance on 3 large 100-portfolio datasets ($m=13781, n=100$).
}
\label{fig:exp_100}
\end{figure*}
\begin{figure*}[h!]
\centering
\begin{subfigure}{0.325\textwidth}
\centering
\includegraphics[width=\textwidth]{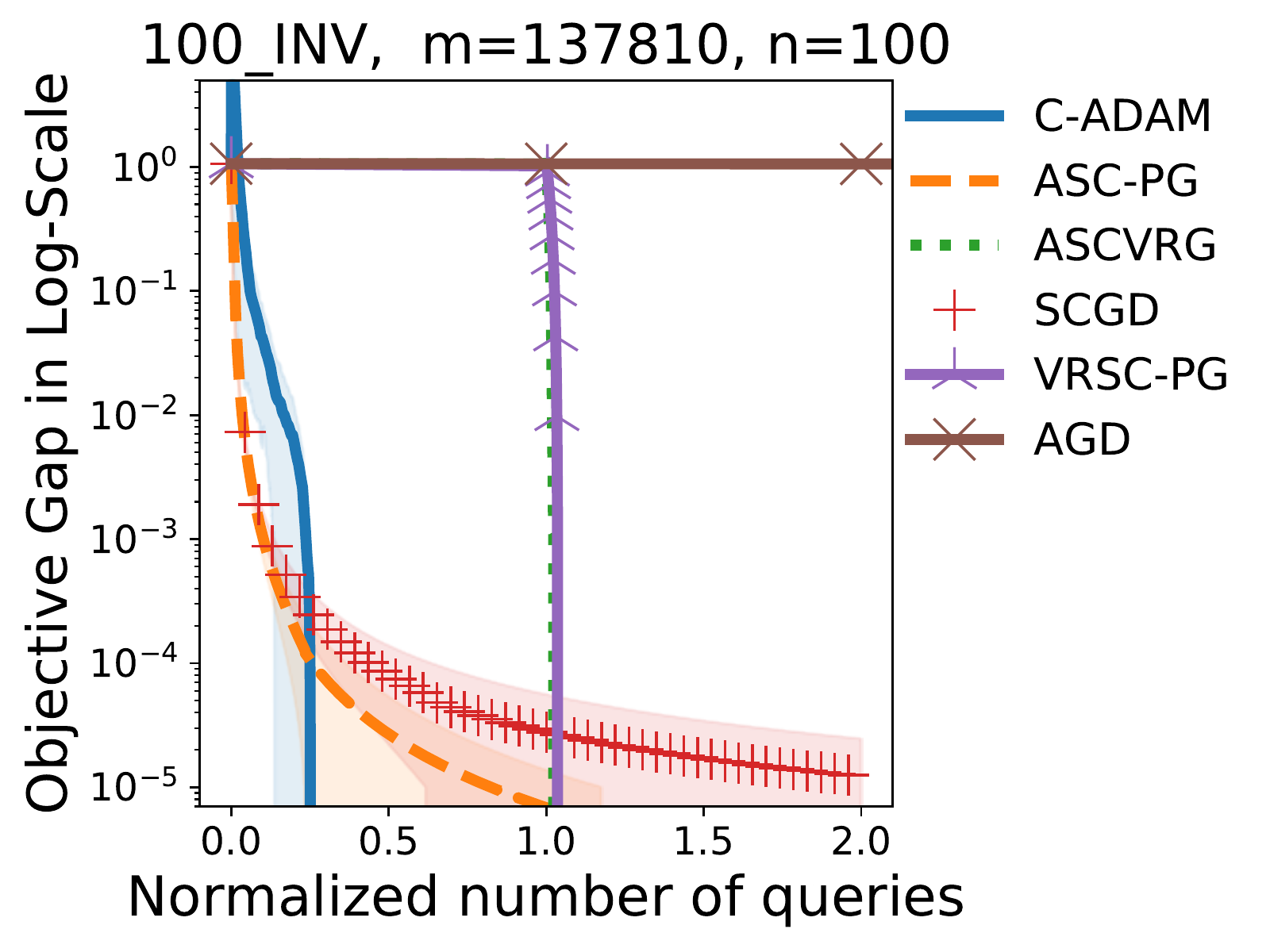}
\caption{}
\label{fig:100_inv_larger}
\end{subfigure}
\begin{subfigure}{0.325\textwidth}
\centering
\includegraphics[width=\textwidth]{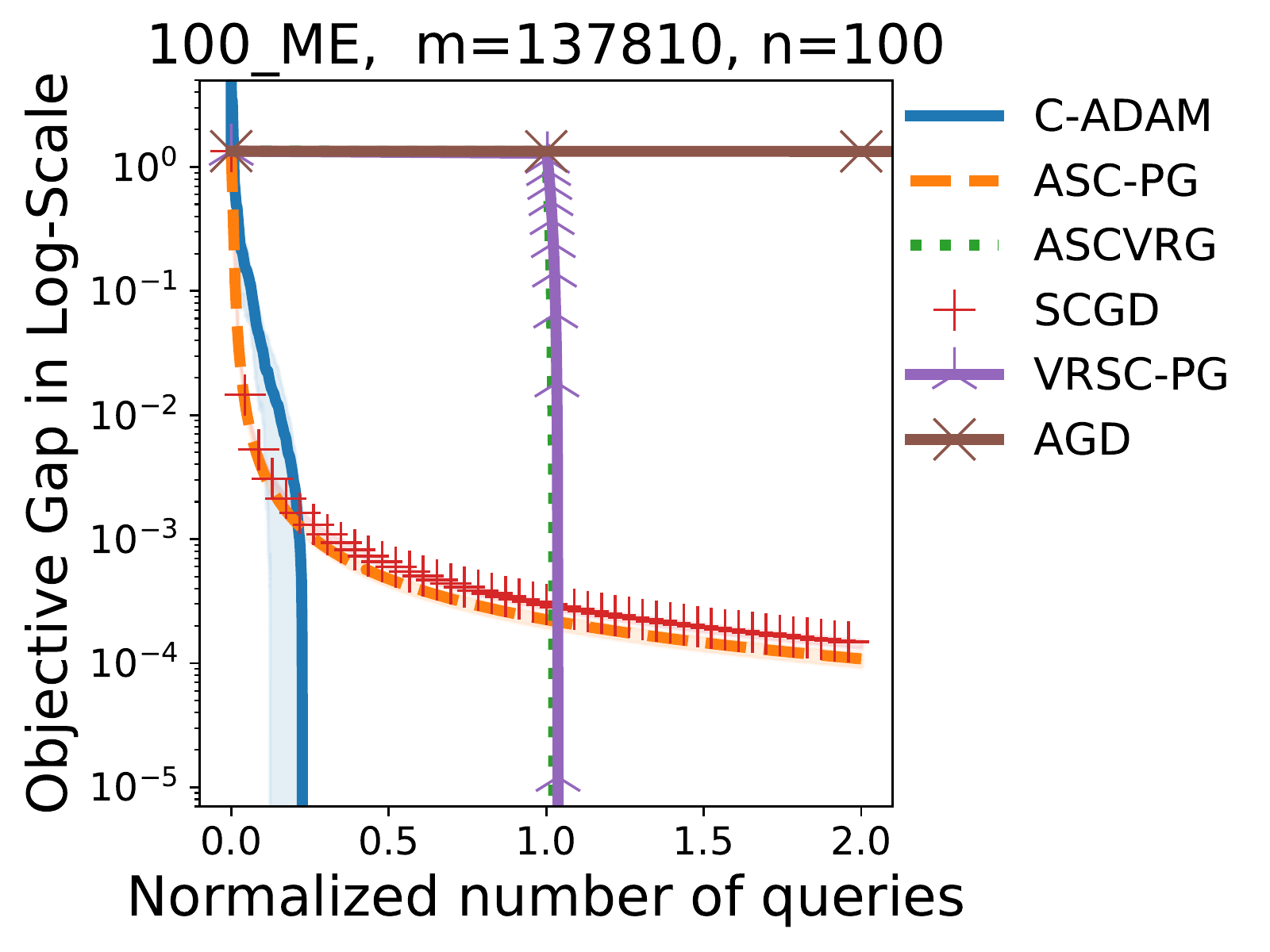}
\caption{}
\label{fig:100_me_larger}
\end{subfigure}
\begin{subfigure}{0.325\textwidth}
\centering
\includegraphics[width=\textwidth]{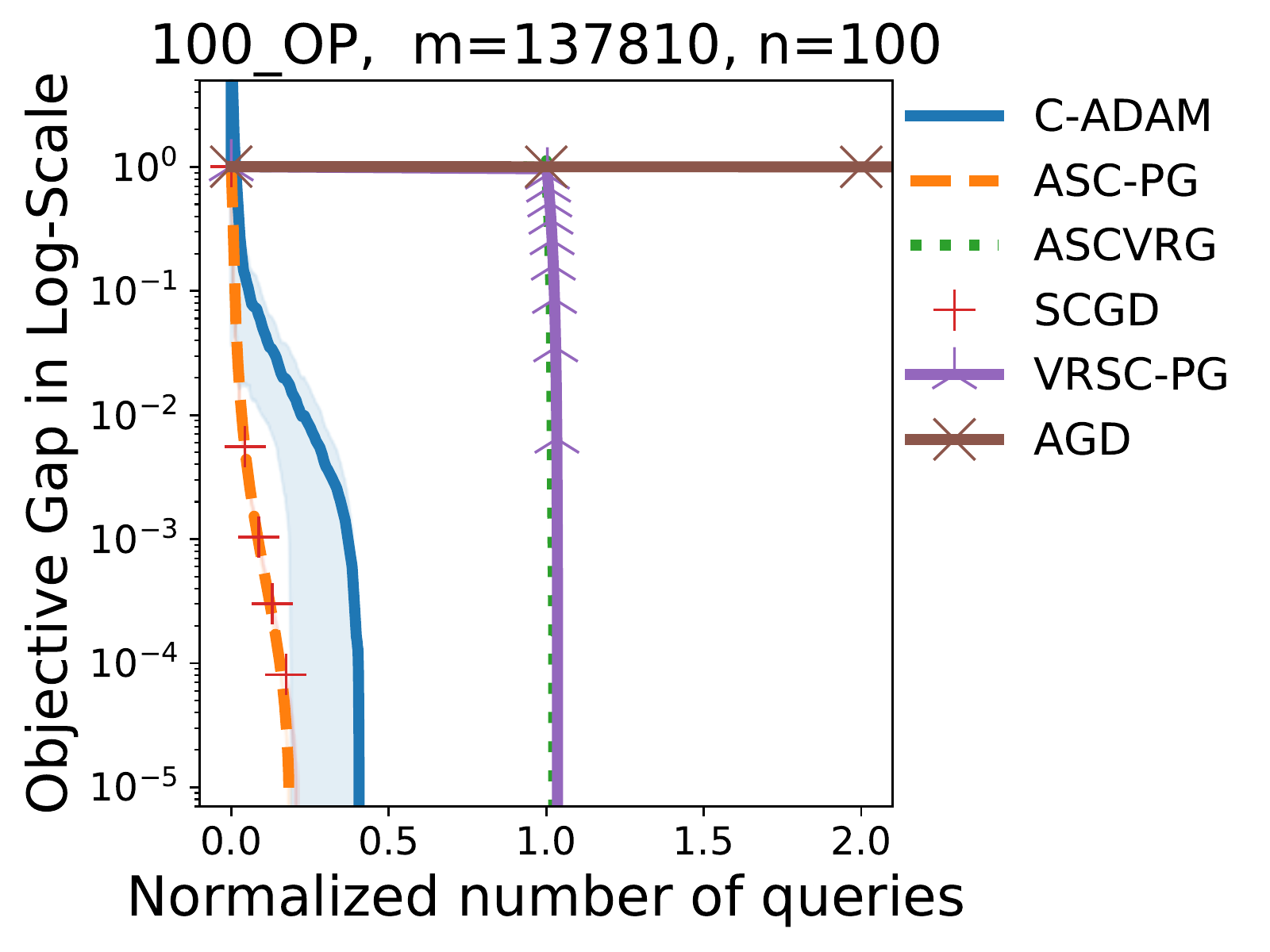}
\caption{}
\label{fig:100_op_larger}
\end{subfigure}
\caption{
Performance on 3 larger 100-portfolio datasets ($m=137810, n=100$).
}
\label{fig:exp_100_larger}
\end{figure*}
\begin{figure*}[h!]
\centering
\begin{subfigure}{0.325\textwidth}
\centering
\includegraphics[width=\textwidth]{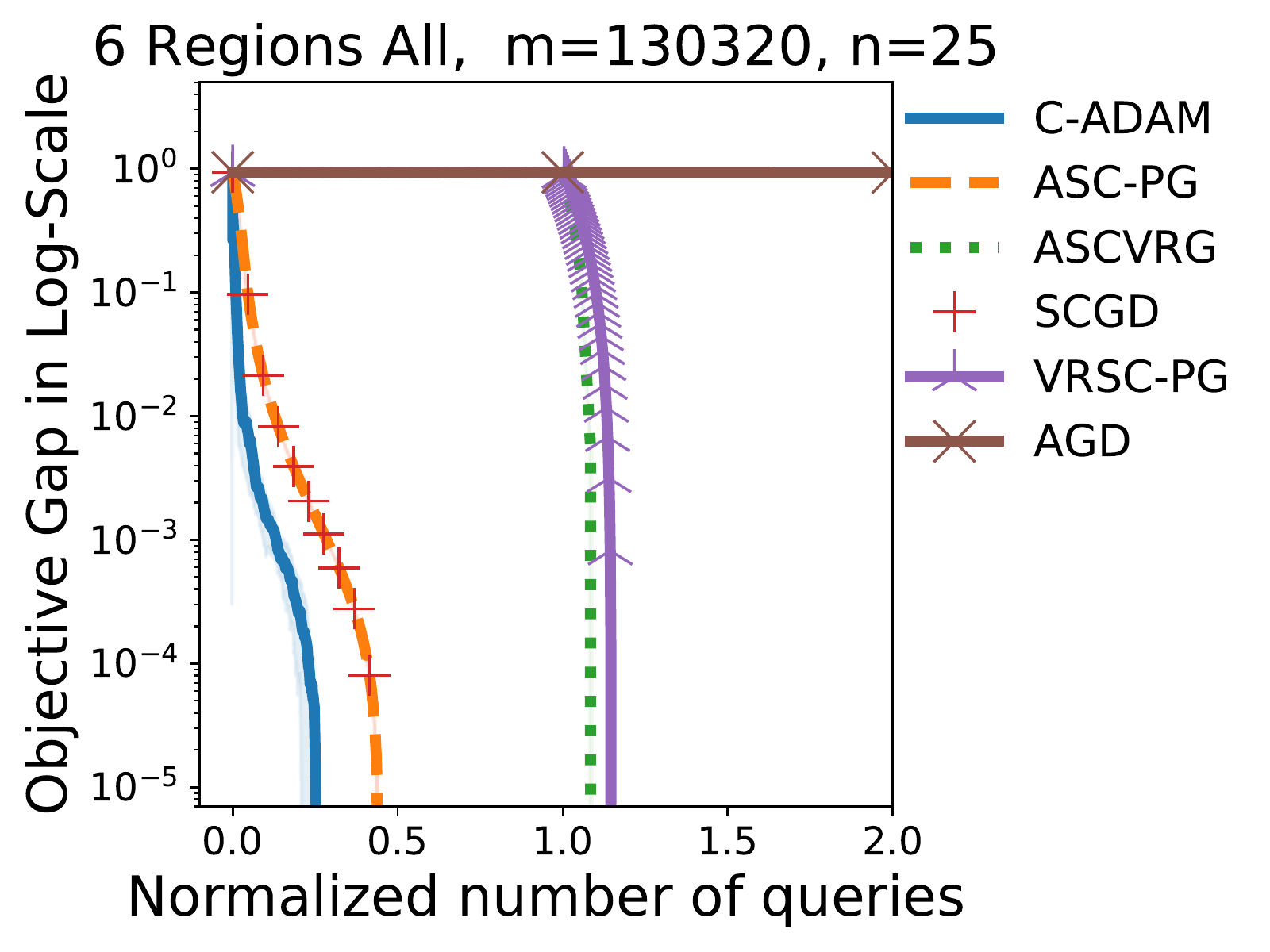}
\caption{}
\label{fig:region_all}
\end{subfigure}
\begin{subfigure}{0.325\textwidth}
\centering
\includegraphics[width=\textwidth]{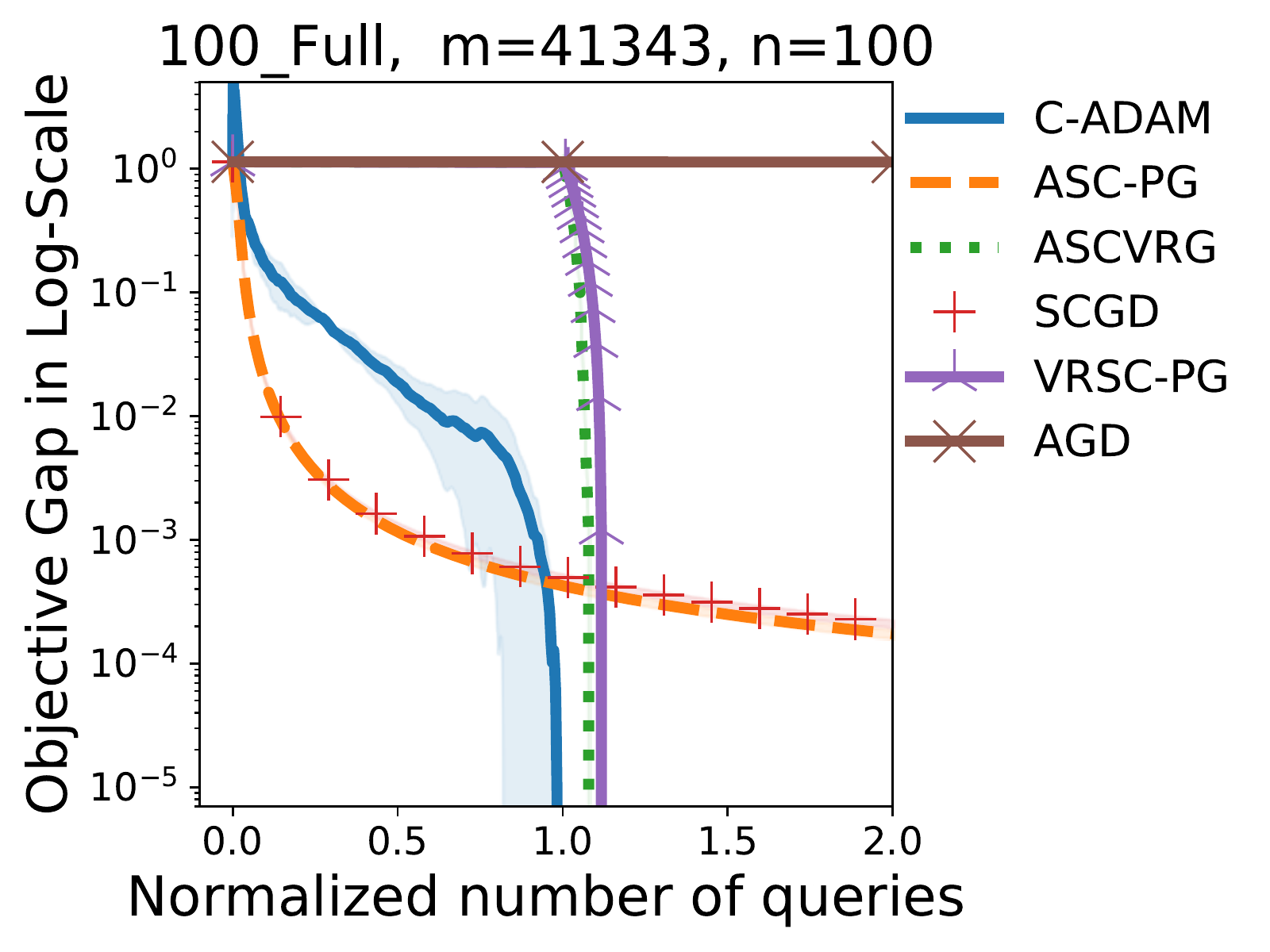}
\caption{}
\label{fig:100_all}
\end{subfigure}
\caption{
Performance on all region-based ($m=130320, n=25$) and all 100-portfolio datasets ($m=41343, n=100$).
}
\label{fig:full}
\end{figure*}

\subsection{Portfolio Mean-Variance}
In this section, we present numerical results of sparse mean-variance optimization problems on real-world portfolio datasets from CRSP\footnote{https://mba.tuck.dartmouth.edu/pages/faculty/ken.french/data\_library.html},
which are formed on size and: 1) Book-to-Market (BM), 2) Operating Profitability (OP), and 3) Investment (INV).
We consider \textbf{three} large 100-portfolio datasets ($m=13781, n=100$),
and \textbf{eighteen} region-based medium 25-portfolio datasets ($m=7240, n=25$).

Given $n$ assets and the reward vectors at $m$ time points,
the goal of sparse mean-variance optimization is to maximize the return of the investment as well as to control the investment risk:
$$
\min _{\boldsymbol{x} \in X} \frac{1}{m} \sum_{i=1}^{m}\left(\mathbf{r}_{i}^{\mathsf{T}} \boldsymbol{x}
-\frac{1}{m} \sum_{i=1}^{m} \mathbf{r}_{i}^{\mathsf{T}} \boldsymbol{x} \right)^{2}
-\frac{1}{m} \sum_{i=1}^{m} \mathbf{r}_{i}^{\mathsf{T}} \boldsymbol{x},
$$
which is in the form of Eq.~(\ref{Eq:Prob}) with
$
p=n,q=n+1,$
\begin{equation*}
    g_{\omega}(\boldsymbol{x})
   =g_{j}(\boldsymbol{x}) = \begin{bmatrix}
    \boldsymbol{x}\\
    -\boldsymbol{r}^{\mathsf{T}}_j\boldsymbol{x}\\
    \end{bmatrix}, \ \ \ \ \ f_{\nu}(g_{\omega}(\boldsymbol{x}))
    =f_{i}(g_{j}(\boldsymbol{x})) = \left([\boldsymbol{r}^{\mathsf{T}}_i, 1]g_{j}(\boldsymbol{x})\right)^2 - [\boldsymbol{r}^{\mathsf{T}}_i, 0]g_j(\boldsymbol{x}).
\end{equation*}
for $\boldsymbol{x} \in \mathbb{R}^{p}$ and a bounded set $X$.

\begin{figure*}[t!]
\centering
\begin{subfigure}{0.325\textwidth}
\centering
\includegraphics[width=\textwidth]{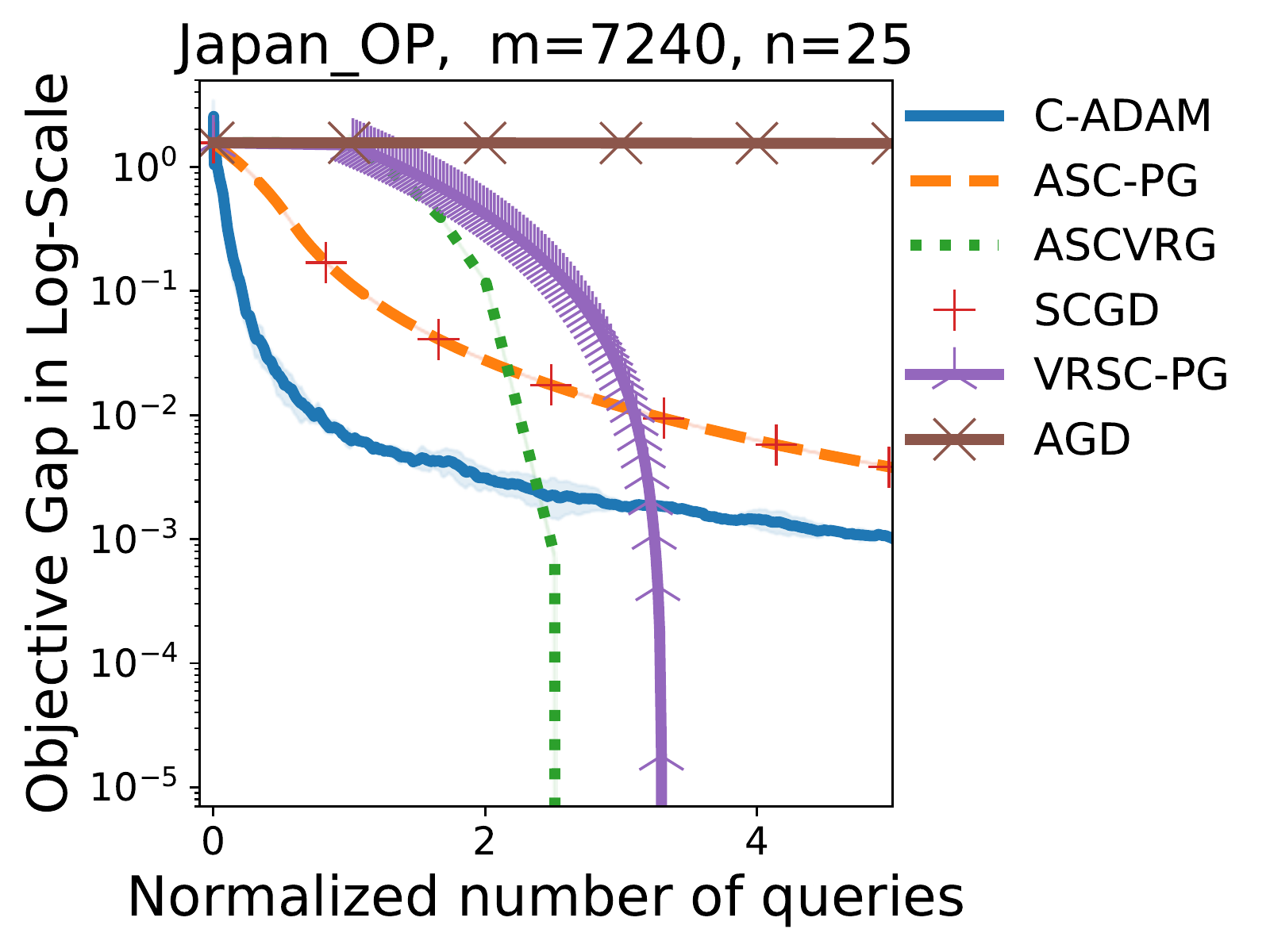}
\caption{}
\end{subfigure}
\begin{subfigure}{0.325\textwidth}
\centering
\includegraphics[width=\textwidth]{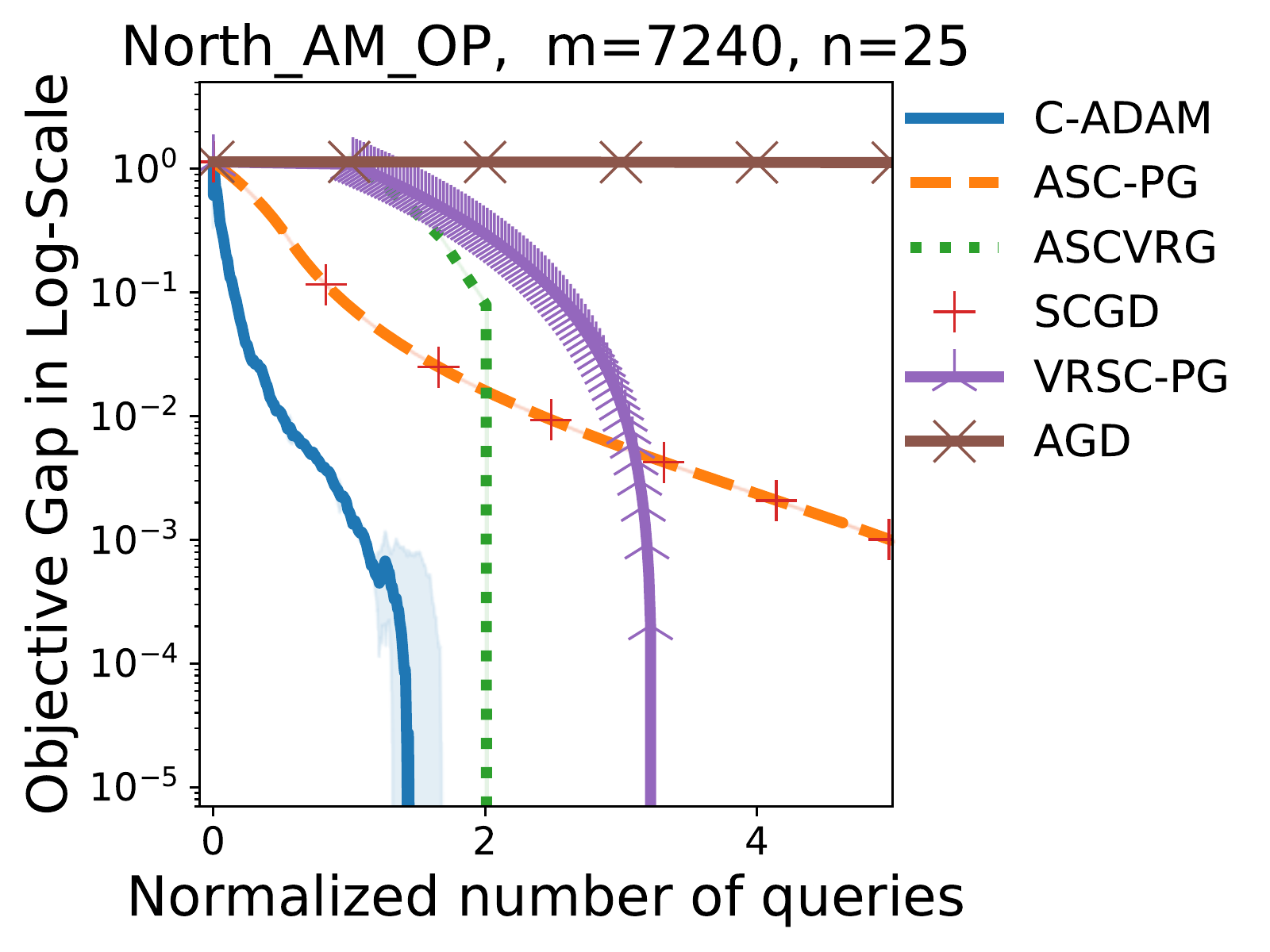}
\caption{}
\end{subfigure}
\begin{subfigure}{0.325\textwidth}
\centering
\includegraphics[width=\textwidth]{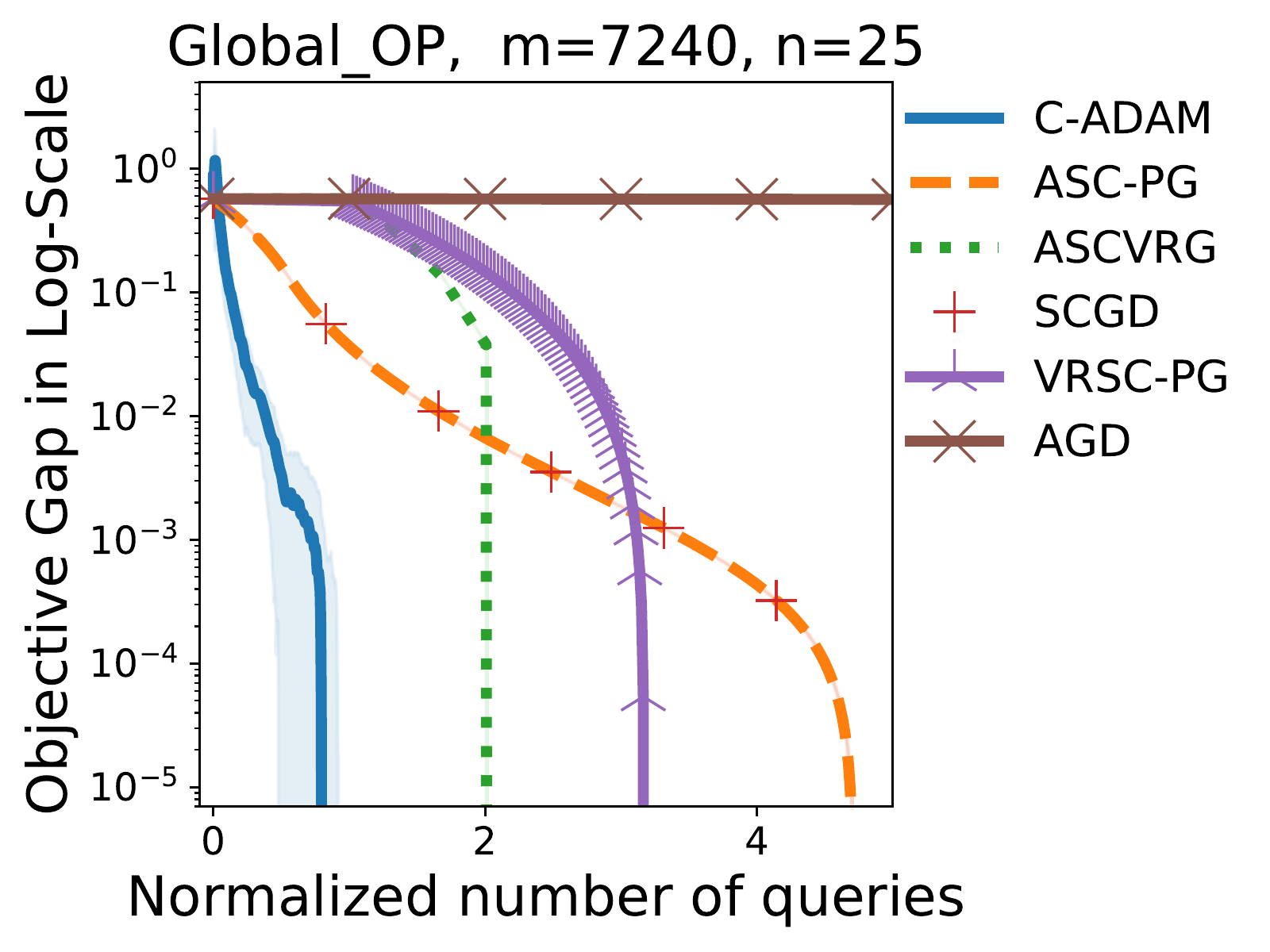}
\caption{}
\end{subfigure}
\begin{subfigure}{0.325\textwidth}
\centering
\includegraphics[width=\textwidth]{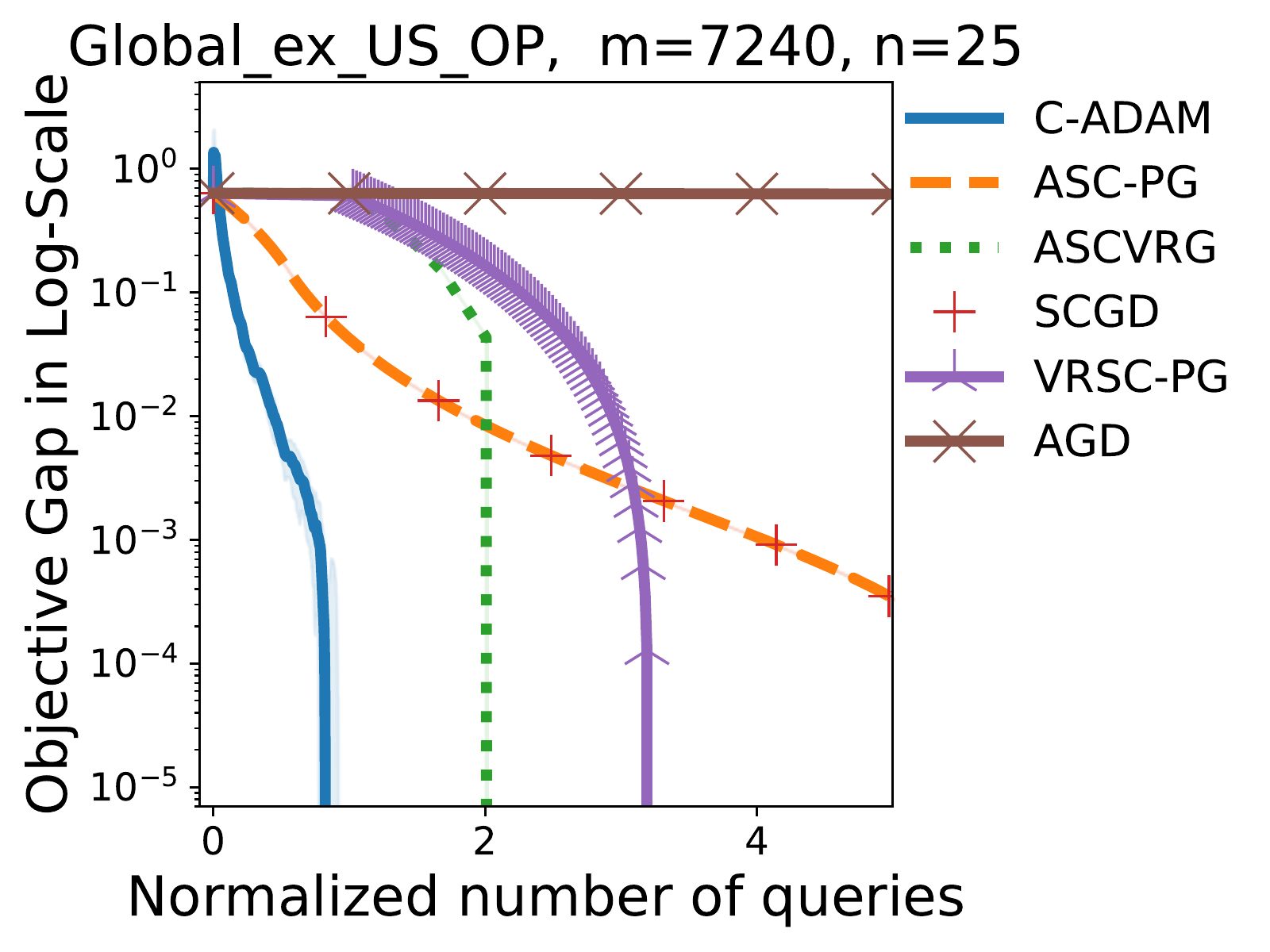}
\caption{}
\end{subfigure}
\begin{subfigure}{0.325\textwidth}
\centering
\includegraphics[width=\textwidth]{figures/Europe_OP.pdf}
\caption{}
\end{subfigure}
\begin{subfigure}{0.325\textwidth}
\centering
\includegraphics[width=\textwidth]{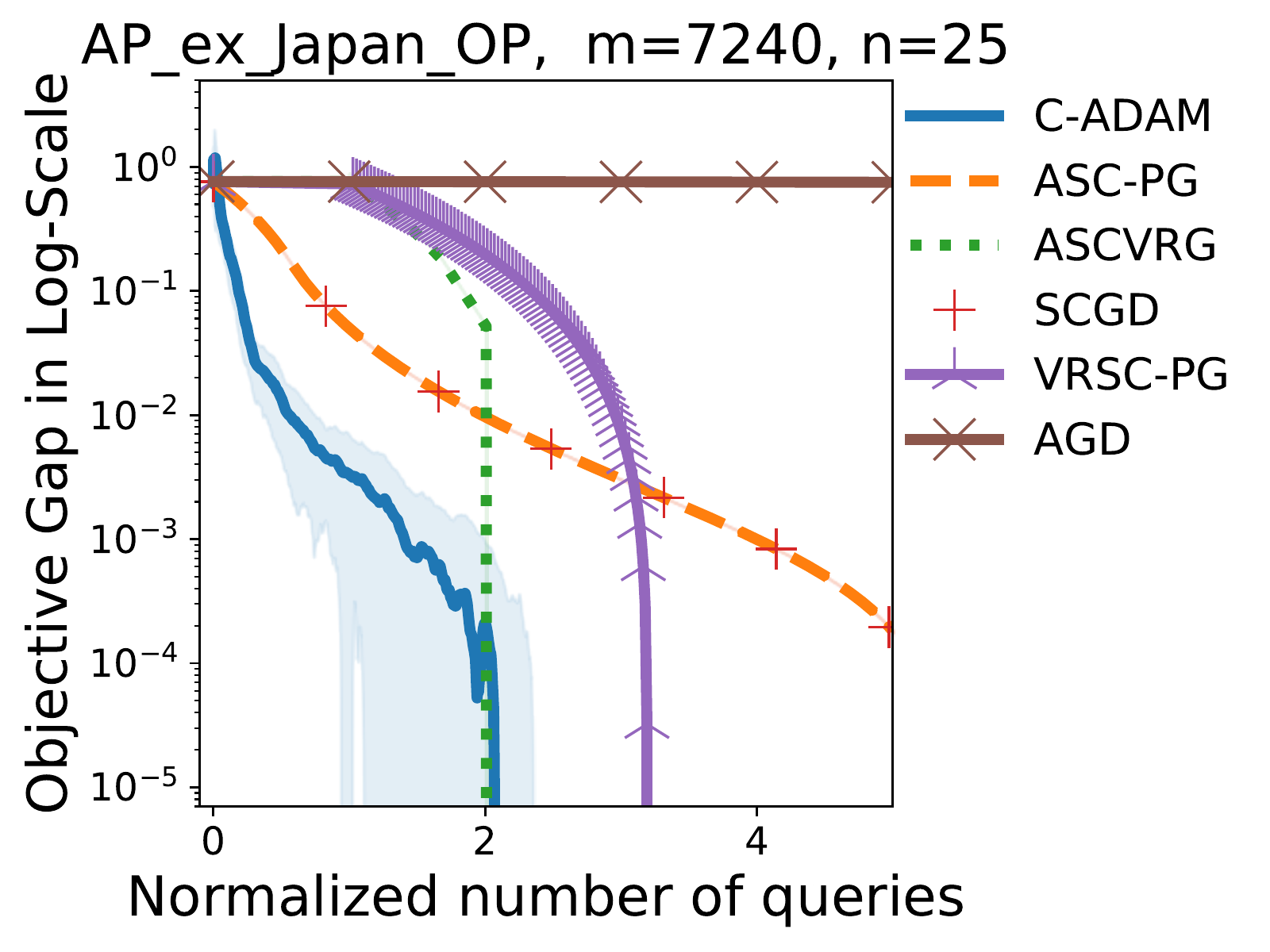}
\caption{}
\end{subfigure}
\caption{
Performance on 6 region-based Operating Profitability datasets ($m=7240, n=25$).
}
\label{fig:exp_25_op}
\end{figure*}
\begin{figure*}[t!]
\centering
\begin{subfigure}{0.325\textwidth}
\centering
\includegraphics[width=\textwidth]{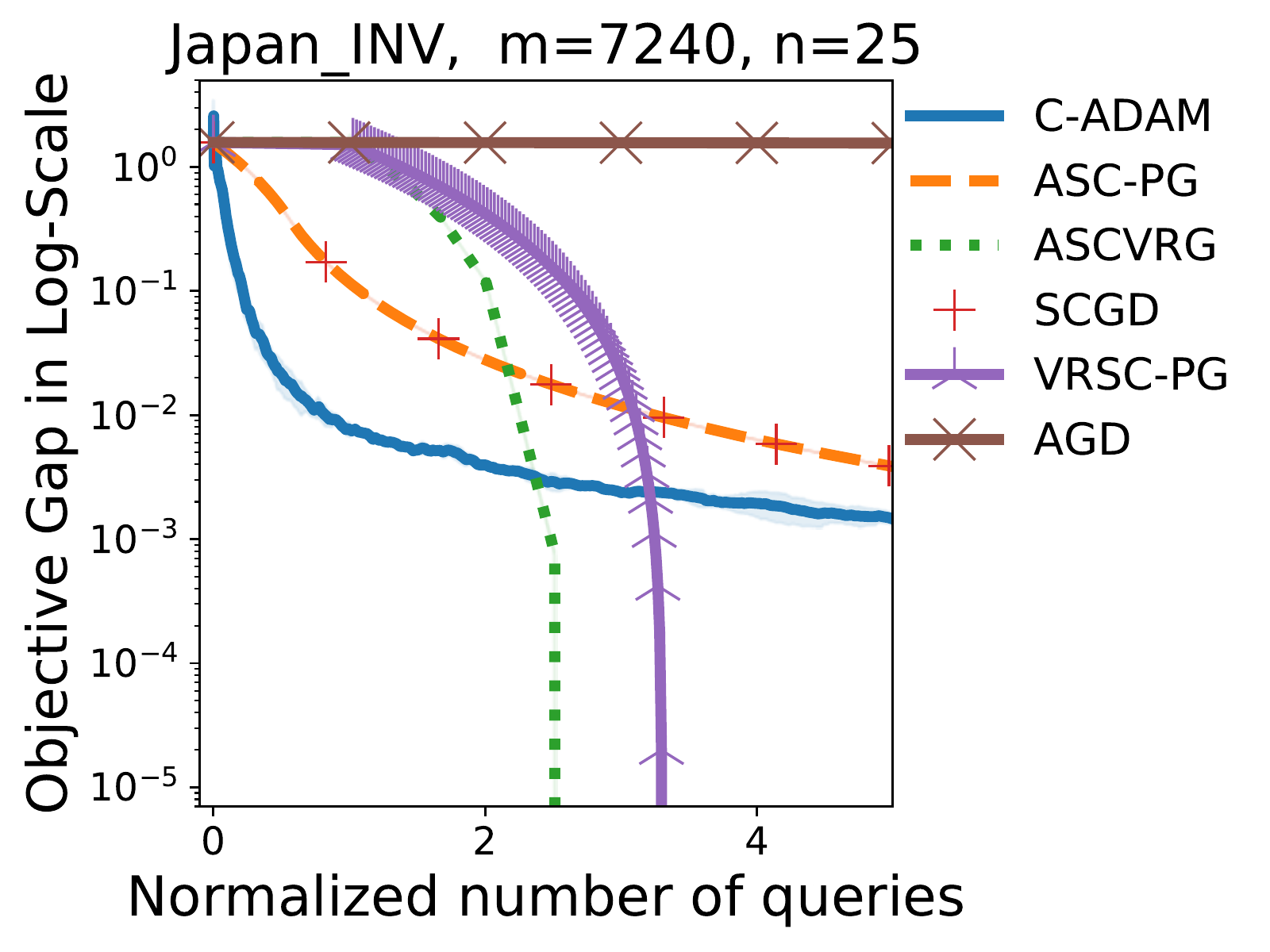}
\caption{}
\end{subfigure}
\begin{subfigure}{0.325\textwidth}
\centering
\includegraphics[width=\textwidth]{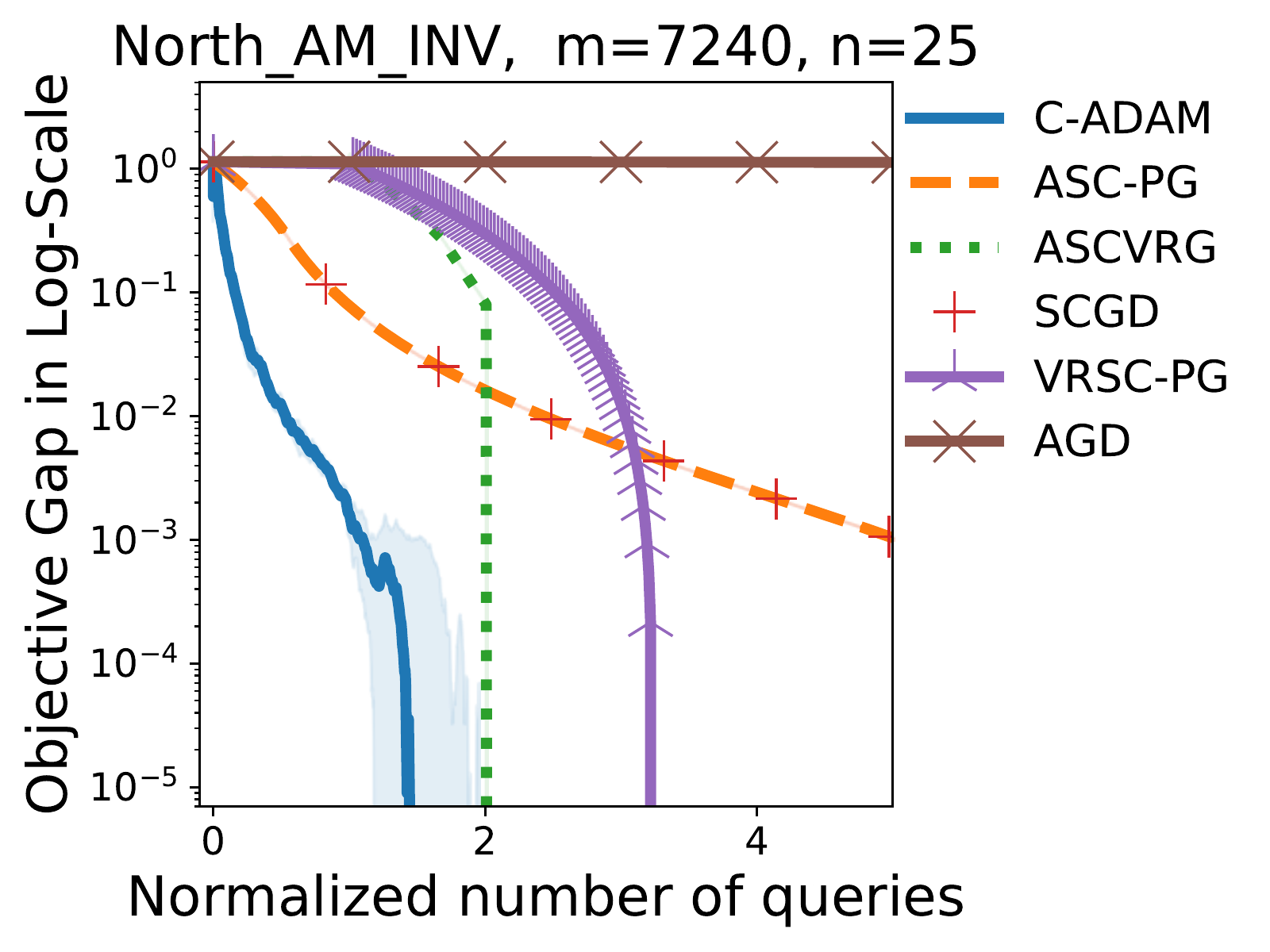}
\caption{}
\end{subfigure}
\begin{subfigure}{0.325\textwidth}
\centering
\includegraphics[width=\textwidth]{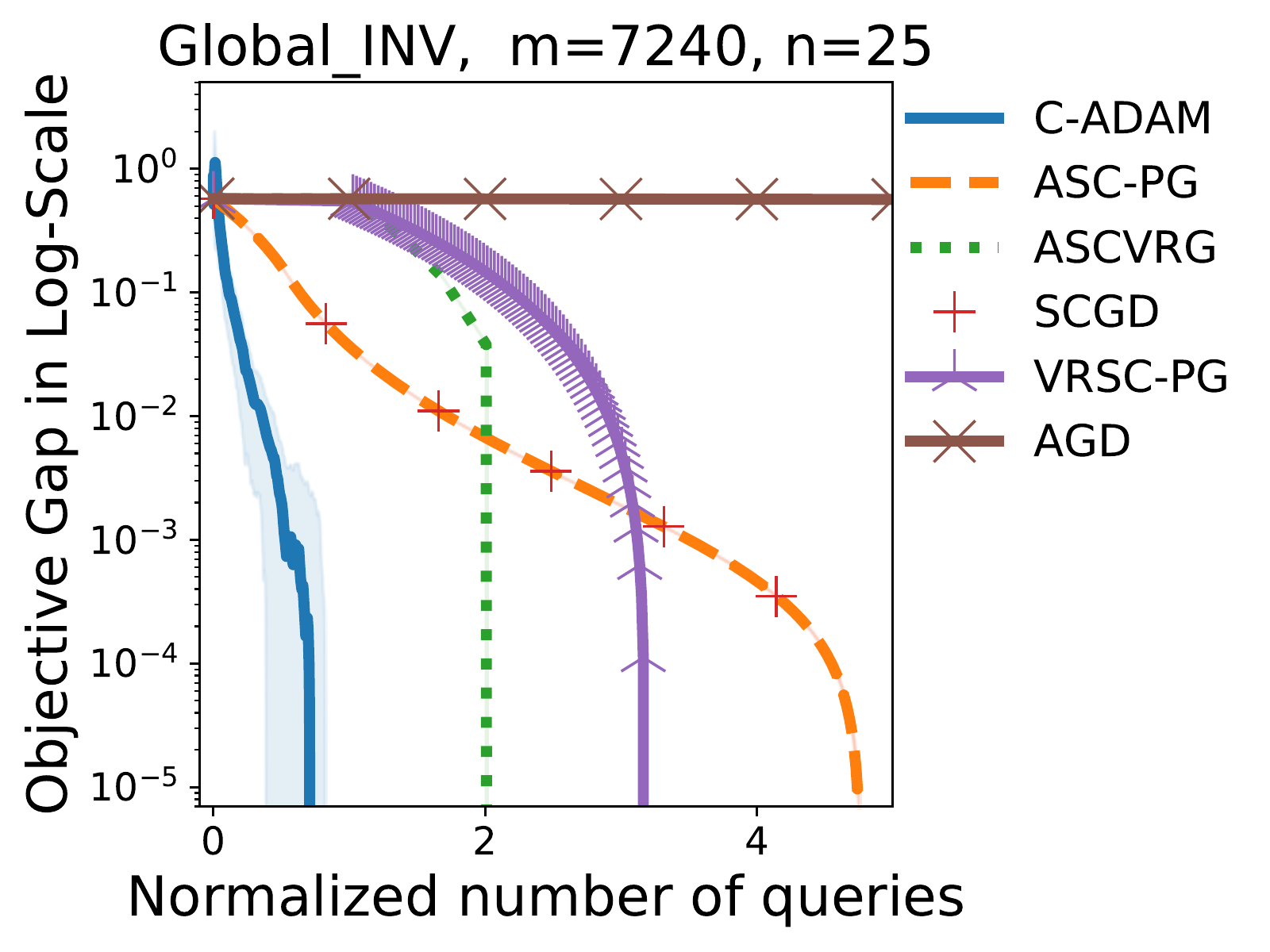}
\caption{}
\end{subfigure}
\begin{subfigure}{0.325\textwidth}
\centering
\includegraphics[width=\textwidth]{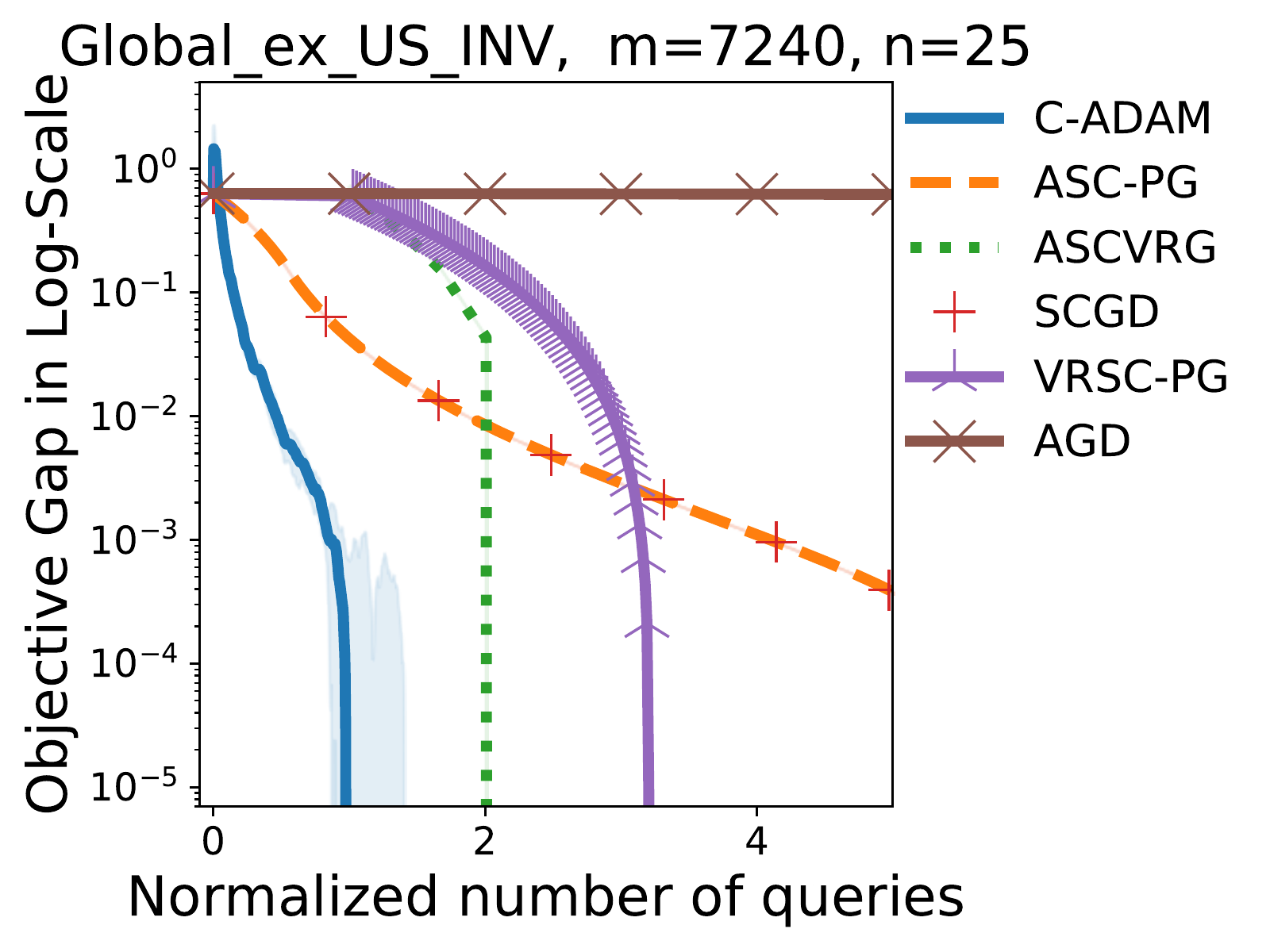}
\caption{}
\end{subfigure}
\begin{subfigure}{0.325\textwidth}
\centering
\includegraphics[width=\textwidth]{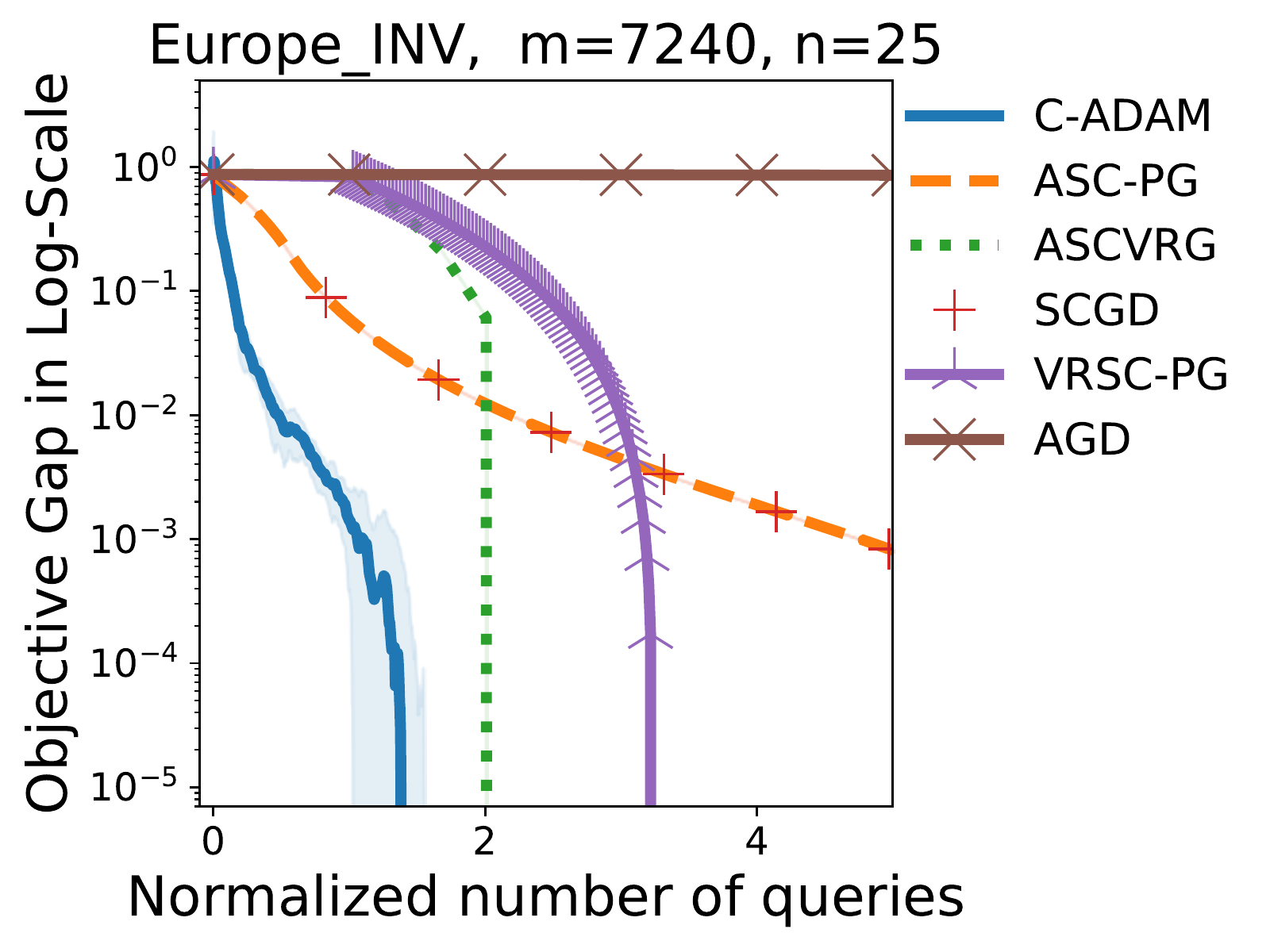}
\caption{}
\end{subfigure}
\begin{subfigure}{0.325\textwidth}
\centering
\includegraphics[width=\textwidth]{figures/Asia_Pacific_ex_Japan_INV.pdf}
\caption{}
\end{subfigure}
\caption{
Performance on 6 region-based Investment datasets ($m=7240, n=25$).
}
\label{fig:exp_25_inv}
\end{figure*}

\begin{figure*}[t!]
\centering
\begin{subfigure}{0.325\textwidth}
\centering
\includegraphics[width=\textwidth]{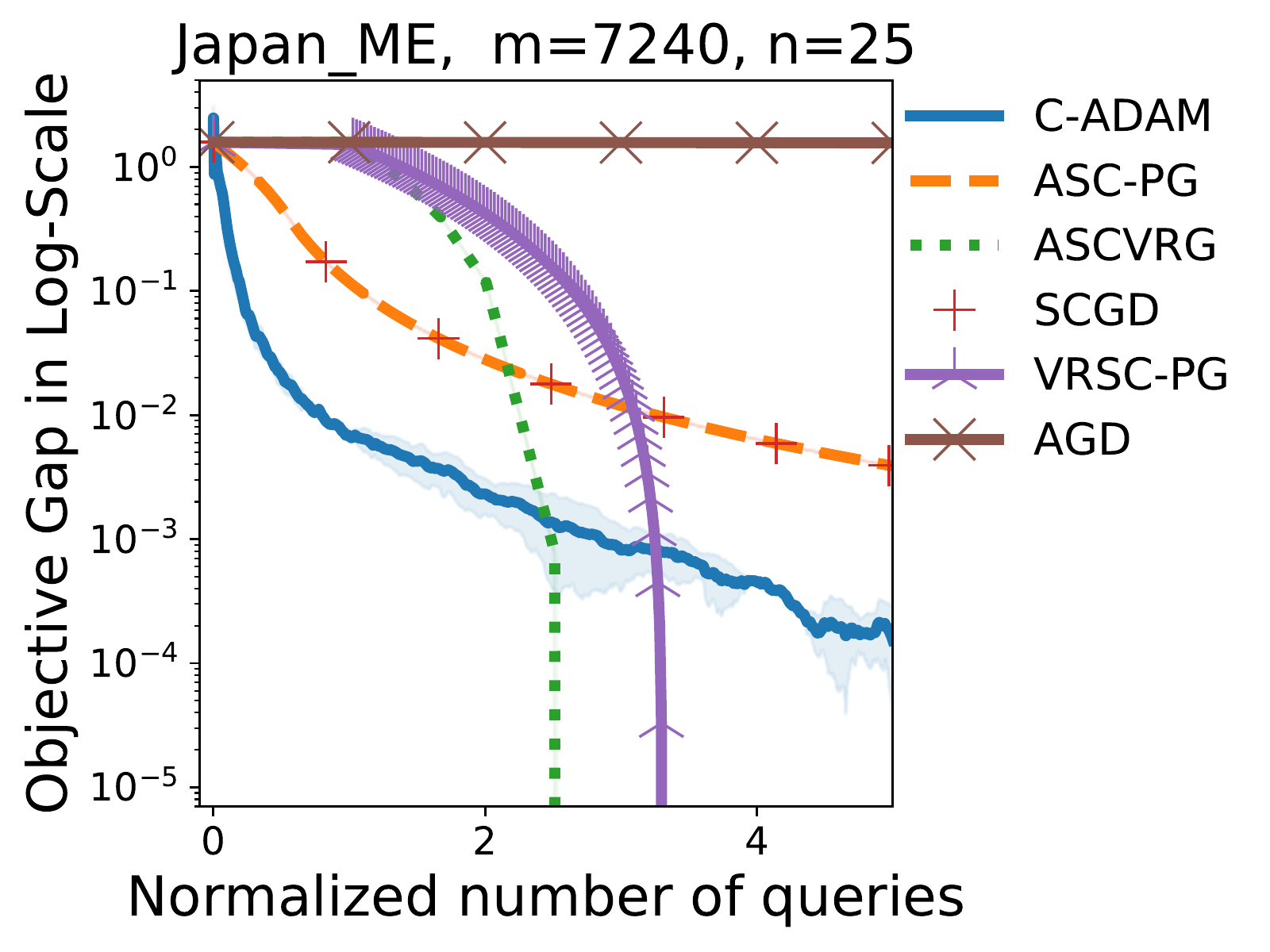}
\caption{}
\end{subfigure}
\begin{subfigure}{0.325\textwidth}
\centering
\includegraphics[width=\textwidth]{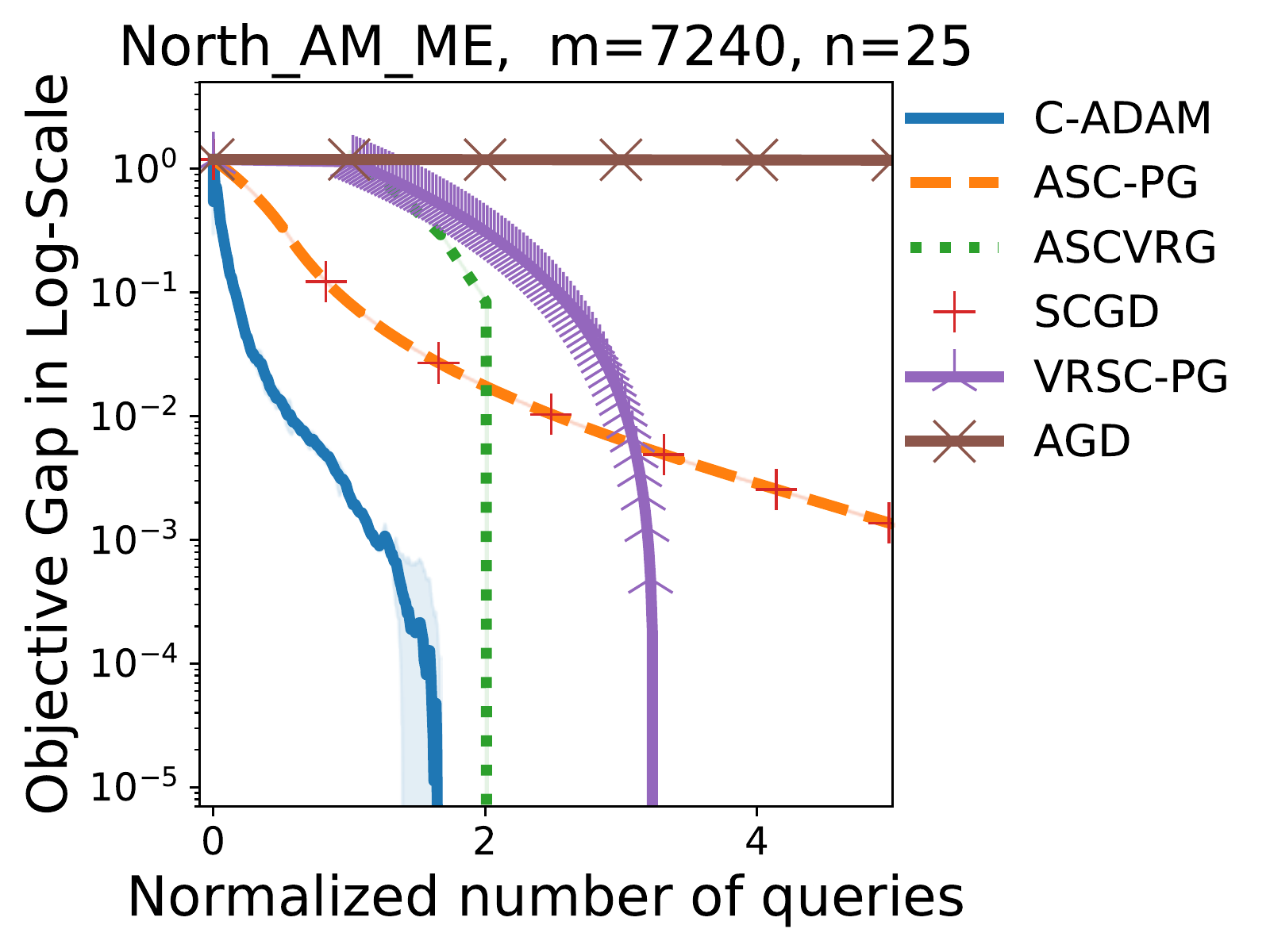}
\caption{}
\end{subfigure}
\begin{subfigure}{0.325\textwidth}
\centering
\includegraphics[width=\textwidth]{figures/Global_ME.pdf}
\caption{}
\end{subfigure}
\begin{subfigure}{0.325\textwidth}
\centering
\includegraphics[width=\textwidth]{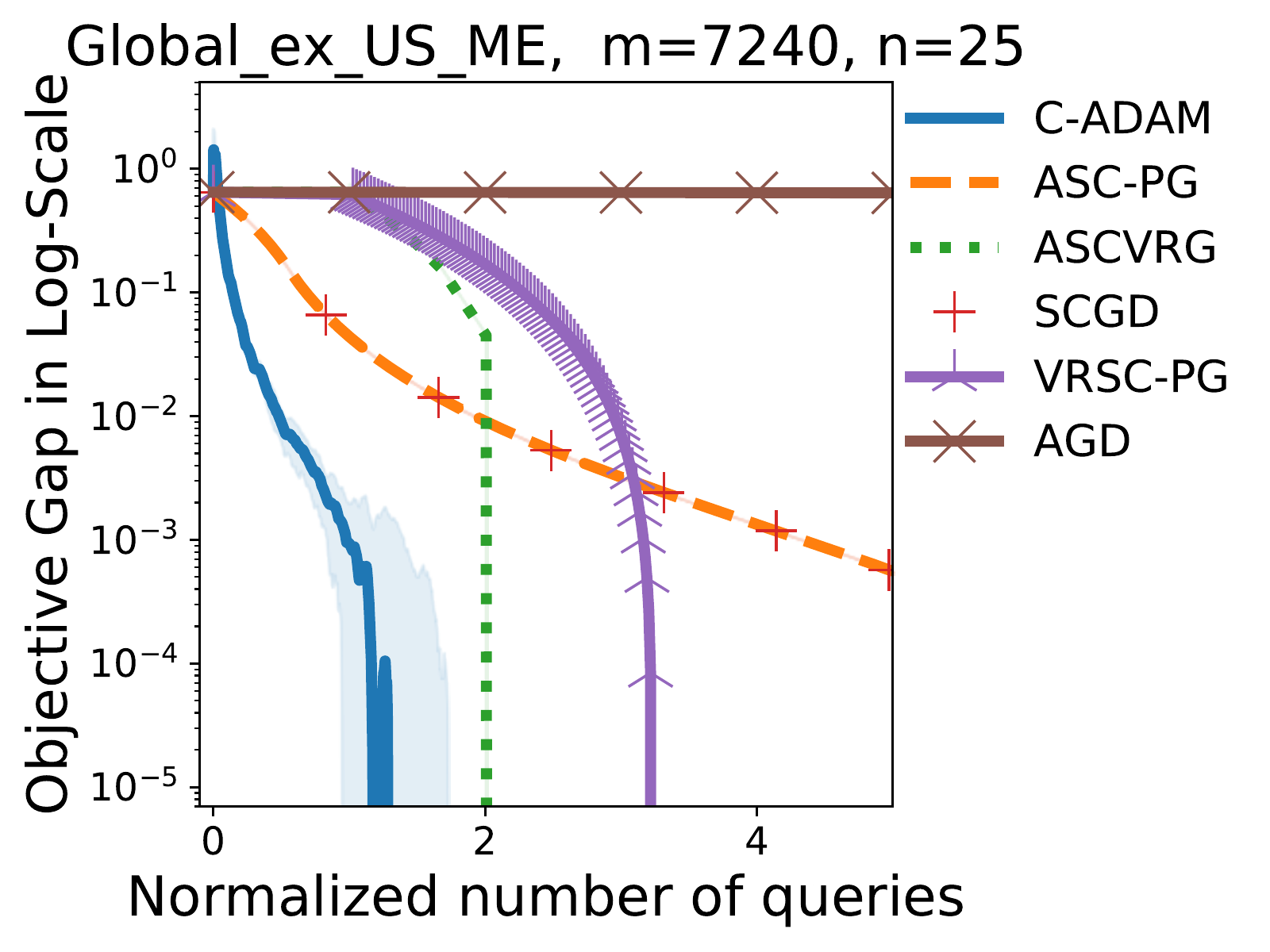}
\caption{}
\end{subfigure}
\begin{subfigure}{0.325\textwidth}
\centering
\includegraphics[width=\textwidth]{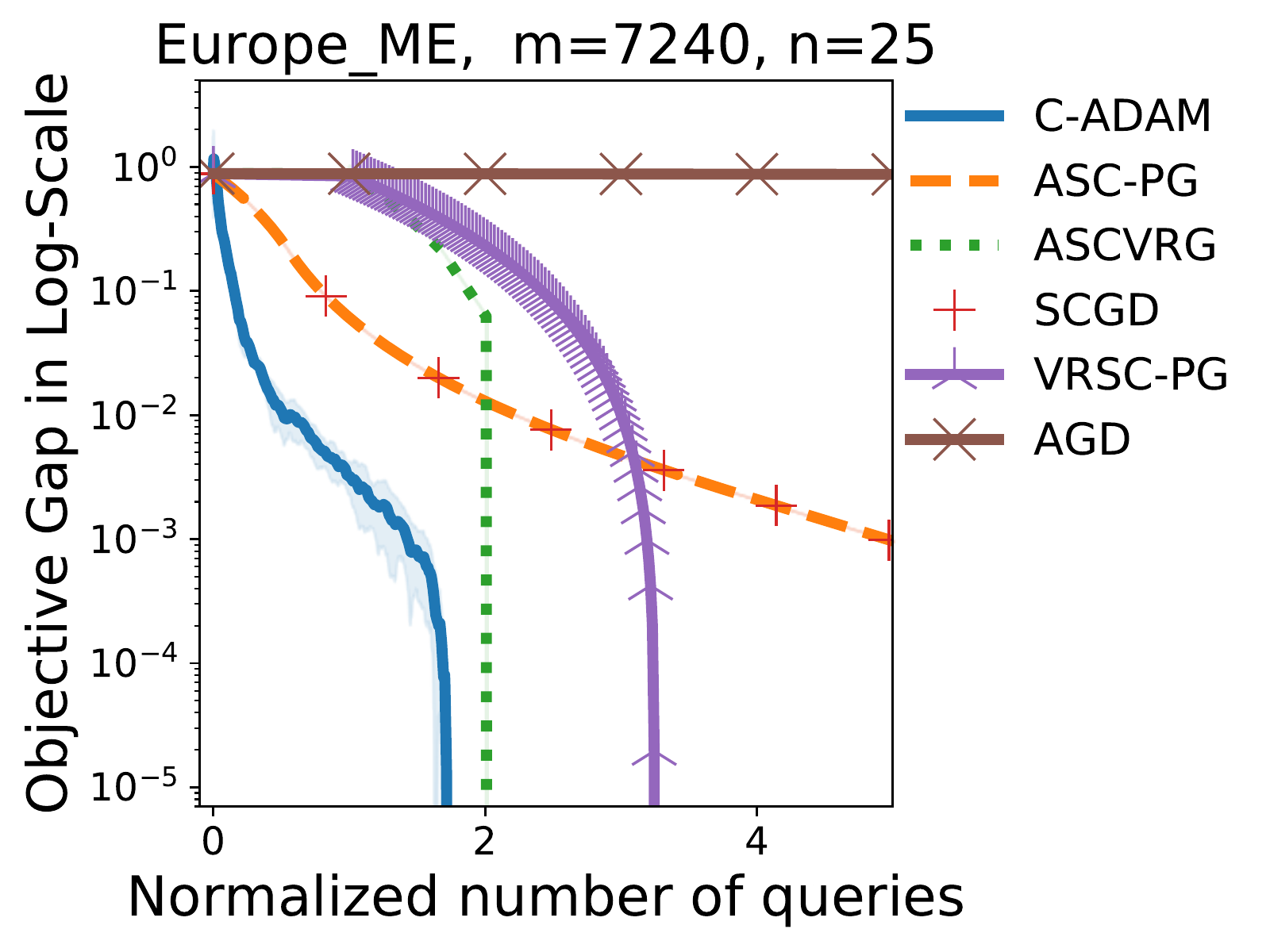}
\caption{}
\end{subfigure}
\begin{subfigure}{0.325\textwidth}
\centering
\includegraphics[width=\textwidth]{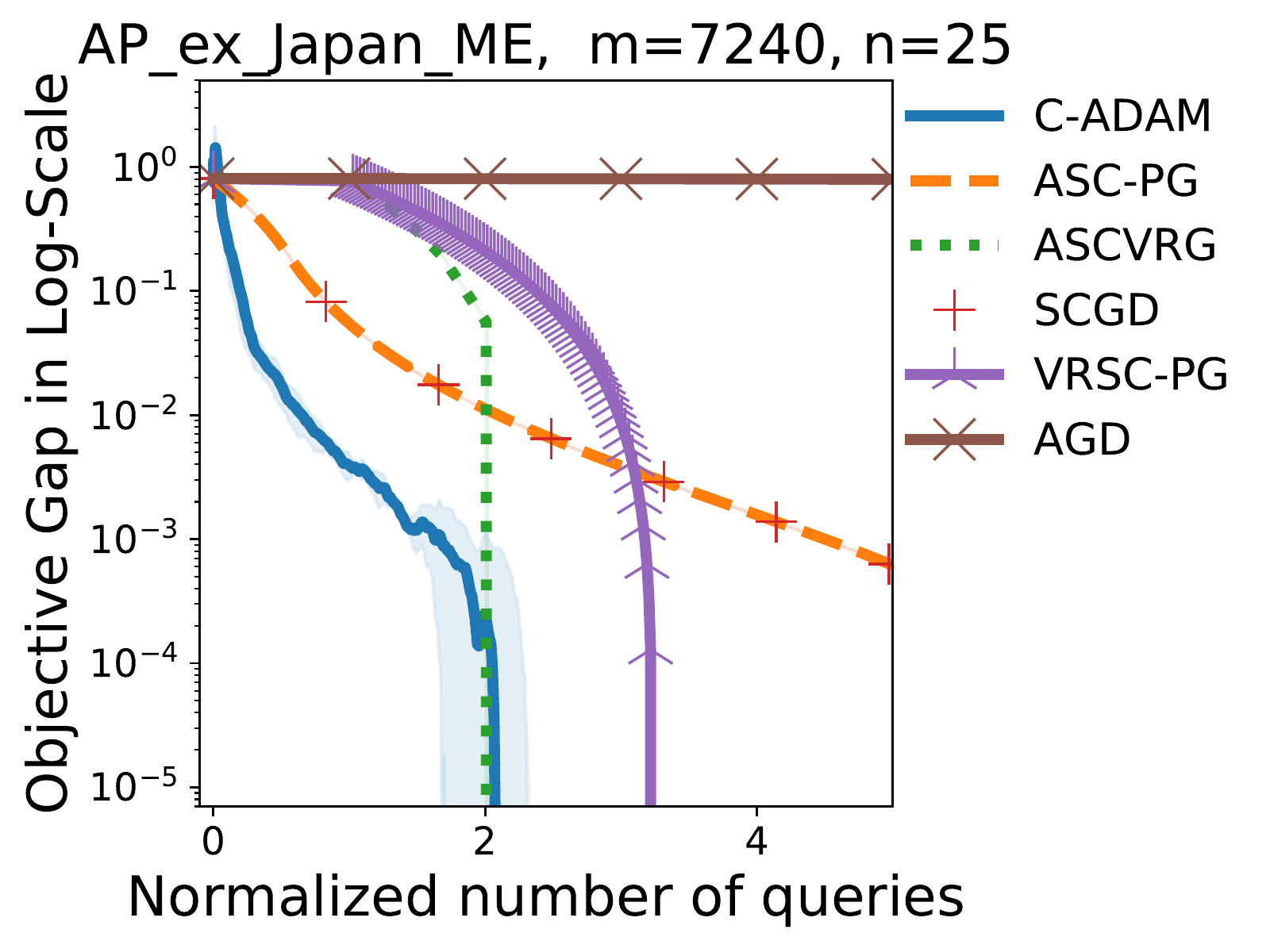}
\caption{}
\end{subfigure}
\caption{
Performance on 6 region-based Book-to-Market datasets ($m=7240, n=25$).
}
\label{fig:exp_25_bm}
\end{figure*}

\label{subsec:exp_portfolio}
Figure~\ref{fig:exp_100} shows that C-ADAM outperforms other algorithms on large datasets.
AGD performs the worst because of the highest per-iteration sample complexity on large datasets.
Figures~\ref{fig:exp_25_op},\ref{fig:exp_25_inv},\ref{fig:exp_25_bm} show that C-ADAM outperforms other algorithms on Operating Profitability, Investment, and Book-to-Market datasets, respectively.
This is consistent with the better complexity bound of C-ADAM.
Overall, C-ADAM has the potential to be a benchmark algorithm for convex composition optimization.

\subsection{Compositional MAML}
In few-shot supervised regression problem,
the goal is to predict the outputs of a sine wave $\mathcal{T}_{k}$ from only a few datapoints $\bm\xi_{obs}$ sampled from $\mathcal{P}^{(\text{data})}_{\mathcal{T}_{k}}(\cdot)$,
after training on many functions $\mathcal{T}_{k}$ using observations of the form $\bm\xi=(\bm\xi_{obs}, \bm\xi_{target})$.
The amplitude and phase of the sinusoid are varied between
tasks.
Specifically, for each task, the underlying function is
$$\xi_{target} = a \sin(\xi_{obs} + b),$$
where $a \in [0.1, 5.0]$ and $b \in [0, 2\pi]$.
The goal is to learn to find $\bm\xi_{target}$ given $\bm\xi_{obs}$ based on $M = 10$ $(\xi_{obs}, \xi_{target})$ pairs.
The loss for task $\mathcal{T}_{k}$ is represented by the mean-squared error between the model’s output for $\xi_{obs}$ and the corresponding target values $\xi_{target}$:
$$
\mathcal{L}_{\mathcal{T}_{k}}\left(\textrm{NN}(\bm{x}); \xi\right) =\sum_{(\xi_{obs}, \xi_{target}) \sim \mathcal{P}^{(\text{data})}_{\mathcal{T}_{k}}(\cdot)}\left\|\textrm{NN}_{\bm{x}}(\xi_{obs})-\xi_{target}\right\|_{2}^{2}.
$$
During training and testing, datapoints $\xi_{obs}$ are sampled uniformly from $[-5.0, 5.0]$.
Free parameters in Algorithm \ref{Algo:ADAM} were set to
$
C_{\alpha}=0.001, C_{\beta}=0.99, K_{t}^{(1)} = K_{t}^{(2)} = K_{t}^{(3)} = 10$, and $C_{\gamma} = 1$.

\section{Image Classification}
We observe the critical impact of meta-steps in MiniImagenet, as our algorithms limitation of only being able to optimize for one meta step significantly reduces overall performance for this task across all optimizers --- we wish to extend our algorithms theoretical guarantees to higher meta step scenarios in the future.

\begin{figure*}[t!]
\centering
\begin{subfigure}{0.245\textwidth}
\centering
\includegraphics[width=\textwidth]{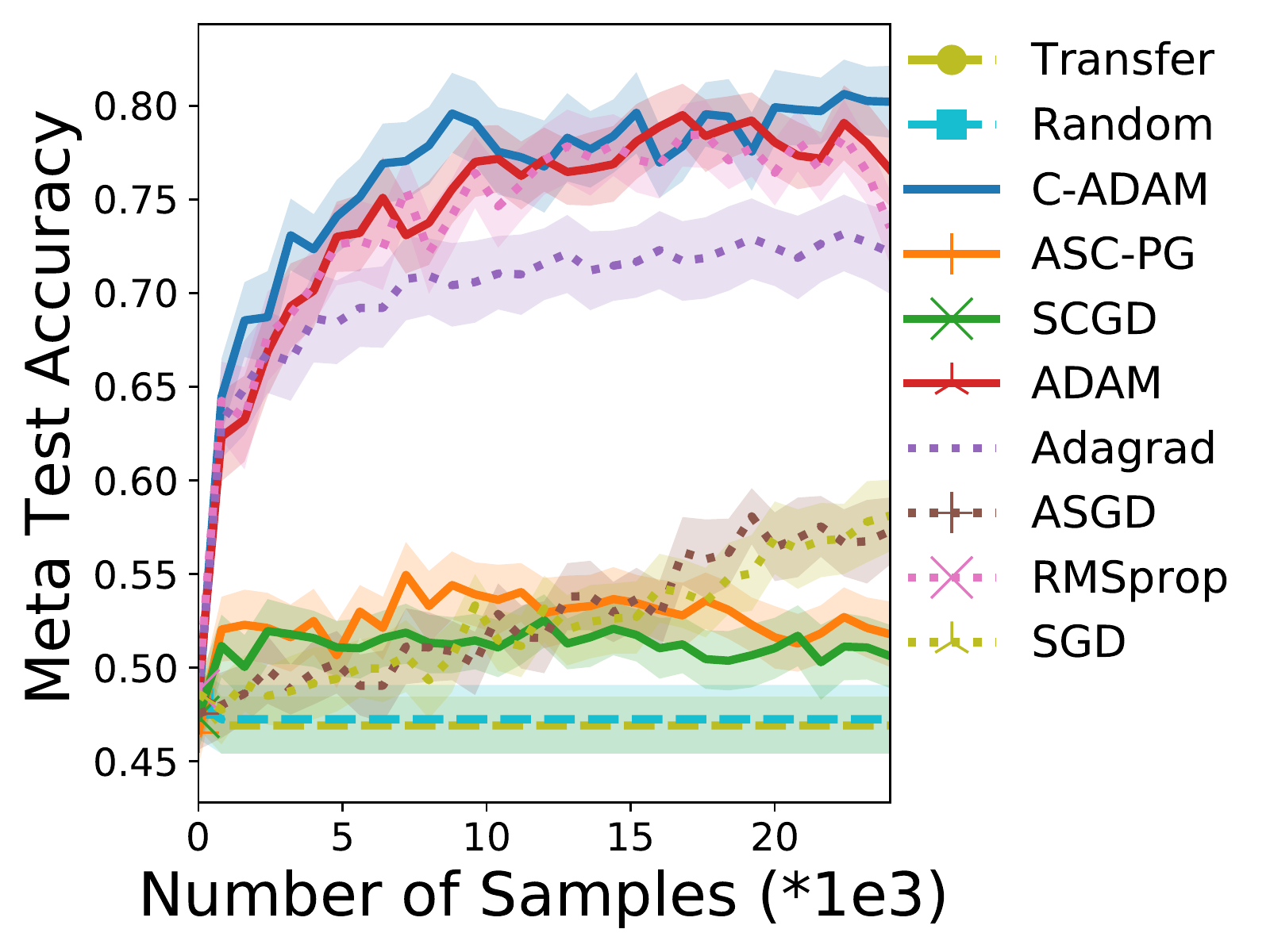}
\caption{}
\end{subfigure}
\begin{subfigure}{0.245\textwidth}
\centering
\includegraphics[width=\textwidth, trim={0 0 0 0.95cm },clip]{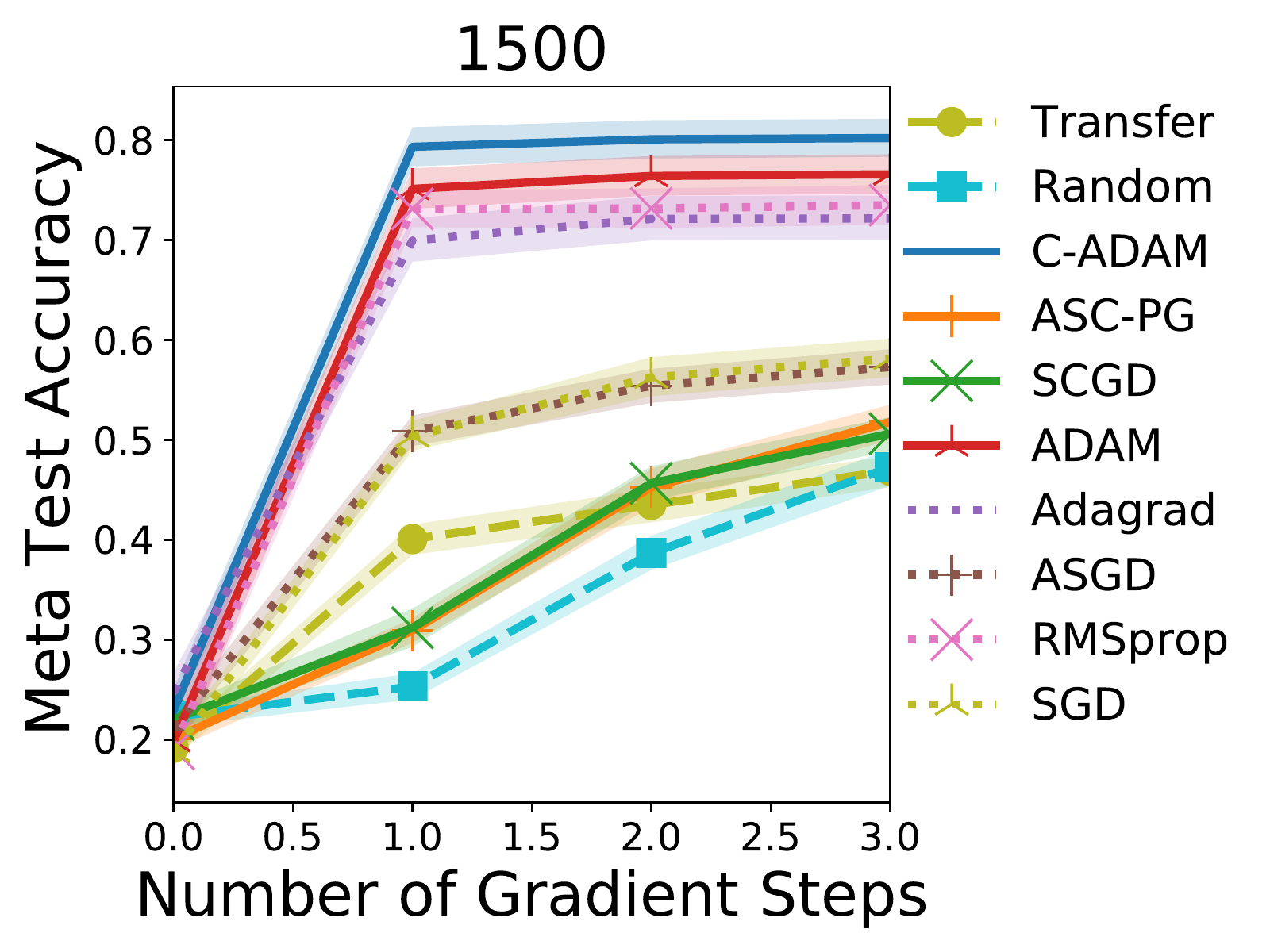}
\caption{}
\end{subfigure}
\begin{subfigure}{0.245\textwidth}
\centering
\includegraphics[width=\textwidth]{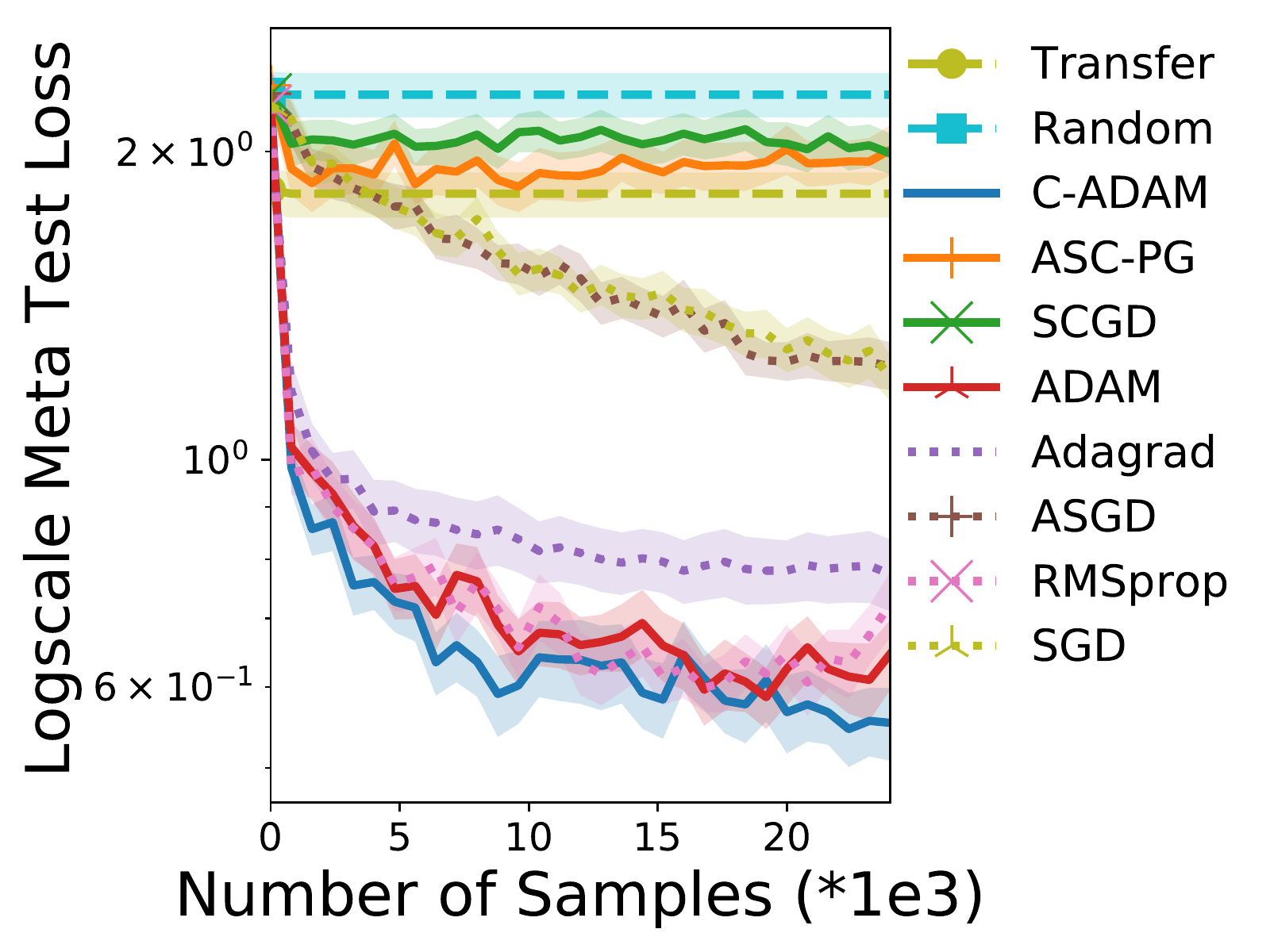}
\caption{}
\end{subfigure}
\begin{subfigure}{0.245\textwidth}
\centering
\includegraphics[width=\textwidth, trim={0 0 0 0.95cm },clip]{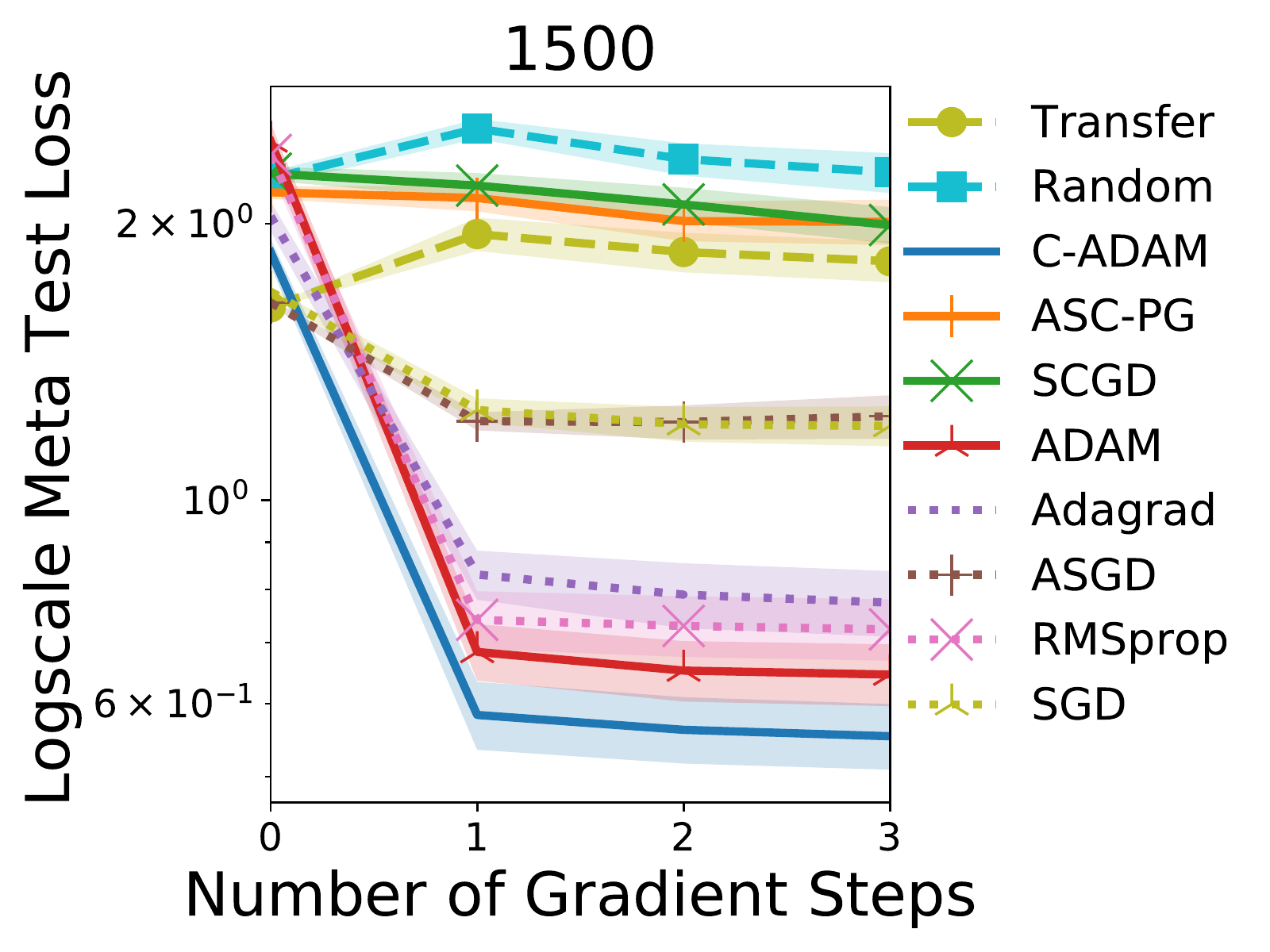}
\caption{}
\end{subfigure}
\begin{subfigure}{0.245\textwidth}
\centering
\includegraphics[width=\textwidth]{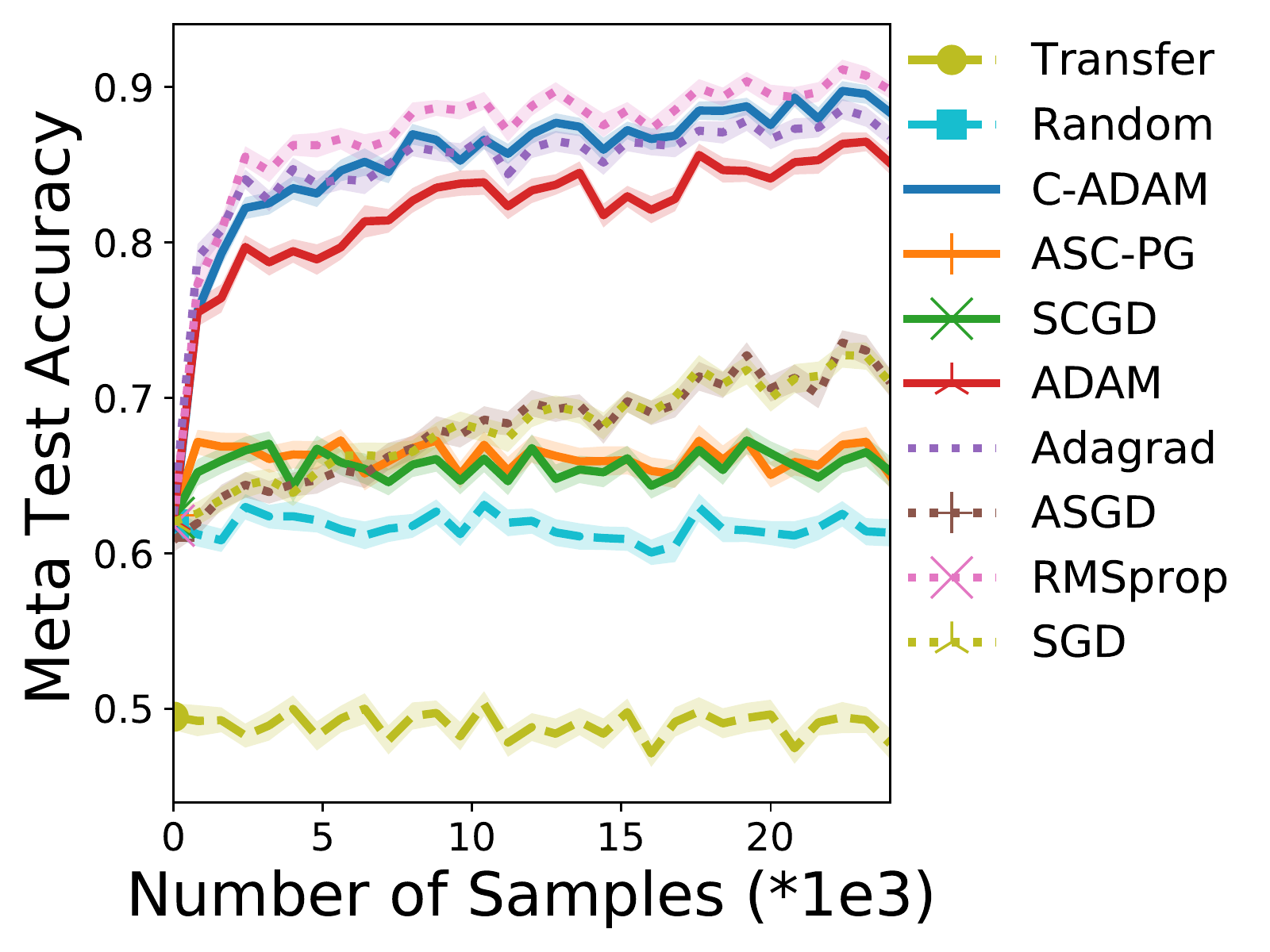}
\caption{}
\end{subfigure}
\begin{subfigure}{0.245\textwidth}
\centering
\includegraphics[width=\textwidth, trim={0 0 0 0.95cm },clip]{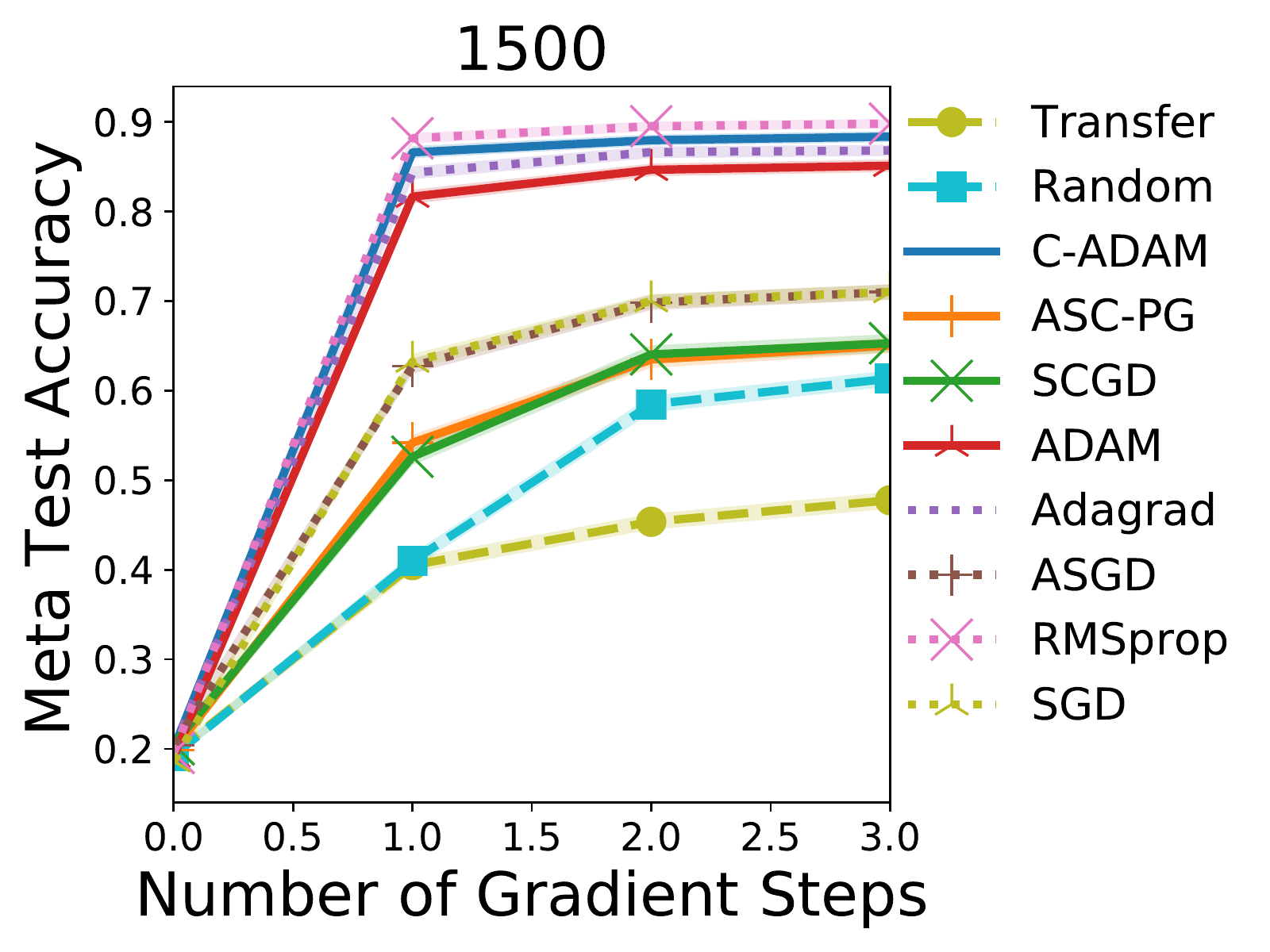}
\caption{}
\end{subfigure}
\begin{subfigure}{0.245\textwidth}
\centering
\includegraphics[width=\textwidth]{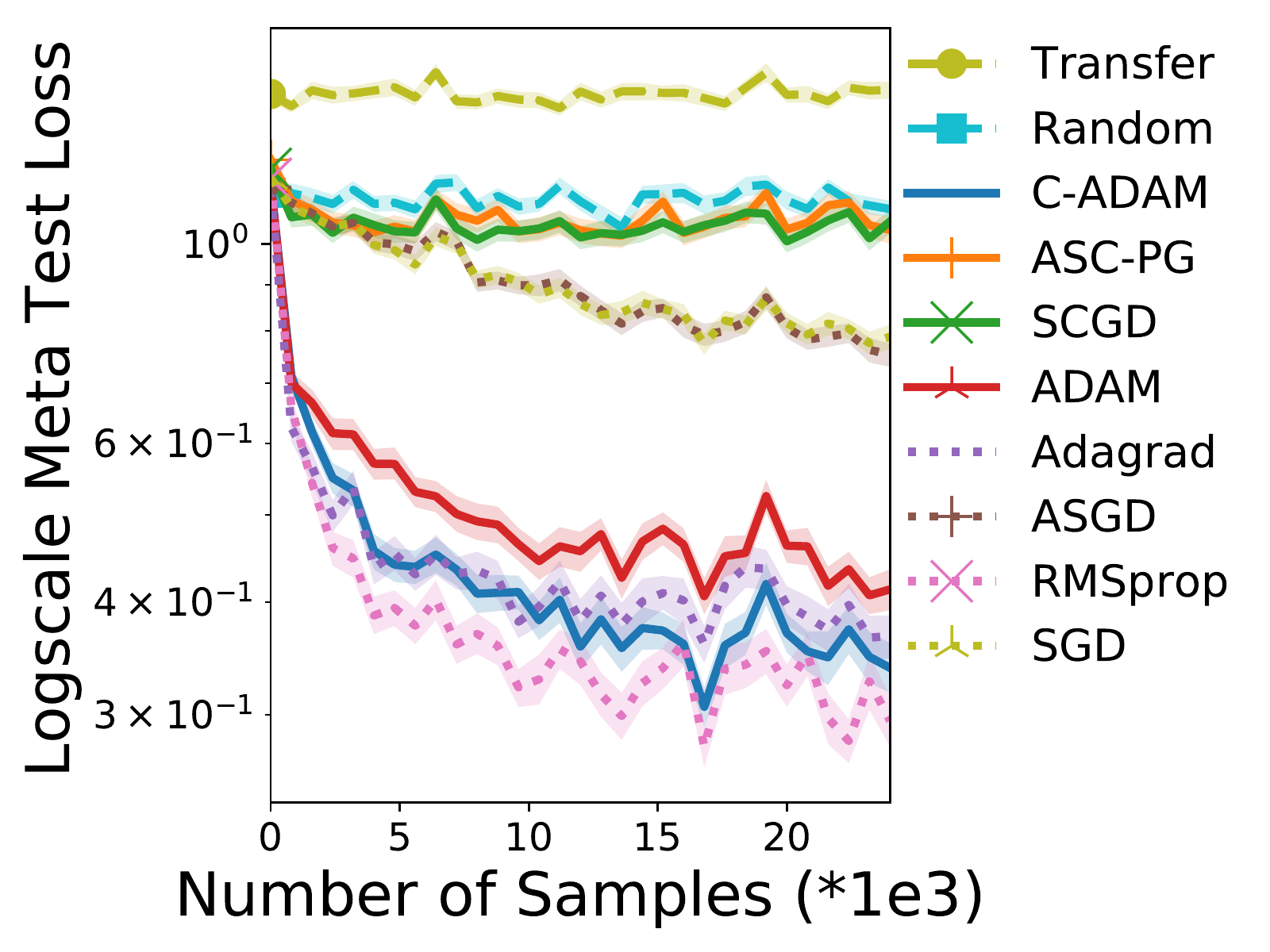}
\caption{}
\end{subfigure}
\begin{subfigure}{0.245\textwidth}
\centering
\includegraphics[width=\textwidth, trim={0 0 0 0.95cm },clip]{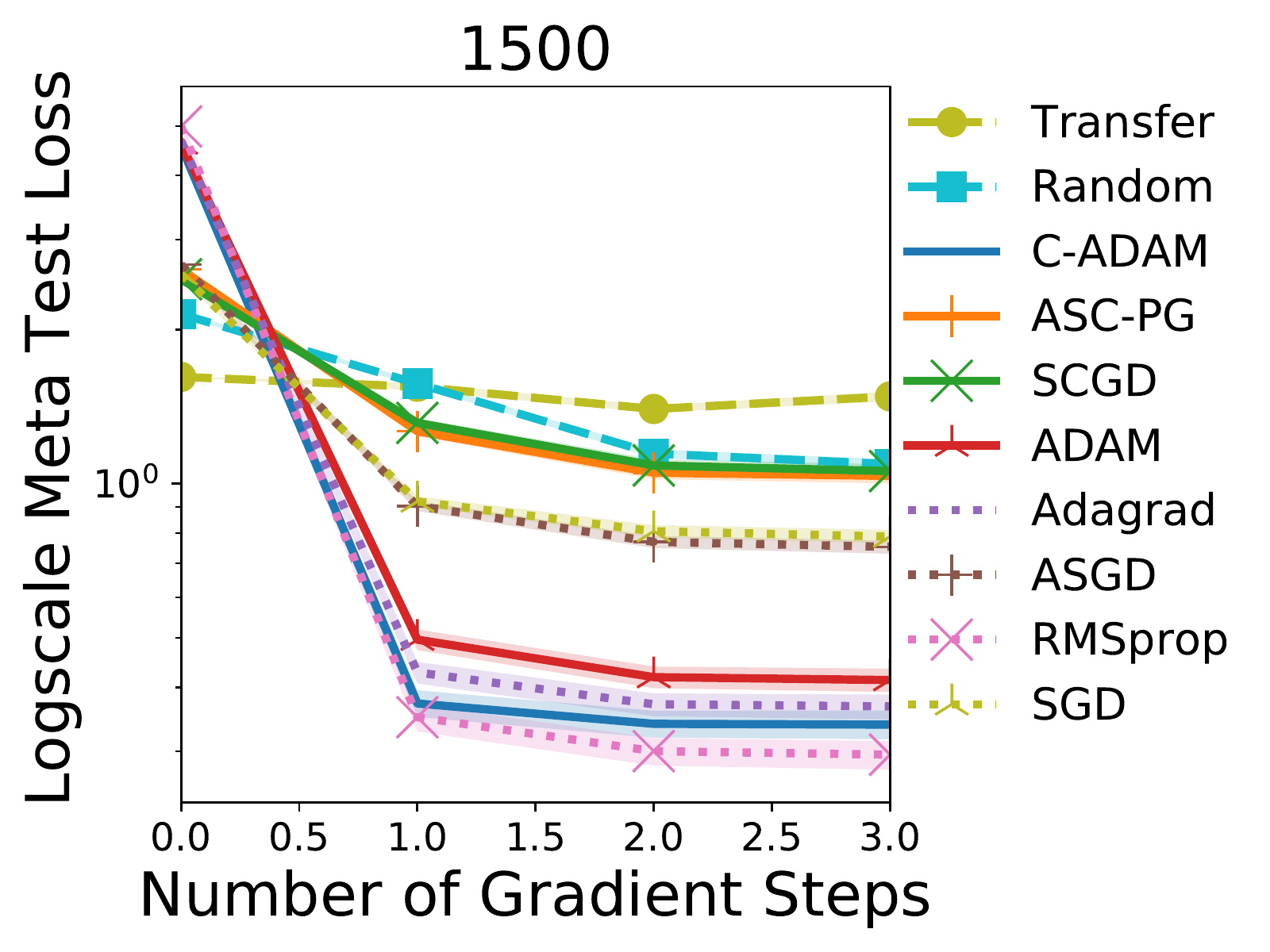}
\caption{}
\end{subfigure}
\begin{subfigure}{0.245\textwidth}
\centering
\includegraphics[width=\textwidth]{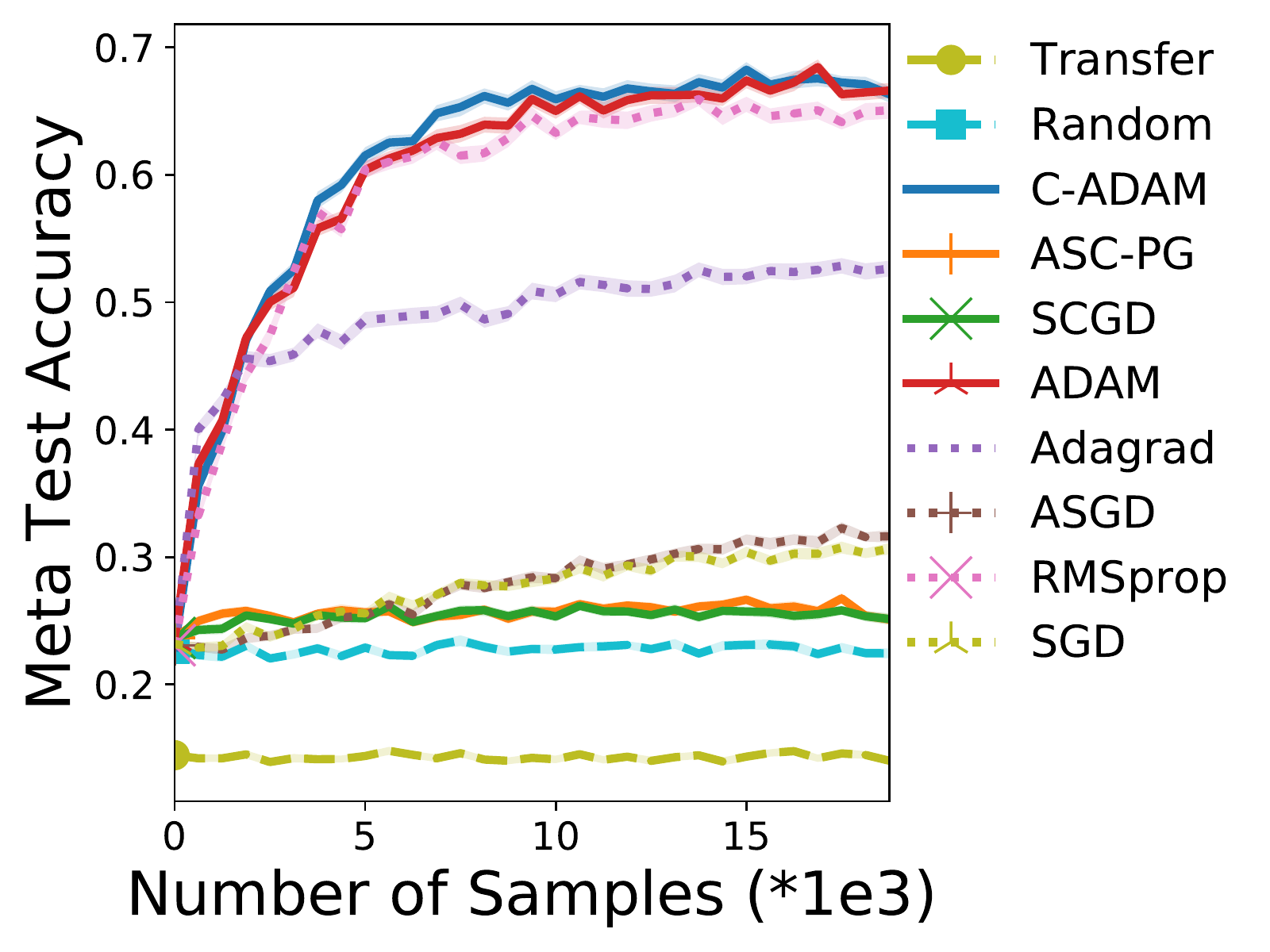}
\caption{}
\end{subfigure}
\begin{subfigure}{0.245\textwidth}
\centering
\includegraphics[width=\textwidth, trim={0 0 0 0.95cm },clip]{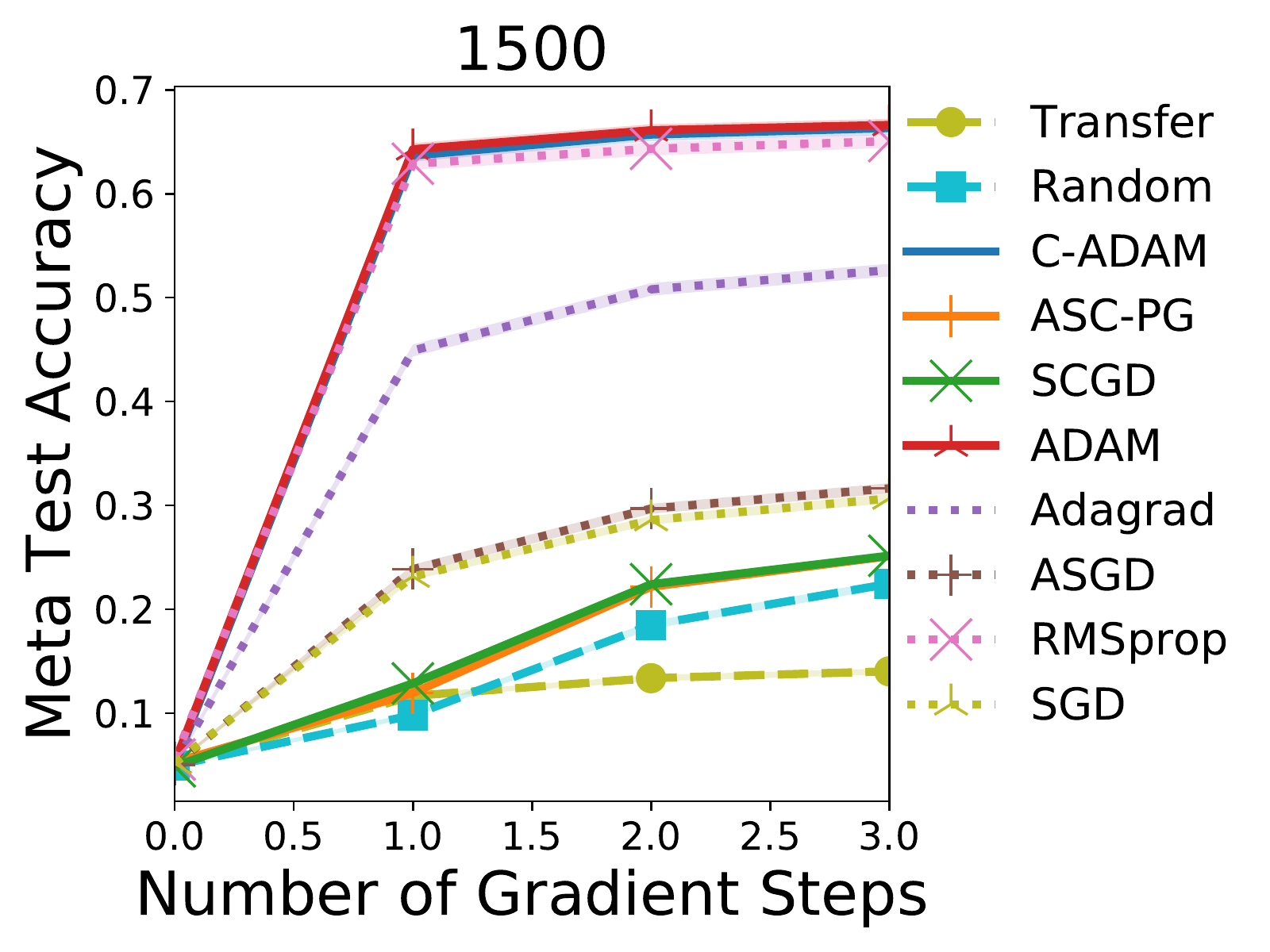}
\caption{}
\end{subfigure}
\begin{subfigure}{0.245\textwidth}
\centering
\includegraphics[width=\textwidth]{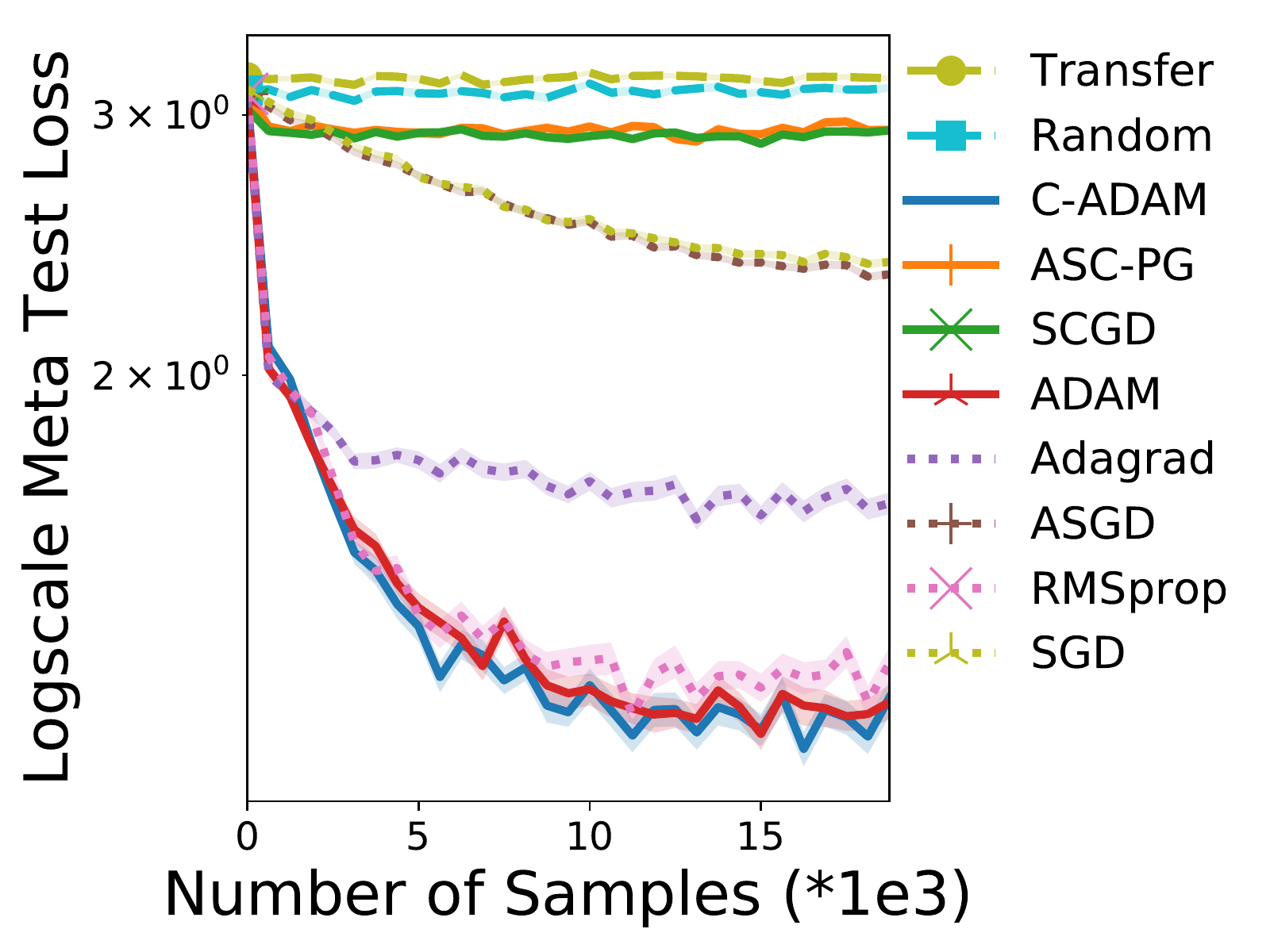}
\caption{}
\end{subfigure}
\begin{subfigure}{0.245\textwidth}
\centering
\includegraphics[width=\textwidth, trim={0 0 0 0.95cm },clip]{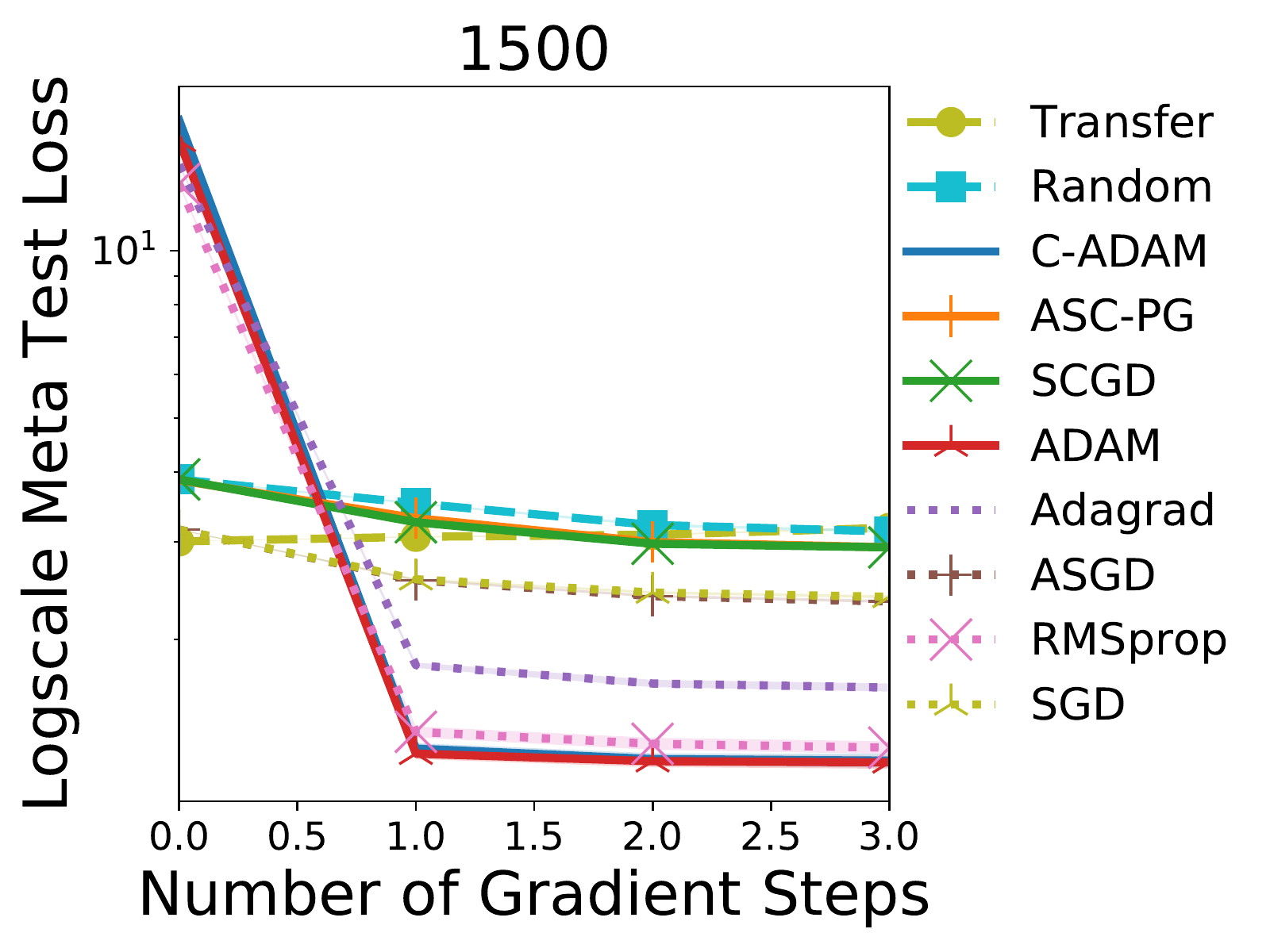}
\caption{}
\end{subfigure}
\begin{subfigure}{0.245\textwidth}
\centering
\includegraphics[width=\textwidth]{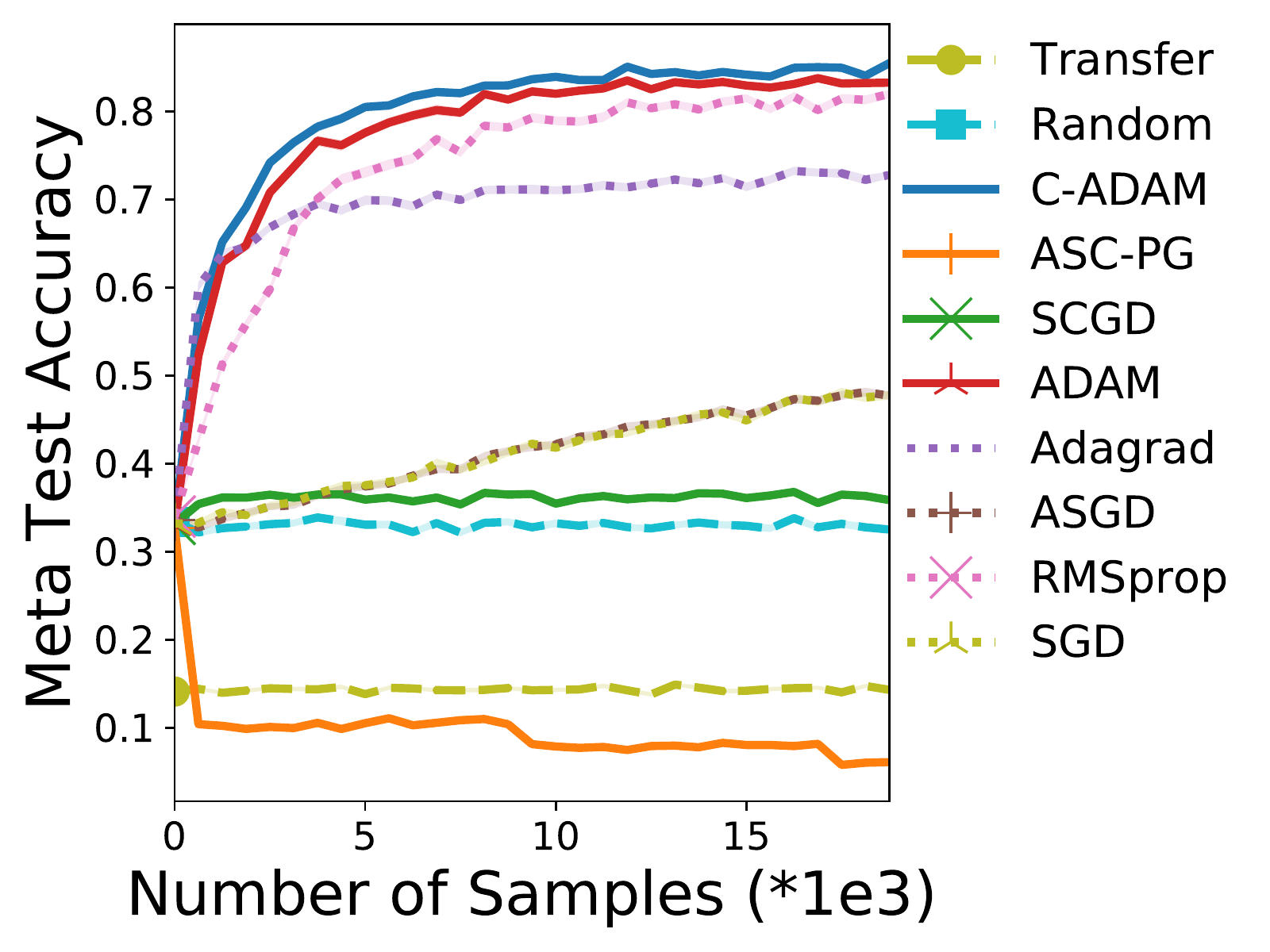}
\caption{}
\end{subfigure}
\begin{subfigure}{0.245\textwidth}
\centering
\includegraphics[width=\textwidth, trim={0 0 0 0.95cm },clip]{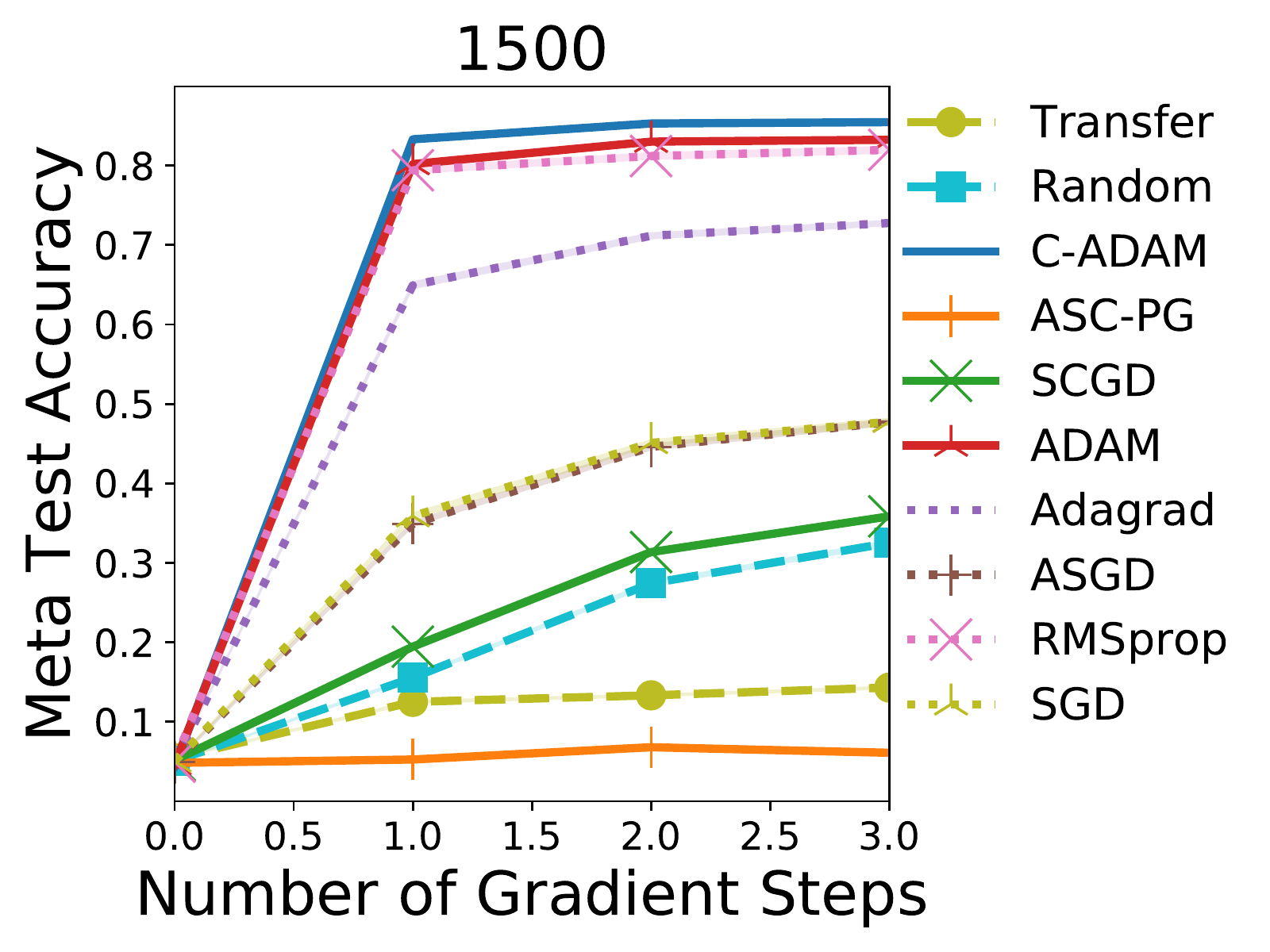}
\caption{}
\end{subfigure}
\begin{subfigure}{0.245\textwidth}
\centering
\includegraphics[width=\textwidth]{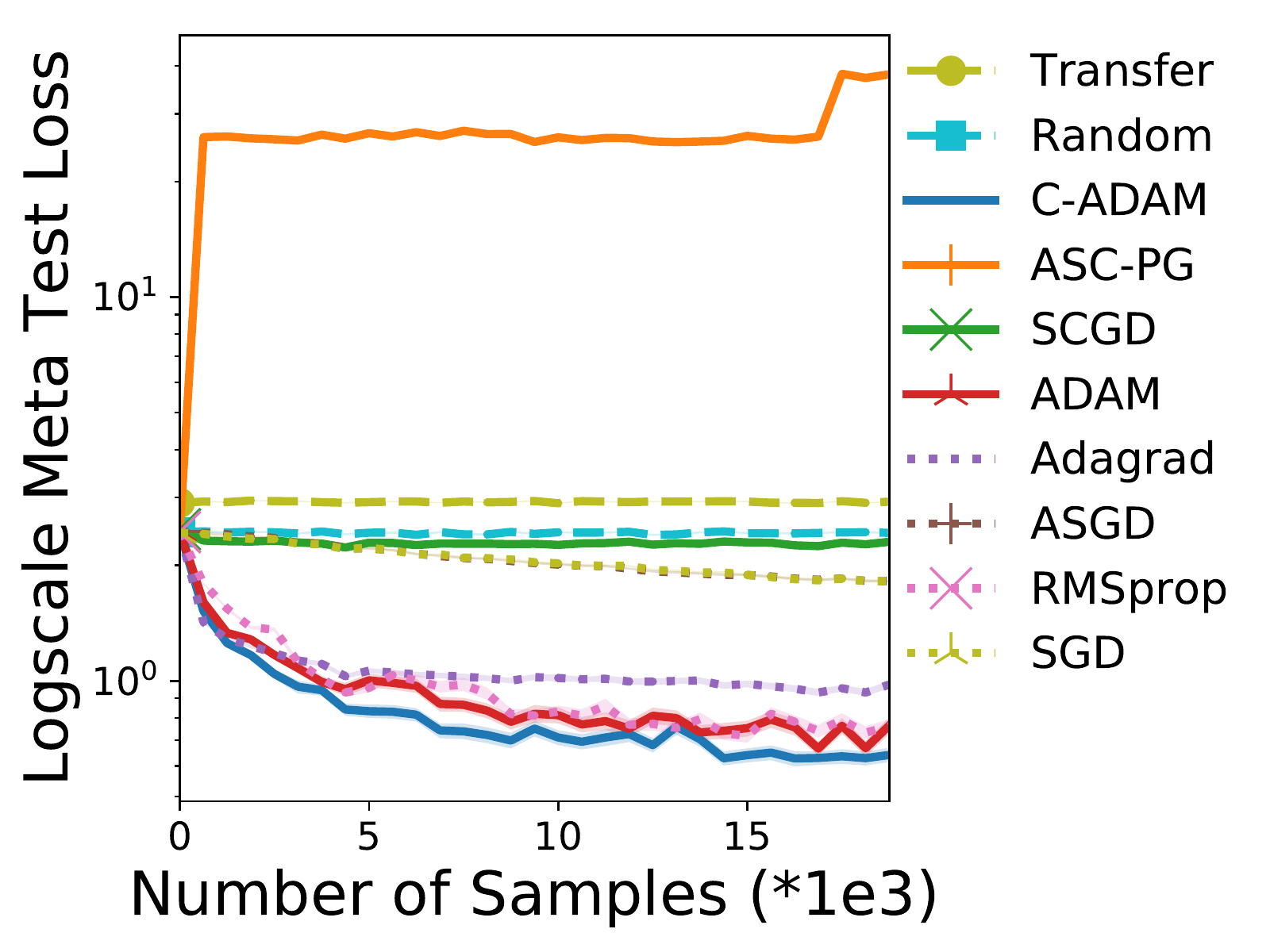}
\caption{}
\end{subfigure}
\begin{subfigure}{0.245\textwidth}
\centering
\includegraphics[width=\textwidth, trim={0 0 0 0.95cm },clip]{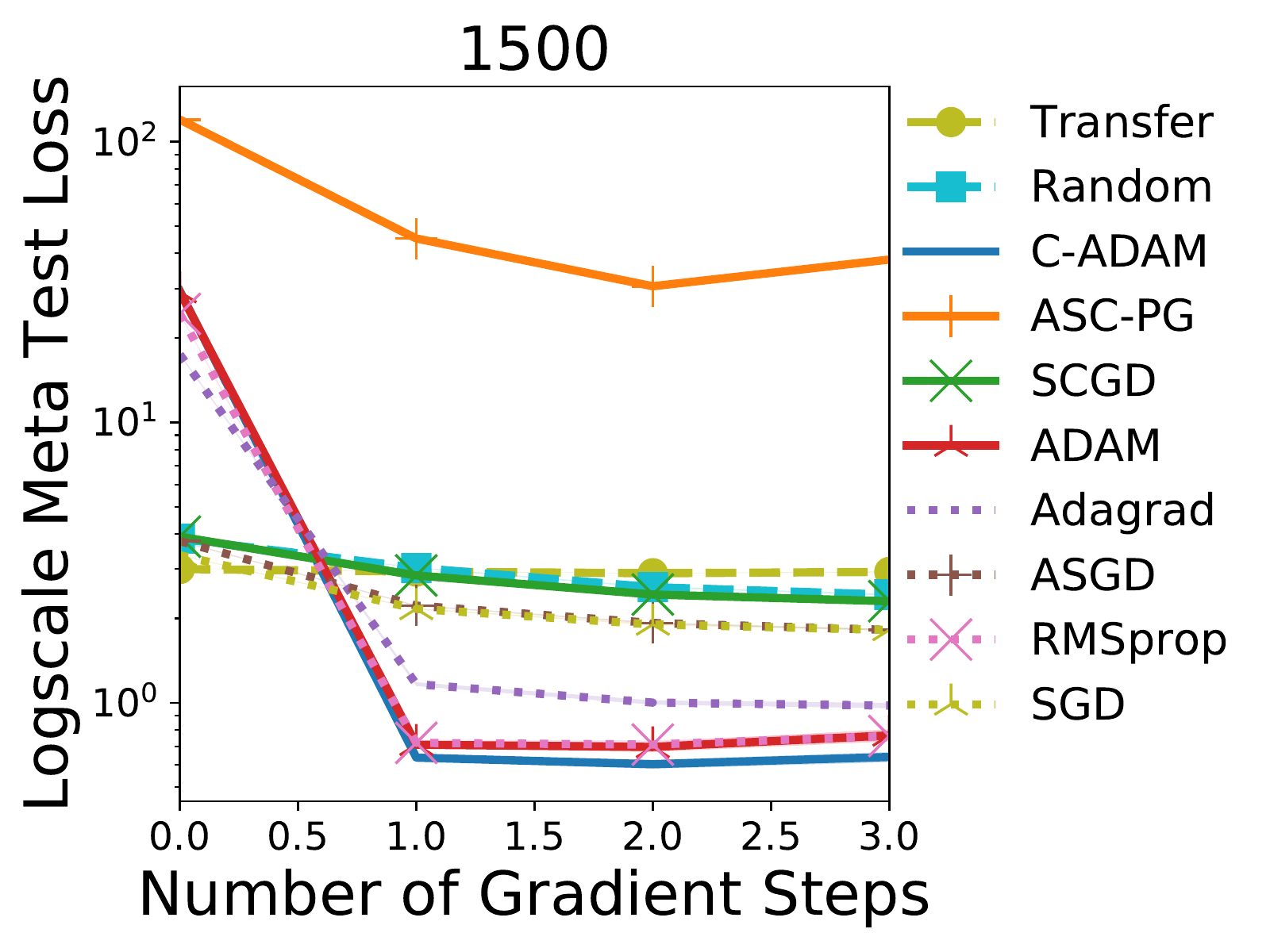}
\caption{}
\end{subfigure}
\begin{subfigure}{0.245\textwidth}
\centering
\includegraphics[width=\textwidth]{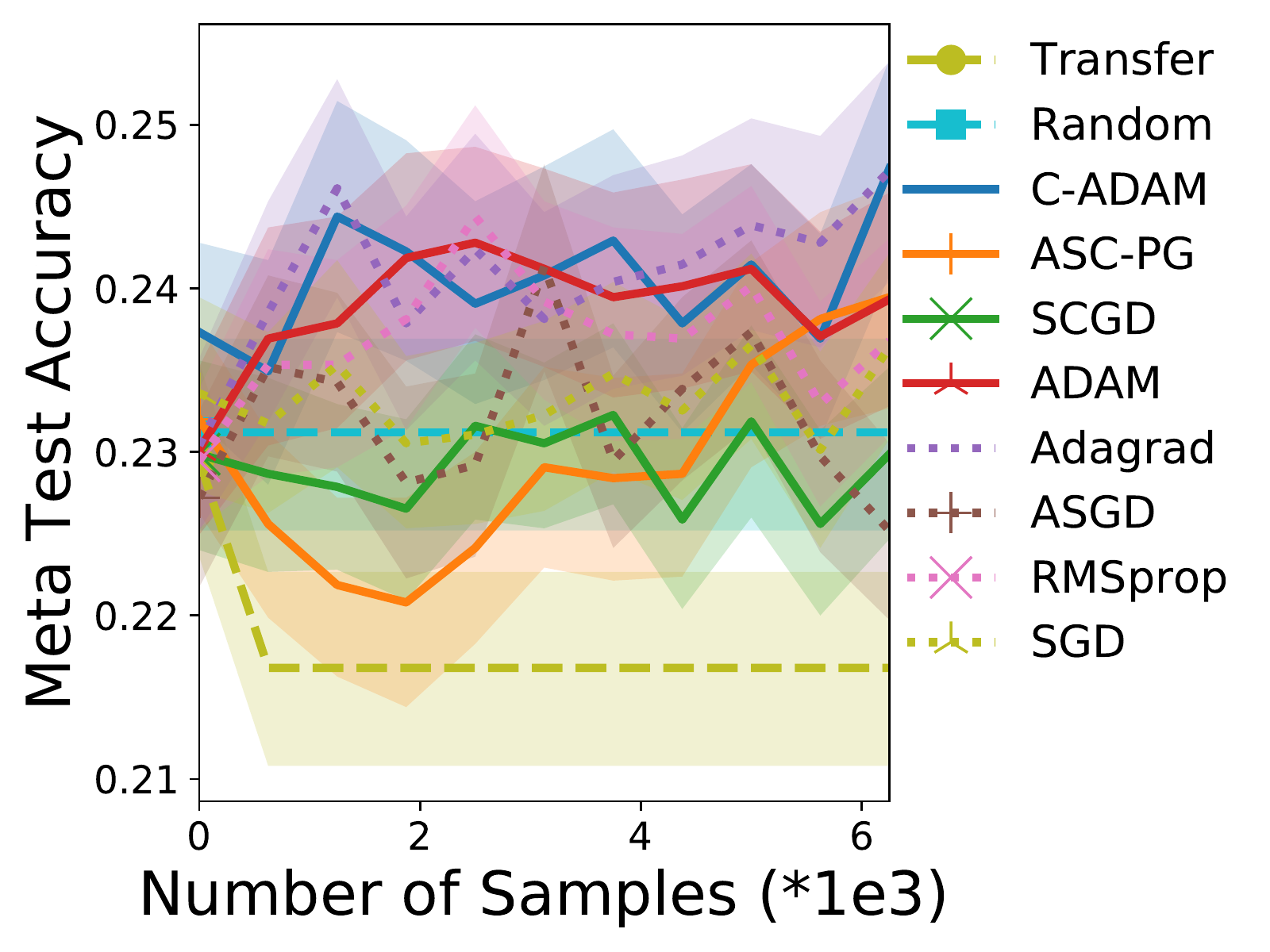}
\caption{}
\end{subfigure}
\begin{subfigure}{0.245\textwidth}
\centering
\includegraphics[width=\textwidth, trim={0 0 0 0.95cm },clip]{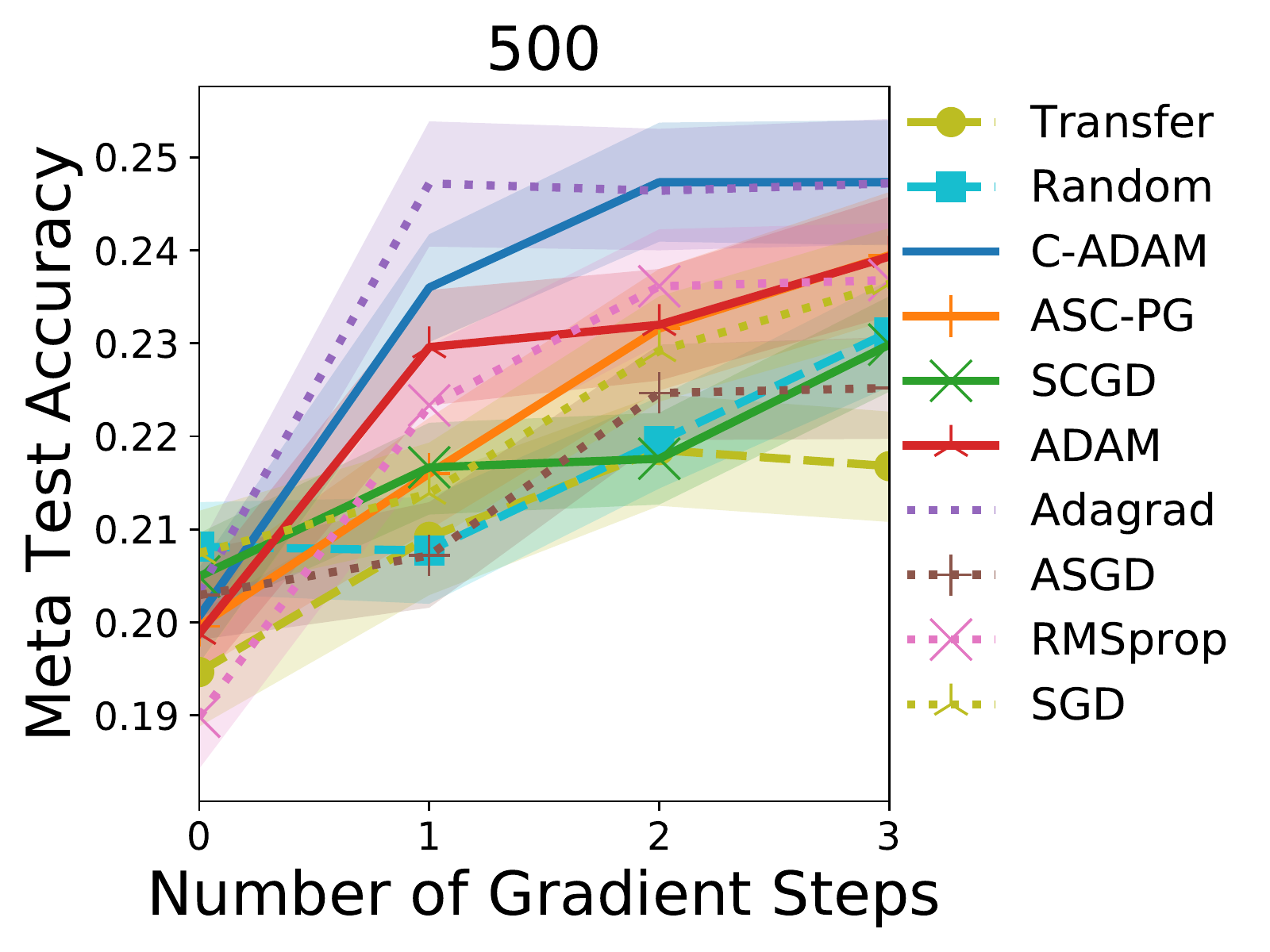}
\caption{}
\end{subfigure}
\begin{subfigure}{0.245\textwidth}
\centering
\includegraphics[width=\textwidth]{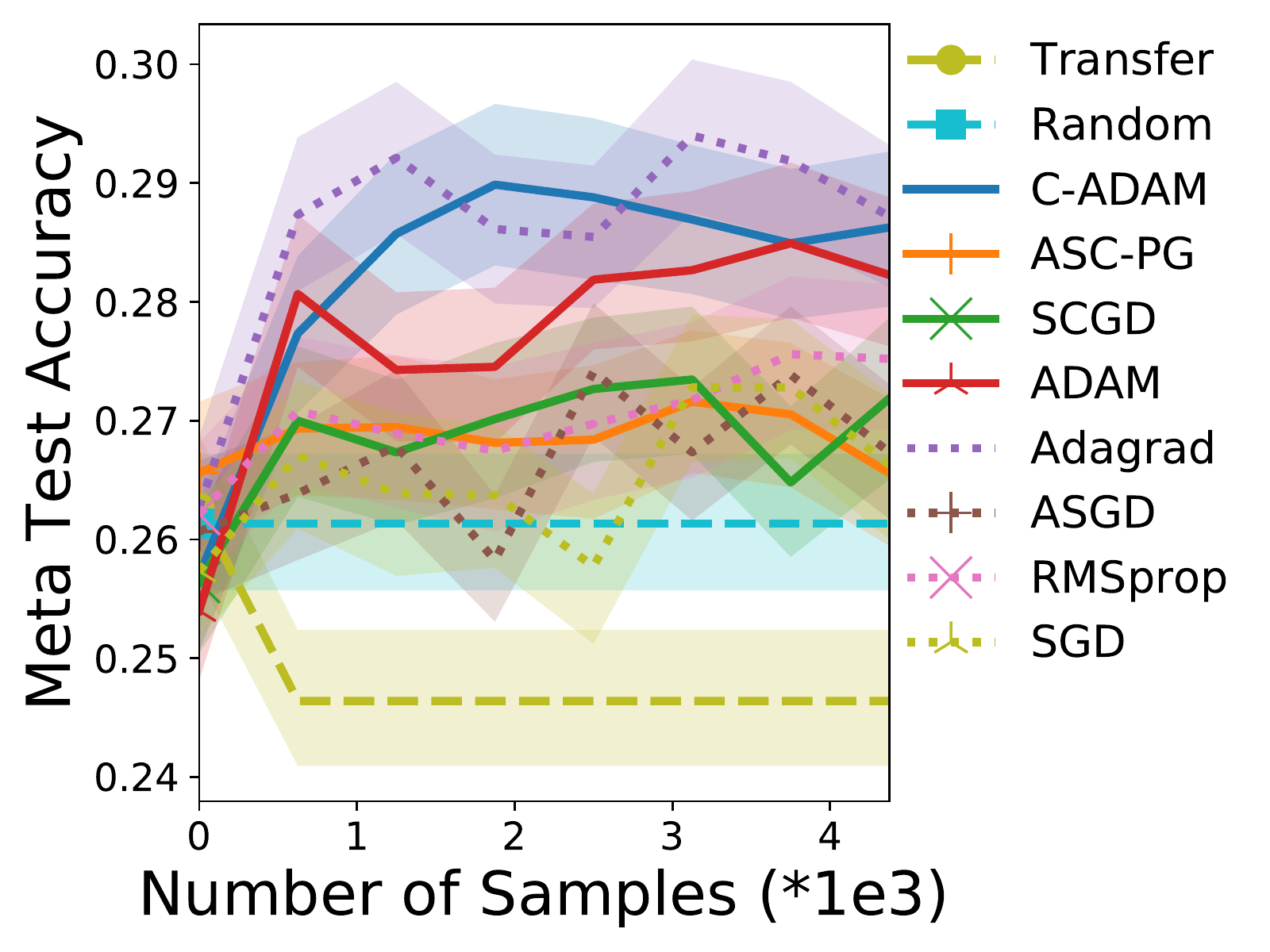}
\caption{}
\end{subfigure}
\begin{subfigure}{0.245\textwidth}
\centering
\includegraphics[width=\textwidth, trim={0 0 0 0.95cm },clip]{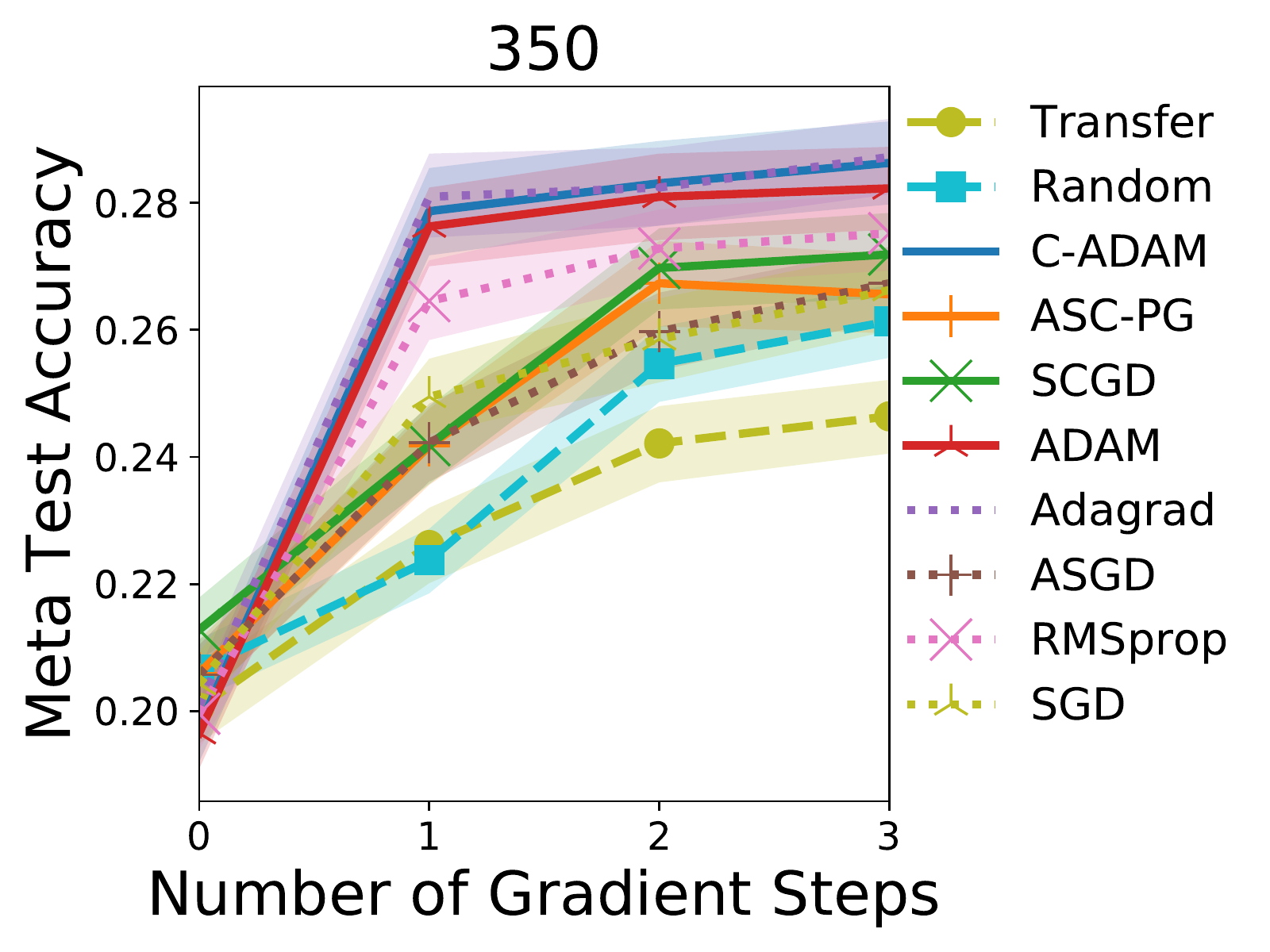}
\caption{}
\end{subfigure}
\caption{We show four scenarios on Omniglot and MiniImagenet data highlighting test statistics of C-MAML compared against other optimisation methods for varying N-shot K-way values using the non-convolutional model. (a)-(d) refer to the Omniglot 1-shot 5-way task, (e)-(h) Omniglot refers to 1-shot 5-way task, (i)-(l) refers to Omniglot 1-shot 20-way task, (m)-(p) refers to the Omniglot 5-shot 20-way task, (q)-(r) refers to the MiniImagenet 1-shot 5-way task and (s)-(t) refers to the MiniImagenet 5-shot 5-way task.}
\label{fig:nshot_kway_omniglot}
\end{figure*}



%% file: main.bbl
\begin{thebibliography}{57}
\providecommand{\natexlab}[1]{#1}
\providecommand{\url}[1]{\texttt{#1}}
\expandafter\ifx\csname urlstyle\endcsname\relax
  \providecommand{\doi}[1]{doi: #1}\else
  \providecommand{\doi}{doi: \begingroup \urlstyle{rm}\Url}\fi

\bibitem[Abdullah et~al.(2019)Abdullah, Ren, Ammar, Milenkovic, Luo, Zhang, and
  Wang]{abdullah2019wasserstein}
Abdullah, M.~A., Ren, H., Ammar, H.~B., Milenkovic, V., Luo, R., Zhang, M., and
  Wang, J.
\newblock Wasserstein robust reinforcement learning, 2019.

\bibitem[Allen-Zhu \& Hazan(2016)Allen-Zhu and Hazan]{allenzhu2016variance}
Allen-Zhu, Z. and Hazan, E.
\newblock Variance reduction for faster non-convex optimization, 2016.

\bibitem[Beck(2017)]{beck2017first}
Beck, A.
\newblock \emph{First-Order Methods in Optimization}.
\newblock MOS-SIAM Series on Optimization. Society for Industrial and Applied
  Mathematics, 2017.
\newblock ISBN 9781611974980.
\newblock URL \url{https://books.google.co.uk/books?id=xLk4DwAAQBAJ}.

\bibitem[Bedi et~al.(2019)Bedi, Koppel, and Rajawat]{bedi2019nonparametric}
Bedi, A.~S., Koppel, A., and Rajawat, K.
\newblock Nonparametric compositional stochastic optimization, 2019.

\bibitem[Bojarski et~al.(2016)Bojarski, Del~Testa, Dworakowski, Firner, Flepp,
  Goyal, Jackel, Monfort, Muller, Zhang, et~al.]{bojarski2016end}
Bojarski, M., Del~Testa, D., Dworakowski, D., Firner, B., Flepp, B., Goyal, P.,
  Jackel, L.~D., Monfort, M., Muller, U., Zhang, J., et~al.
\newblock End to end learning for self-driving cars.
\newblock \emph{arXiv preprint arXiv:1604.07316}, 2016.

\bibitem[Bou{-}Ammar et~al.(2014)Bou{-}Ammar, Eaton, Ruvolo, and
  Taylor]{Ammar1}
Bou{-}Ammar, H., Eaton, E., Ruvolo, P., and Taylor, M.~E.
\newblock Online multi-task learning for policy gradient methods.
\newblock In \emph{Proceedings of the 31th International Conference on Machine
  Learning, {ICML} 2014, Beijing, China, 21-26 June 2014}, pp.\  1206--1214,
  2014.

\bibitem[Bou{-}Ammar et~al.(2015{\natexlab{a}})Bou{-}Ammar, Eaton, Luna, and
  Ruvolo]{Ammar3}
Bou{-}Ammar, H., Eaton, E., Luna, J., and Ruvolo, P.
\newblock Autonomous cross-domain knowledge transfer in lifelong policy
  gradient reinforcement learning.
\newblock In \emph{Proceedings of the Twenty-Fourth International Joint
  Conference on Artificial Intelligence, {IJCAI} 2015, Buenos Aires, Argentina,
  July 25-31, 2015}, pp.\  3345--3351, 2015{\natexlab{a}}.

\bibitem[Bou{-}Ammar et~al.(2015{\natexlab{b}})Bou{-}Ammar, Tutunov, and
  Eaton]{Ammar2}
Bou{-}Ammar, H., Tutunov, R., and Eaton, E.
\newblock Safe policy search for lifelong reinforcement learning with sublinear
  regret.
\newblock \emph{CoRR}, abs/1505.05798, 2015{\natexlab{b}}.
\newblock URL \url{http://arxiv.org/abs/1505.05798}.

\bibitem[Boyd \& Vandenberghe(2004)Boyd and Vandenberghe]{BoydBook}
Boyd, S. and Vandenberghe, L.
\newblock \emph{Convex Optimization}.
\newblock Cambridge University Press, USA, 2004.
\newblock ISBN 0521833787.

\bibitem[Duchi et~al.(2011)Duchi, Hazan, and Singer]{ADAgrad}
Duchi, J., Hazan, E., and Singer, Y.
\newblock Adaptive subgradient methods for online learning and stochastic
  optimization.
\newblock \emph{J. Mach. Learn. Res.}, 12\penalty0 (null):\penalty0
  2121–2159, July 2011.
\newblock ISSN 1532-4435.

\bibitem[Faes et~al.(2019)Faes, Wagner, Fu, Liu, Korot, Ledsam, Back, Chopra,
  Pontikos, Kern, et~al.]{faes2019automated}
Faes, L., Wagner, S.~K., Fu, D.~J., Liu, X., Korot, E., Ledsam, J.~R., Back,
  T., Chopra, R., Pontikos, N., Kern, C., et~al.
\newblock Automated deep learning design for medical image classification by
  health-care professionals with no coding experience: a feasibility study.
\newblock \emph{The Lancet Digital Health}, 1\penalty0 (5):\penalty0
  e232--e242, 2019.

\bibitem[Fallah et~al.(2019)Fallah, Mokhtari, and
  Ozdaglar]{fallah2019convergence}
Fallah, A., Mokhtari, A., and Ozdaglar, A.
\newblock On the convergence theory of gradient-based model-agnostic
  meta-learning algorithms, 2019.

\bibitem[Fang et~al.(2018)Fang, Li, Lin, and Zhang]{fang2018spider}
Fang, C., Li, C.~J., Lin, Z., and Zhang, T.
\newblock Spider: Near-optimal non-convex optimization via stochastic path
  integrated differential estimator, 2018.

\bibitem[Finn et~al.(2017)Finn, Abbeel, and Levine]{FinnAL17}
Finn, C., Abbeel, P., and Levine, S.
\newblock Model-agnostic meta-learning for fast adaptation of deep networks.
\newblock \emph{CoRR}, abs/1703.03400, 2017.
\newblock URL \url{http://arxiv.org/abs/1703.03400}.

\bibitem[Finn et~al.(2018)Finn, Xu, and Levine]{finn2018probabilistic}
Finn, C., Xu, K., and Levine, S.
\newblock Probabilistic model-agnostic meta-learning.
\newblock In \emph{Advances in Neural Information Processing Systems}, pp.\
  9516--9527, 2018.

\bibitem[Finn et~al.(2019)Finn, Rajeswaran, Kakade, and Levine]{OnlineMAML}
Finn, C., Rajeswaran, A., Kakade, S.~M., and Levine, S.
\newblock Online meta-learning.
\newblock \emph{CoRR}, abs/1902.08438, 2019.
\newblock URL \url{http://arxiv.org/abs/1902.08438}.

\bibitem[Gabillon et~al.(2019)Gabillon, Tutunov, Valko, and
  Ammar]{gabillon2019derivative}
Gabillon, V., Tutunov, R., Valko, M., and Ammar, H.~B.
\newblock Derivative-free \& order-robust optimisation.
\newblock \emph{arXiv preprint arXiv:1910.04034}, 2019.

\bibitem[Golmant(2019)]{golmantconvergence}
Golmant, N.
\newblock On the convergence of model-agnostic meta-learning.
\newblock 2019.
\newblock URL \url{http://noahgolmant.com/writings/maml.pdf}.

\bibitem[Grant et~al.(2018)Grant, Finn, Levine, Darrell, and
  Griffiths]{grant2018recasting}
Grant, E., Finn, C., Levine, S., Darrell, T., and Griffiths, T.
\newblock Recasting gradient-based meta-learning as hierarchical bayes.
\newblock \emph{arXiv preprint arXiv:1801.08930}, 2018.

\bibitem[Grau-Moya et~al.(2018)Grau-Moya, Leibfried, and
  Bou-Ammar]{graumoya2018balancing}
Grau-Moya, J., Leibfried, F., and Bou-Ammar, H.
\newblock Balancing two-player stochastic games with soft q-learning, 2018.

\bibitem[Hu et~al.(2019)Hu, Li, Lian, Liu, and Yuan]{Hu_WenqingNIPS2019}
Hu, W., Li, C.~J., Lian, X., Liu, J., and Yuan, H.
\newblock Efficient smooth non-convex stochastic compositional optimization via
  stochastic recursive gradient descent.
\newblock In Wallach, H., Larochelle, H., Beygelzimer, A., d\textquotesingle
  Alch\'{e}-Buc, F., Fox, E., and Garnett, R. (eds.), \emph{Advances in Neural
  Information Processing Systems 32}, pp.\  6926--6935. Curran Associates,
  Inc., 2019.

\bibitem[Hu et~al.(2016)Hu, Prashanth, Gy{\"o}rgy, and
  Szepesv{\'a}ri]{hu2016bandit}
Hu, X., Prashanth, L., Gy{\"o}rgy, A., and Szepesv{\'a}ri, C.
\newblock (bandit) convex optimization with biased noisy gradient oracles.
\newblock In \emph{Artificial Intelligence and Statistics}, pp.\  819--828,
  2016.

\bibitem[Huo et~al.(2017)Huo, Gu, and Huang]{ZHuo2017}
Huo, Z., Gu, B., and Huang, H.
\newblock Accelerated method for stochastic composition optimization with
  nonsmooth regularization.
\newblock \emph{CoRR}, abs/1711.03937, 2017.
\newblock URL \url{http://arxiv.org/abs/1711.03937}.

\bibitem[Kim et~al.(2018)Kim, Yoon, Dia, Kim, Bengio, and Ahn]{kim2018bayesian}
Kim, T., Yoon, J., Dia, O., Kim, S., Bengio, Y., and Ahn, S.
\newblock Bayesian model-agnostic meta-learning.
\newblock \emph{arXiv preprint arXiv:1806.03836}, 2018.

\bibitem[Kingma \& Ba(2014)Kingma and Ba]{kingma2014adam}
Kingma, D.~P. and Ba, J.
\newblock Adam: A method for stochastic optimization, 2014.

\bibitem[Lake et~al.(2011)Lake, Salakhutdinov, Gross, and
  Tenenbaum]{lake2011one}
Lake, B., Salakhutdinov, R., Gross, J., and Tenenbaum, J.
\newblock One shot learning of simple visual concepts.
\newblock In \emph{Proceedings of the annual meeting of the cognitive science
  society}, volume~33, 2011.

\bibitem[Li \& Lin(2015)Li and Lin]{APG}
Li, H. and Lin, Z.
\newblock Accelerated proximal gradient methods for nonconvex programming.
\newblock In \emph{Proceedings of the 28th International Conference on Neural
  Information Processing Systems - Volume 1}, NIPS’15, pp.\  379–387,
  Cambridge, MA, USA, 2015. MIT Press.

\bibitem[Lian et~al.(2016)Lian, Wang, and Liu]{lian2016finitesum}
Lian, X., Wang, M., and Liu, J.
\newblock Finite-sum composition optimization via variance reduced gradient
  descent, 2016.

\bibitem[Lin et~al.(2018)Lin, Fan, Wang, and Jordan]{lin2018improved}
Lin, T., Fan, C., Wang, M., and Jordan, M.~I.
\newblock Improved sample complexity for stochastic compositional variance
  reduced gradient, 2018.

\bibitem[Liu et~al.(2017)Liu, Liu, and Tao]{liu2017variance}
Liu, L., Liu, J., and Tao, D.
\newblock Variance reduced methods for non-convex composition optimization,
  2017.

\bibitem[Liu et~al.(2018)Liu, Liu, Hsieh, and Tao]{liu2018stochastically}
Liu, L., Liu, J., Hsieh, C.-J., and Tao, D.
\newblock Stochastically controlled stochastic gradient for the convex and
  non-convex composition problem, 2018.

\bibitem[Menghan(2019)]{ClassFirstOrder}
Menghan, Z.
\newblock A review of classical first order optimization methods in machine
  learning.
\newblock In Deng, K., Yu, Z., Patnaik, S., and Wang, J. (eds.), \emph{Recent
  Developments in Mechatronics and Intelligent Robotics}, pp.\  1183--1190,
  Cham, 2019. Springer International Publishing.
\newblock ISBN 978-3-030-00214-5.

\bibitem[Mguni(2018)]{mguni2018viscosity}
Mguni, D.
\newblock A viscosity approach to stochastic differential games of control and
  stopping involving impulsive control, 2018.

\bibitem[Mnih et~al.(2015)Mnih, Kavukcuoglu, Silver, Rusu, Veness, Bellemare,
  Graves, Riedmiller, Fidjeland, Ostrovski, et~al.]{mnih2015human}
Mnih, V., Kavukcuoglu, K., Silver, D., Rusu, A.~A., Veness, J., Bellemare,
  M.~G., Graves, A., Riedmiller, M., Fidjeland, A.~K., Ostrovski, G., et~al.
\newblock Human-level control through deep reinforcement learning.
\newblock \emph{Nature}, 518\penalty0 (7540):\penalty0 529--533, 2015.

\bibitem[Mukkamala \& Hein(2017)Mukkamala and Hein]{RMSprop}
Mukkamala, M.~C. and Hein, M.
\newblock Variants of rmsprop and adagrad with logarithmic regret bounds.
\newblock In \emph{Proceedings of the 34th International Conference on Machine
  Learning - Volume 70}, ICML’17, pp.\  2545–2553. JMLR.org, 2017.

\bibitem[Nesterov(2014)]{Nesterov}
Nesterov, Y.
\newblock \emph{Introductory Lectures on Convex Optimization: A Basic Course}.
\newblock Springer Publishing Company, Incorporated, 1 edition, 2014.
\newblock ISBN 1461346916.

\bibitem[Nesterov \& Polyak(2006)Nesterov and Polyak]{Nesterov2006}
Nesterov, Y. and Polyak, B.
\newblock Cubic regularization of newton method and its global performance.
\newblock \emph{Mathematical Programming}, 108\penalty0 (1):\penalty0 177--205,
  Aug 2006.
\newblock ISSN 1436-4646.
\newblock \doi{10.1007/s10107-006-0706-8}.
\newblock URL \url{https://doi.org/10.1007/s10107-006-0706-8}.

\bibitem[Polyak \& Juditsky(1992)Polyak and Juditsky]{polyak1992acceleration}
Polyak, B.~T. and Juditsky, A.~B.
\newblock Acceleration of stochastic approximation by averaging.
\newblock \emph{SIAM journal on control and optimization}, 30\penalty0
  (4):\penalty0 838--855, 1992.

\bibitem[Ravi \& Larochelle(2016)Ravi and Larochelle]{ravi2016optimization}
Ravi, S. and Larochelle, H.
\newblock Optimization as a model for few-shot learning.
\newblock 2016.

\bibitem[Ravikumar et~al.(2007)Ravikumar, Lafferty, Liu, and
  Wasserman]{ravikumar2007sparse}
Ravikumar, P., Lafferty, J., Liu, H., and Wasserman, L.
\newblock Sparse additive models, 2007.

\bibitem[Reddi et~al.(2019)Reddi, Kale, and Kumar]{OnConvAdamand_Beyond}
Reddi, S.~J., Kale, S., and Kumar, S.
\newblock On the convergence of adam and beyond.
\newblock \emph{CoRR}, abs/1904.09237, 2019.
\newblock URL \url{http://arxiv.org/abs/1904.09237}.

\bibitem[Ruvolo \& Eaton(2013)Ruvolo and Eaton]{Ruvolo2013ELLA}
Ruvolo, P. and Eaton, E.
\newblock Ella: An efficient lifelong learning algorithm.
\newblock In \emph{Proceedings of the 30th International Conference on Machine
  Learning (ICML-13)}, June 2013.

\bibitem[Shamir(2017)]{shamir2017optimal}
Shamir, O.
\newblock An optimal algorithm for bandit and zero-order convex optimization
  with two-point feedback.
\newblock \emph{The Journal of Machine Learning Research}, 18\penalty0
  (1):\penalty0 1703--1713, 2017.

\bibitem[Silver et~al.(2017)Silver, Hubert, Schrittwieser, Antonoglou, Lai,
  Guez, Lanctot, Sifre, Kumaran, Graepel, et~al.]{silver2017mastering}
Silver, D., Hubert, T., Schrittwieser, J., Antonoglou, I., Lai, M., Guez, A.,
  Lanctot, M., Sifre, L., Kumaran, D., Graepel, T., et~al.
\newblock Mastering chess and shogi by self-play with a general reinforcement
  learning algorithm.
\newblock \emph{arXiv preprint arXiv:1712.01815}, 2017.

\bibitem[Sun et~al.(2019)Sun, Cao, Zhu, and Zhao]{FirstOrder_review_1}
Sun, S., Cao, Z., Zhu, H., and Zhao, J.
\newblock A survey of optimization methods from a machine learning perspective.
\newblock \emph{CoRR}, abs/1906.06821, 2019.
\newblock URL \url{http://arxiv.org/abs/1906.06821}.

\bibitem[Thrun \& Pratt(1998)Thrun and Pratt]{thrun1998learning}
Thrun, S. and Pratt, L.
\newblock Learning to learn: Introduction and overview.
\newblock In \emph{Learning to learn}, pp.\  3--17. Springer, 1998.

\bibitem[Tian et~al.(2018)Tian, Zou, Warr, Wu, and Wang]{tian2018learning}
Tian, Z., Zou, S., Warr, T., Wu, L., and Wang, J.
\newblock Learning to communicate implicitly by actions.
\newblock \emph{arXiv preprint arXiv:1810.04444}, 2018.

\bibitem[Tutunov et~al.(2016)Tutunov, Bou{-}Ammar, and Jadbabaie]{TutunovBJ16}
Tutunov, R., Bou{-}Ammar, H., and Jadbabaie, A.
\newblock A distributed newton method for large scale consensus optimization.
\newblock \emph{CoRR}, abs/1606.06593, 2016.
\newblock URL \url{http://arxiv.org/abs/1606.06593}.

\bibitem[Vuorio et~al.(2018)Vuorio, Sun, Hu, and Lim]{vuorio2018toward}
Vuorio, R., Sun, S.-H., Hu, H., and Lim, J.~J.
\newblock Toward multimodal model-agnostic meta-learning.
\newblock \emph{arXiv preprint arXiv:1812.07172}, 2018.

\bibitem[Wang et~al.(2014)Wang, Fang, and Liu]{wang2014stochastic}
Wang, M., Fang, E.~X., and Liu, H.
\newblock Stochastic compositional gradient descent: Algorithms for minimizing
  compositions of expected-value functions, 2014.

\bibitem[Wang et~al.(2016)Wang, Liu, and Fang]{Lui2016}
Wang, M., Liu, J., and Fang, E.
\newblock Accelerating stochastic composition optimization.
\newblock In Lee, D.~D., Sugiyama, M., Luxburg, U.~V., Guyon, I., and Garnett,
  R. (eds.), \emph{Advances in Neural Information Processing Systems 29}, pp.\
  1714--1722. Curran Associates, Inc., 2016.

\bibitem[Wang et~al.(2017)Wang, Fang, and Liu]{Mendgi_2017}
Wang, M., Fang, E.~X., and Liu, H.
\newblock Stochastic compositional gradient descent: Algorithms for minimizing
  compositions of expected-value functions.
\newblock \emph{Math. Program.}, 161\penalty0 (1–2):\penalty0 419–449,
  January 2017.
\newblock ISSN 0025-5610.
\newblock \doi{10.1007/s10107-016-1017-3}.
\newblock URL \url{https://doi.org/10.1007/s10107-016-1017-3}.

\bibitem[Wang et~al.(2020)Wang, Nie, Wang, Yang, and Long]{wang2020deep}
Wang, R., Nie, K., Wang, T., Yang, Y., and Long, B.
\newblock Deep learning for anomaly detection.
\newblock In \emph{Proceedings of the 13th International Conference on Web
  Search and Data Mining}, pp.\  894--896, 2020.

\bibitem[Wang \& Yao(2019)Wang and Yao]{Few}
Wang, Y. and Yao, Q.
\newblock Few-shot learning: {A} survey.
\newblock \emph{CoRR}, abs/1904.05046, 2019.
\newblock URL \url{http://arxiv.org/abs/1904.05046}.

\bibitem[Wen et~al.(2019)Wen, Yang, Luo, and Wang]{wen2019modelling}
Wen, Y., Yang, Y., Luo, R., and Wang, J.
\newblock Modelling bounded rationality in multi-agent interactions by
  generalized recursive reasoning, 2019.

\bibitem[Zaheer et~al.(2018)Zaheer, Reddi, Sachan, Kale, and
  Kumar]{NonAdaptiveGR}
Zaheer, M., Reddi, S., Sachan, D., Kale, S., and Kumar, S.
\newblock Adaptive methods for nonconvex optimization.
\newblock In Bengio, S., Wallach, H., Larochelle, H., Grauman, K.,
  Cesa-Bianchi, N., and Garnett, R. (eds.), \emph{Advances in Neural
  Information Processing Systems 31}, pp.\  9793--9803. Curran Associates,
  Inc., 2018.

\bibitem[Zeiler(2012)]{zeiler2012adadelta}
Zeiler, M.~D.
\newblock Adadelta: An adaptive learning rate method, 2012.

\end{thebibliography}
